\documentclass{article}
\PassOptionsToPackage{numbers, sort&compress}{natbib}
\usepackage[final]{neurips_2023}

\usepackage{graphicx}%
\usepackage{multirow}%
\usepackage{amsmath,amssymb,amsfonts}%
\usepackage{amsthm}%
\usepackage{mathrsfs}%
\usepackage{xcolor}%
\usepackage{textcomp}%
\usepackage{manyfoot}%
\usepackage{booktabs}%
\usepackage{algorithm}%
\usepackage{algorithmicx}%
\usepackage{algpseudocode}%
\usepackage{listings}%
\usepackage{rotating}%

\graphicspath{{figs/}}
\usepackage{amssymb}
\usepackage{amsmath,amsfonts}
\usepackage{amsopn,bm,mathtools}

\usepackage{nicefrac}       
\newcommand{\vct}[1]{\boldsymbol{#1}} 

\newcommand{\ProbOpr}[1]{\mathbb{#1}}

\newcommand{\expect}[2]{%
\ifthenelse{\equal{#2}{}}{\ProbOpr{E}_{#1}}
{\ifthenelse{\equal{#1}{}}{\ProbOpr{E}\left[#2\right]}{\ProbOpr{E}_{#1}\left[#2\right]}}} 
\newcommand{\var}[2]{%
\ifthenelse{\equal{#2}{}}{\ProbOpr{VAR}_{#1}}
{\ifthenelse{\equal{#1}{}}{\ProbOpr{VAR}\left[#2\right]}{\ProbOpr{VAR}_{#1}\left[#2\right]}}} 

\newcommand{\vc}{\vct{c}}

\newcommand{\vu}{\vct{u}}
\newcommand{\vv}{\vct{v}}
\newcommand{\vV}{\vct{V}}
\newcommand{\vw}{\vct{w}}
\newcommand{\vx}{{\vct{x}}}

\usepackage{xspace}

\let\olditemize=\itemize
\let\endolditemize=\enditemize
\renewenvironment{itemize}{
    \olditemize
    \setlength{\itemsep}{0pt}
    \setlength{\parskip}{0pt}
}{\endolditemize}

\let\oldenumerate=\enumerate
\let\endoldenumerate=\endenumerate
\renewenvironment{enumerate}{
    \oldenumerate
    \setlength{\itemsep}{0pt}
    \setlength{\parskip}{0pt}
}{\endoldenumerate}

\makeatletter
\DeclareRobustCommand\onedot{\futurelet\@let@token\@onedot}
\def\@onedot{\ifx\@let@token.\else.\null\fi\xspace}

\def\ie{\emph{i.e}\onedot}

\makeatother

\usepackage[colorinlistoftodos,linecolor=blue,backgroundcolor=white,bordercolor=blue,textsize=tiny,textwidth=28mm]
{todonotes}
\newcounter{task}
\newcounter{fs}
\newcounter{larry}
\newcounter{ig}
\newcounter{rc}
\newcommand{\eat}[1]{}

\newcommand{\ourmethod}{Scalable Ensemble Envelope Diffusion Sampler \xspace}
\newcommand{\ourmethodshort}{SEEDS\xspace}
\newcommand{\ourmethodlower}{seeds\xspace}

\newcommand{\gefsfull}{\textsc{gefs-full}\xspace}
\newcommand{\gefsK}{\textsc{gefs-2}\xspace}
\newcommand{\era}{\textsc{era5}\xspace}
\newcommand{\erafull}{\textsc{era5-10}\xspace}

\newcommand{\reforecastfull}{\textsc{gefs-rf5}\xspace}
\newcommand{\ourgee}{\textsc{\ourmethodlower-gee}\xspace}
\newcommand{\ourgpp}{\textsc{\ourmethodlower-gpp}\xspace}

\newcommand{\rmse}{\textsc{rmse}\xspace}
\newcommand{\acc}{\textsc{acc}\xspace}

\newcommand{\crps}{\textsc{crps}\xspace}
\newcommand{\brier}{\textsc{Brier}\xspace}
\newcommand{\corr}{\textsc{corr}\xspace}
\newcommand{\logloss}{\textsc{logloss}\xspace}
\usepackage{soul}
\usepackage[hidelinks, colorlinks=true, citecolor=blue]{hyperref} 

\begin{document}

\title{SEEDS:  Emulation of Weather Forecast Ensembles with Diffusion Models}
\author{%
Lizao Li\\
Google Research\\
\texttt{lizaoli@google.com}\\
\And
Robert Carver\thanks{Contributed equally}\\
Google Research\\
\texttt{carver@google.com}\\
\And
Ignacio Lopez-Gomez$^*$\\\
Google Research\\
\texttt{ilopezgp@google.com}
\And
Fei Sha\thanks{to whom correspondence should be addressed}\\
Google Research\\
\texttt{fsha@google.com}\\
\And
John Anderson\\
Google Research\\
\texttt{janders@google.com}\\
}

\maketitle

\begin{abstract}

Uncertainty quantification is crucial to decision-making. A prominent example is probabilistic forecasting in numerical weather prediction. The dominant approach to representing uncertainty in weather forecasting is to generate an ensemble of forecasts. This is done by running many physics-based simulations under different conditions, which is a computationally costly process. We propose to amortize the computational cost by emulating these forecasts with deep generative diffusion models learned from historical data.  The learned models are highly scalable with respect to high-performance computing accelerators and can sample hundreds to tens of thousands of realistic weather forecasts at low cost. When designed to emulate operational ensemble forecasts, the generated ones are similar to physics-based ensembles in important statistical properties and predictive skill.  When designed to  correct biases present in the operational forecasting system, the generated ensembles show improved probabilistic forecast metrics. They are more reliable and forecast probabilities of extreme weather events more accurately. While this work demonstrates the utility of the methodology by focusing on weather forecasting, the generative artificial intelligence methodology can be extended for uncertainty quantification in climate modeling, where we believe the generation of very large ensembles of climate projections will play an increasingly important role in climate risk assessment.
\end{abstract}

\section{Introduction}
\label{sIntro}

Weather is inherently uncertain, making the assessment of forecast uncertainty a vital component of operational weather forecasting~\citep{Bauer2015, Zhu2002}. Given a numerical weather prediction (NWP) model, the standard way to quantify this uncertainty is to stochastically perturb the model's initial conditions and its representation of small-scale physical processes to create an ensemble of weather trajectories~\citep{Palmer2019}. These trajectories are then regarded as Monte Carlo samples of the underlying probability distribution of weather states.

Given the computational cost of generating each ensemble member, weather forecasting centers can only afford to generate 10 to 50 members for each forecast cycle~\citep{CY47R3-ch5, Leutbecher2019, Zhou2022-nv}.
This limitation is particularly problematic for users concerned with the likelihood of high-impact extreme or rare weather events, which typically requires much larger ensembles to assess~\citep{Palmer2002, Palmer2017, Fischer2023}. For instance, one would need a 10,000-member calibrated ensemble
to forecast events with $1\%$ probability of occurrence with a relative error less than $10\%$.
Large ensembles are even more necessary for forecasting compound extreme events~\citep{Bevacqua2023, Leutbecher2019}.
Besides relying on increases in available computational power to generate larger ensembles in the future, it is imperative to explore more efficient approaches for generating ensemble forecasts.

In this context, recent advances in generative artificial intelligence (GAI) offer a potential path towards massive reductions in the cost of ensemble forecasting. GAI models extract statistical priors from datasets, and enable conditional and unconditional sampling from the learned probability distributions. Through this mechanism, GAI techniques reduce the cost of ensemble forecast generation: once learning is complete, the sampling process is far more computationally efficient than time-stepping a physics-based NWP model.

In this work, we propose a technique that is based on probabilistic diffusion models, which have recently revolutionized GAI use cases such as image and video generation \citep{dhariwal_nichol_2021, ho2022imagen, StableDiffusion}. Our \ourmethod (\ourmethodshort) can generate an arbitrarily large ensemble conditioned on as few as one or two forecasts from an operational NWP system. We compare the generated ensembles to ground-truth ensembles from the operational systems, and to ERA5 reanalysis \citep{Hersbach2020-iu}. The generated ensembles not only yield weather-like forecasts but also match or exceed physics-based ensembles in skill metrics such as the rank histogram, the root-mean-squared error (RMSE) and continuous ranked probability score (CRPS). In particular, the generated ensembles assign more accurate likelihoods to the tail of the distribution, such as $\pm 2\sigma$ and $\pm 3\sigma$ weather events. Most importantly, the computational cost of the model is negligible; it has a throughput of 256 ensemble members (at $2^\circ$ resolution) per 3 minutes
on Google Cloud TPUv3-32 instances and can easily scale to higher throughput by deploying more accelerators.
We apply our methodology to uncertainty quantification in weather forecasting due to the wealth of data available and the ability to validate models on reanalysis. Nevertheless, the same approach could be used to augment climate projection ensembles.

Previous work leveraging artificial intelligence to augment and post-process ensemble or deterministic forecasts has focused on improving the aggregate output statistics of the prediction system. Convolutional neural networks have been used to learn a global measure of forecast uncertainty given a single deterministic forecast, trained using as labels either the error of previous forecasts or the spread of an ensemble system~\citep{Scher2018-ne}. This approach has been generalized to predict the ensemble spread at each location of the input deterministic forecast; over both small regions using fully connected networks~\citep{sacco22MLens}, or over the entire globe using conditional generative adversarial networks~\citep{Brecht2023} based on the \textsc{pix2pix} architecture~\citep{Isola2017}. Deep learning has also proved effective in calibrating limited-size ensembles. For instance, self-attentive transformers can be used to calibrate the ensemble spread~\citep{Finn2021-mu}. More related to our work, deep learning models have been successfully used to correct the probabilistic forecasts of ensemble prediction systems such that their final skill exceeds that of pure physics-based ensembles with at least double the number of members~\citep{Gronquist2021-am}.

Our work differs from all previous studies in that our probabilistic generative model outputs \emph{high-dimensional weather-like} samples from the target forecast distribution, akin to generative precipitation downscaling models \cite{Harris2022-af}. Thus, our approach offers added value beyond improved estimates of the ensemble mean and spread: the drawn samples can be used to characterize spatial patterns associated with weather extremes \citep{Scher2021}, or as input to targeted weather applications that depend on variable and spatial correlations \citep{Palmer2002}.

\section{Method}
\label{sMethod}

We start by framing the learning tasks. We then outline the data and neural network learning algorithm we use.  Details, including background, data processing and preparation, and learning architectures and procedures, are presented in Supplementary Information~\ref{sBackground} and~\ref{sMethodAppendix}.

\subsection{Setup}

To address the computational challenge of generating large weather forecast ensembles, we consider two learning tasks: \textbf{generative ensemble emulation} and \textbf{generative post-processing}. In both tasks, we are given as inputs a few examples sampled from a probability distribution $p(\vv)$, where $\vv$ stands for the atmospheric state variables. In our case, these examples represent physics-based weather forecasts. We seek to generate additional samples that either approximate the same distribution, or a related desired distribution. The central theme of statistical modeling for both tasks is to construct a computationally fast and scalable sampler for the target distributions.  

Generative ensemble emulation leverages $K$ input samples to conditionally generate $N >  K$ samples such that they approximate the original distribution $p(\vv)$ from which the input samples are drawn. Its main purpose is to augment the ensemble size inexpensively without the need to compute and issue more than $K$ physics-based forecasts.

In generative post-processing, the sampler generates $N>K$ samples such that they approximate a mixture distribution where $p(\vv)$ is just one of the components. We consider the case where the target distribution is $\alpha p(\vv) + (1-\alpha) p'(\vv)$, with $\alpha \in [0, 1)$ being the mixture weight and $p'(\vv)$ a different distribution. The generative post-processing task aims not only to augment the ensemble size, but also to bias the new samples towards $p'(\vv)$, which we take to be a distribution that more closely resembles actual weather. The underlying goal is to generate ensembles that are less biased than those provided by the physics-based model, while still quantifying the forecast uncertainty captured by $p(\vv)$. We emphasize that while this task has the flavor and also achieves the effect of debiasing to some degree,  we focus on generating samples instead of minimizing the difference between their mean and a given reanalysis or observations. In both the emulation and post-processing tasks, the smaller the value of $K$ is, the greater the computational savings. 

\begin{figure}[t]
    \centering
    \includegraphics[width=0.6\columnwidth,draft=false]{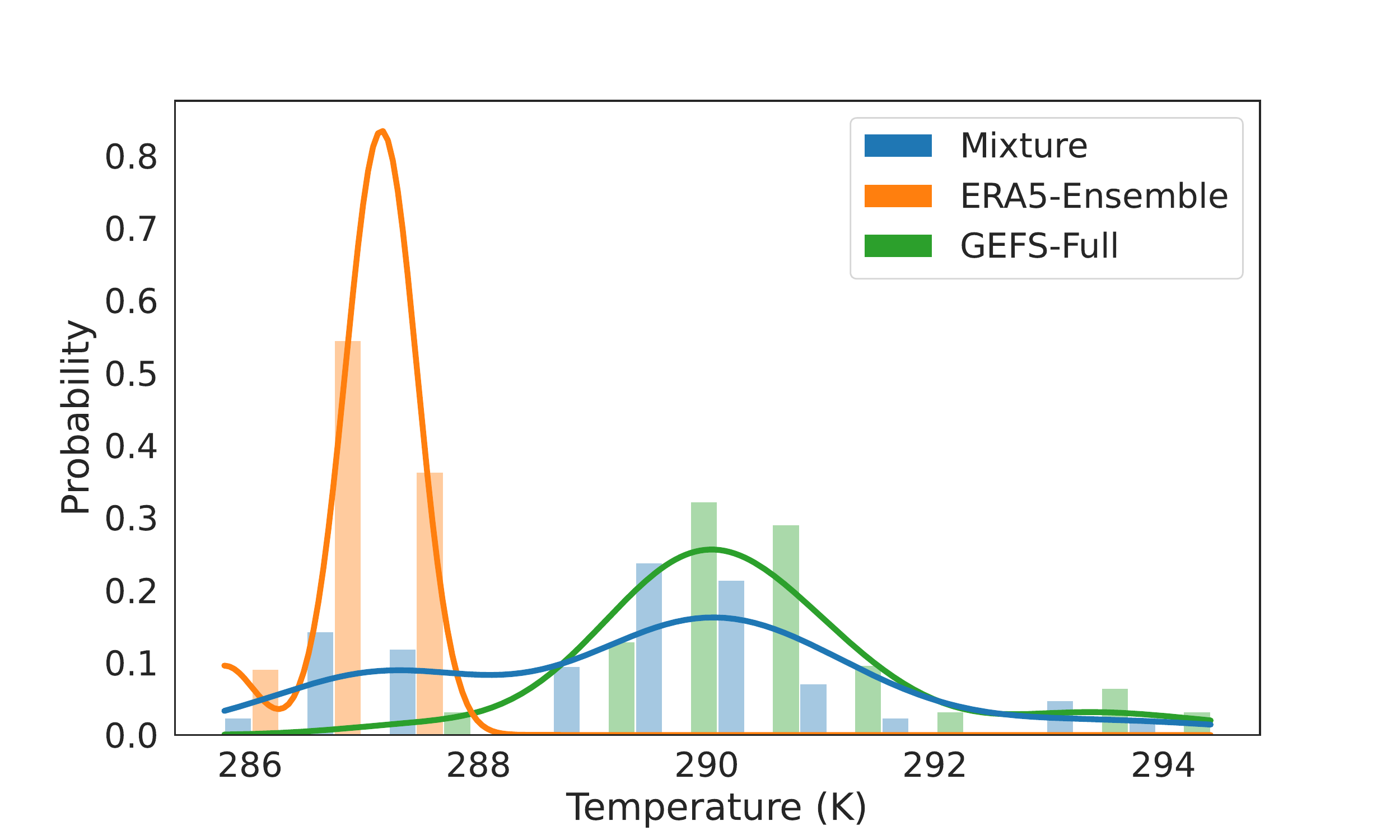}
    \caption{Illustration of the target distributions of generative ensemble emulation (\gefsfull) and post-processing (Mixture). Shown are the histograms (bars: frequencies with 12 shared bins, curves: Gaussian kernel density estimators fit to the bars), \ie, the empirical distributions of the surface temperature near Mountain View, CA on 2021/07/04 in the GEFS and ERA5 ensembles.  The goal common to both tasks is to generate additional ensemble members to capture the statistics of the desired distribution conditioned on a few GEFS samples. Note the small ``bump'' at the temperature of 287K in the mixture distribution.}
    \label{fDistributionConcepts}
\end{figure}

Figure~\ref{fDistributionConcepts} illustrates the concepts behind these two tasks. There, $p(\vv)$ is the distribution of the  surface temperature near Mountain View, CA on 2021/07/04 as predicted by the GEFS 13-day forecast ensemble \citep{Zhou2022-nv}, and $p'(\vv)$ the corresponding ERA5 reanalysis ensemble \citep{Hersbach2020-iu}. While the GEFS ensemble has 31 members, our goal is to use $K\ll31$ GEFS ensemble members to steer our samplers to generate additional forecast members that are consistent with either GEFS's statistics or the mixture distribution's statistics. Inspired by terminology from natural language understanding and computer vision, we refer to those $K$ input examples from $p(\vv)$ as ``seeds''. The desirability to have a small $K$ is in spirit similar to few-shot learning setups in those works.

We stress that the primary goal of both tasks is to improve the computational efficiency of ensemble weather forecasting, not to replace physics-based models. The generated samples should be not only consistent with the underlying distribution of atmospheric states (each sample is ``weather-like''), but also validated by standard forecast verification metrics. In this work, we examine the generated ensembles by comparing them to other physics-based ensembles using the rank histogram, the anomaly correlation coefficient (ACC), RMSE, CRPS, and rare event classification metrics, as defined in~\ref{sAppendixResults}.

\subsection{Data for Learning and Evaluation}
\label{sTrainData}

We target the GEFS (version 12) ensemble forecasting system for the generative ensemble emulation task~\citep{Zhou2022-nv}. We use 20 years of GEFS 5-member reforecasts~\citep{Guan2022-vl}, denoted hereafter as \reforecastfull, to learn $p(\vv)$. Generative post-processing attempts to remove systematic biases of the original forecasting system from the learned emulator. To this end, we take the ERA5 10-member Reanalysis Ensemble~\citep{Hersbach2020-iu}, denoted  as \erafull, to represent $p'(\vv)$ in the target mixture distribution. We also use ERA5 HRES reanalysis as a proxy for real historical observations when evaluating the skill of our generated ensemble predictions.

All data are derived from the publicly available sources listed in Table~\ref{tb:data_sources}. Table~\ref{tb:fields} lists the atmospheric state variables that are considered by our models. They are extracted and spatially regridded to the same cubed sphere mesh of size $6\times 48\times 48$ ($\approx2^\circ$ resolution) using inverse distance weighting with 4 neighbors~\citep{Ronchi1996}. We only retain the 00h-UTC time snapshots of the fields in Table~\ref{tb:fields} for each day.

The climatology is computed from the ERA5 HRES dataset, using the reference period 1990-2020. The daily climatological mean and standard deviation are obtained by smoothing these two time series with a 15-day centered window over the year with periodic boundary conditions. The mean and standard deviation for February 29th is the average of those for February 28th and March 1st.

Our models take as inputs and produce as outputs the standardized climatological anomalies of variables in Table~\ref{tb:fields}, defined as the standardized anomalies using the aforementioned climatological mean and standard deviation for the day of year and location, which facilitates learning~\citep{Dabernig2017, lg2023, Qian2021}.  The  outputs are converted back to raw values for evaluation.

For each unique pair of forecast lead time and number of seeds $K$, we train a diffusion model for the generative ensemble emulation task. For each unique triplet of lead time, $K$ and mixture weight $\alpha$, we train a model for the generative post-processing task. We provide results for lead times of $\{1, 4, 7, 10, 13, 16\}$~days, $K=2$ seeds, and generated ensembles with $N=512$ members. For the post-processing task, we consider the mixing ratio $\alpha=0.5$. The sensitivity to $K$, $N$, and $\alpha$ is explored in~\ref{sAppendixResults}.

We evaluate our models against the operational GEFS 31-member ensemble~\citep{Zhou2022-nv} (\gefsfull) and the ERA5 HRES reanalysis. Note that we can do so because the \gefsfull and \reforecastfull datasets are considered to have similar distributions --- the reforecasts are reruns of the operational GEFS model using historical initial conditions \citep{Guan2022-vl}. We use the 20 years from 2000 to 2019 for training, year 2020 and 2021 for validation, and year 2022 for evaluation. In particular, to accommodate the longest lead time of 16 days, we evaluate using the forecasts initialized from 2022/01/01 to 2022/12/15 (349 days in total) and the ERA5 HRES data aligned with the corresponding days.

\begin{table}[t]
\centering
\caption{Data Used for Training and Evaluation}
\label{tb:data_sources}

\begin{tabular}{@{}lccc@{}}
\toprule
Name & Date Range & Ensemble size & Citation\\
\midrule
ERA5-HRES & 1959/01/01 -- 2022/12/31 & 1 & \citep{Hersbach2020-iu} \\
ERA5-Ensemble & 1959/01/01 -- 2021/12/31 & 10 & \citep{Hersbach2020-iu}\\
GEFS & 2020/09/23 -- 2022/12/31 & 31 & \citep{Zhou2022-nv}\\
GEFS-Reforecast & 2000/01/01 - 2019/12/31 & 5 & \citep{Guan2022-vl}\\
\bottomrule
\end{tabular}
\end{table}

\begin{table}[t]
\centering
\caption{List of Atmospheric State Variables That Are Modeled}
\label{tb:fields}
\begin{tabular}{@{}ll@{}}
\toprule
Quantity & Processed Units \\
\midrule
Mean sea level pressure & $Pa$ \\
Temperature at 2 meters & $K$ \\
Eastward wind speed at 850hPa & $m/s$ \\
Northward wind speed at 850hPa & $m/s$ \\
Geopotential at 500hPa & $m^2/s^2$ \\
Temperature at 850hPa & $K$ \\
Total column water vapour & $kg/m^2$ \\
Specific humidity at 500 hPa & $kg/kg$ \\
\bottomrule
\end{tabular}
\end{table}

\subsection{Learning Method and Architecture}

The use of probabilistic diffusion models to parameterize the target distributions, conditioned on a few ``seeds'', is at the core of our statistical modeling algorithm for both tasks.

Probabilistic diffusion models are generative models of data.  The generative process follows a Markov chain. It starts with a random draw from an initial noise distribution -- often an isotropic multivariate Gaussian. Then it iteratively transforms and denoises the sample until it resembles a random draw from the data distribution \citep{Ho2020}. The iteration steps advance the diffusion time, which is independent from the real-world time. The denoising operation relies on the instantiation of a diffusion-time-dependent score function, which is the Jacobian of the log-likelihood of the data at a given diffusion time \citep{song2020score}. Score functions often take the form of deep learning architectures whose parameters are learned from training data.

Typically, the score is a function of the noisy sample and the diffusion time. In this case, the resulting data distribution is a model of the unconditional distribution of the training data.  When additional inputs are passed to the score function, such as $K$ seeding forecasts in our setting, the sampler constructs the  distribution conditioned on these inputs.

In this work, our choice of the score function is inspired by the Vision Transformer (ViT), which has been successfully applied to a range of computer vision tasks~\citep{Dosovitskiy2021ViT}. It is intuitive to view atmospheric data as a temporal sequence of snapshots, which are in turn viewed as ``images''. Each snapshot is formed by ``pixels'' covering the globe with ``color'' channels. In this case, the channels correspond to the collection of atmospheric variables at different vertical levels. These can easily exceed in number the very few color channels of a typical image, e.g. 3 in the case of an RGB image. Due to this, we use a variant of ViT via axial attention~\citep{ho2019axial}, so that the model remains moderate in size and can be trained efficiently.

Irrespective of the lead times and the number of seeds, all the models share the same architecture and  have about 114M trainable parameters.  They are trained with a batch size of 128 for 200K steps.  The training of each model takes slightly less than 18 hours on a $2\times 2 \times 4$ TPUv4 cluster. Inference (namely, ensemble generation) runs at batch size 512 on a $4\times 8$ TPUv3 cluster at less than 3 minutes per batch. It is thus very efficient and easily scalable to generate thousands of members.

\section{Results}
\label{sResults}

Using the \ourmethodshort methodology, we have developed two generative models. The \ourgee model learns to emulate the distribution of the U.S. operational ensemble NWP system, the Global Ensemble Forecast System (GEFS) Version 12~\citep{Zhou2022-nv}. The \ourgpp model learns to emulate a blended distribution that combines the GEFS ensemble with historical data from the ERA5 reanalysis of the European Centre for Medium-Range Weather Forecasts (ECMWF), aiming to correct underlying biases in the operational GEFS system (\ie, post-processing).

\ourgee is trained using 20 years of GEFS 5-member retrospective forecasts~\citep{Guan2022-vl}, and \ourgpp additionally learns from ECMWF's ERA5 10-member Reanalysis Ensemble over the same period~\citep{Hersbach2020-iu}. Once learned, both models take as inputs a few randomly selected member forecasts from the operational GEFS ensemble, which has 31 members. We refer to the selected members as the seeding forecasts. These seeds provide the physical basis used by the generative models to conditionally sample additional plausible weather states. Both \ourgee and \ourgpp can be used to generate ensembles with a significantly larger number of forecasts than operational physics-based systems, easily reaching hundreds to tens of thousands of members.

\begin{figure}[t]
    \centering
    \includegraphics[width=0.95\columnwidth]{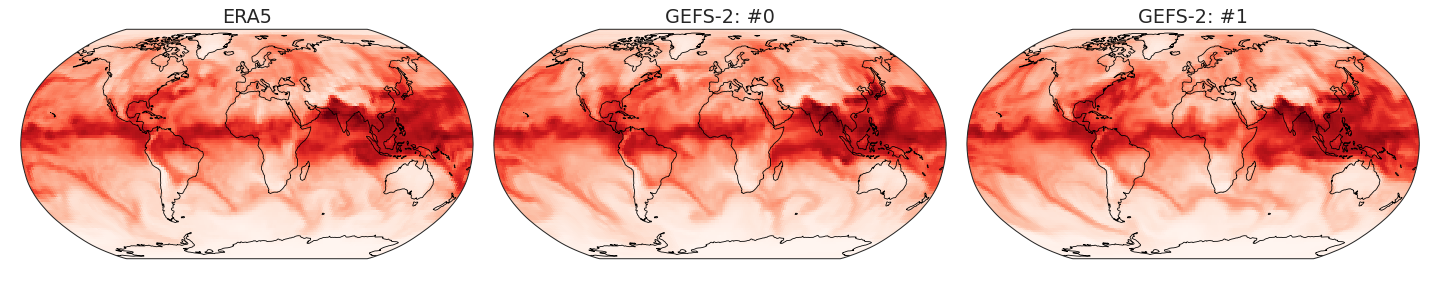}
    \includegraphics[width=0.95\columnwidth]{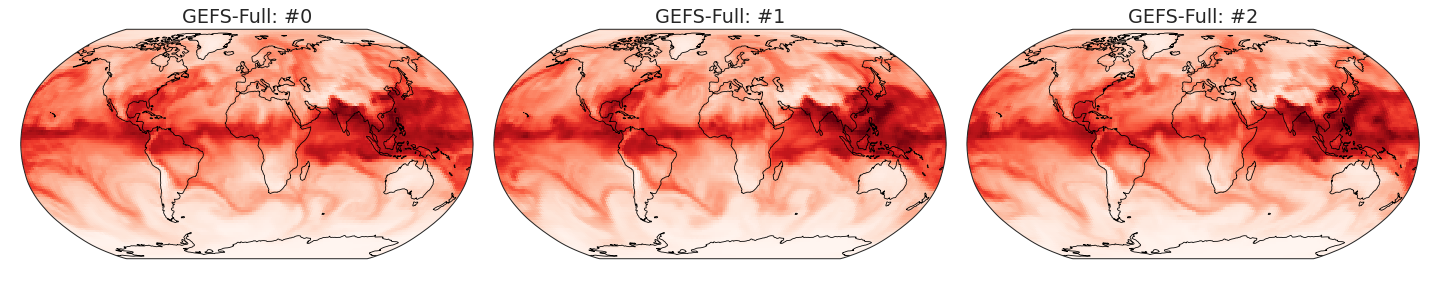}
    \includegraphics[width=0.95\columnwidth]{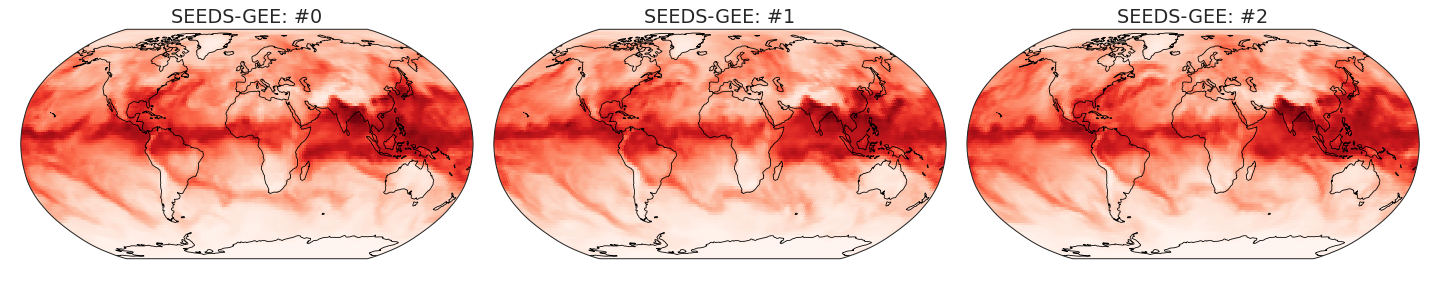}
    \includegraphics[width=0.7\columnwidth]{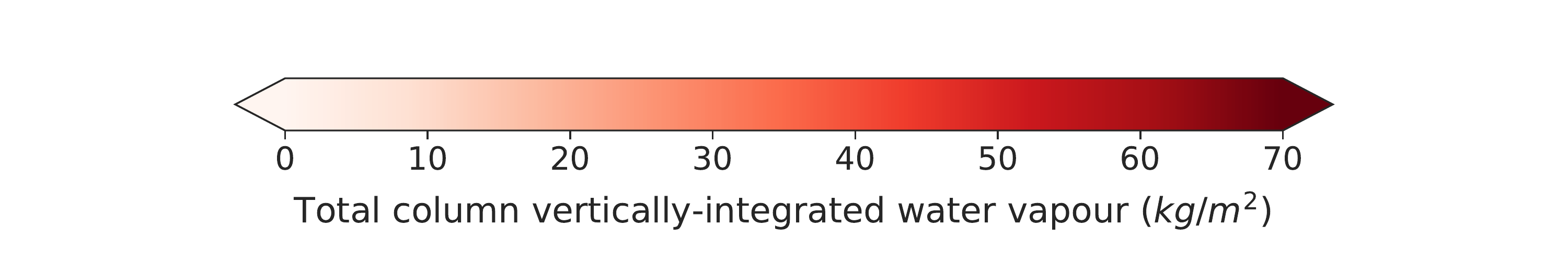}
    \caption{Maps of total column vertically-integrated water vapor ($kg/m^2$) for 2022/07/14, as captured by (top left) the ERA5 reanalysis, (top right and middle row) 5 members of the \gefsfull forecast issued with a 7-day lead time, and (bottom) 3 samples from \ourgee. The top 2 GEFS forecasts were used to seed the \ourgee sampler.}
  \label{fGlobalTCWV}
\end{figure}

Figure \ref{fGlobalTCWV} compares samples from the GEFS operational system, the ERA5 reanalysis, and the generative emulator \ourgee. We also assess the quality of the generated ensembles in terms of multiple important characteristics of useful ensemble prediction systems. First, we analyze whether the forecasts in the generative ensembles display spatial coherence, multivariate correlation structures, and wavenumber spectra consistent with actual weather states. Second, we compare the pointwise predictive skill of the generative ensembles and the full operational physics-based GEFS ensemble, measured against the ERA5 high resolution (HRES) reanalysis~\citep{Hersbach2020-iu}.

We report results on a subset of field variables: the mean sea level pressure, the temperature $2~\mathrm{m}$ above the surface, and the zonal wind speed at pressure level $850~\mathrm{hPa}$. Results for all modeled fields, listed in Table~\ref{tb:fields}, are presented in the Supplementary Information (SI). We use \gefsfull to refer to the full 31-member GEFS ensemble, and \gefsK to an ensemble made of $2$ randomly selected seeding forecasts. Unless noted, our generated ensembles have 512 members.

\subsection{Generated Weather Forecasts Are Plausible Weather Maps}
\label{sSpectra}

Ensemble forecasting systems are most useful when individual weather forecasts resemble real weather maps~\citep{Lorenz1982-gp}. This is because for many applications, such as ship routing, energy forecasting, or compound extreme event forecasting, capturing cross-field and spatial correlations is fundamental \citep{Palmer2002, Scher2021, worsnop_2018}.

\begin{figure}[pt]
    \centering
    \includegraphics[width=0.85\columnwidth]{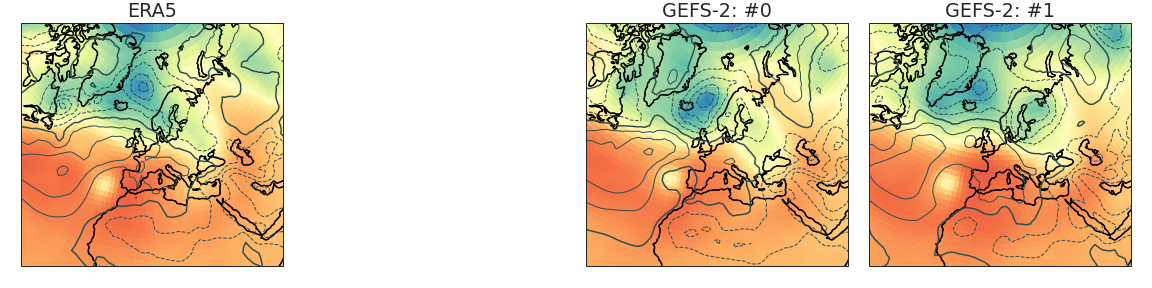}
    \includegraphics[width=0.85\columnwidth]{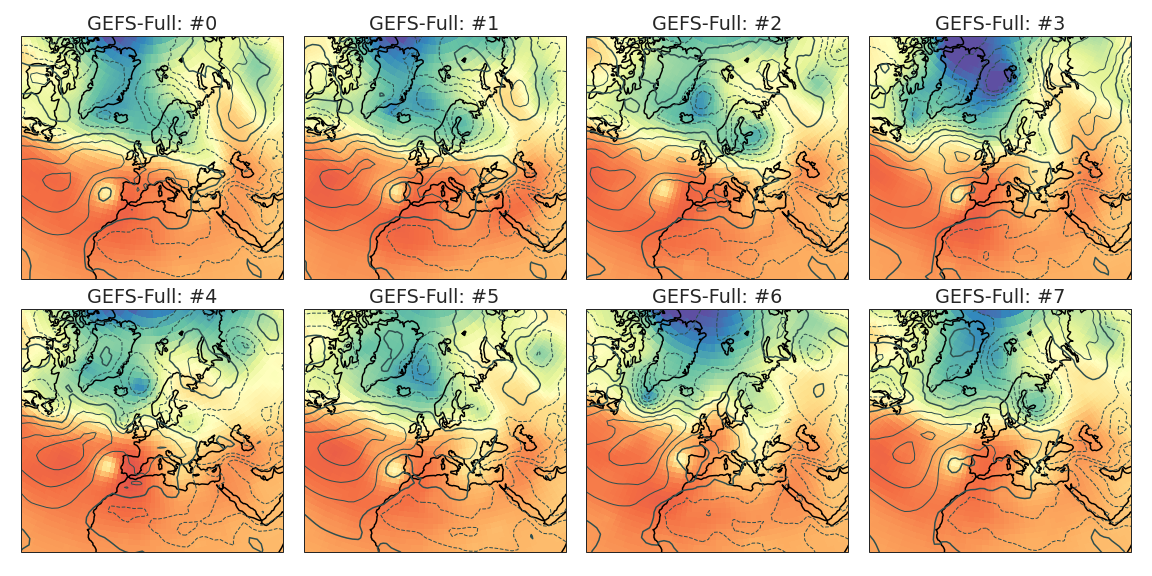}
    \includegraphics[width=0.85\columnwidth]{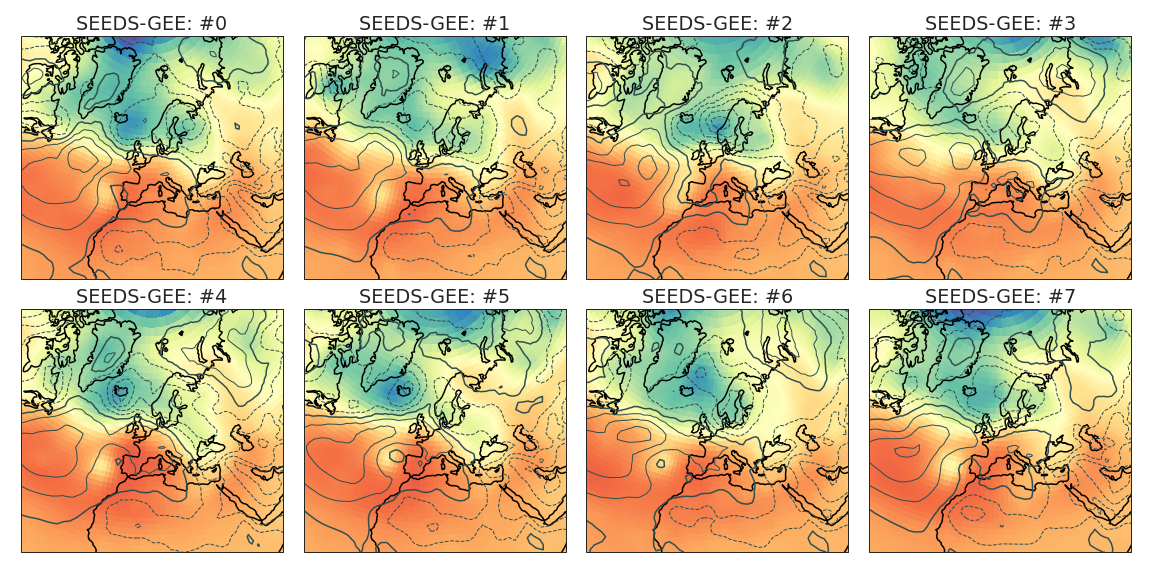}
    \includegraphics[width=0.85\columnwidth]{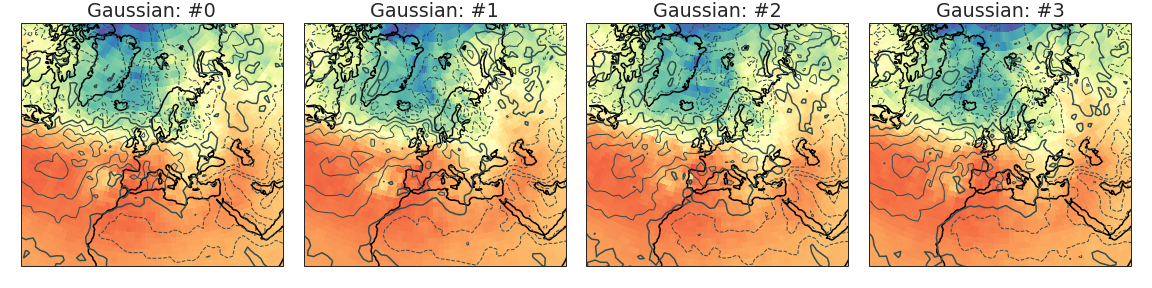}
    \includegraphics[width=0.6\columnwidth]{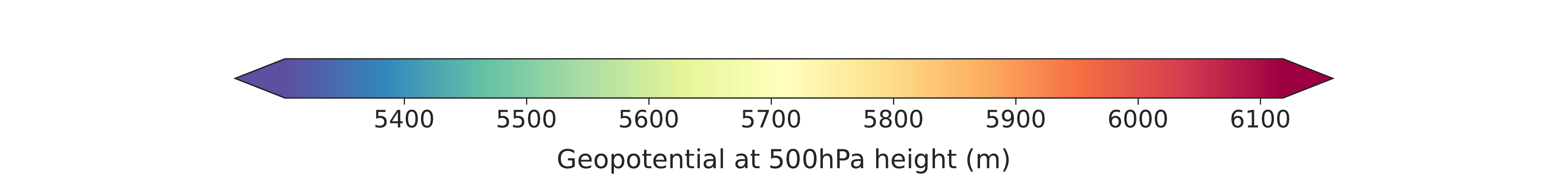}
    \caption{Visualization of spatial coherence in forecasted weather charts for 2022/07/14,  with a 7-day lead time. The contours are for mean sea level pressure (dashed lines mark isobars below 1010~hPa) while the heatmap depicts the geopotential height at the 500~hPa pressure level. Row 1: ERA5 reanalysis, then  2 forecast members from \gefsfull  used as seeds to our model.  Row 2--3: Other  forecast members from \gefsfull.    Row 4--5: 8 samples drawn from \ourgee. Row 6:   Samples from a pointwise Gaussian model parameterized by the \gefsfull ensemble mean and variance.}
  \label{fVisual}
\end{figure}

To investigate this aspect of weather forecasts, we compare the covariance structure of the generated samples to those from the ERA5 Reanalysis and GEFS through a stamp map over Europe for a date during the 2022 European heatwave in Figure~\ref{fVisual} \citep{witze2022extreme}. The global atmospheric context of a few of these samples is shown in Figure~\ref{fGlobalTCWV} for reference. We also present in Figure~\ref{fVisual} weather samples obtained from a Gaussian model that predicts the univariate mean and standard deviation of each atmospheric field at each location, such as the data-driven model proposed in \citep{Brecht2023}. This Gaussian model is meant to characterize the output of pointwise post-processing~\citep{Scher2018, sacco22MLens, Brecht2023}, which ignore correlations and treat each grid point as an independent random variable.

\ourgee captures well both the spatial covariance and the correlation between midtropospheric geopotential and mean sea level pressure, since it directly models the joint distribution of the atmospheric state. The generative samples display a geopotential trough west of Portugal with spatial structure similar to that found in samples from \gefsfull or the reanalysis. They also depict realistic correlations between geopotential and sea level pressure anomalies. Although the Gaussian model predicts the marginal univariate distributions adequately, it fails to capture cross-field or spatial correlations. This hinders the assessment of the effects that these anomalies may have on hot air intrusions from North Africa, which can exacerbate heatwaves over Europe~\citep{sanchezbenitez2018}.

\begin{figure}[tp]
    \centering
    \includegraphics[width=0.32\columnwidth,draft=false]{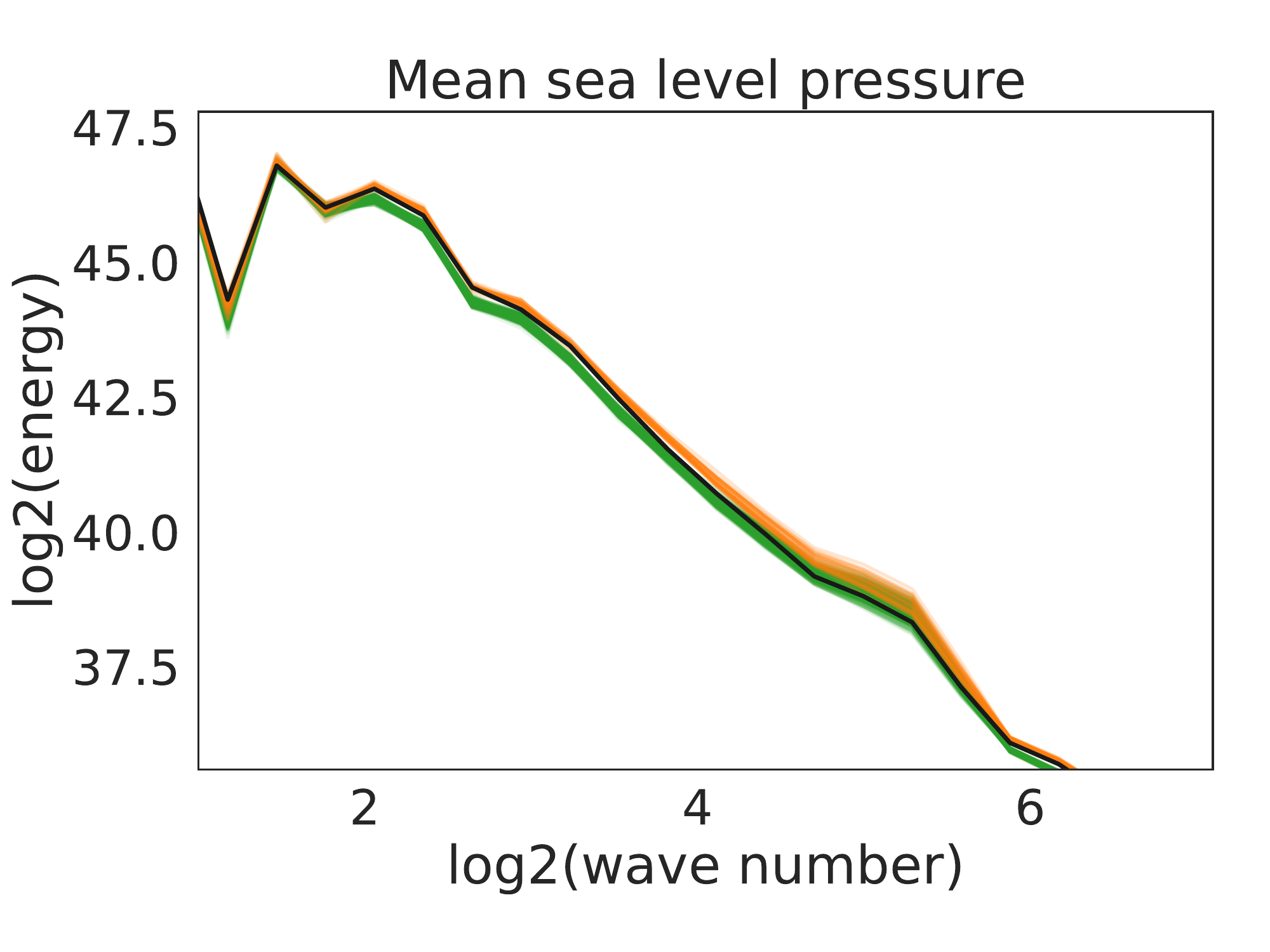}  
    \includegraphics[width=0.32\columnwidth,draft=false]{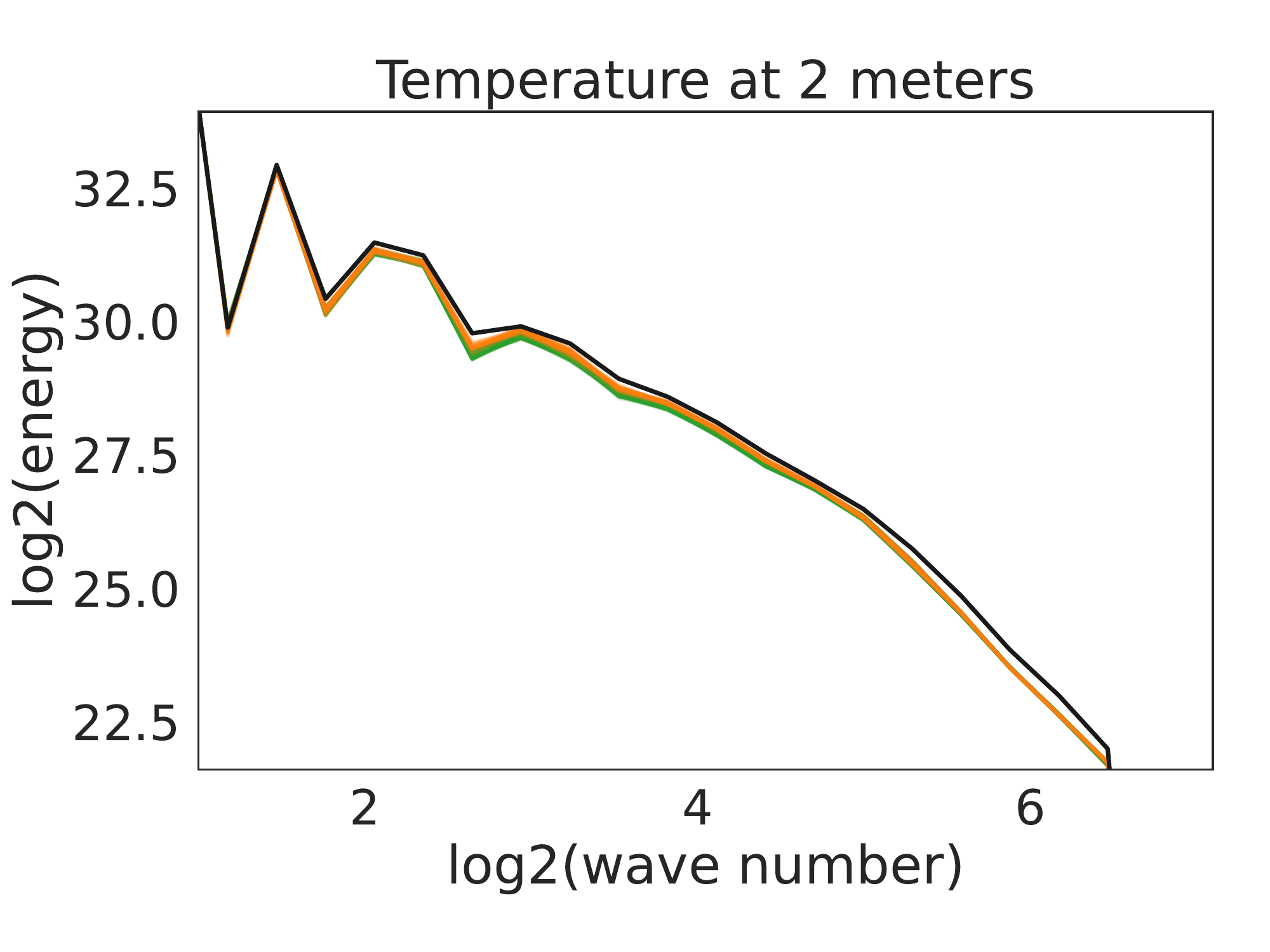}  
    \includegraphics[width=0.32\columnwidth,draft=false]{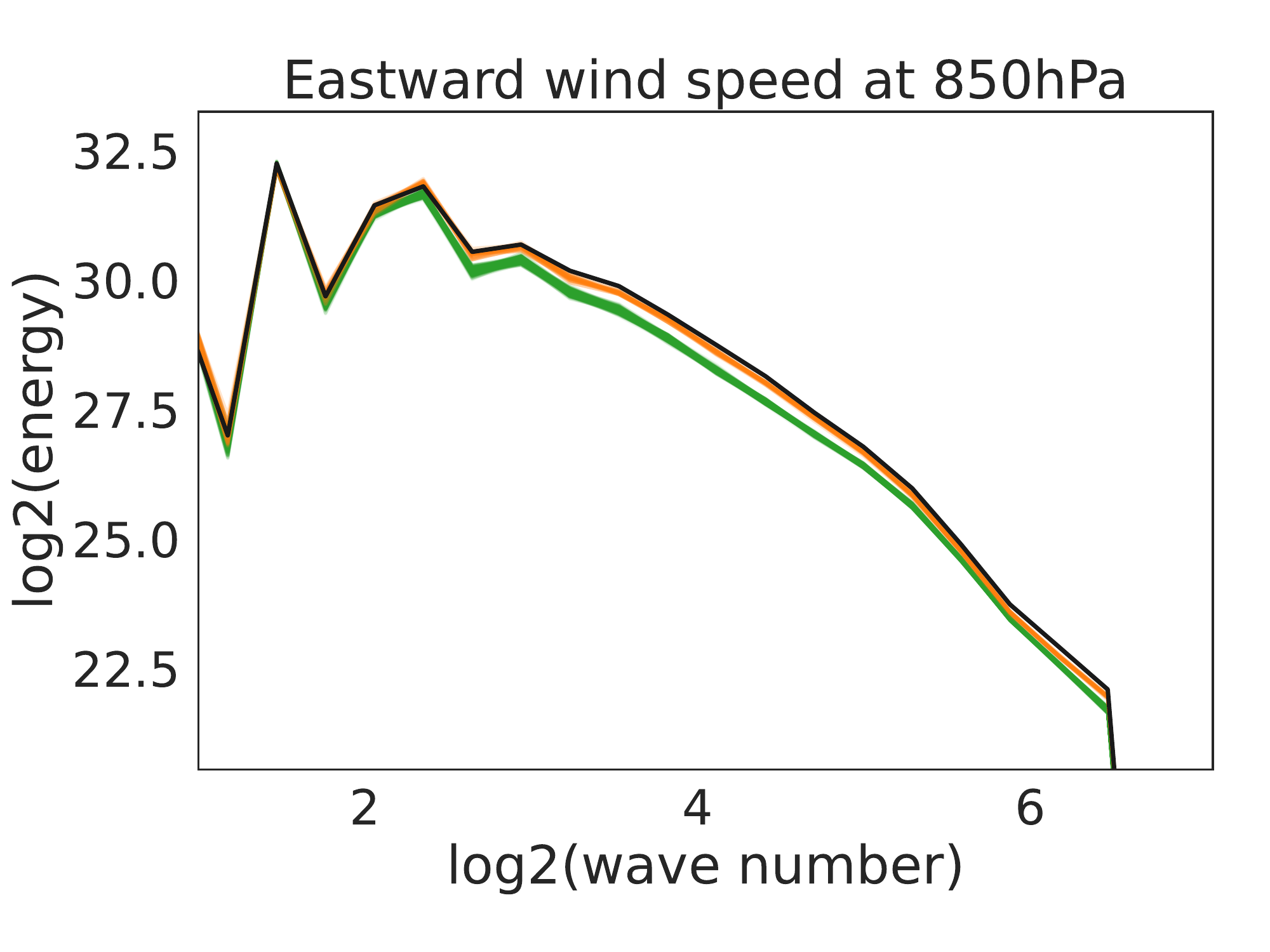}  
    \includegraphics[width=0.4\columnwidth,draft=false]{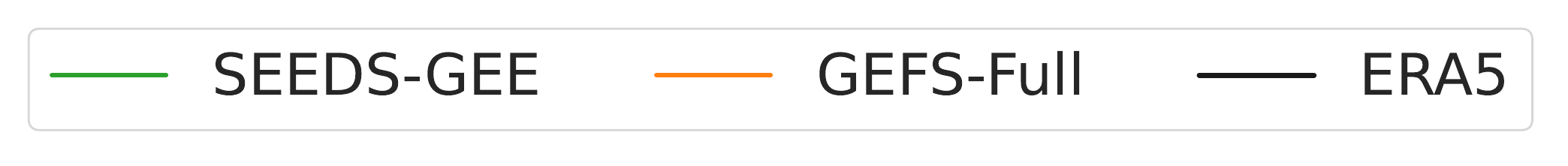}  
    \caption{The energy spectra of several global atmospheric variables for January of 2022 from the ERA5 reanalysis (thick black), members of the \gefsfull 7-day forecast (orange), and samples from \ourgee (green).
    The forecasts for each day are re-gridded to a latitude-longitude rectangular grid of the same angular resolution prior to computing the spectra. The computed spectra are averaged over the entire month. Each ensemble member is plotted separately.}
    \label{fSpectra}
\end{figure}

Figure~\ref{fSpectra} contrasts the energy spectra of \ourgee forecasts with that of ERA5 and \gefsfull. The large overlap between samples from both forecast systems and the reanalysis demonstrates that the two ensembles have similar spatial structure. Small systematic differences can be observed in some variables like the zonal wind in the low troposphere, but for most variables the differences between \ourgee and \gefsfull are similar to the differences between the operational system and the ERA5 reanalysis.

\begin{figure}[tp]
    \centering
    \includegraphics[width=0.32\columnwidth,draft=false]{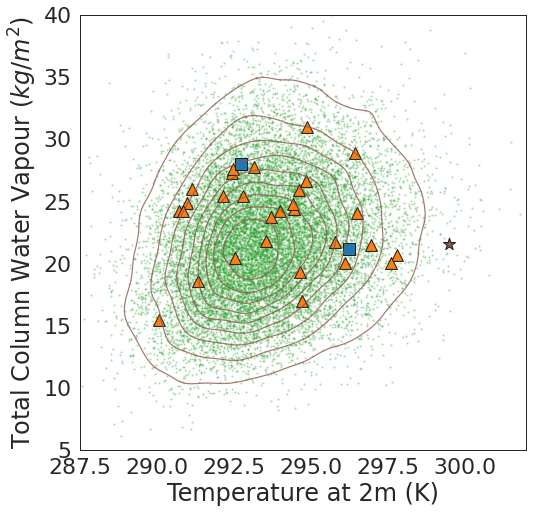}
    \includegraphics[width=0.32\columnwidth,draft=false]{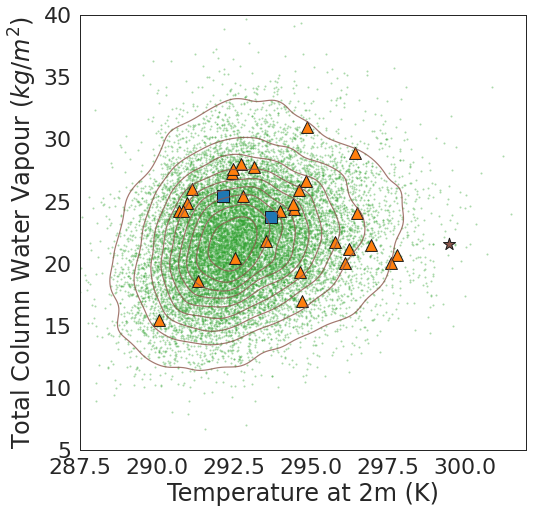}
    \includegraphics[width=0.32\columnwidth,draft=false]{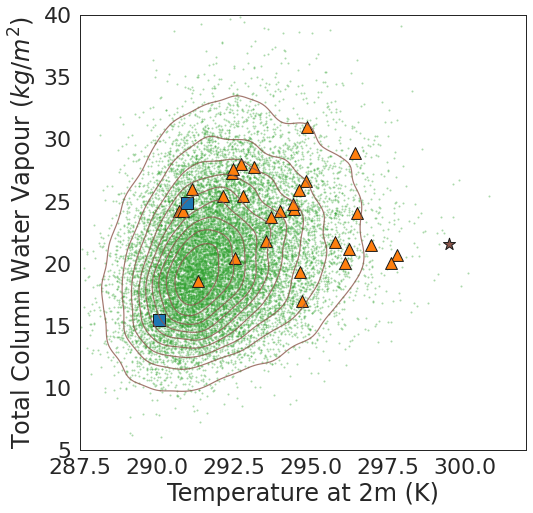}
    \includegraphics[width=0.6\columnwidth,draft=false]{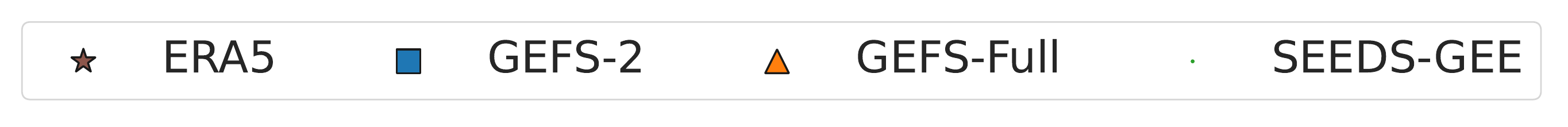}
    
    \includegraphics[width=0.32\columnwidth,draft=false]{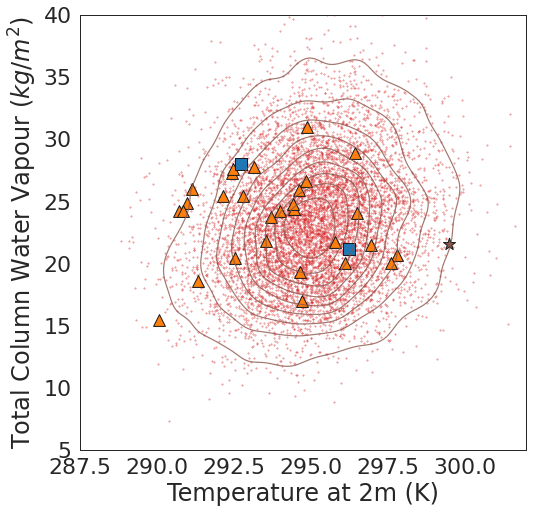}
    \includegraphics[width=0.32\columnwidth,draft=false]{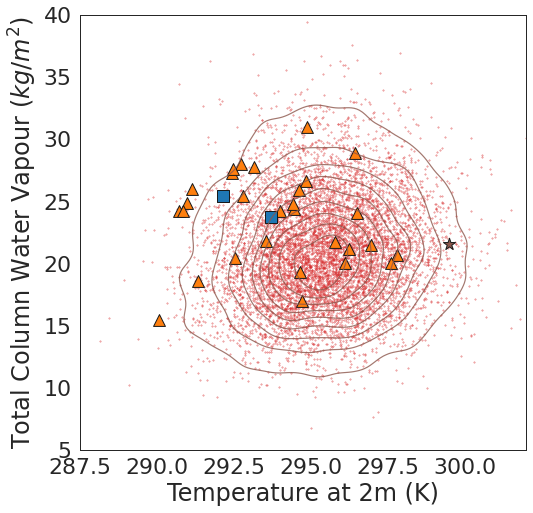}
    \includegraphics[width=0.32\columnwidth,draft=false]{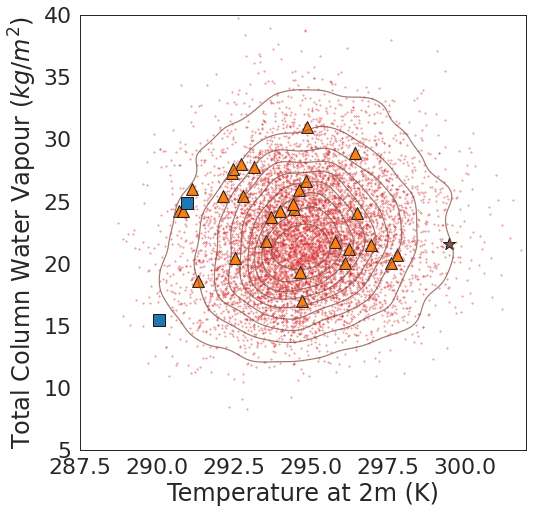}
    \includegraphics[width=0.6\columnwidth,draft=false]{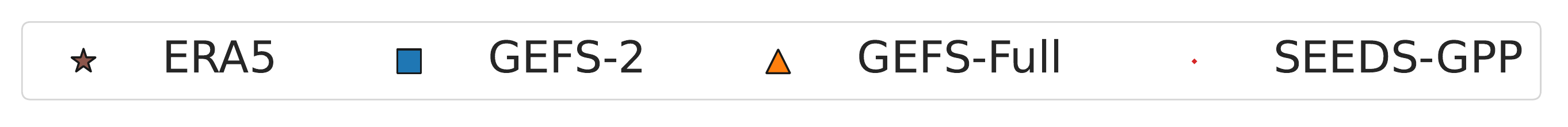}
    \caption{Generated ensembles provide better statistical coverage of the extreme heat event over Portugal. Each plot displays 16,384 generated forecasts from our method, extrapolating from the two seeding forecasts randomly taken from the operational forecasts. Contour curves of iso-probability are also shown. The first row is from \ourgee and the second from \ourgpp.  \ourgpp characterizes the event best. Most notably, in the two rightmost plots of the bottom row, \ourgpp is able to generate well-dispersed forecast envelopes that cover the extreme event, despite the two seeding ones deviating substantially from the observed event.}
    \label{fPortugal}
\end{figure}

In addition to examining the coherence of regional structures and the global spectra of the generative samples, we also examine the multivariate correlation structure of generative samples locally. 
Figure~\ref{fPortugal} depicts the joint distributions of temperature at 2 meters and total column water vapour at the grid point near Lisbon during the extreme heat event on 2022/07/14. We used the 7-day forecasts made on 2022/07/07. For each plot, we generate  16,384-member ensembles. The observed weather event from ERA5 is denoted by the star. The operational ensemble, denoted by the squares (also used as the seeding forecasts) and triangles (the rest of the GEFS ensemble), fails to predict the intensity of the extreme temperature event. This highlights that the observed event was so unlikely 7 days prior that none of the 31 forecast members from \gefsfull attained near-surface temperatures as warm as those observed. In contrast, the generated ensembles are able to extrapolate from the two seeding forecasts, providing an envelope of possible weather states with much better coverage of the event. This allows quantifying the probability of the event taking place (see Figure~\ref{fLogLoss2sigma} and~\ref{sAppendixResults}). Specifically, our highly scalable generative approach enables the creation of very large ensembles that can capture the likelihood of very rare events that would be characterized with a null probability by limited-size ensembles. Moreover, we observe that the distributions of the generated ensembles do not depend critically on the (positioning of the) seeding forecasts. This suggests that the generative approach is plausibly learning the intrinsic dynamical structure, i.e., the attractor of the atmosphere, in order to expand the envelopes of the phase of the dynamical systems to include extreme events that deviate strongly from the seeds.

\subsection{Forecast Reliability and Predictive Skills}
\label{sExtreme}

\begin{figure}[tp]
    \centering
    \includegraphics[width=0.32\columnwidth,draft=false]{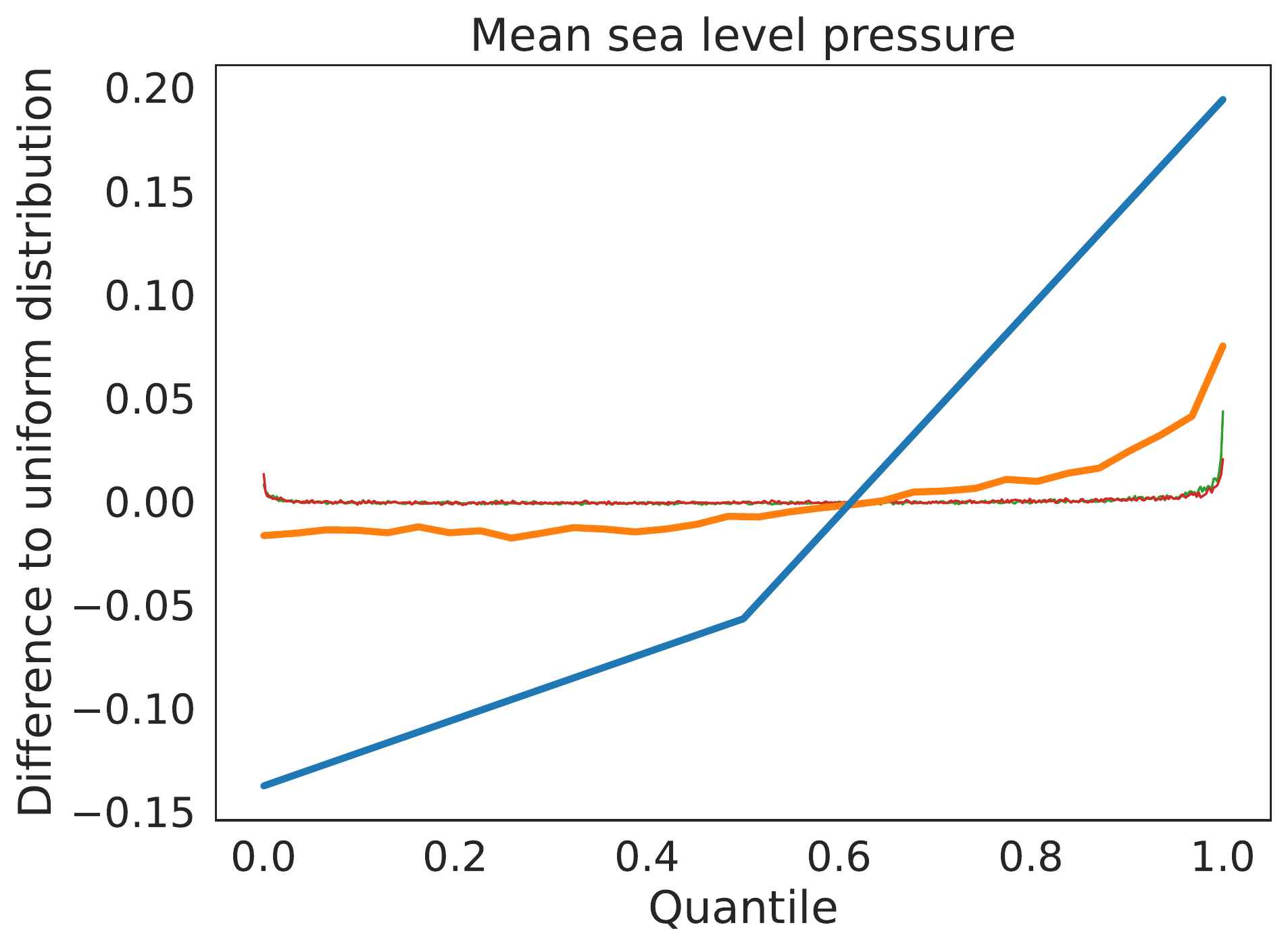}
    \includegraphics[width=0.32\columnwidth,draft=false]{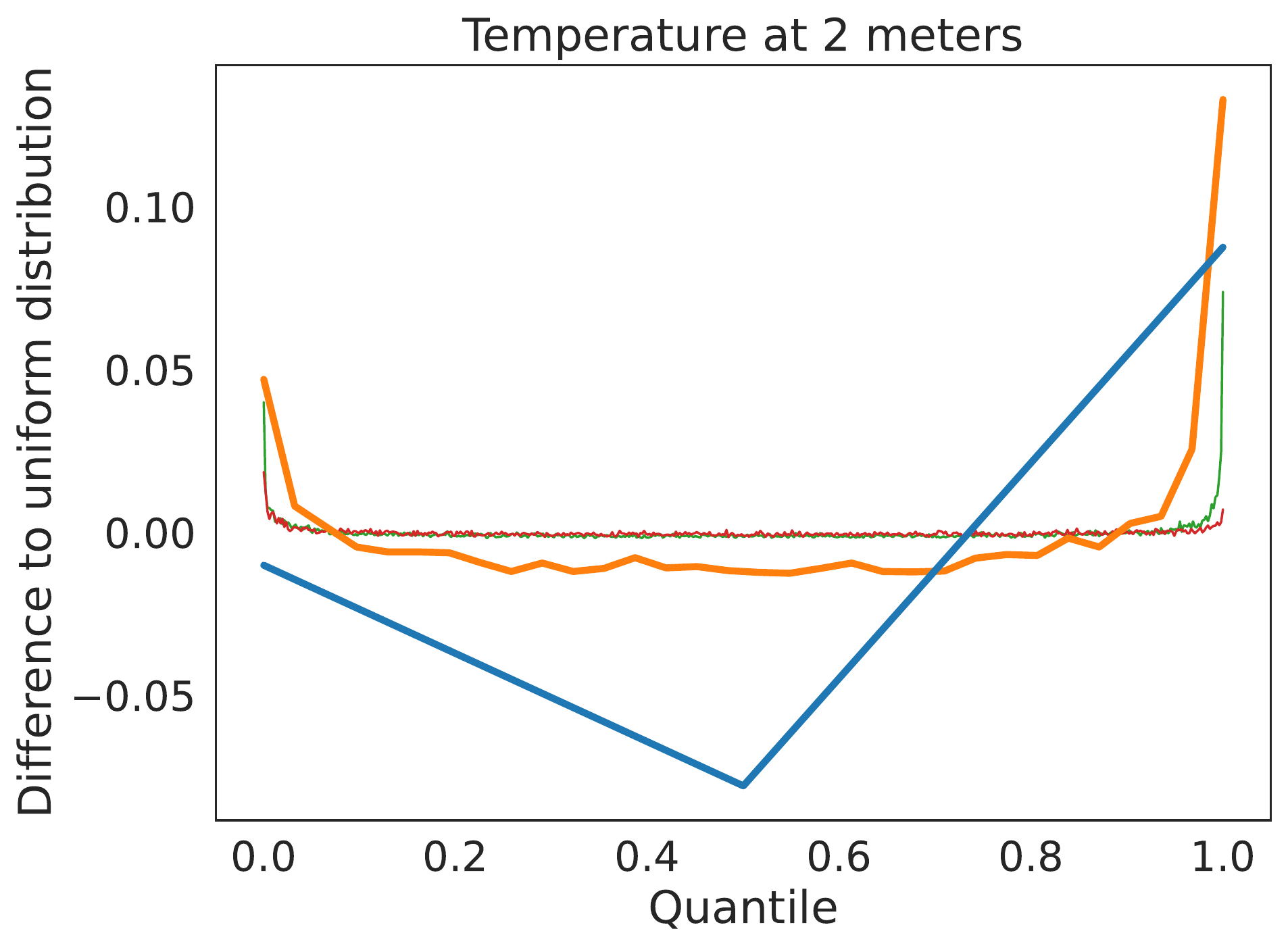}
    \includegraphics[width=0.32\columnwidth,draft=false]{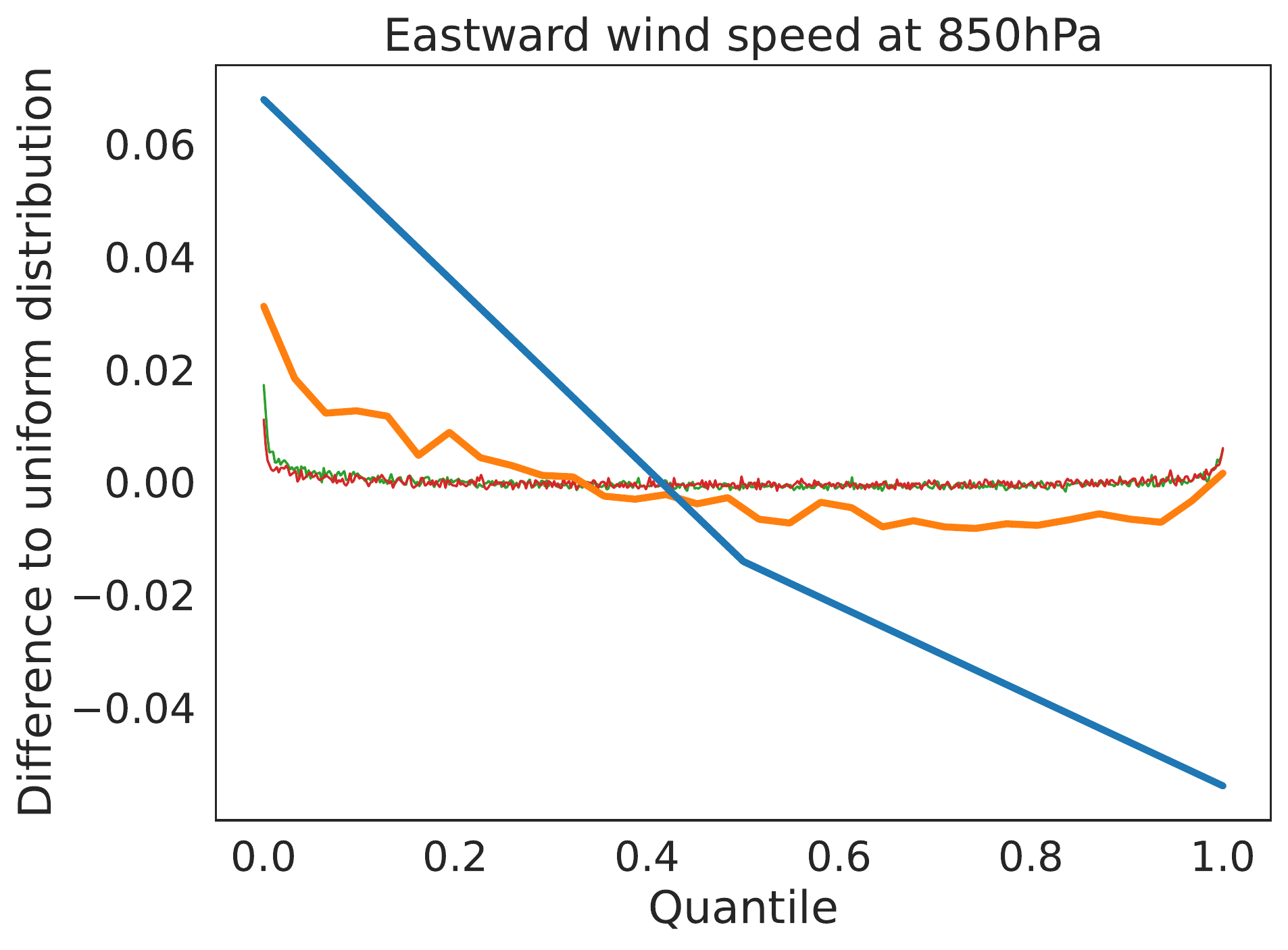}
    \includegraphics[width=0.5\columnwidth,draft=false]{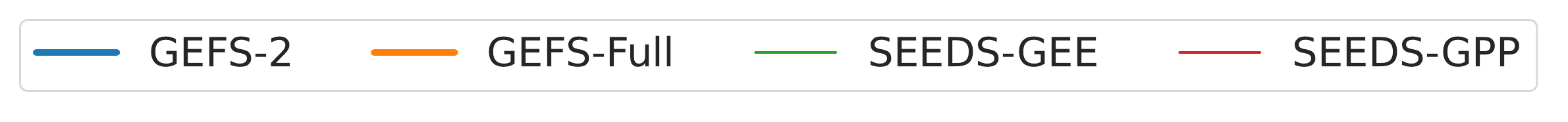}
    \\
    \includegraphics[width=0.32\columnwidth,draft=false]{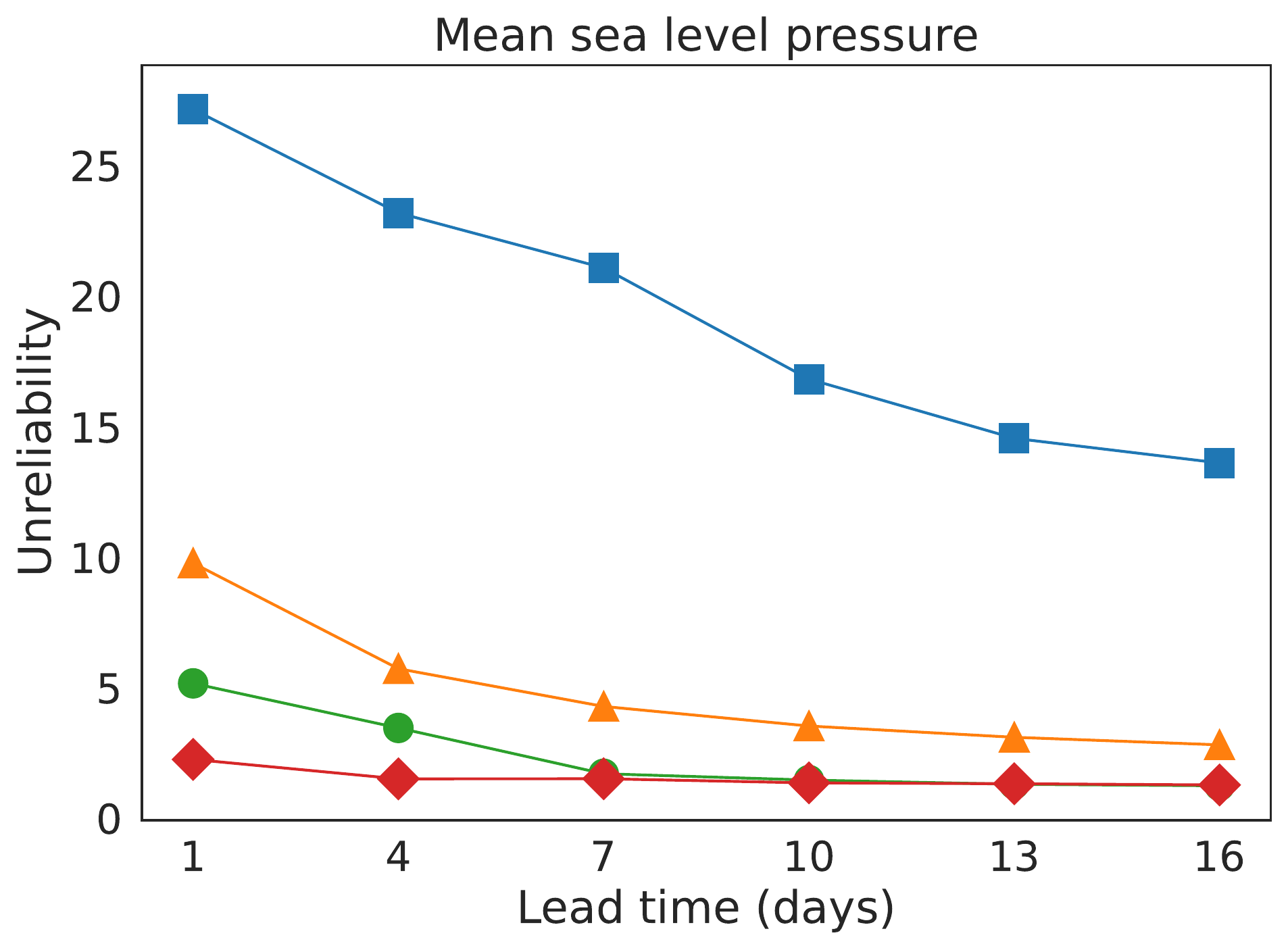}
    \includegraphics[width=0.32\columnwidth,draft=false]{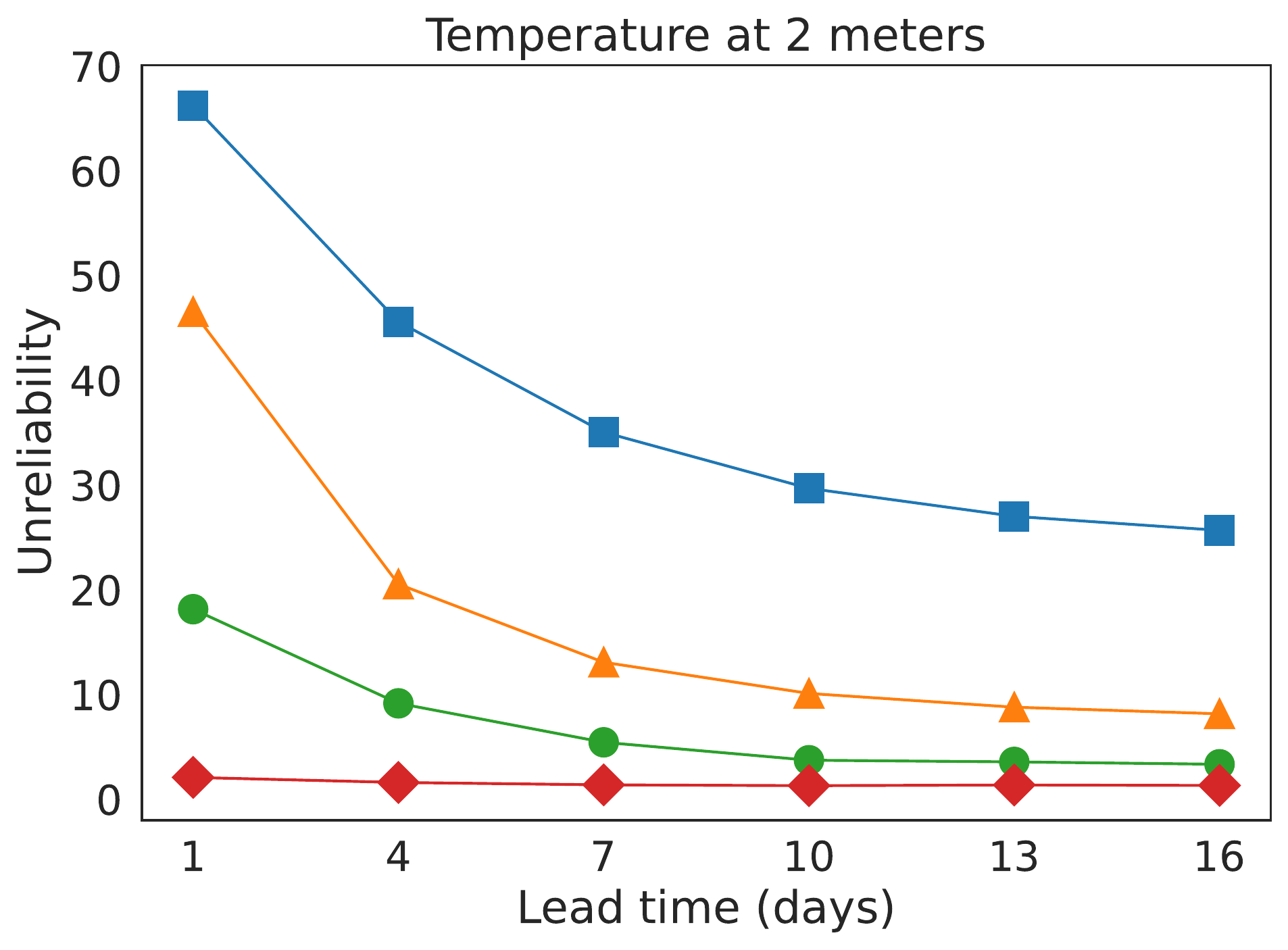}
    \includegraphics[width=0.32\columnwidth,draft=false]{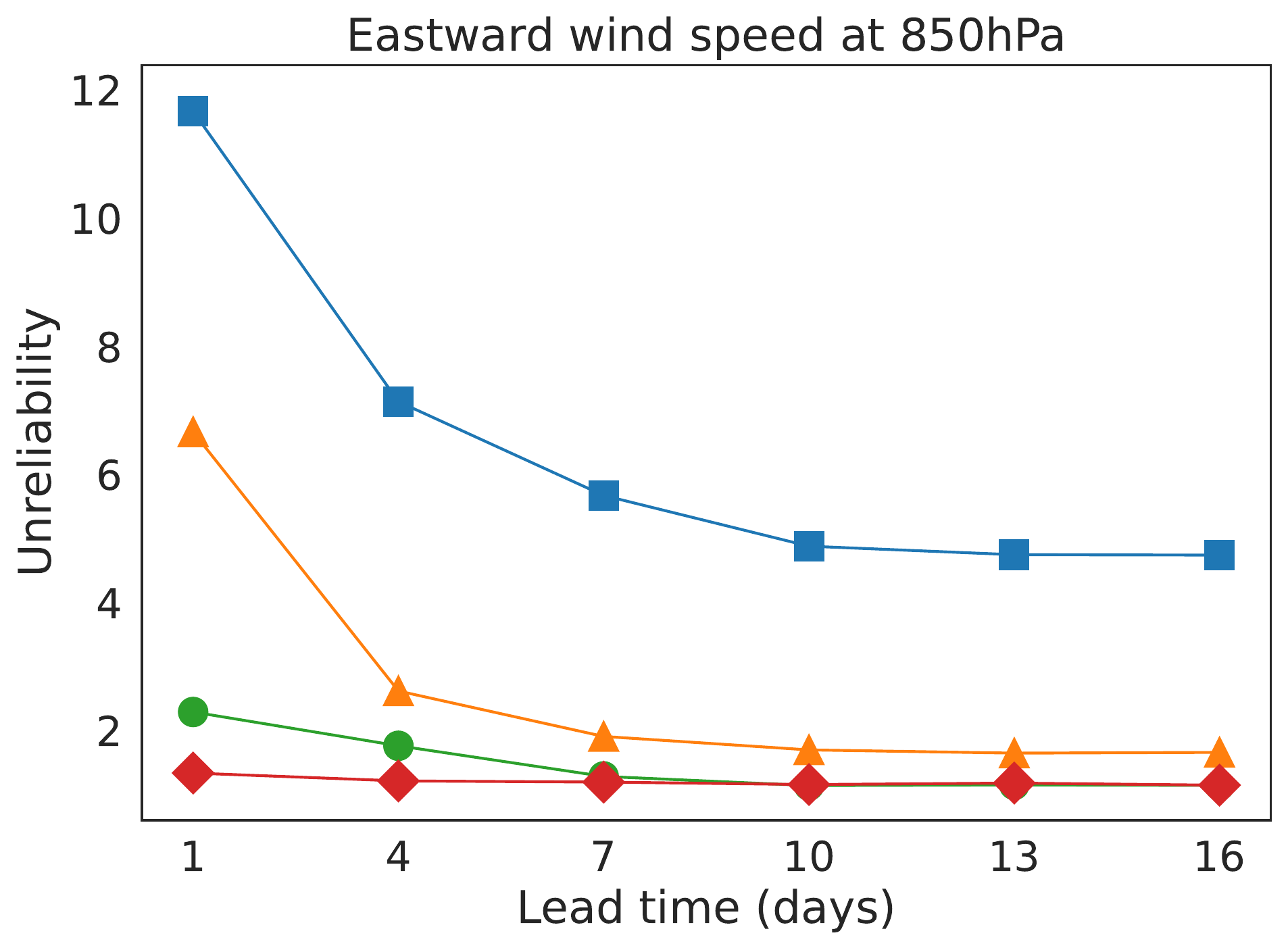}
    \includegraphics[width=0.5\columnwidth,draft=false]{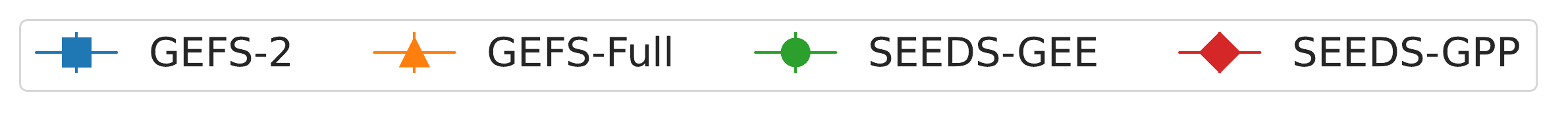}  
\caption{Top: Rank histograms from 7-day forecasts for grid points in the region bounded by parallels 34N and 42N, and meridians 124W and 114W, for the year 2022. This region roughly encompasses California and Nevada, USA. To compare the histograms of ensembles of different size, the $x$ axis is normalized to quantiles instead of ranks, and the $y$ axis shows the difference to the uniform distribution. A perfectly calibrated ensemble forecast should have a flat curve at $0$. Bottom: Unreliability parameter $\delta$~\citep{Candille2005} as a function of lead time, computed for the same year and averaged globally.}
\label{fReliability}
\end{figure}

An important characteristic of ensemble forecast systems is their ability to adequately capture the full distribution of plausible weather states. This characteristic is known as forecast calibration or reliability~\citep{Wilks2019}. Forecast reliability can be characterized for a given lead time in terms of the rank histogram \citep{Anderson1996, talagrand1997}. Deviations from flatness of this histogram indicate systematic differences between the ensemble forecast distribution and the true weather distribution. Rank histograms for 7-day forecasts from \gefsfull, \gefsK, \ourgee, and \ourgpp over California and Nevada are shown in Figure~\ref{fReliability}. The GEFS ensembles display systematic negative biases in mean sea level pressure and near-surface temperature over the region, as well as an underestimation of near-surface temperature uncertainty. Our model ensembles are more reliable than \gefsK and \gefsfull, due in part to the larger number of members that can be effortlessly generated. \ourgpp shows the highest reliability of all, validating generative post-processing as a useful debiasing methodology. In particular, Figure~\ref{fReliability} shows how \ourgpp substantially reduces the ensemble under-dispersion for 2-meter temperature forecasts.

The reliability information contained in the rank histogram can be further summarized in terms of its bulk deviation from flatness, which we measure using the unreliability metric $\delta$ introduced by Candille and Talagrand~\citep{Candille2005}. Higher values of $\delta$ indicate higher deviations from flatness, or a lower reliability of the forecasts. Figure~\ref{fReliability} confirms that the generated ensembles are on a global average more reliable than \gefsfull for all lead times. In addition, the refined calibration of \ourgpp is more noticeable in the first forecast week.

The predictive skill of the generated ensembles is measured in terms of the root-mean-squared-error (\rmse) and the anomaly correlation coefficient (\acc) of the ensemble mean, as well as the continuous ranked probability score (\crps), treating the ERA5 HRES reanalsyis as the reference ground-truth.
These metrics are computed and averaged over the grid points every forecast day in the test set and then aggregate over the test days. ~\ref{sAppendixResults} details how these metrics are defined.

\begin{figure}[t]
    \centering
    \includegraphics[width=0.32\columnwidth,draft=false]{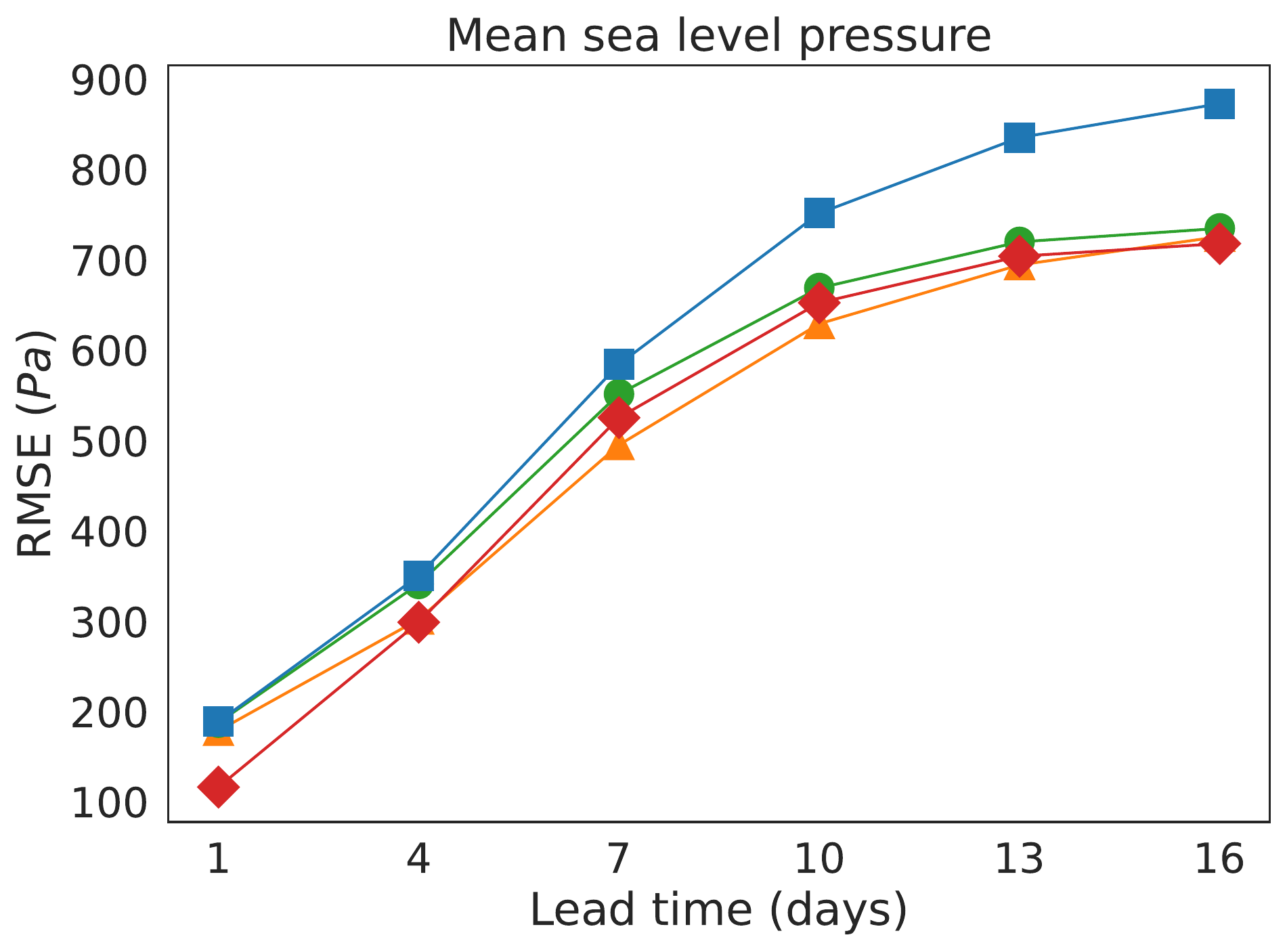}
    \includegraphics[width=0.32\columnwidth,draft=false]{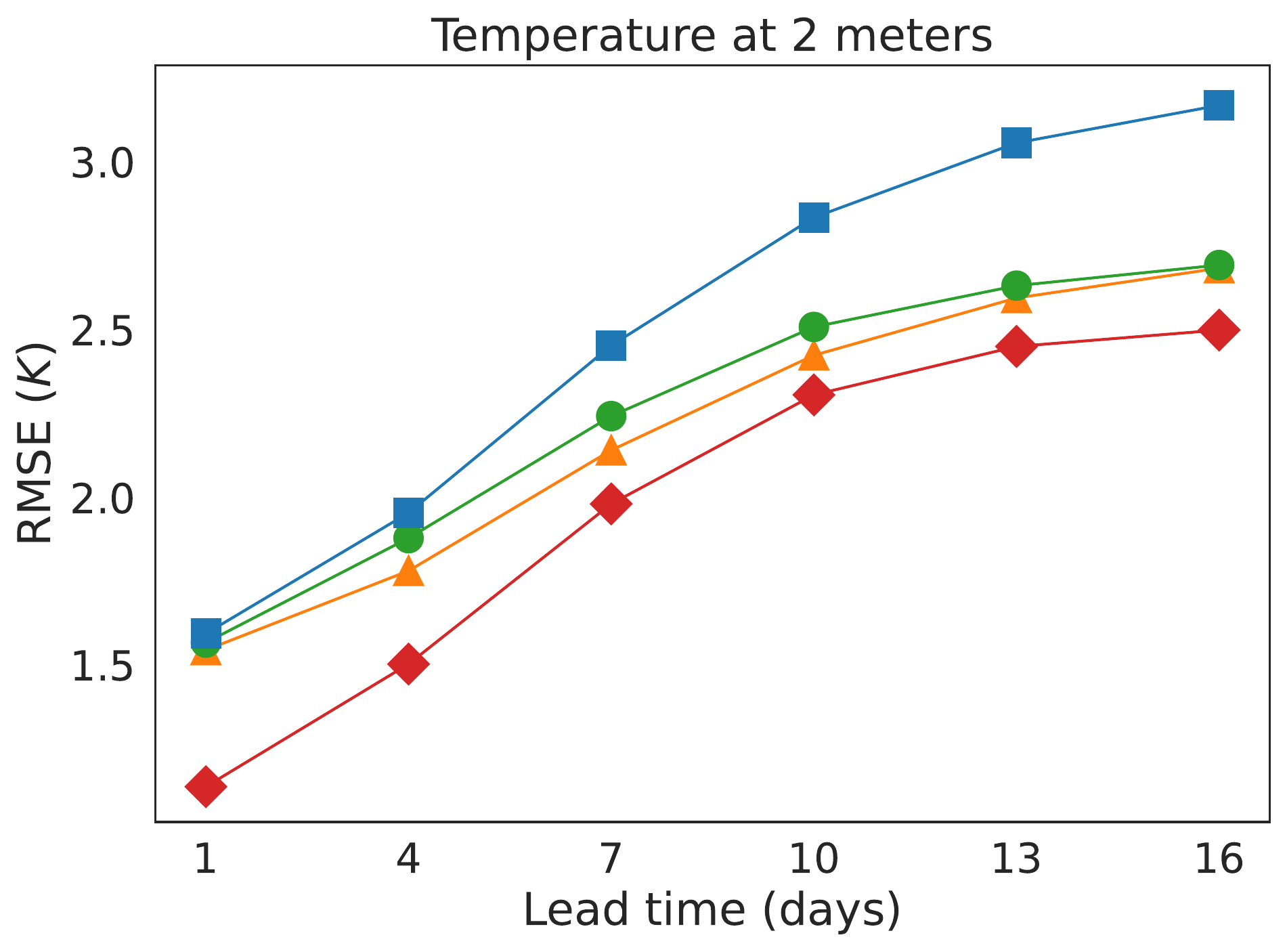}
    \includegraphics[width=0.32\columnwidth,draft=false]{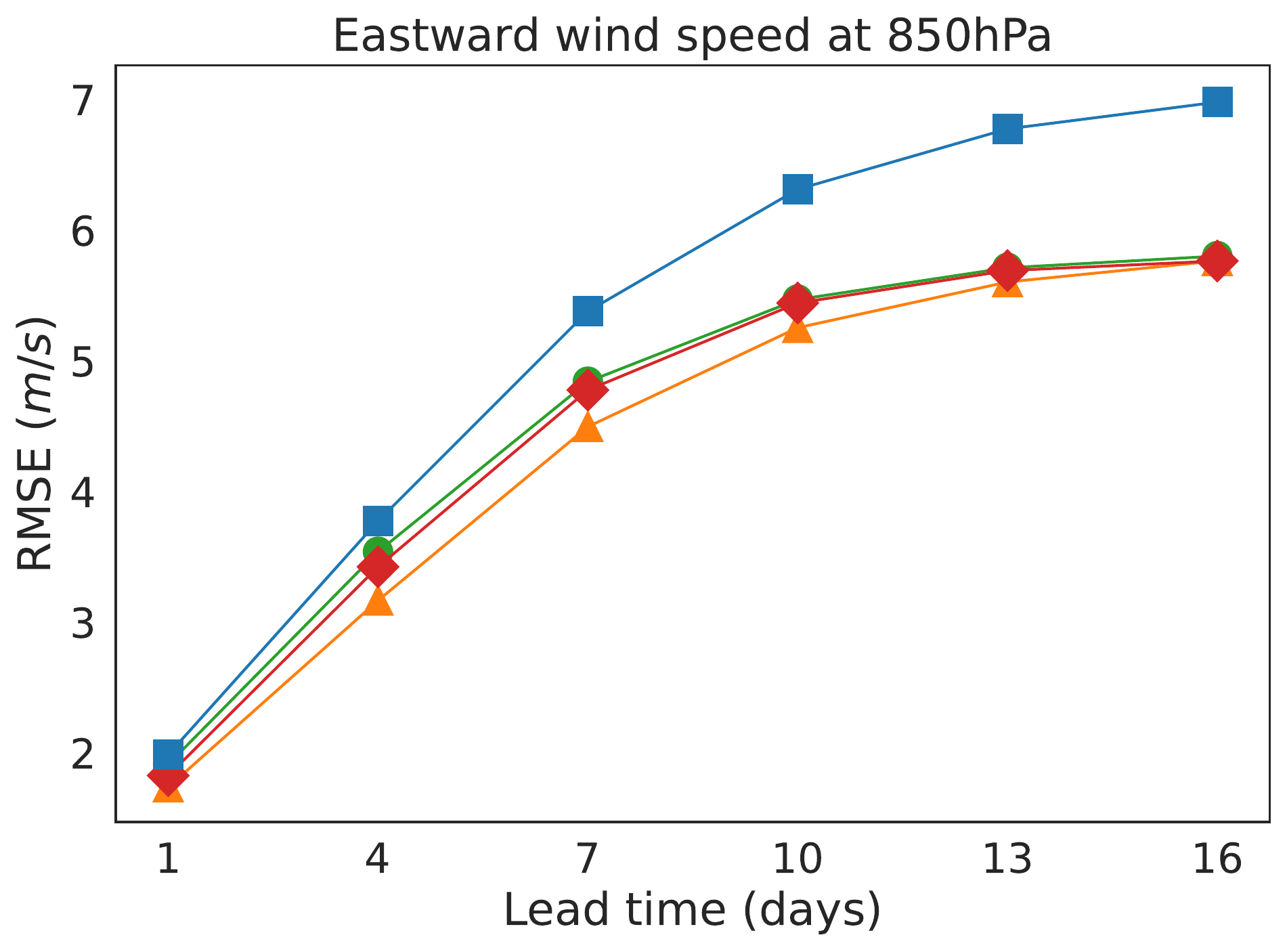}  
    \includegraphics[width=0.32\columnwidth,draft=false]{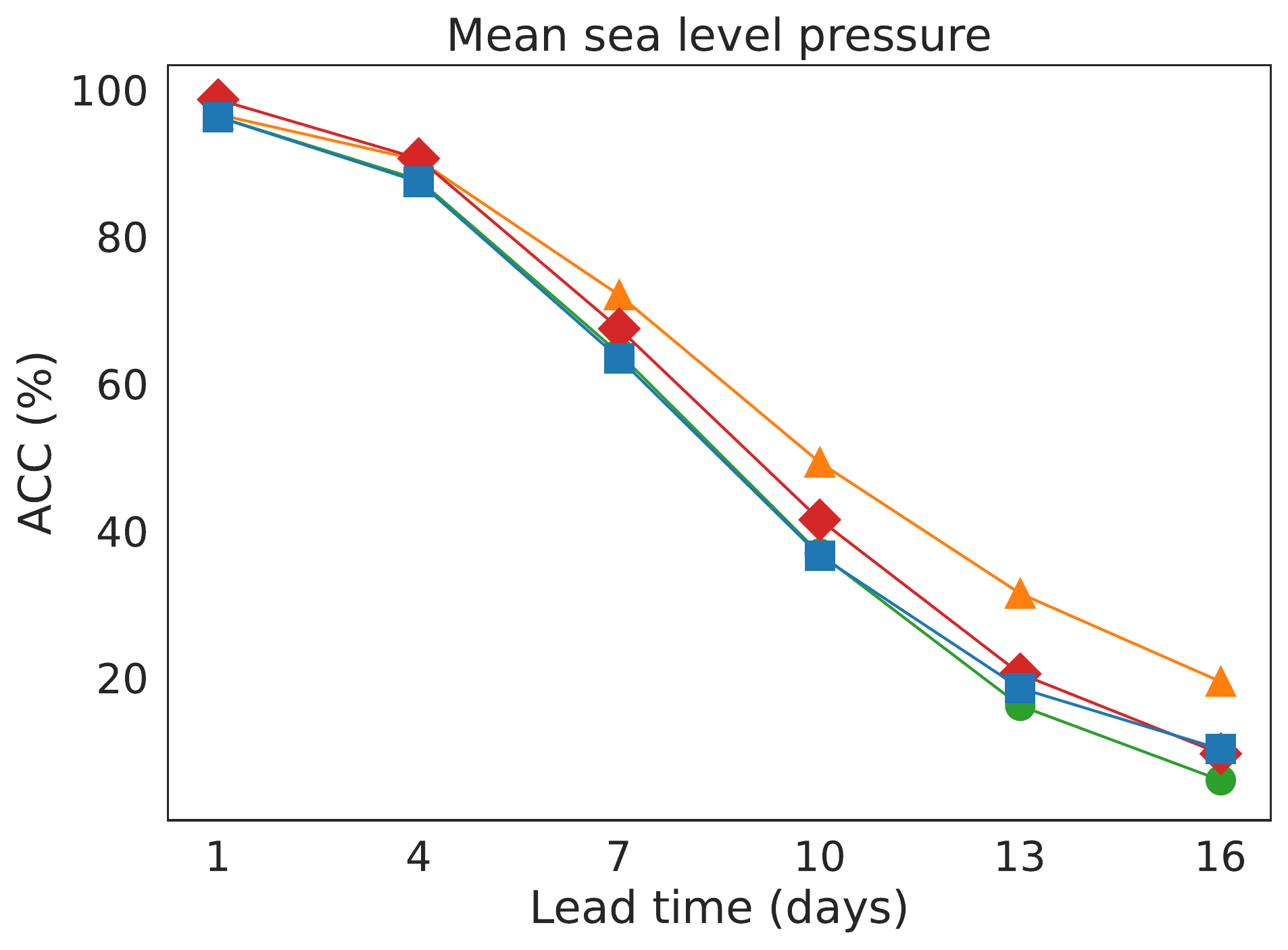} 
    \includegraphics[width=0.32\columnwidth,draft=false]{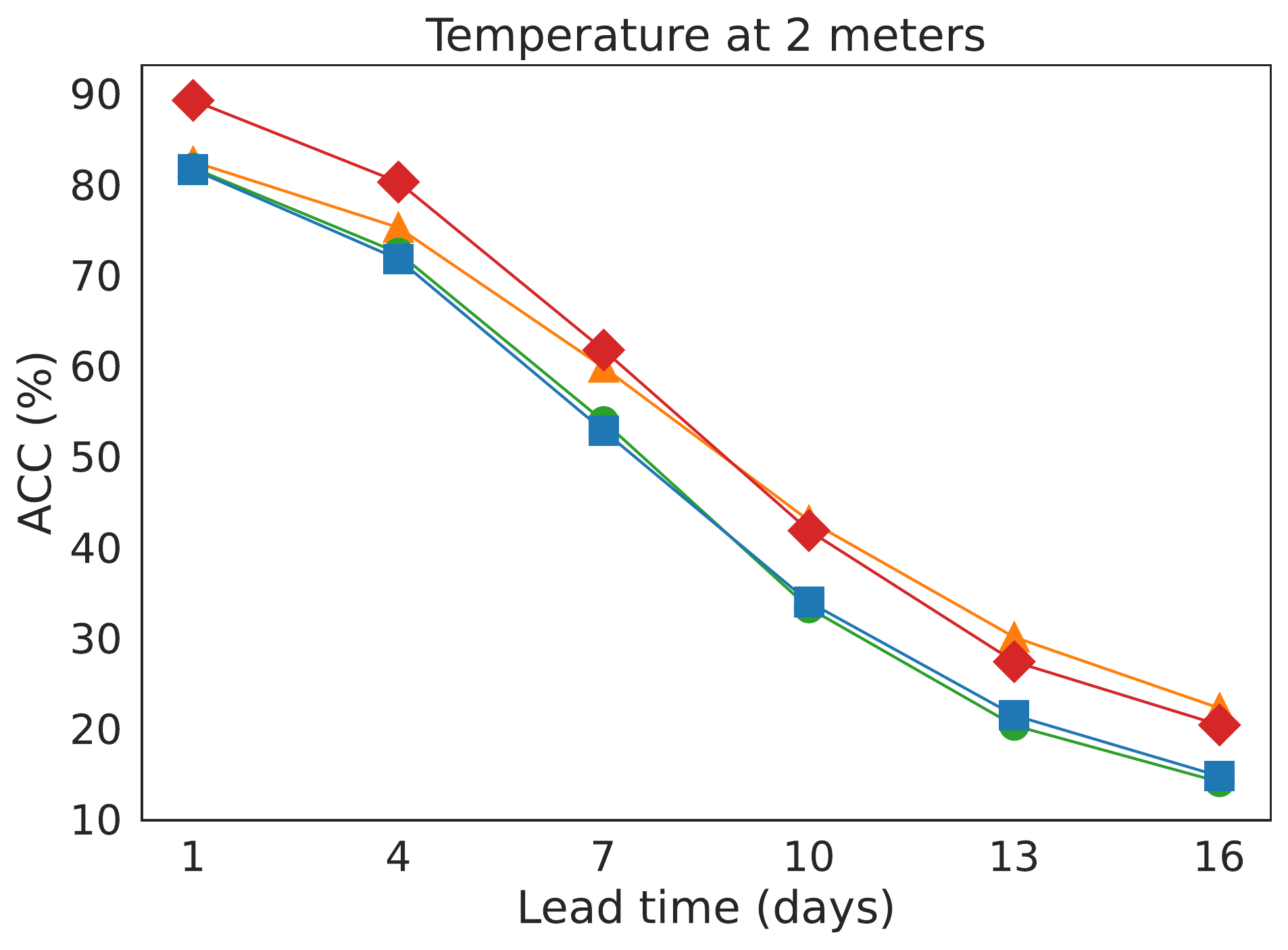} 
    \includegraphics[width=0.32\columnwidth,draft=false]{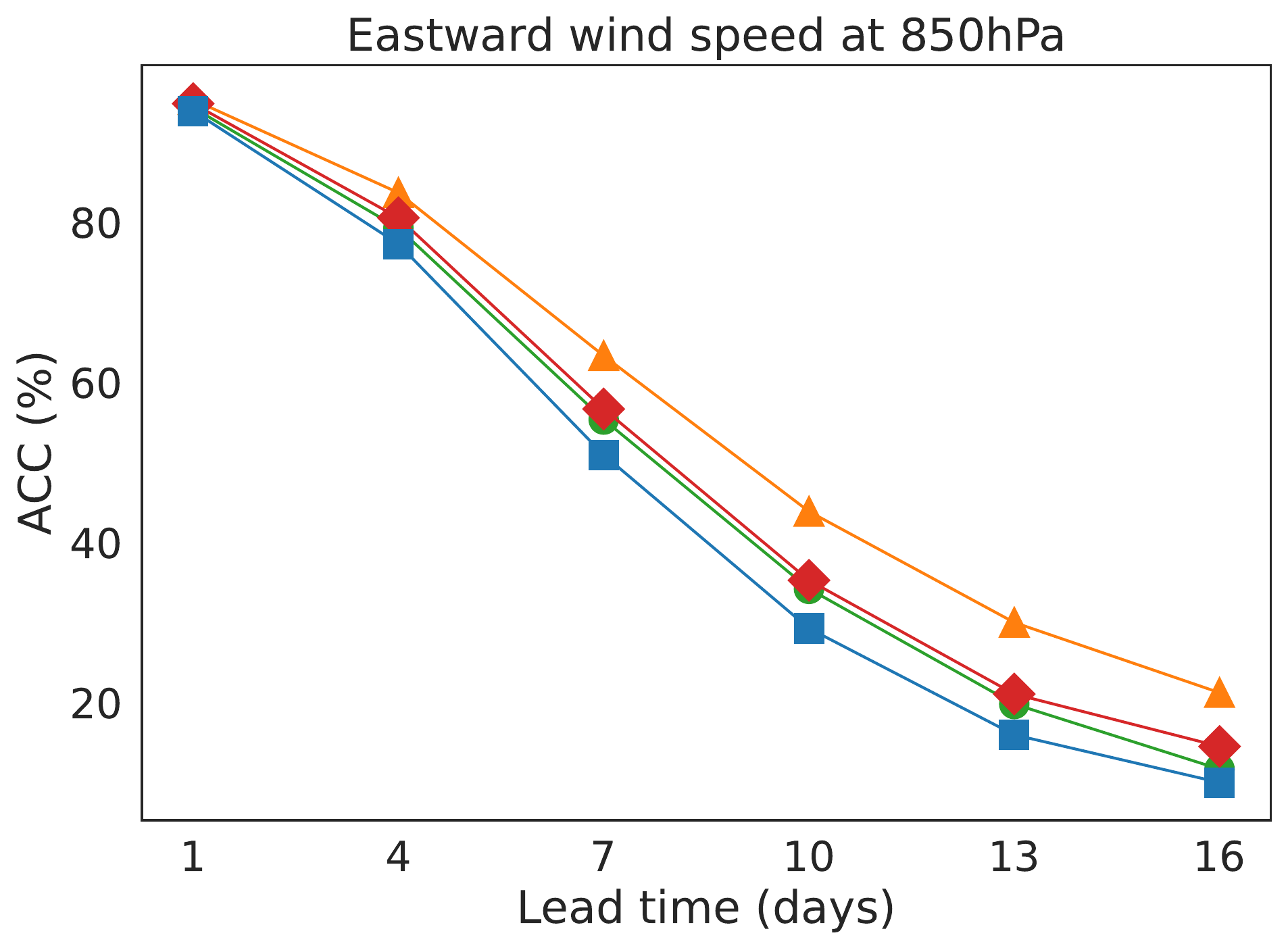}  
    \includegraphics[width=0.32\columnwidth,draft=false]{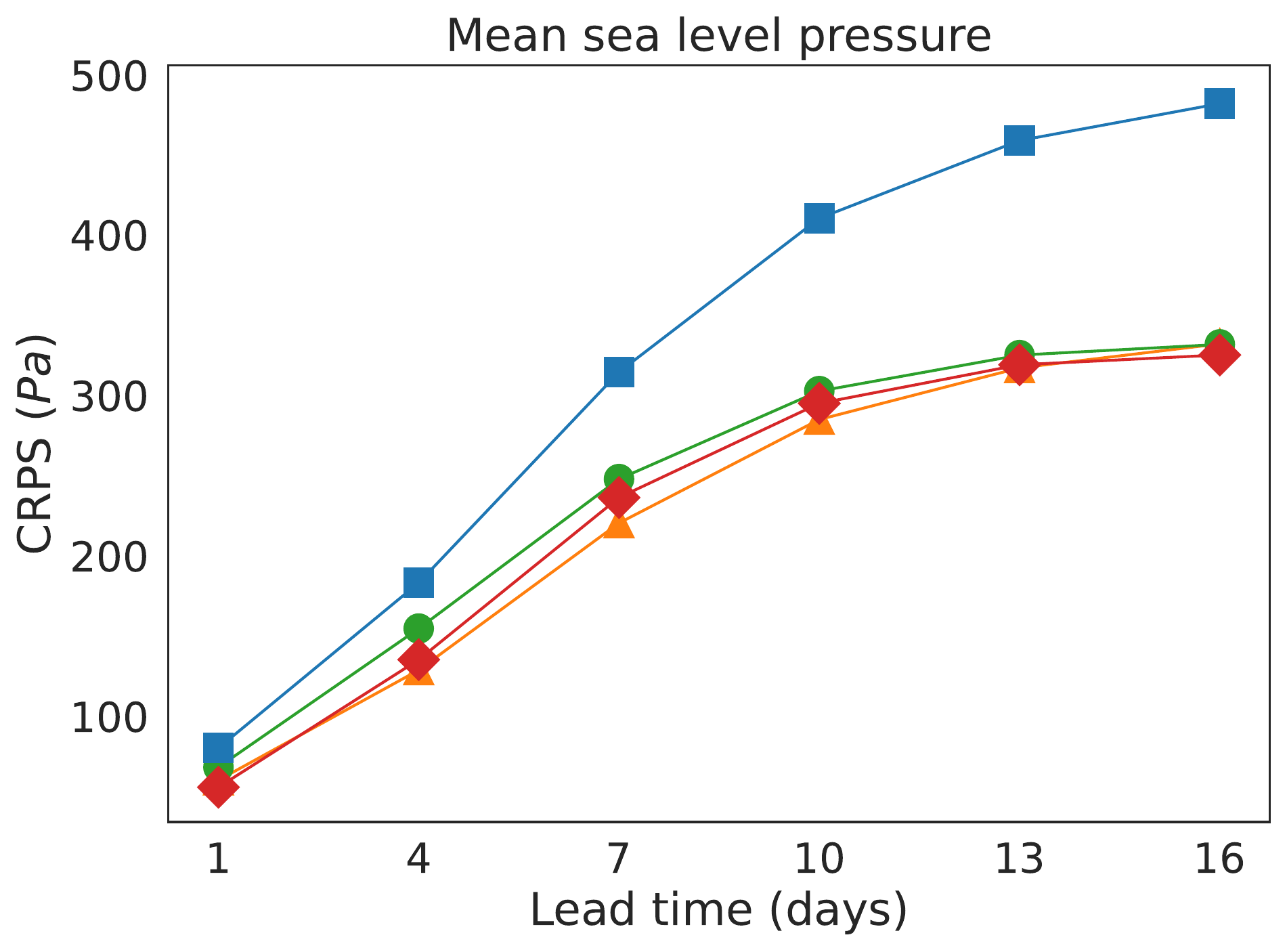} 
    \includegraphics[width=0.32\columnwidth,draft=false]{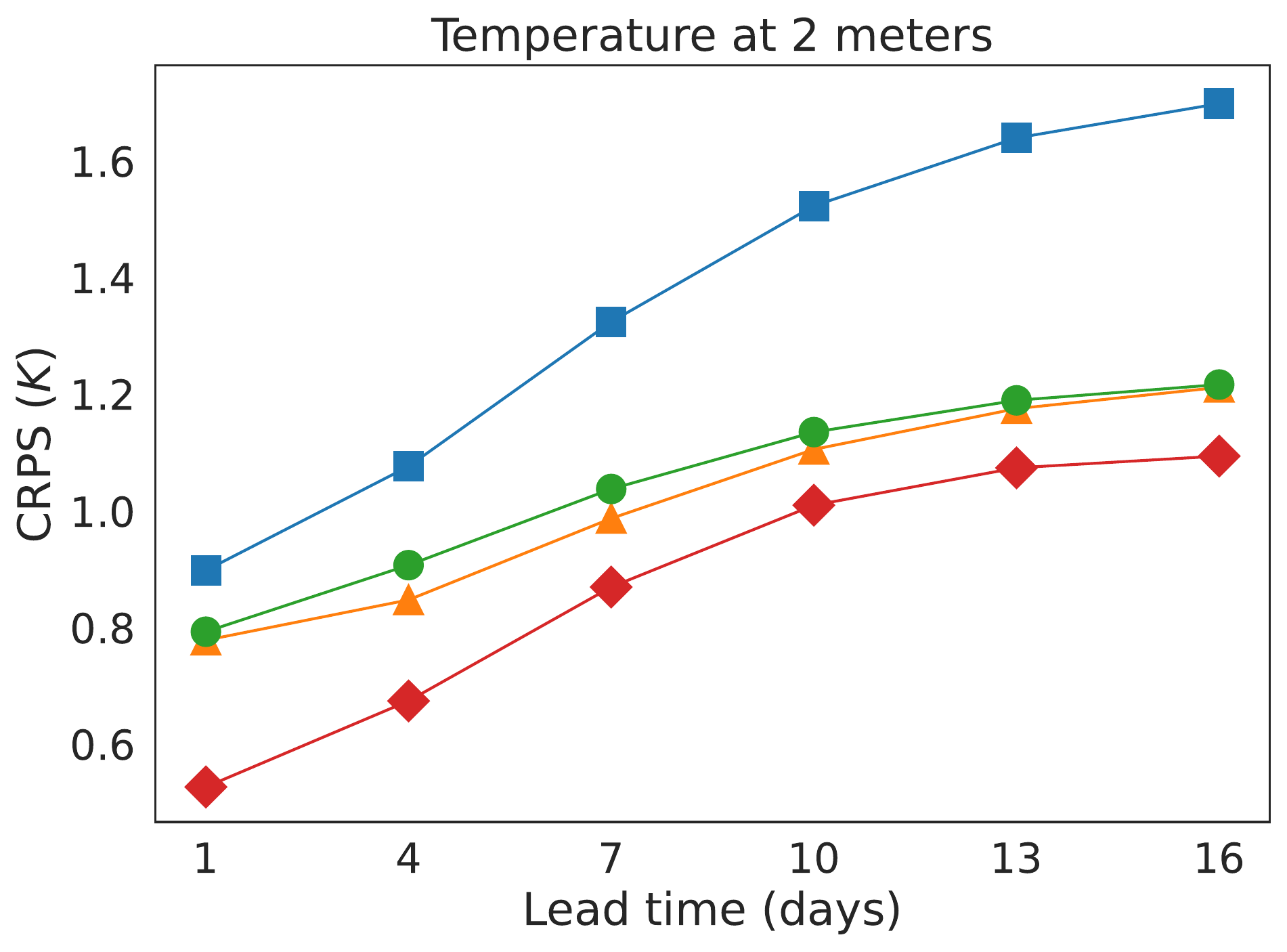}  
    \includegraphics[width=0.32\columnwidth,draft=false]{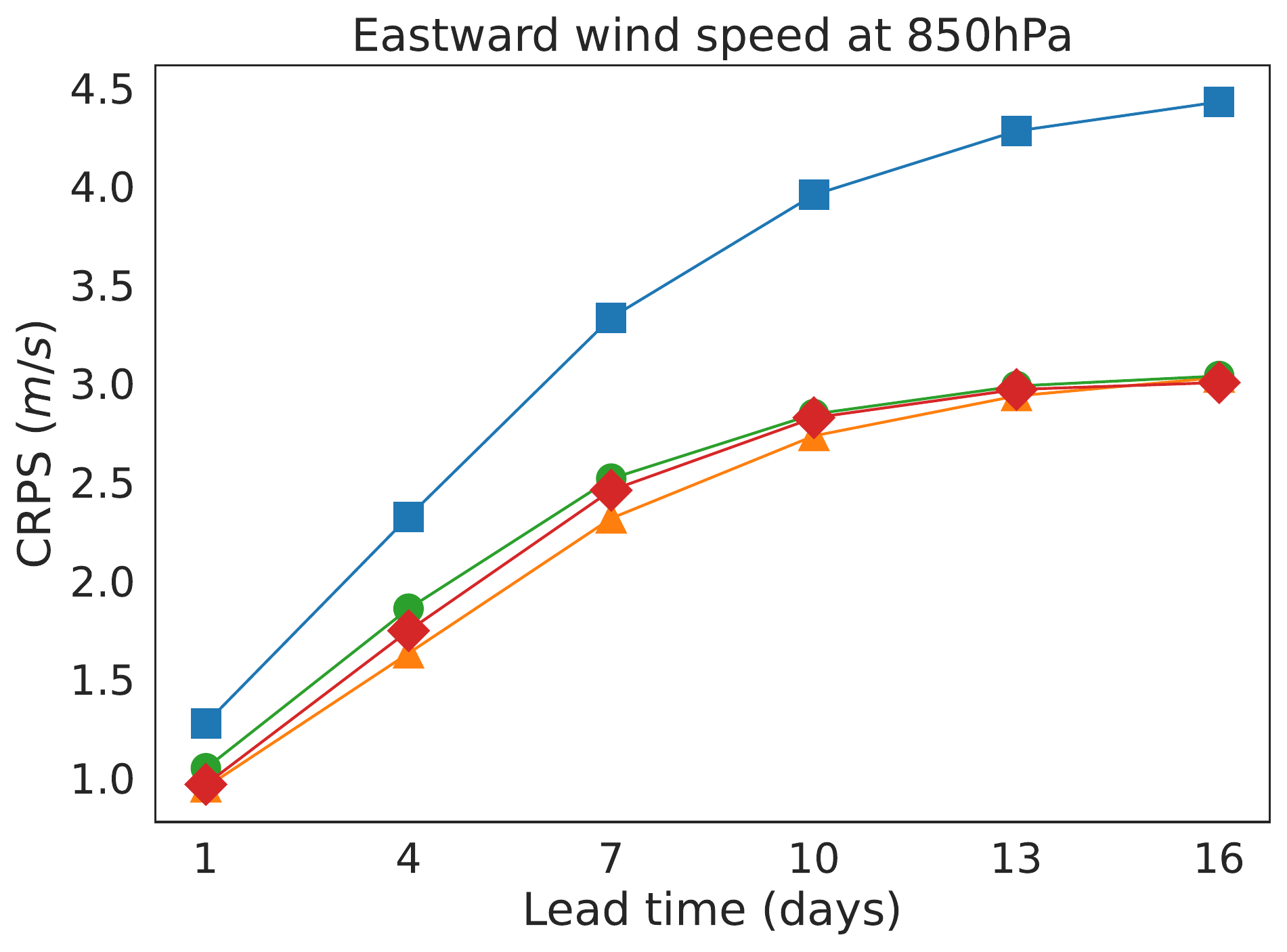}  
    \includegraphics[width=0.5\columnwidth,draft=false]{figs/gee_gpp_c2/legend}  
   \caption{Metrics of point-wise skill (\rmse, \acc and \crps) of the generative and physics-based ensemble forecasts, measured against the ERA5 HRES reanalysis as ground-truth. Shown are results for mean sea level pressure (left), $2$-meter temperature (center), and zonal velocity at $850$~hPa (right). A detailed description of these metrics is included in ~\ref{sAppendixResults}.}
\label{fEnsembleAccuracy}
\end{figure}

Figure~\ref{fEnsembleAccuracy} reports these metrics for 3 atmospheric fields: the mean sea level pressure, the temperature 2 meters above the ground, and the eastward wind speed at 850hPa. Both \ourgee or \ourgpp perform significantly better than the seeding \gefsK ensemble across all metrics. The emulator \ourgee shows similar but slightly lower skill than \gefsfull across all metrics and variables. Our generative post-processing \ourgpp is noticeably better than the physics-based \gefsfull at predicting near-surface temperature, roughly matching its skill for the other two fields. Intuitively, the potential benefits of statistical blending with a corrective data source are determined by the variable-dependent biases of the emulated forecast model. In this case, the GEFS model is known to have a cold bias near the surface \citep{Zhou2022-nv}.

\begin{figure}[t]
    \centering
    \includegraphics[width=0.32\columnwidth,draft=false]{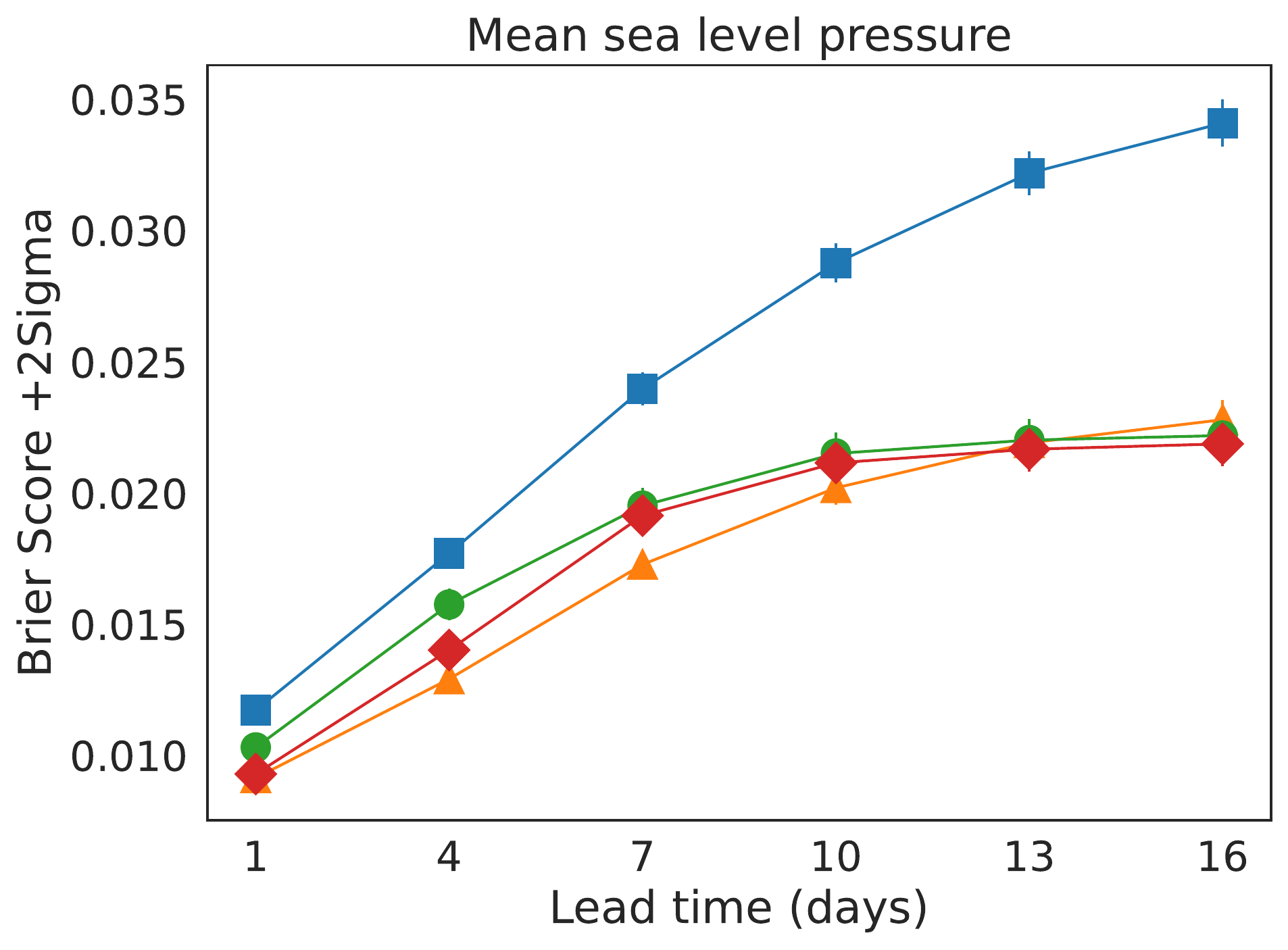} 
    \includegraphics[width=0.32\columnwidth,draft=false]{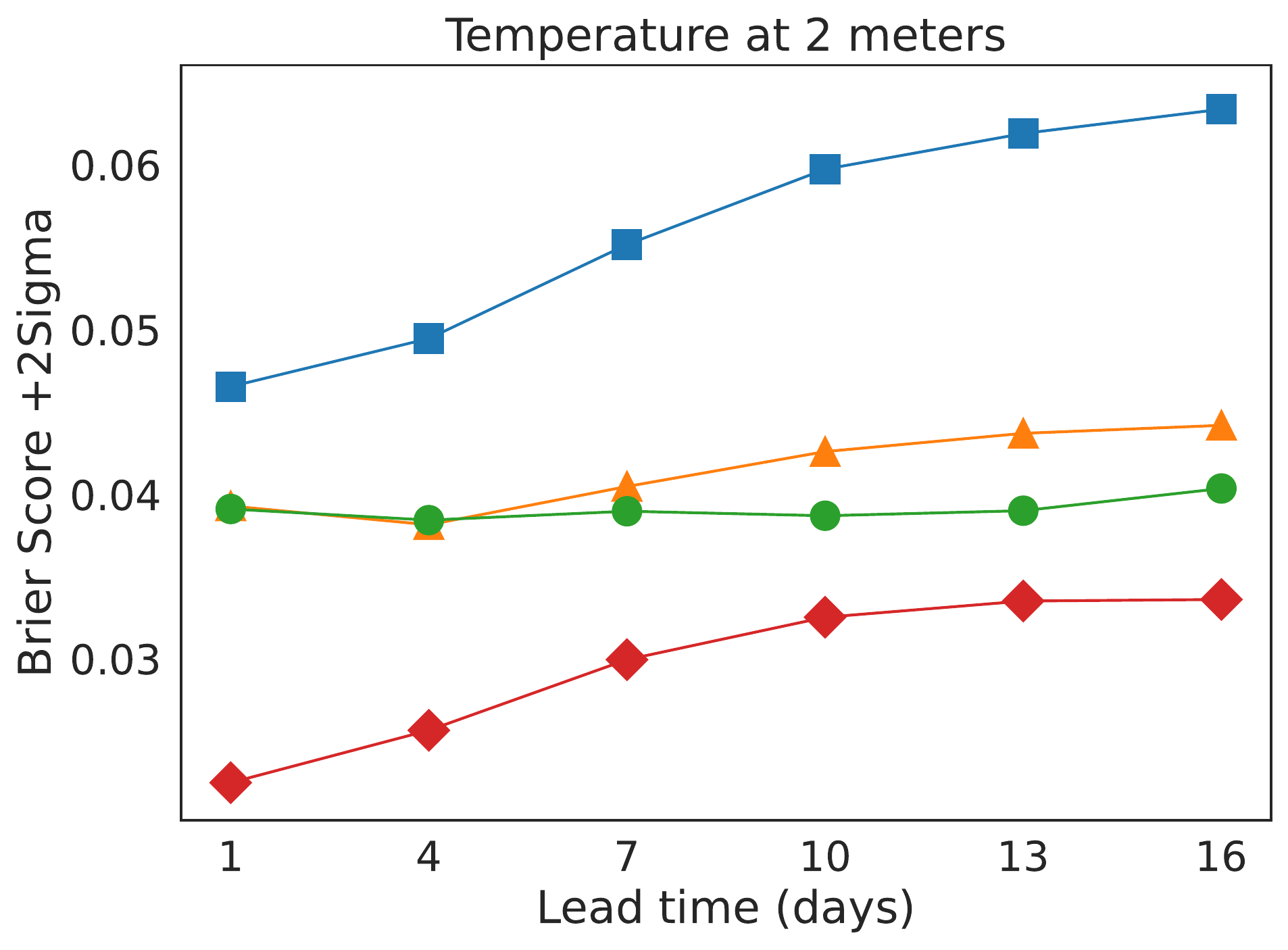} 
    \includegraphics[width=0.32\columnwidth,draft=false]{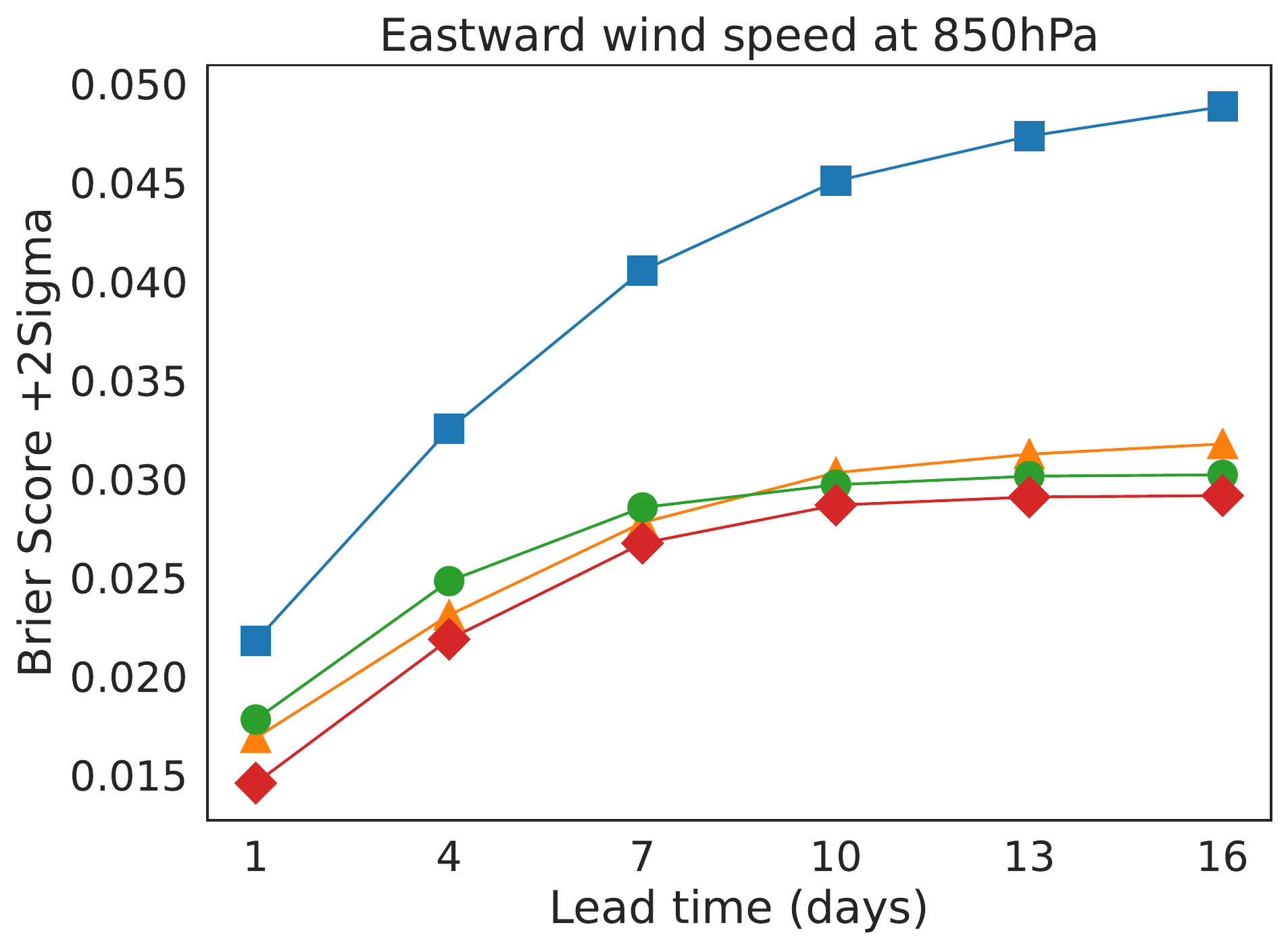} 
    \includegraphics[width=0.32\columnwidth,draft=false]{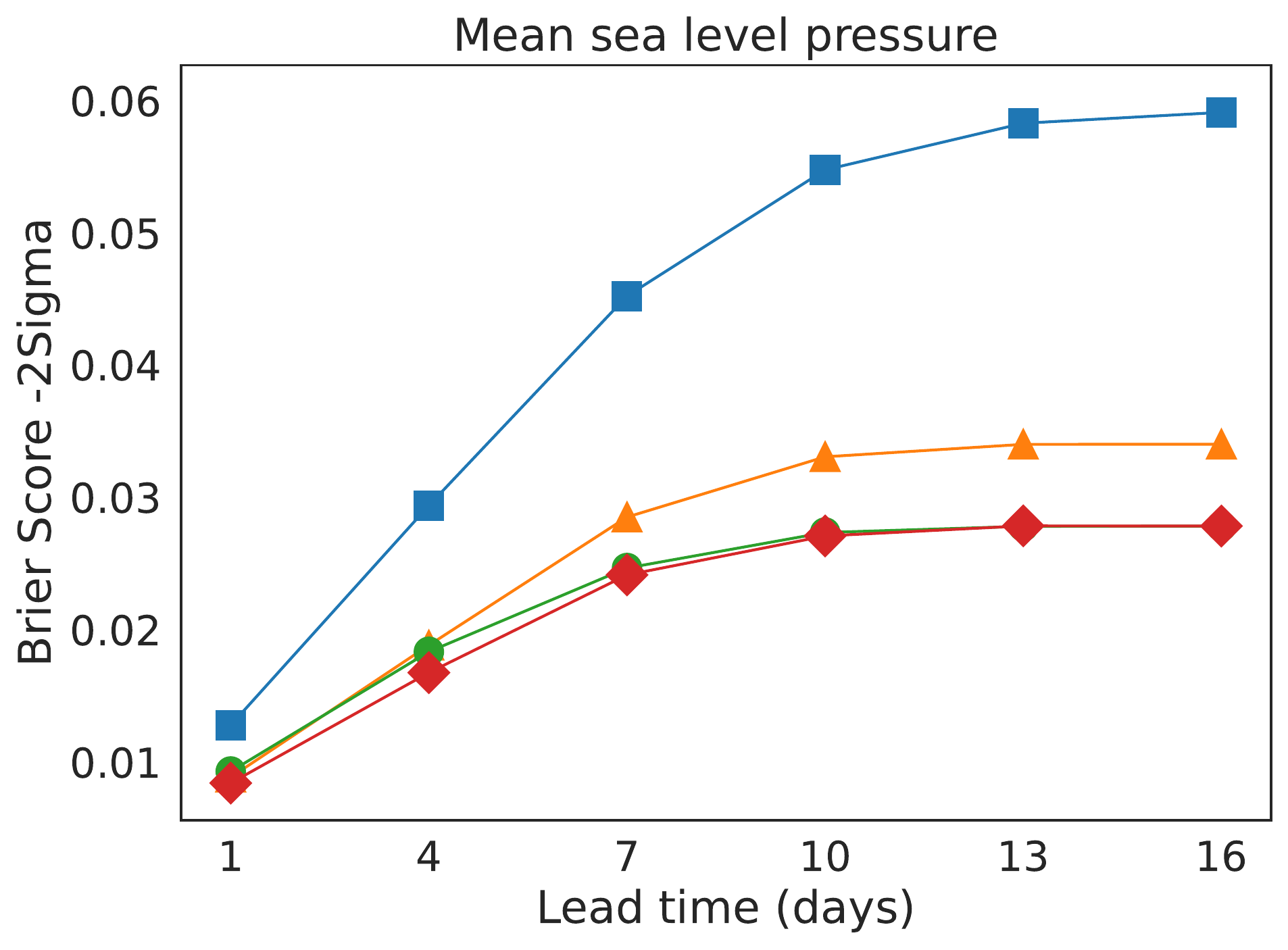} 
    \includegraphics[width=0.32\columnwidth,draft=false]{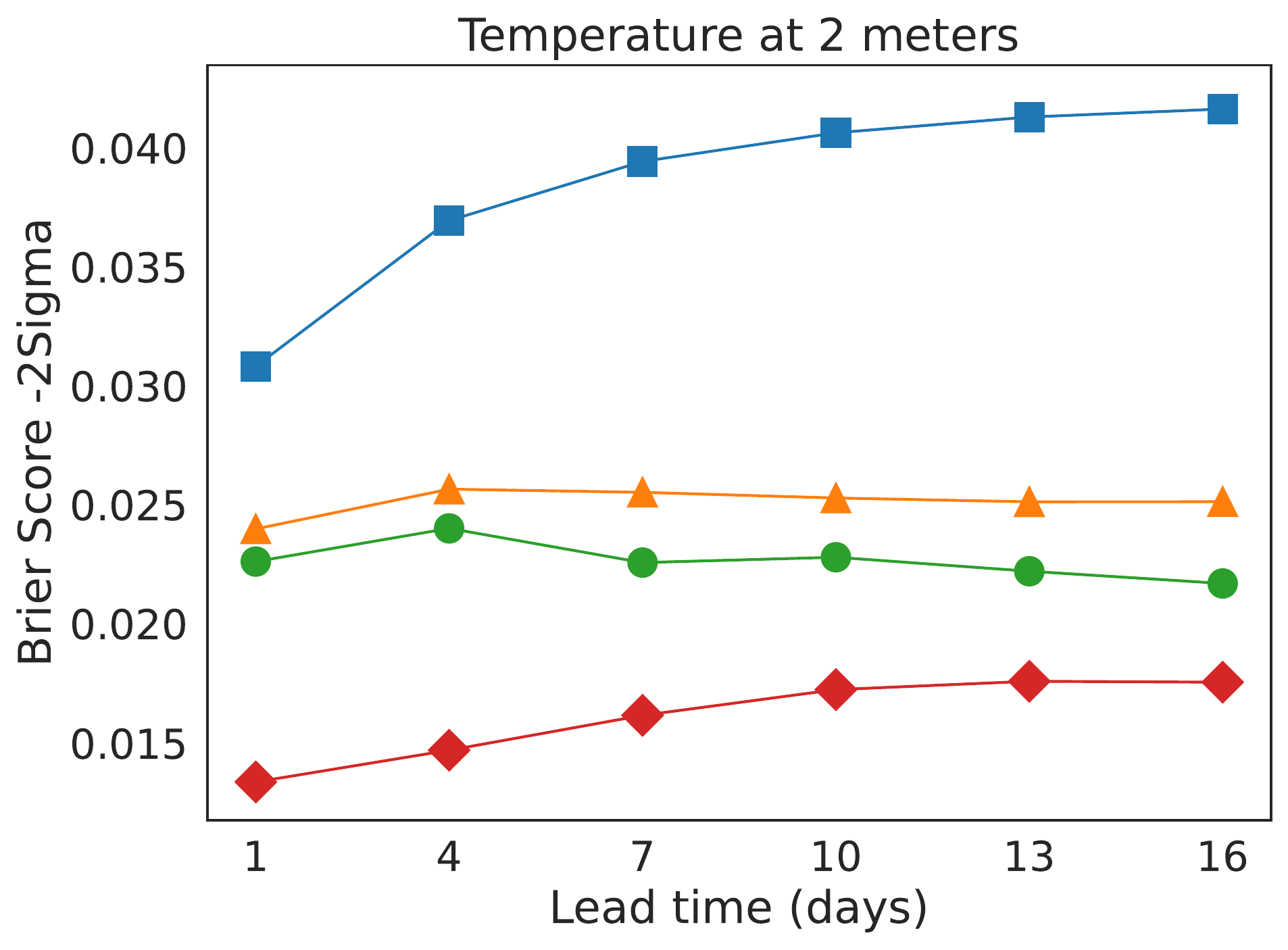}  
    \includegraphics[width=0.32\columnwidth,draft=false]{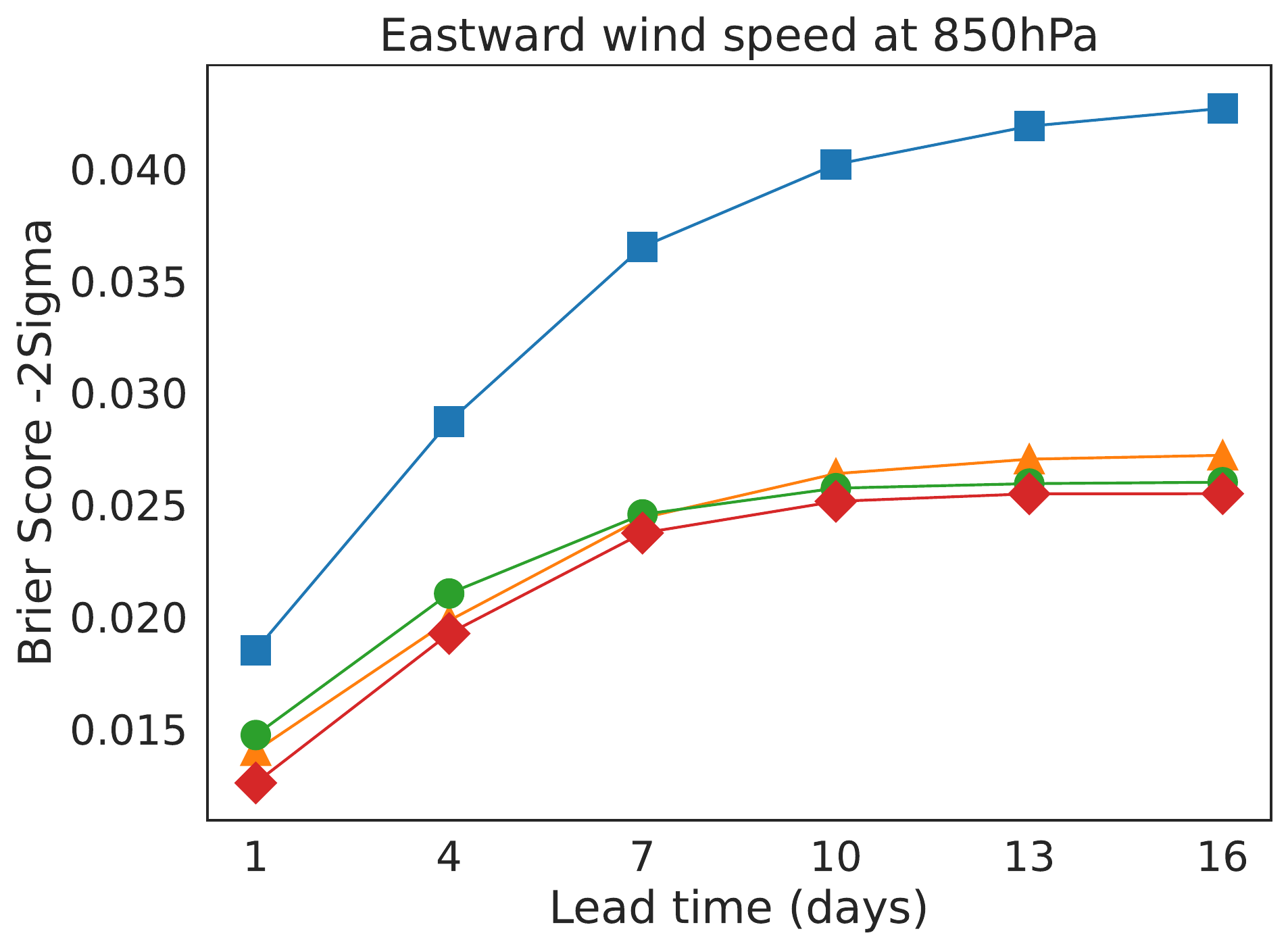}  
    \includegraphics[width=0.5\columnwidth,draft=false]{figs/gee_gpp_c2/legend}  
\caption{Binary classification skill of the different ensembles regarding extreme events ($\pm 2\sigma$ from climatology) in mean slea level pressure, $2$-m temperature, and zonal velocity at $850$~hPa, using ERA5 HRES as the ground-truth. Skill is measured in terms of the cross-entropy; lower values are indicative of higher skill. First row: Brier score for $+2\sigma$. Second row: Brier score for $-2\sigma$.}
\label{fLogLoss2sigma}
\end{figure}

A particularly challenging but important task of ensemble forecasts is being able to forecast extreme events and assign meaningful likelihoods to them \citep{Palmer2002}.   Figure~\ref{fLogLoss2sigma} compares the skill of the same 4 ensembles in predicting events deviating at least $\pm 2\sigma$ from the mean climatology. We measure binary classification skill by computing the Brier score of occurrence using ERA5 HRES as the binary reference, and assigning a probability of occurrence to the ensemble forecasts equal to the fraction of occurrences within the ensemble.

We observe that \ourgee is comparable in skill to the full ensemble \gefsfull and far exceeds the skill of the seeding forecast ensemble \gefsK.  In the forecast of 2-meter temperature, \ourgpp performs noticeably better than the other ensembles. For other variables, despite the less apparent advantage, \ourgpp remains the best extreme forecast system for most lead times and variables. This highlights the relevance of our generative approach for forecasting tasks focused on extremes.

\subsection{Hallucination or In-filling?}

One frequently cited issue of generative AI technology is its tendency to ``hallucinate information''. We conclude this section by exploring the nature of the distribution information that the generative ensembles are able to represent, beyond what is present in the two seeding forecasts from the GEFS full ensemble. As shown previously, the generated ensembles outperform the seeding forecast ensembles in all metrics and often match or improve over the physics-based full ensemble.

\begin{figure}[tp]
    \centering
    \includegraphics[width=0.32\columnwidth,draft=false]{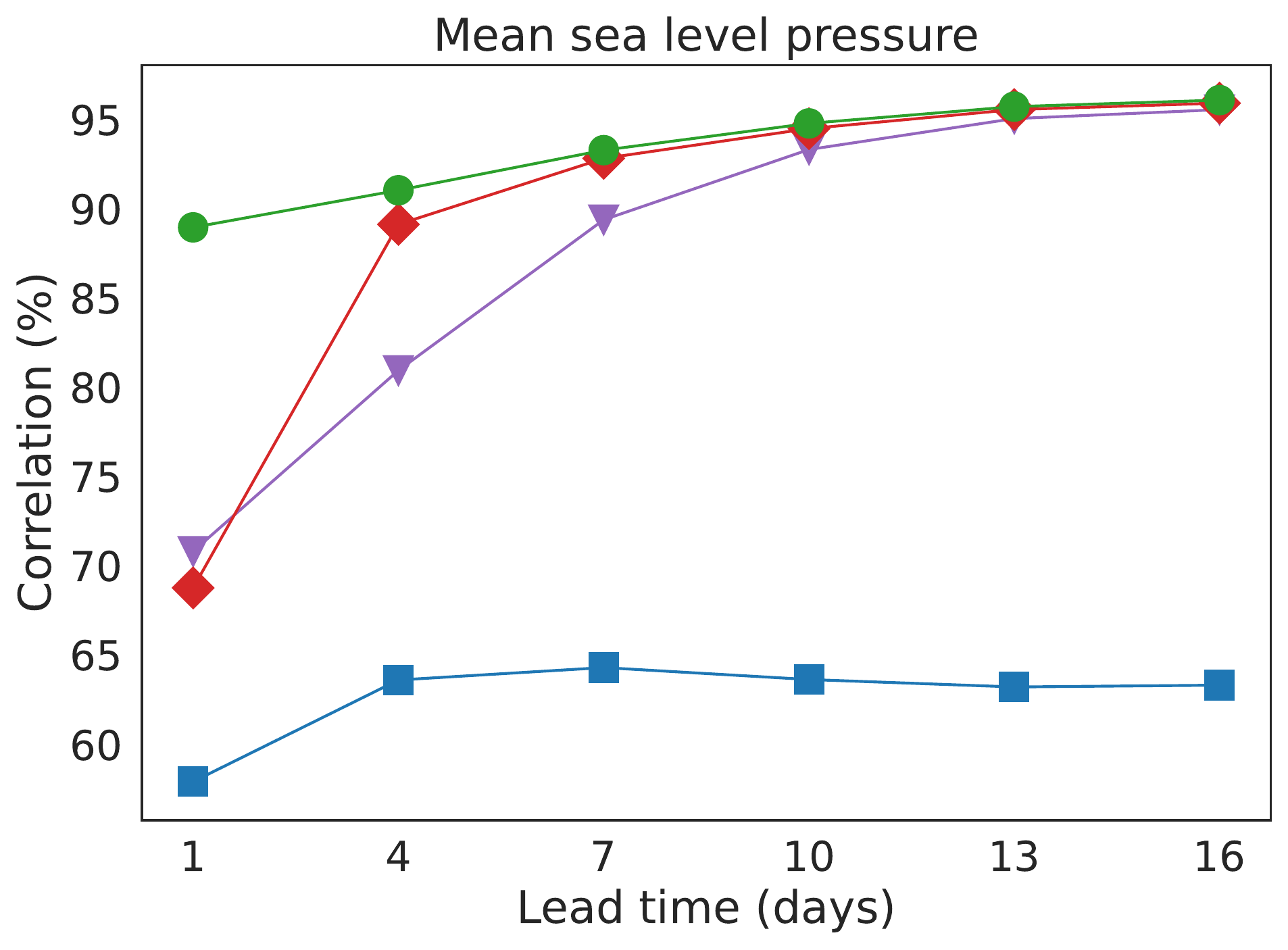} 
    \includegraphics[width=0.32\columnwidth,draft=false]{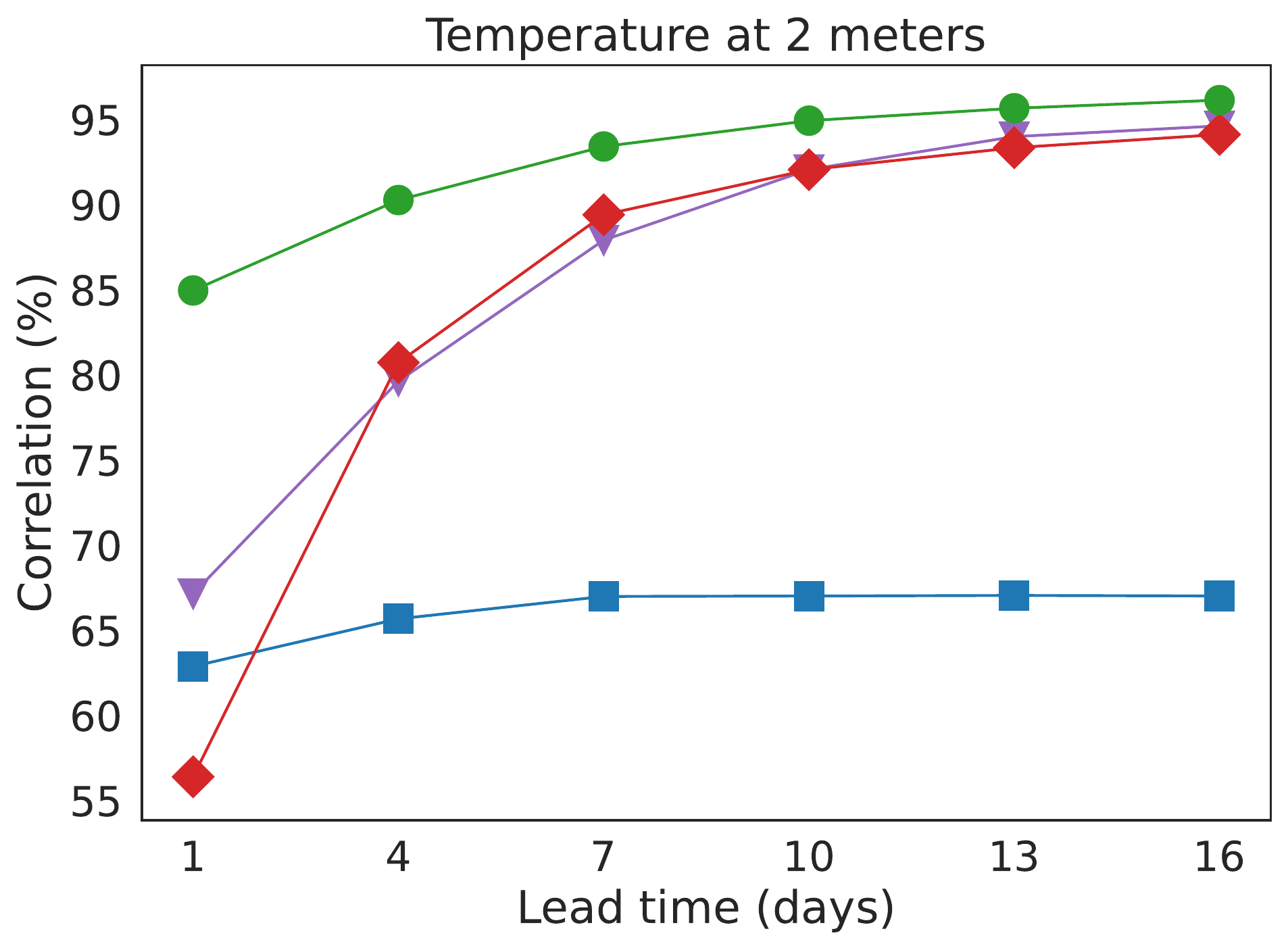}  
    \includegraphics[width=0.32\columnwidth,draft=false]{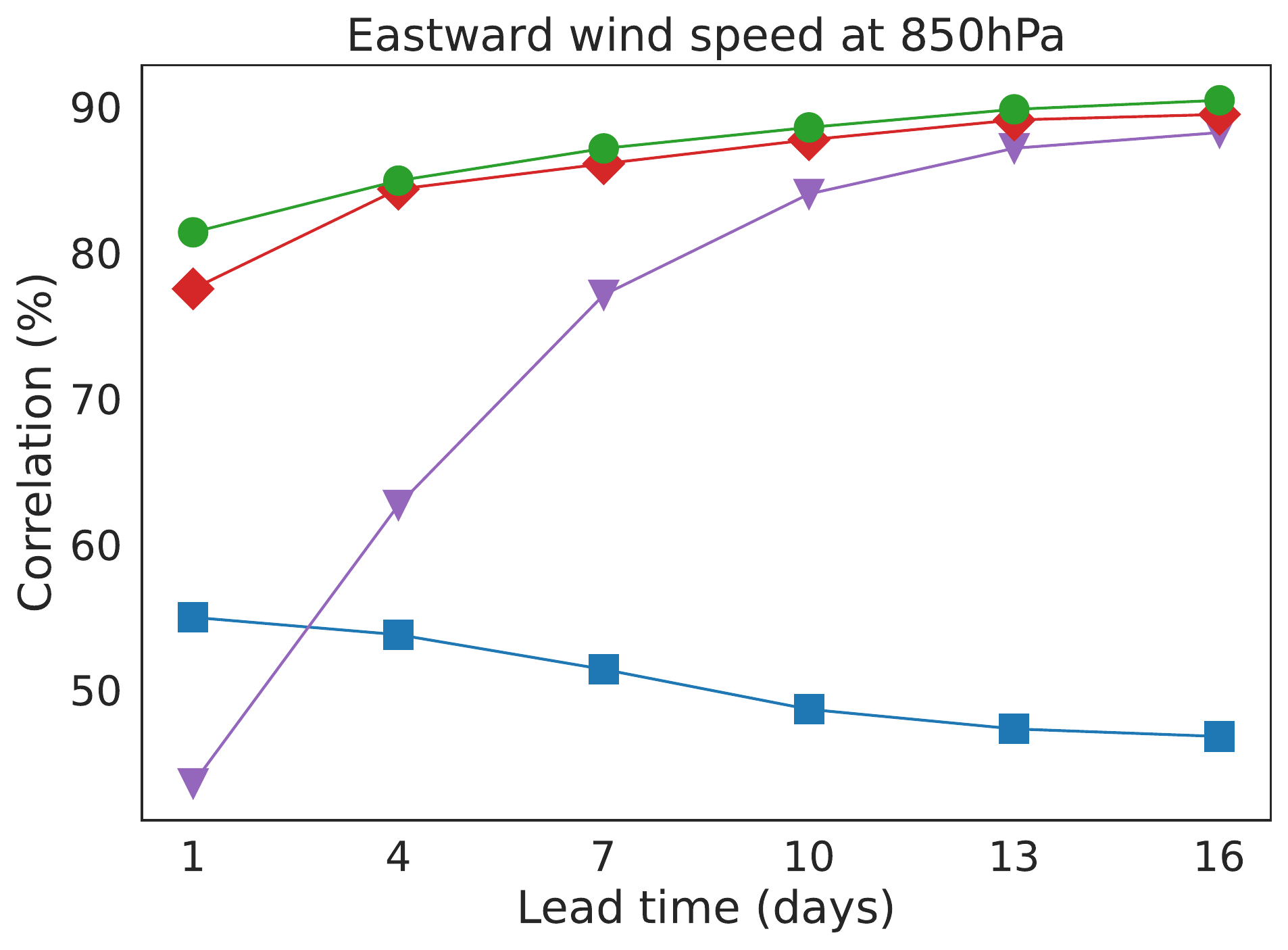}
    \includegraphics[width=0.5\columnwidth,draft=false]{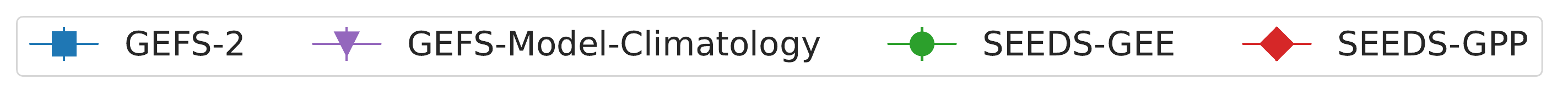}
\caption{Comparing the ensembles and the model climatology to \gefsfull in terms of how the ensemble spreads are correlated with those from \gefsfull.  The plots show that in medium-range between 4 to 10 days, the model has leveraged the  two seeding forecasts to generate different, yet informative, ensemble members to represent uncertainty.}
\label{fDispersionCorr}
\end{figure}

Figure~\ref{fDispersionCorr} measures the correlation of the generative ensembles (\ourgee and \ourgpp), the seeding ensemble \gefsK, and the GEFS model climatology, with respect to the \gefsfull ensemble forecasts.  While comparing full joint distributions remains infeasible, we compute how well the spread of each ensemble forecast correlates with that of the full physics-based ensemble \gefsfull.  The plots show that at long-lead times ($\ge 10$ days),  all ensembles but \gefsK converge to high correlations ($\ge 95\%$) with \gefsfull.  This is also true for the model climatology.  However, in the medium range (more than 4 days but less than 10 days ahead), the generative ensembles display a higher correlation with the \gefsfull than both the model climatology and \gefsK. This suggests that the generative models are indeed able to generate information about forecast uncertainty beyond the two seeding forecasts. In addition, the fact that generative ensembles can capture a higher correlation with \gefsfull than the model climatology in the short and medium range shows that the diffusion models are learning to emulate dynamically-relevant features beyond model biases; they have resolution beyond climatology. 

Thus, we put forward a reasonable hypothesis that the generated ensembles in-fill probability density gaps in the small seeding ensembles.  They also extend the (tails of the) envelopes of the full ensembles such that extreme events are well represented in the envelops.

\section{Related Work}
\label{sRelated}

Previous work leveraging artificial intelligence to augment and post-process ensemble forecasts has focused on improving the aggregate output statistics of the prediction system. \citeauthor{Scher2018-ne} trained a convolutional neural network to quantify forecast uncertainty given a single deterministic forecast~\citep{Scher2018-ne}. They learned a global measure of uncertainty in a supervised setting, using as labels either the error of previous forecasts or the spread of an ensemble system. \citeauthor{Brecht2023} generalized this approach by predicting the ensemble spread at each forecast location, given a deterministic forecast~\citep{Brecht2023}. For this task, they used a conditional generative adversarial network based on the \textsc{pix2pix} architecture~\citep{Isola2017}. \citeauthor{Gronquist2021-am} trained a deep learning system to post-process a 5-member ensemble forecast, resulting in a lower CRPS than a 10-member ensemble from the same operational system~\citep{Gronquist2021-am}. \citeauthor{sacco22MLens} extended this work to build a system capable of predicting the ensemble mean and spread over a limited domain~\citep{sacco22MLens}.

Our work differs from that of \cite{Brecht2023}, \citep{Gronquist2021-am}, and \citep{sacco22MLens} in that our probabilistic generative model outputs actual samples from the target forecast distribution. Thus, our approach offers added value beyond the ensemble mean and spread: the drawn samples can be used to characterize spatial patterns associated with weather extremes \citep{Scher2021}, or as input to targeted weather applications that depend on variable and spatial correlations \citep{Palmer2002}.
\section{Discussion}
\label{sDiscuss}

The \ourmethod (\ourmethodshort) proposed in this work leverages the power of generative artificial intelligence to produce ensemble forecasts comparable to those from the operational GEFS system at accelerated pace -- the results reported in this paper need only 2 seeding forecasts from the operational system, which generates 31 forecasts in its current version \citep{Zhou2022-nv}. This leads to a hybrid forecasting system where a few weather trajectories computed with a physics-based model are used to seed a diffusion model that can generate additional forecasts much more efficiently. This methodology provides an alternative to the current operational weather forecasting paradigm, where the computational resources saved by the statistical emulator could be allocated to increasing the resolution of the physics-based model \citep{Ma2012}, or issuing forecasts more frequently.

\ourmethodshort is trained on historical retrospective forecasts (\ie, reforecasts) issued with the operational physics-based model, which are already required for post-processing in the current paradigm \citep{Hagedorn2008}. Our framework is also flexible enough to enable direct generation of debiased ensembles when the generative post-processing task is considered during training; the only additional requirement is access to historical reanalysis for the reforecast period.

For future work, we will conduct case studies of high-impact weather events to further evaluate \ourmethodshort' performance, and consider specific ensemble forecast applications such as tropical and extratropical cyclone tracking \citep{Froude2007-ep, Lin2020-in}. We will also explore more deeply the statistical modeling mechanisms that such models employ to extract information from weather data and in-fill the ensemble forecast distribution. It is our belief that our application of generative AI to weather forecast emulation represents just one way of many that will accelerate progress in operational NWP in coming years. Additionally, we hope the established utility of generative AI technology for weather forecast emulation and post-processing will spur its application in research areas such as climate risk assessment, where generating a large number of ensembles of climate projections is crucial to accurately quantifying the uncertainty about future climate \citep{Deser2020}.

\begin{ack}
Our colleagues at Google Research have provided invaluable advice.  Among them, we thank Stephan Rasp, Stephan Hoyer, and Tapio Schneider for their inputs and useful discussion on the manuscript. We thank Carla Bromberg and Tyler Russell for technical program management, as well as Alex Merose for data coordination and support. We also thank Cenk Gazen, Shreya Agrawal and Jason Hickey for discussions with them in the early stage of this work.
\end{ack}

\bibliographystyle{plainnat}
\bibliography{ensemble}

\newpage
\appendix

\section{Probabilistic Diffusion Models}
\label{sBackground}

We give a brief introduction to probabilistic diffusion models. Diffusion models are powerful methods for learning distributions from data. They have recently become one of the most popular approaches for image generation and video synthesis \citep{ho2022imagen}. For a detailed description, see~\cite{song2020score}. 
 
Consider a multivariate random variable $\vV$ with an underlying distribution $p_{\text{data}}(\vv)$.  Intuitively, diffusion-based generative models iteratively transform samples from an initial noise distribution $p_\mathcal{T}$ into samples from the target data distribution $p_{\text{data}}$ through a denoising operator. By convention, $\mathcal{T}$ is set to 1 as it is just nominal and does not correspond to the real time in the physical world\footnote{For notation clarity, we will be using $\tau$ and its capitalized version $\mathcal{T}$ to denote diffusion times, which are different from the physical times.}. Additionally, as it will be clear from the explanation below, this initial distribution is a multivariate Gaussian distribution.

Noise is removed  such that the samples follow a family of diffusion-time-dependent marginal distributions $p_\tau(\vv_\tau; \sigma_\tau)$ for decreasing diffusion times $\tau$ and noise levels $\sigma_\tau$~\citep{Ho2020}. The distributions are given by a forward blurring process that is described by the following stochastic differential equation (SDE)~\cite{karras2022elucidating,song2020score}
\begin{equation}
\label{eq:fwd_sde}
  d\vV_\tau  = f(\vV_\tau, \tau) d\tau +g(\vV_\tau, \tau) dW_\tau, 
\end{equation}
with drift $f$, diffusion coefficient $g$, and the standard Wiener process $W_\tau$. Following~\cite{karras2022elucidating}, we set
\begin{equation}
    f(\vV_\tau, \tau) = f(\tau) \vV_\tau := \frac{\dot{s}_\tau}{s_\tau}\vV_\tau,  \qquad \text{and} \qquad
   g(\vV_\tau, \tau) = g(\tau) := s_\tau\sqrt{2\dot{\sigma}_\tau\sigma_\tau},
\end{equation}
where the overhead dot denotes the time derivative.
Solving the SDE in~\eqref{eq:fwd_sde} forward in time with an initial condition $\vv_0$ leads to the Gaussian perturbation kernel $p_\tau(\vV_\tau|\vv_0) = \mathcal{N}( s_\tau \vv_0, s_\tau^2\sigma_\tau^2\textbf{I})$. Integrating the kernel over the data distribution $p_0(\vv_0) = p_{\text{data}}$, we obtain the marginal distribution $p_\tau(\vv_\tau)$ at any $\tau$. As such, one may prescribe the profiles of $s_\tau$ and $\sigma_\tau$ so that $p_0 = p_{\text{data}}$ (with $s_0 = 1, \sigma_0 = 0$),
and more importantly
\begin{equation}
 \label{eq:dist_t}
     p_\mathcal{T}(\vV_\mathcal{T}) \approx \mathcal{N}( 0, s_\mathcal{T}^2\sigma_\mathcal{T}^2\mathbf{I}),
 \end{equation}
i.e., the distribution at the (nominal) terminal time $\mathcal{T}=1$ becomes indistinguishable from an isotropic, zero-mean Gaussian. To sample from $p_{\text{data}}$, we utilize the fact that the  reverse-time SDE 
\begin{equation}
\label{eq:bwd_sde}
 d\vV_\tau = \big[f(\tau) \vV_\tau - g(\tau)^2\nabla_{\vV_\tau} \log p_\tau(\vV_\tau)\big]d\tau +g(\tau) dW_\tau
\end{equation}
has the same marginals as eq.~\eqref{eq:fwd_sde}. Thus, by solving \eqref{eq:bwd_sde} \emph{backwards} using \eqref{eq:dist_t} as the initial condition, we obtain samples from $p_{\text{data}}$ at $\tau=0$.

There are several approaches to learn the score function $\nabla_{\vV_\tau} \log p_\tau(\vV_\tau)$. The key step is to parameterize it with a neural network $s_\theta(\vv_\tau,  \sigma_\tau)$ such that 
\begin{equation}
s_\theta(\vv_\tau,  \sigma_\tau) \approx \nabla _{\vv_\tau}\log p_\tau(\vv_\tau| \vv_0) = -\frac{s_\tau \vv_0 - \vv_\tau}{\sigma^2_{\tau}} =z_\tau/\sigma_\tau
\label{eScoreNetwork}
\end{equation}
for any $\vv\sim p_\tau(\vV_\tau|\vv_0 )$ and $\sigma_\tau$. In other words, the \emph{normalized} score network $\sigma_\tau s_\theta(\vv_\tau,  \sigma_\tau)$ identifies the noise $z_\tau$ injected into the data $\vv_0$. The parameters $\theta$ of the score network are trained to reduce the L2 difference between  them. Once this score network is learned, it is used to \emph{denoise} a noised version of the desired data $\vv_0$.   We refer to this process as learning the \emph{normalized score function}. In this work, we use the hyperparameters $f(\tau) = 0$ and $g(\tau)=100^\tau$.
\section{Method Details}
\label{sMethodAppendix}

\subsection{Problem Setup and Main Idea}

Here we provide a detailed description of the two generative tasks considered. Formally, let $p(\vv)$ denote an unknown distribution of the atmospheric state $\vV$. In the task of \textbf{generative ensemble emulation}, we are given a few examples or \textit{seeds} sampled from $p(\vv)$ and our task is to generate more samples from the same distribution.

Given $K$ samples $\mathcal{E}^K=(\vv^1, \vv^2, \cdots, \vv^K)$ from $p(\vv)$, we construct an easy-to-sample \emph{conditional} distribution
\begin{equation}
\hat{p}(\vv) = p( \vv ; \mathcal{E}^K),
\label{eCondition}
\end{equation}
which approximates $p(\vv)$.  The conditional distribution $\hat{p}(\vv)$ needs to have two desiderata: it approximates well $p(\vv)$, and it is much less costly to sample than $p(\vv)$.  Note that this problem extends the typical density estimation problem, where there is no conditioning, \ie, $K=0$.

In the task of \textbf{generative post-processing}, we construct an efficient sampler to approximate a mixture distribution. As another example,  let $p'(\vv)$ be the reanalysis distribution of the atmospheric state corresponding to the forecast $p(\vv)$.  We construct a conditional distribution $p( \vv ; \mathcal{E}^K)$ to approximate the following mixture of distributions:
\begin{equation}
   p^\textsc{mix}(\vv) =  \alpha p(\vv) + (1-\alpha)p'(\vv).
   \label{eMix}
\end{equation}
In practice, the information about $p'(\vv)$ is given to the learning algorithm in the form of a representative $K'$-member ensemble $\mathcal{E}'^{K'}$. The mixture weight $\alpha$ is then given by
\begin{equation}
    \alpha = \frac{M-K}{M-K+K'}.
\end{equation}
where $M>K $ is the number of samples we have drawn from $p(\vv)$.
In the following, we describe how to apply the technique of generative modeling introduced in~\ref{sBackground} to these two tasks.  We first mention how data is organized.

\subsection{Data for Training and Evaluation}
\label{sTrainDataAppendix}

Each ensemble $\mathcal{E}_{tl}$ is identified by the forecast initialization day $t$ and the lead time $l$. Here, $t \in T^{\textsc{train}}$ indexes all the days in the training set and $l \in L$, where $L= \{1, 3, 6, 10, 13, 16\}$~days is the set of lead times we consider.  We group ensembles of  the same lead time $l$ as a training dataset $\mathcal{D}^\textsc{train}_l = \{\mathcal{E}_{1l}, \mathcal{E}_{2l}, \cdots, \mathcal{E}_{T^\textsc{train}l}\}$. In generative ensemble emulation, $\mathcal{D}^\textsc{train}_l$ contains samples from \reforecastfull exclusively, since the target distribution is \gefsfull. In generative post-processing, $\mathcal{D}^\textsc{train}_l$ contains samples from both \reforecastfull and  \erafull, to captue the target distribution eq.~\eqref{eMix}.

Note that the $K$ seeds can be included in the generated ensembles for evaluation. However, in this work, we decide to adopt a stricter evaluation protocol where the seeds are excluded from the generated ensembles; including them would mildly improve the quality of the generated ensembles, since the number of generated forecasts is far greater than $K$.

As a shorthand, we use $\vv_t$ to denote all members in the ensemble forecast made at time $t$, and $\vv_{tm}$ to denote the $m$\textit{th} forecast at the same time. 

\subsubsection{Generative Ensemble Emulation}

In the ensemble emulation task we use $K$ samples to construct a sampler of the conditional distribution $\hat{p}(\vv)$. To this end, for each lead time $l$ and value of $K$, we learn a distinct conditional distribution model $\mathcal{M}_{lK}$. We choose $K=1,2,3,4$; and use $\mathcal{D}^\textsc{train}_l$ to learn $\mathcal{M}_{lK}$.

To train the conditional generation model $\mathcal{M}_{lK}$, we use as input data $K$ randomly chosen members from each ensemble forecast $\vv_t$ in $\mathcal{D}^\textsc{train}_l$, and another member $\vv_{tm_0}$ from the remaining $(M-K)$ members as our target. This procedure augments the data in $\mathcal{D}^\textsc{train}_l$ by $C_M^K$-fold, where $C_M^K$ is the number of $K$-combinations of the $M$ members. 

The learning goal is then to optimize the score network eq.~\eqref{eScoreNetwork}, so that the resulting sampler gives rise to:
\begin{equation}
    p_{lK}( \vv_{tm_0} |  \vv_{tm_1},   \vv_{tm_2}, \cdots,   \vv_{tm_K}),
    \label{eConditionalDistribution}
\end{equation}
where $m_1, m_2, \cdots, m_K,$ and $m_0$ index the chosen ensemble members.  Specifically, the score network is given by
\begin{equation}\label{scoreEmulation}
  s_\theta( \vv_{tm,\tau}, \vv_{tm_1}, \vv_{tm_2}, \cdots, \vv_{tm_K}, \mathbf{c}_t, \sigma_\tau)
\end{equation}
where $\vv_{tm,\tau}$ is a perturbed version of $\vv_{tm_0}$.  We include the climatological mean $\mathbf{c}_t$ as an input to the score function eq.~\eqref{scoreEmulation}, since the atmospheric state is non-stationary with respect to the time of the year.

\subsubsection{Generative Post-Processing}

For the task of generative post-processing, the setup is similar except that we seek to approximate the mixture distribution of $p(\vv)$ and $p'(\vv)$ defined in eq.~\eqref{eMix}. The training dataset $\mathcal{D}_l^{\textsc{train}}$ thus contains ensembles from two different distributions, $\vv_t$ and $\vv'_t$.

In this case, we train conditional distribution models $\mathcal{M}_{lKK'}$ for each triplet $(l, K, K')$ using the same input as the conditional emulator $\mathcal{M}_{lK}$, \ie, $K$ out of $M$ members of each ensemble $\vv_t$. However, the target sample is now drawn randomly from either $p(\vv)$ or $p'(\vv)$. Consistent with eq.~\eqref{eMix}, we draw a sample from the remaining $(M-K)$ members of ensemble $\vv_t$ with probability $(M-K)/(M-K+K')$, or a member from $\vv'_t$ with probability $K'/(M-K+K')$. Denoting the final selection as $\vx_{t0}$, the score function becomes
\begin{equation}\label{scorePost}
  s_\theta( \vx_{t0,\tau}, \vv_{tm_1}, \vv_{tm_2}, \cdots, \vv_{tm_K}, \mathbf{c}_t, \sigma_\tau)
\end{equation}
where $\vx_{t0, \tau}$ is a perturbed version of $\vx_{t0}$.

In this task, $K'$ is a hyperparameter that controls the mixture distribution being approximated. We choose  $K'$ from 2, 3 or 4.

\subsection{Model Architecture}

We employ an axial variant of the vision Transformer (ViT) \cite{Dosovitskiy2021ViT} to model the normalized score functions $\sigma_\tau s_\theta(\cdot, \cdot)$ in eqns.~\eqref{scoreEmulation} and \eqref{scorePost}, adapted to the particular characteristics of atmospheric modeling.

Each atmospheric state snapshot $\vv$ is represented as a tensor $v_{qp}$, where $q$ indexes the physical quantities and $p$ the locations on the cubed sphere. In particular, the altitude levels of the same physical quantity are also indexed by $q$. The overall model operates on sequences of atmospheric state snapshots $v_{sqp}$ where $s$ indexes the sequence position. In the context of computer vision, this is similar to a video model, where $p$ indexes the position in the image, $q$ the ``RGB'' channels, and $s$ the time.

The number of channels, which we use to model physical fields at different atmospheric levels, can be $q\sim\mathcal{O}(100)$ in our setting.
For this reason, we resort to axial attention, applied to $p$, $q$, and $s$ separately \citep{ho2019axial}.
Each sequence dimension has its own learned position embedding instead of the usual fixed positional encodings.

\subsubsection{Spatial Embedding}

\begin{figure}[t]
    \centering
    \includegraphics[width=0.85\columnwidth,draft=false]{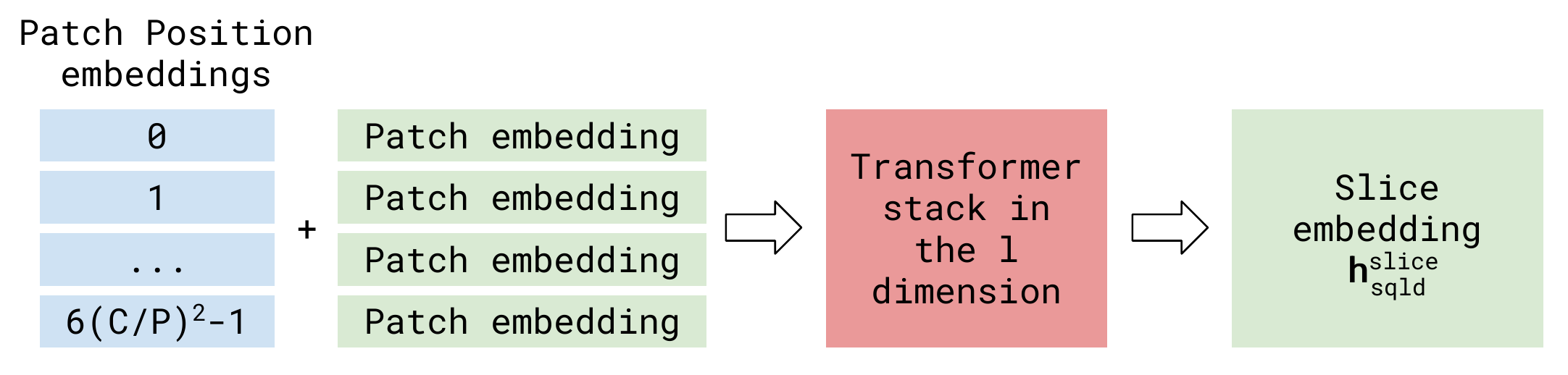}  
    \caption{Schematic of the spatial attention mechanism. Patches of cubed sphere grid points are mapped to patch embeddings, and a transformer stack transforms their combination with patch position embeddings in the patch ($l$) dimension\label{fTl}.}
\end{figure}

We call a piece of data assigned to each grid point of the cubed sphere as a \textbf{slice}.
Regarding spatial attention, the $6\times C\times C$ values of a slice associated with the cubed sphere grid points indexed by $p$ are partitioned into $6(C/P)^2$ square patches of size $P\times P$. We employ $C=48$ in our work, corresponding roughly to $2^\circ$ resolution. Each patch is then flattened and linearly embedded into $D$ dimensions with learned weights, leading to a tensor $h_{sqld}$, where $l=0,\ldots 6(C/P)^2-1$ indexes the patches and $d$ the in-patch dimension.
Due to the nontrivial neighborhood structure of the cubed sphere, a shared learned positional encoding $e^{\text{pos}}_{ld}$ is used for the $l$ dimension.
This is then fed into a transformer of $T_L$ layers operating with $l$ as the sequence dimension and $d$ as the embedding dimension. The output is the the  slice embedding corresponding to each patch:
\begin{equation}
h^{\text{slice}}_{sqld} = T_L(h_{sqld} + e^{\text{pos}}_{ld}).
\end{equation}
For a schematic illustration, see Figure~\ref{fTl}.

\begin{figure}[t]
    \centering
    \includegraphics[width=0.85\columnwidth,draft=false]{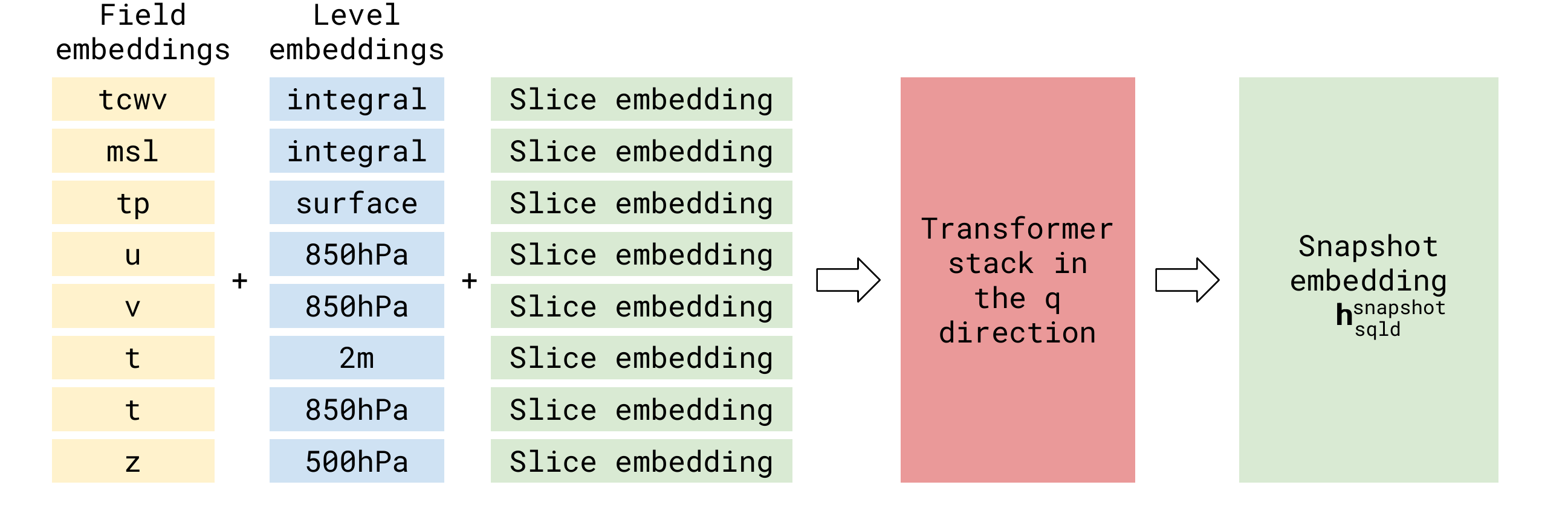}  
    \caption{Transformer stacked in the $q$ dimension\label{fTf}}
\end{figure}

\subsubsection{Atmospheric Field Embedding}

The physical quantities have discrete semantic meanings, so we encode the positions in the
$q$ dimension using a sum of two shared learned embeddings $e^{\text{field}}_{qd}$ and $e^{\text{level}}_{qd}$.
We treat field labels like ``mean sea-level pressure'' and ``eastward wind speed'', and generalized levels like ``surface'', ``2m'', ``500hPa'', or ``integratal'' as strings. We then encode them with a categorical embedding.
This is then fed into a transformer of $T_F$ layers operating with $q$ as the sequence dimension and $d$ as the embedding dimension. The output is the snapshot embedding
\begin{equation}
h^{\text{snapshot}}_{sqld} = T_F(e^{\text{field}}_{qd} + e^{\text{level}}_{qd} + h^{\text{slice}}_{sqld}).
\end{equation}
For the schematic illustration, see Figure~\ref{fTf}.

\subsubsection{Sequence Embedding}

The neural network operates on a sequence of snapshots.
We tag each snapshot by adding two embeddings: a learned categorical embedding $e^{\text{type}}_{sd}$
with values like ``Denoise'', ``GEFS'', ``Climatology'' for the type of the snapshot and a random Fourier embedding \cite{Tancik2020Fourier} $e^{\text{time}}_{sd}$ for the relative physical time.
In particular, the label ``Denoise'' denotes a snapshot for the score function input $\vv$.
For examples, GEFS ensemble members are labeled as type ``GEFS'' and the same relative physical time embedding, so to the sequence transformer, they are exchangeable as intended.

\begin{figure}[t]
    \centering
    \includegraphics[width=0.85\columnwidth,draft=false]{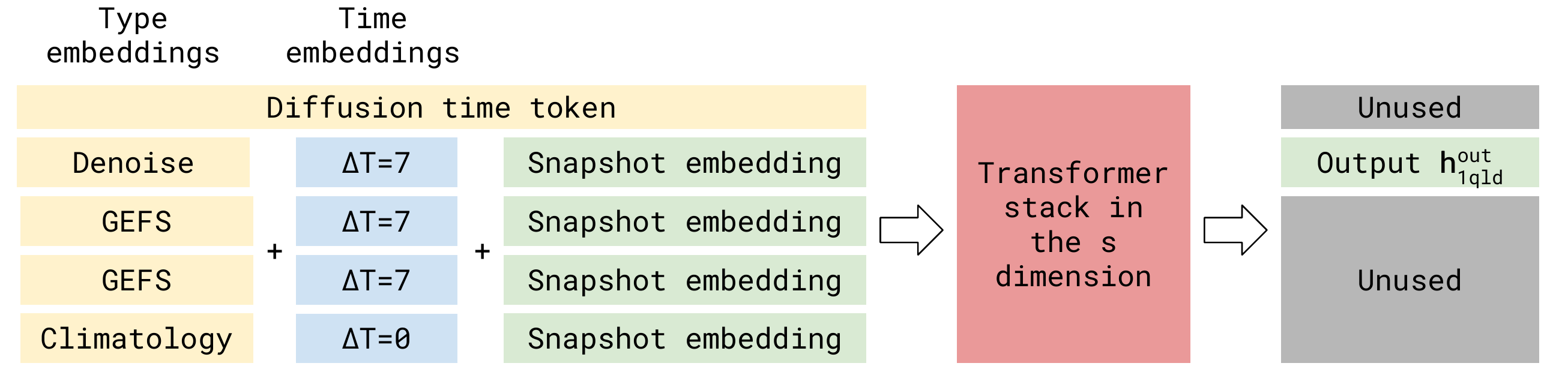}  
    \caption{Transformer stacked in the $s$ dimension\label{fTs}}
\end{figure}

The score function of a diffusion model requires the diffusion time $\tau$ as an additional input,
which we model by prepending to the sequence the token $e^{\tau}_{d}$, derived from a random Fourier embedding of $\tau$.
The sequence and its embeddings are then fed into a transformer of $T_S$ layers operating with $s$ as the sequence dimension and $d$ as the embedding dimension. The output embedding is
\begin{equation}
h^{\text{out}}_{sqld} = T_S(e^{\tau}_{d} \oplus_s (e^{\text{type}}_{sd} + e^{\text{time}}_{sd} + h^{\text{snapshot}}_{sqld})),
\end{equation}
where $\oplus_s$ means concatenation in the $s$ dimension.  This is illustrated in Figure~\ref{fTs}.

To obtain the score function $s_\theta(\vv, \tau)$, we take the output token $h^{\text{out}}_{1qld}$ matching the position of the denoise input and project it with learned weights and reshape back to a tensor $v^{\text{out}}_{qp}$ as the final output.

\subsubsection{Hyperparameters}

The most salient hyperparameters of this architecture are the patch size $P$, the model embedding dimension $D$, and the number of layers $L_L,L_F,L_S$ of transformer stacks $T_L,T_F,T_S$. In our work, we use patch size $P=12$ (number of patches $96$), embedding dimension $D=768$, and transformer layers $(T_L,T_F,T_S)=(6,4,6)$. The hidden layers in the feed-forward network have dimension $4D$ as usual.
The model has 113,777,296 trainable parameters. These parameters are learned by training over randomly sampled days from the 20-year GEFS reforecast dataset~\citep{Guan2022-vl}, using a batch size 128 for 200,000 steps.

\section{Evaluation Metrics and More Results}
\label{sAppendixResults}

Similar to how we organize the training data (cf.~\ref{sTrainDataAppendix}), we partition all evaluation data according to the lead time $l$, and generate one evaluation dataset $\mathcal{D}^\textsc{eval}_l$ for each  training dataset $\mathcal{D}^\textsc{train}_l$. In the main text, we report results for ensembles generated by models $\mathcal{M}_{l2}$ and $\mathcal{M}_{l2K'}$, conditioned on $K=2$ ensemble forecasts. 

\subsection{Evaluation Metrics}

We use $\vv$ (or $\vw$) to denote ensemble forecasts, which are indexed by $t$ (time), $m$ (ensemble member ID), $q$ (the atmospheric variable of interest), and $p$ (the geolocation). We use $M_{\vv}$ to denote the number of members in the ensemble, $\overline{\vv}$ to denote the ensemble mean forecast, and $s_{\vv}$ to denote the spread:

\begin{align}
\overline{\vv} & = \frac{1}{M_{\vv}} \sum_{m=1}^{M_{\vv}} \vv_m,\\
    s_{\vv} & = \sqrt{\frac{1}{M_{\vv}-1}\sum_{m}(\vv_{m} -\overline{\vv})^2}.
\end{align}
We omit other indices ($t$, $q$ and $p$) to avoid excessive subscripts, assuming the operations are applied element-wise to all of them. 

\subsubsection{RMSE}

The (spatial) root-mean-square-error (RMSE) between two (ensemble) means $\overline{\vv}$ and $\overline{\vw}$  is defined as 
\begin{equation}
    \rmse_t(\overline{\vv}, \overline{\vu}) = \sqrt{\frac{1}{P} \sum_{p=1}^P (\overline{\vv}_{tp} - \overline{\vw}_{tp})^2},
\end{equation}
where $p$ indexes all the geospatial locations. Again, this is computed element-wise with respect to every variable $q$. We summarize the RMSE along the temporal axis by computing the sample mean and variance,
\begin{align}
    \overline{\rmse}(\overline{\vv}, \overline{\vu}) & =  \frac{1}{T}\sum_{t=1}^T  \rmse_t(\overline{\vv}, \overline{\vu}),     \label{eEnsembleMeansRMSE}\\
    s^2_{\rmse(\overline{\vv}, \overline{\vu})} & = \frac{1}{T-1} \sum_{t=1}^T \left( \rmse_t(\overline{\vv}, \overline{\vu}) - \overline{\rmse}(\overline{\vv}, \overline{\vu})\right)^2.
\end{align}
Assuming the errors at different time are \textit{i.i.d.}, the sample variance can be used to estimate the standard error of the mean estimate eq.~\eqref{eEnsembleMeansRMSE}.

\subsubsection{Correlation Coefficients and ACC}

The spatial correlation between two (ensemble) means $\overline{\vv}$ and $\overline{\vw}$  is defined as 
\begin{equation}
  \corr_t(\overline{\vv}, \overline{\vu}) =\frac{\sum_p (\overline{\vv}_{tp} -\overline{\overline{\vv}_{tp}}) (\overline{\vw}_{tp} -\overline{\overline{\vw}_{tp}} ) }{\sqrt{\sum_p(\overline{\vv}_{tp} -\overline{\overline{\vv}_{tp}})^2} \sqrt{\sum_p(\overline{\vw}_{tp} -\overline{\overline{\vw}_{tp}})^2} },
    \label{eEnsembleMeansPearson}
\end{equation}
where $\overline{\overline{\cdot}_{tp}}$ refers to averages in space.

The centered anomaly correlation coefficient (\acc) is defined as the spatial correlation between the climatological anomalies \citep{Wilks2019},
\begin{equation}
\acc_t(\overline{\vv}, \overline{\vu})= \frac{\sum_p (\overline{\vv}_{tp}' -\overline{\overline{\vv}_{tp}'}) (\overline{\vw}_{tp}' -\overline{\overline{\vw}_{tp}'} ) }{\sqrt{\sum_p(\overline{\vv}_{tp}' -\overline{\overline{\vv}_{tp}'})^2} \sqrt{\sum_p(\overline{\vw}_{tp}' -\overline{\overline{\vw}_{tp}'})^2} },
\label{eEnsembleMeansACC}
\end{equation}
where the anomalies are defined as the raw values minus the corresponding climatological mean $\vc_{tp}$,
\begin{equation}
\overline{\vv}_{tp}' = \overline{\vv}_{tp} -\vc_{tp}, \;\;\;\; \overline{\vw}_{tp}' = \overline{\vw}_{tp} -\vc_{tp}.
\label{eAnom}
\end{equation}

Both estimates eq.~\eqref{eEnsembleMeansPearson} and \eqref{eEnsembleMeansACC} are then averaged in time over the evaluation set.

\subsubsection{CRPS}
The continuous ranked probability score (CRPS)~\cite{Gneiting07ProperScoringRules} between an ensemble and  \era is given by
\begin{equation}
    \crps(\vv, \era) = \frac{1}{M_{\vv}} \sum_{m=1}^{M_{\vv}} |\vv_m - \era| - \frac{1}{2M_{\vv}^2}\sum_{m=1}^{M_{\vv}}\sum_{m'=1}^{M_{\vv}} |\vv_m - \vv_{m'}|.
    \label{eEnsembleCRPSAppendix}
\end{equation}
This is computed for all time and geo-spatial locations, and then averaged.

We use the traditional CRPS instead of the ensemble-adjusted CRPS$^*$~\cite{Leutbecher2019}. The latter examines properties of ensembles in the theoretical  asymptotic limit where their sizes are infinite.  Since our approach aims at \emph{augmenting} the physics-based ensembles, the definition eq.~\eqref{eEnsembleCRPSAppendix} is  more operationally relevant; even perfect emulation would obviously not improve the CRPS$^*$ score.
Indeed, our model \ourgee can be seen as a practical way to bridge the skill gap between a small ensemble and the asymptotic limit.

\subsubsection{Rank Histogram and Reliability Metric}
\label{siReliability}
We assess the reliability of the ensemble forecasts in terms of their rank histograms~\citep{Anderson1996}.
An ideal ensemble forecast should have a flat rank histogram, so deviations
from a flat rank histogram indicate unreliability.

We aggregate rank histograms for all $n_{\text{testing}}$ dates in the evaluation set of  days to obtain the average rank histogram $\{s_i\}_{i=0}^{M_{\vv}}$, where $s_i$ is the number of times the label has rank $i$ and $M_{\vv}$ is the ensemble size.
Following Candille and Talagrand~\citep{Candille2005}, the unreliability $\Delta$ of the ensemble is defined
as the squared distance of this histogram to a flat histogram
\begin{equation}
    \Delta=\sum_{i=0}^{M_{\vv}} \left(s_i-\frac{n_{\text{testing}}}{1+M_{\vv}}\right)^2.
\end{equation}
This metric has the expectation $\Delta_0=n_{\text{testing}}\frac{M_{\vv}}{1+M_{\vv}}$ for a perfectly reliable forecast.
The unreliability metric $\delta$ is defined as
\begin{equation}
    \delta=\frac{\Delta}{\Delta_0},
\end{equation}
which has the advantage that it can be used to compare ensembles of different sizes. Once $\delta$ is computed for each location, a global $\delta$ value is obtained by averaging over all locations.
 
\subsubsection{Brier Score for Extreme Event Classification}

To measure how well the ensembles can predict extreme events,  we first apply a binarization criterion $b:\mathbb{R}\rightarrow \{0,1\}$ (such as ``if \textsc{t2m} is 2$\sigma$ away from its mean, issue an extreme heat warning.''), and verify against the classification of the same event by the ERA5 HRES reanalysis,
\begin{equation}
b^{\era}_{tqp}=b(\era_{tqp}),
\end{equation}
by converting the ensemble forecast into a probabilistic (Bernoulli) prediction
\begin{equation}
b^{\vv}_{tqp}=\frac{1}{M_{\vv}}\sum_{m=1}^{M_{\vv}} b(\vv_{tmqp}).
\end{equation}
We can evaluate how  well the  probabilistic predictions align with the discrete labels by evaluating the Brier score~\citep{BRIER1950}
\begin{equation}
\brier^b(\vv, \era)_{tq} =\frac{1}{P}\sum_{p=1}^P
(b^{\era}_{tqp} - b^{\vv}_{tqp} )^2.
\label{eBrierScore}
\end{equation}
Alternatively, the probabilistic predictions can be evaluated by their cross-entropy, also known as the logarithmic loss~\citep{Benedetti2010-jo}
\begin{equation}
\logloss^b(\vv, \era)_{tq} =-\frac{1}{P}\sum_{p=1}^P
b^{\vv}_{tqp} \ln b^{\era}_{tqp} + (1 - b^{\vv}_{tqp} ) \ln (1 - b^{\era}_{tqp}).
\label{eLogLoss}
\end{equation}
It is possible for all ensemble members to predict the wrong result (especially for small ensembles), leading to the $\logloss^b = -\infty$ due to $\ln(0)$.
In practice, a small number $\epsilon=10^{-7}$ is added to both logarithms since no forecast is truly impossible \citep{Benedetti2010-jo}.
The value of the metric is thus affected by the choice of $\epsilon$, a known inconvenience of this otherwise strictly proper metric.

\subsubsection{GEFS Model Climatological Spread}
The baseline GEFS-Climatology spread in Figure~\ref{fDispersionCorr} is computed from the training dataset, which is the 20-year GEFS reforecast \citep{Guan2022-vl}. The climatological spread is computed for each lead time independently. For a fixed lead time, the pointwise spread of the 5-member ensemble is computed for each day. Then the model climatological spread for that lead time is defined as the day-of-year average over the 20 years.

Note that we slightly abuse standard notation, since this spread accounts not only for the internal variability of the model, but also for forecast uncertainty~\citep{Slingo2011}.

\subsection{Detailed and Additional Results}

Unless noted, the results reported here are obtained from $N=512$ generated ensembles from our models, with $K=2$ seeding forecasts and $K'=3$ for \ourgpp. 

\subsubsection{RMSE, ACC, CRPS, Reliability for All Fields}
\label{siAllFields}

\begin{figure}[t]
    \centering
    \includegraphics[width=0.85\columnwidth,draft=false]{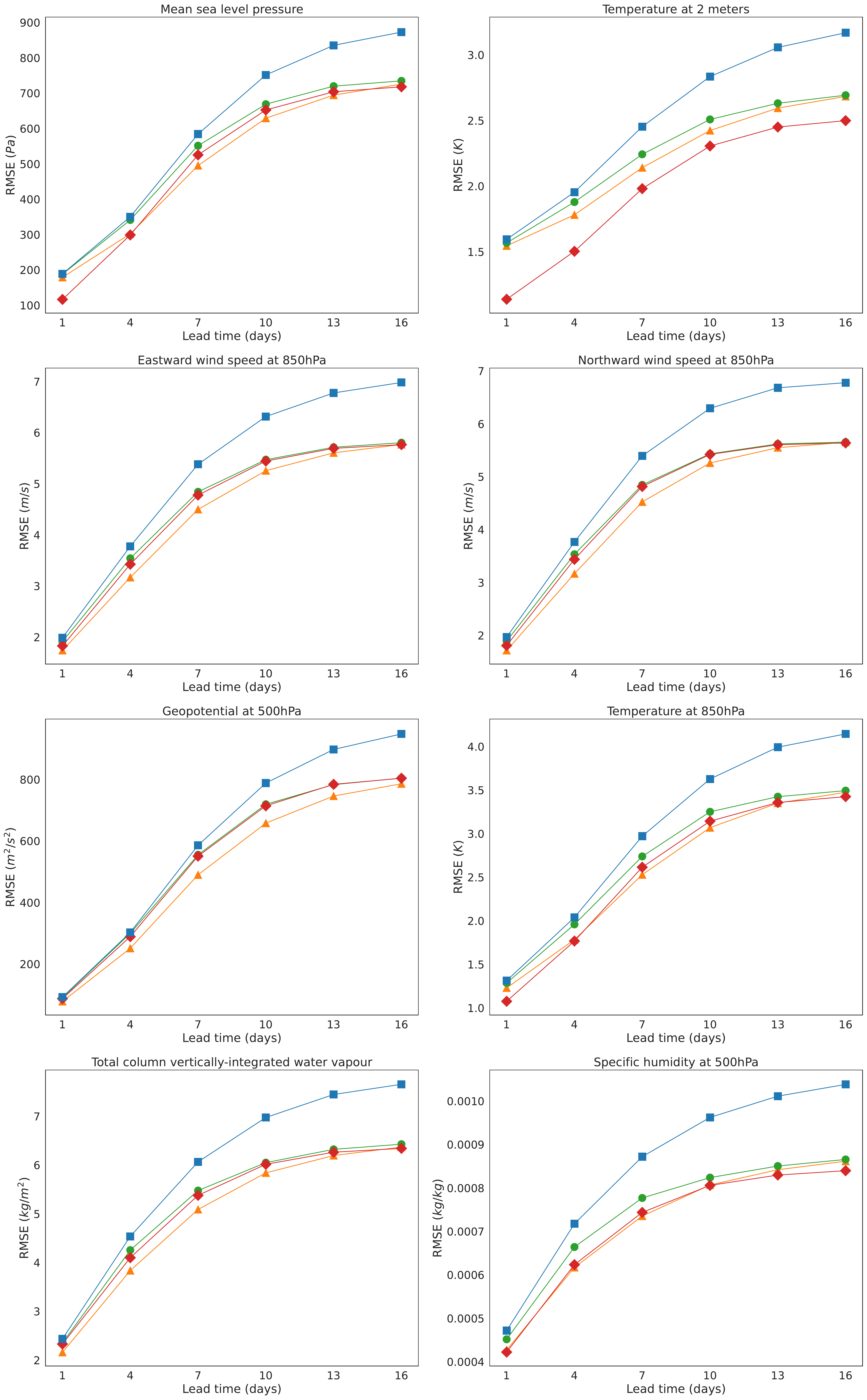}
    \includegraphics[width=0.5\columnwidth,draft=false]{figs/gee_gpp_c2/legend}  
    \caption{RMSE of the ensemble means with ERA5 HRES as the ``ground-truth'' labels.}
    \label{frmse_all}
\end{figure}

\begin{figure}[t]
    \centering
    \includegraphics[width=0.85\columnwidth,draft=false]{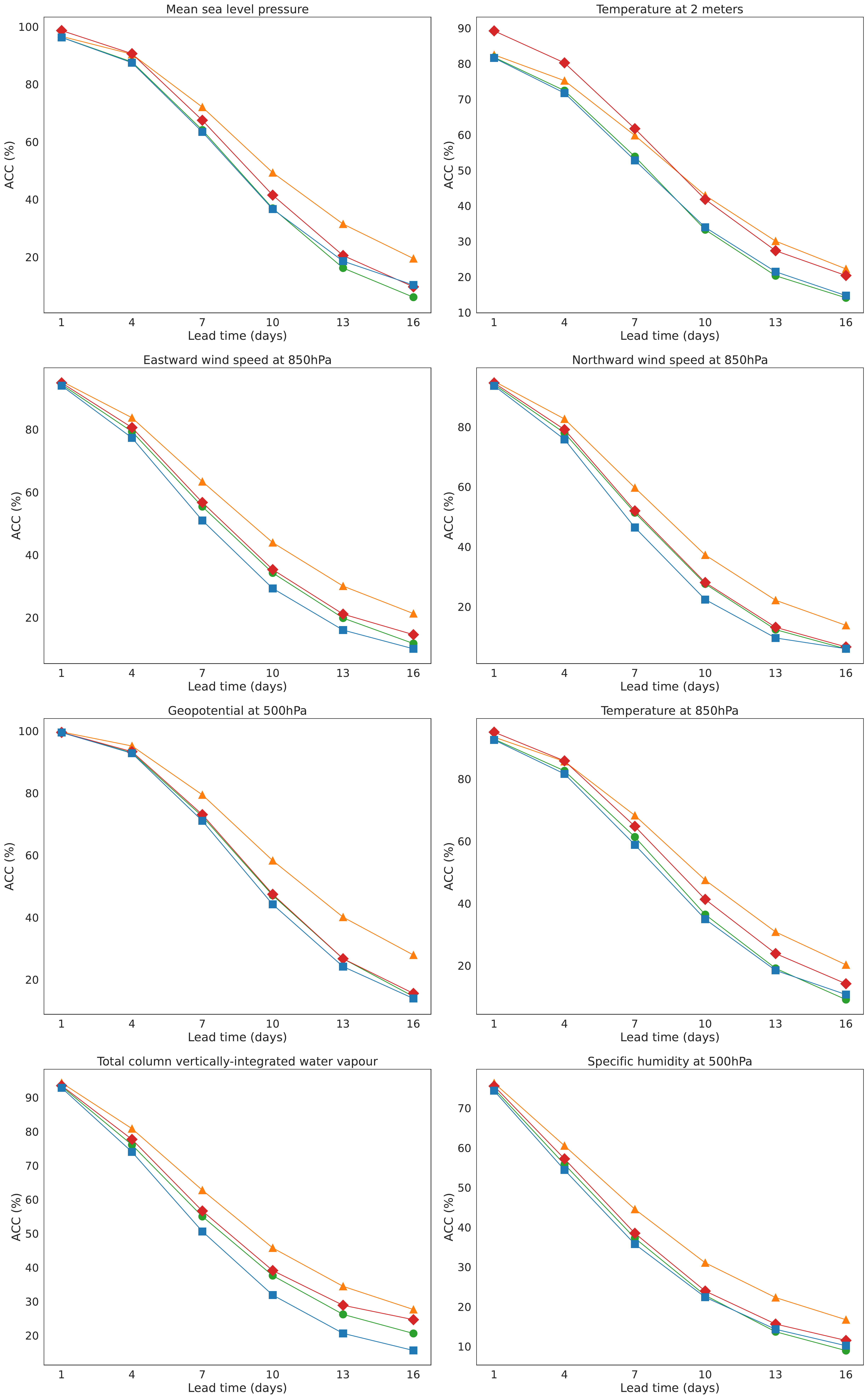}
    \includegraphics[width=0.5\columnwidth,draft=false]{figs/gee_gpp_c2/legend}  
    \caption{ACC of the ensemble means with ERA5 HRES as ``ground-truth'' labels.}
    \label{facc_all}
\end{figure}

\begin{figure}[t]
    \centering
    \includegraphics[width=0.85\columnwidth,draft=false]{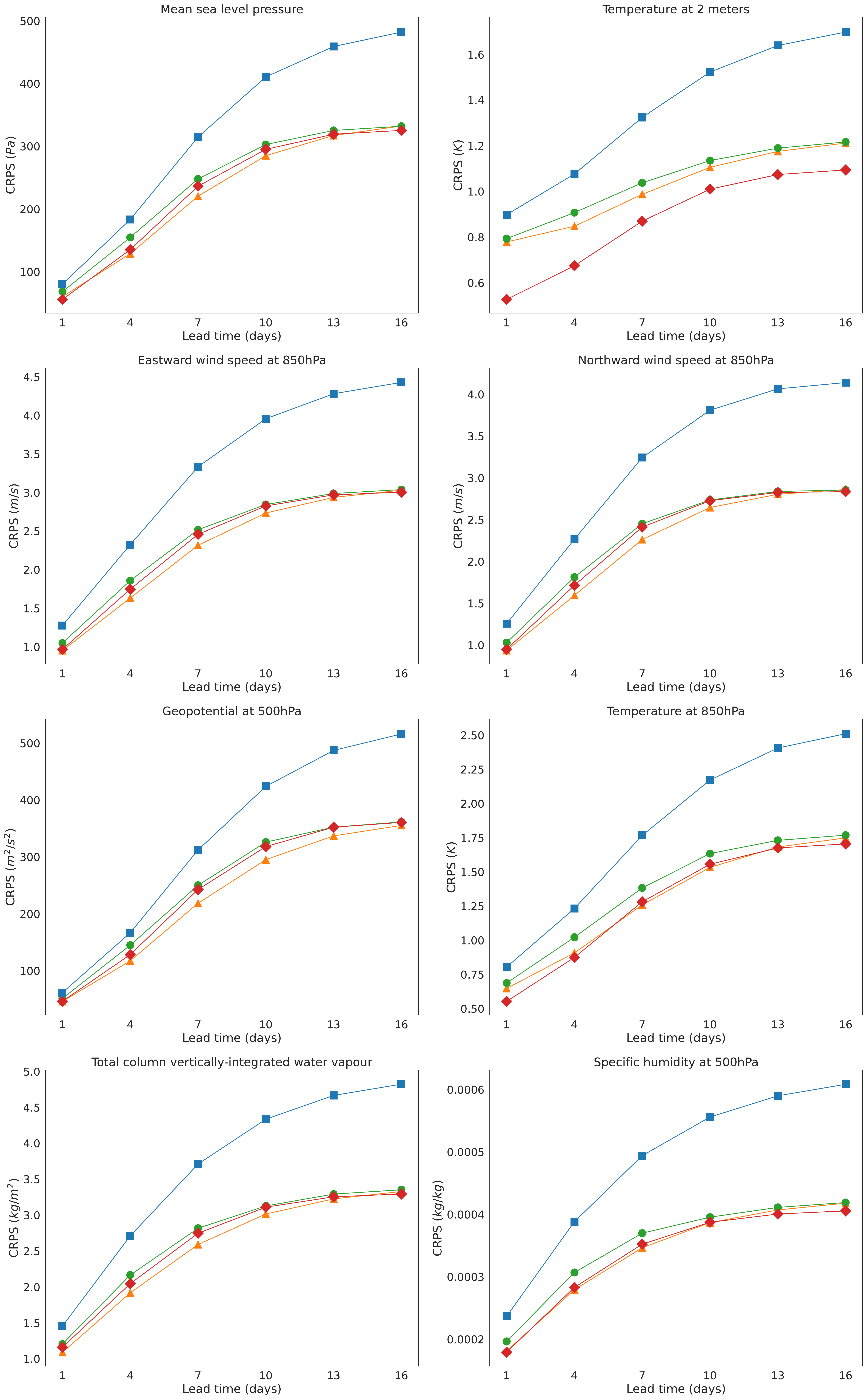}
    \includegraphics[width=0.5\columnwidth,draft=false]{figs/gee_gpp_c2/legend}  
    \caption{CRPS with ERA5 HRES  as the ``ground-truth'' labels.}
    \label{fcrps_all}
\end{figure}

\begin{figure}[t]
    \centering
    \includegraphics[width=0.85\columnwidth,draft=false]{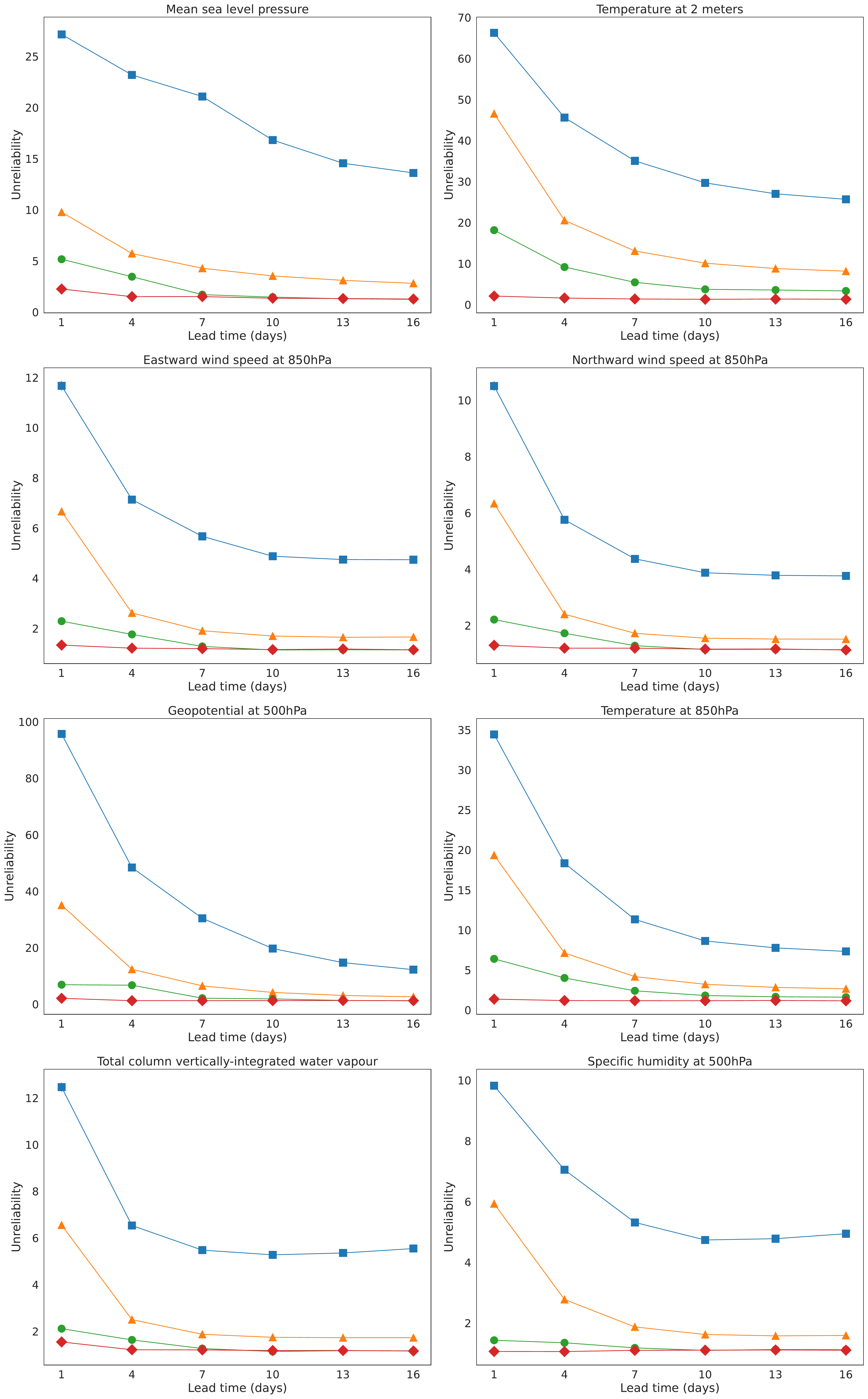}  
    \includegraphics[width=0.5\columnwidth,draft=false]{figs/gee_gpp_c2/legend}  
    \caption{Unreliability $\delta$ with ERA5 as the label.}
    \label{fdelta_all}
\end{figure}

We report values for \rmse (Figure~\ref{frmse_all}), \acc (Figure~\ref{facc_all}), \crps (Figure~\ref{fcrps_all}), and $\delta$ (Figure~\ref{fdelta_all}) for all 8 modeled fields in Table \ref{tb:fields}. The results  are consistent with those reported in the main text: \ourmethodshort ensembles attain similar or better skill than the physics-based ensembles, with \ourgpp performing better than \ourgee for variables that are biased in \gefsfull. In addition, \ourgpp is in general more reliable than \gefsfull and \ourgee, particularly within the first forecast week.

\subsubsection{Brier score of all fields at various thresholds}
\label{siExtremes}

\begin{figure}[t]
    \centering
    \includegraphics[width=0.85\columnwidth,draft=false]{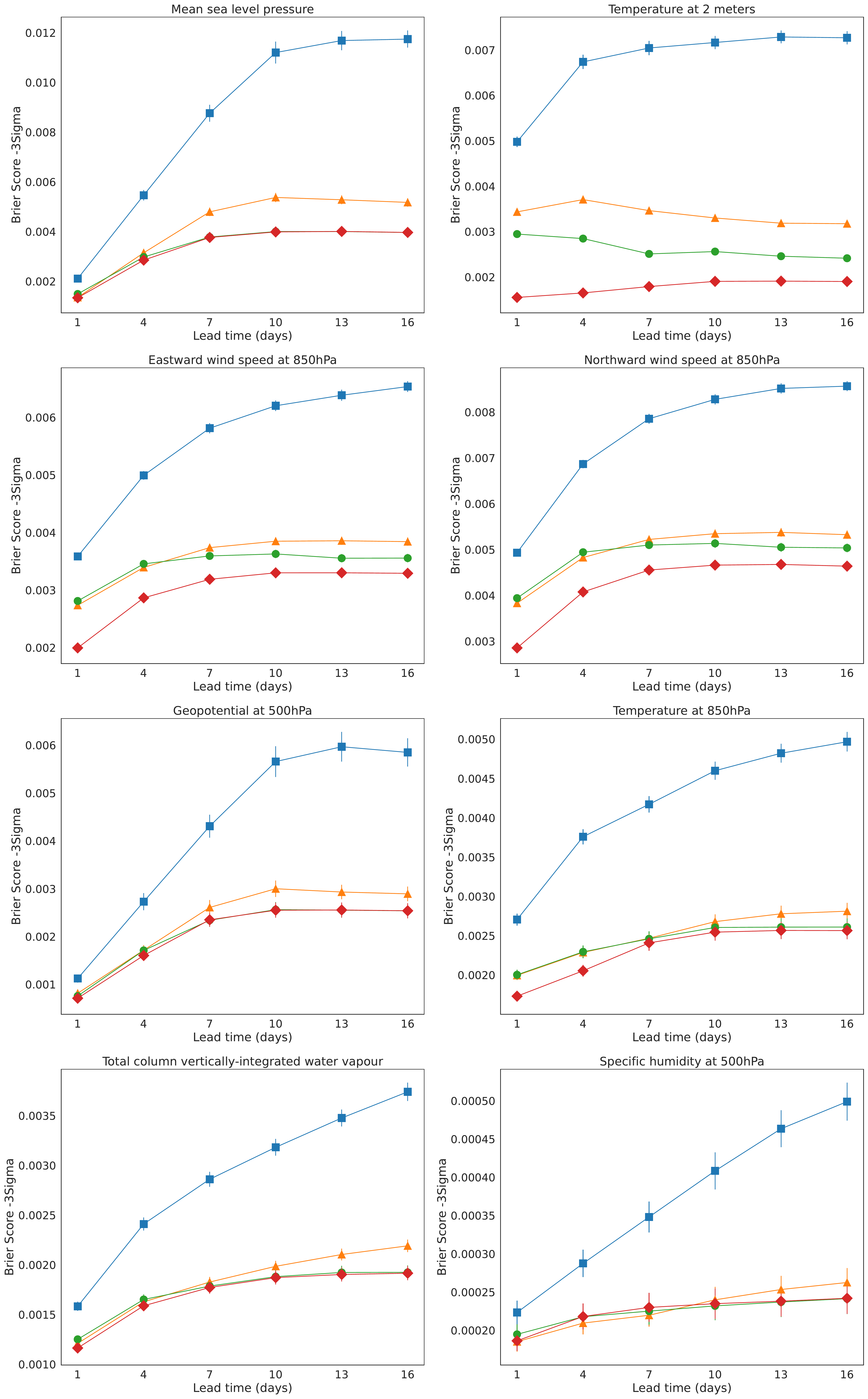}
    \includegraphics[width=0.5\columnwidth,draft=false]{figs/gee_gpp_c2/legend}  
    \caption{Brier score for $-3\sigma$ with ERA5 HRES as the ``ground-truth'' label.}
    \label{fbrier_m3}
\end{figure}

\begin{figure}[t]
    \centering
    \includegraphics[width=0.85\columnwidth,draft=false]{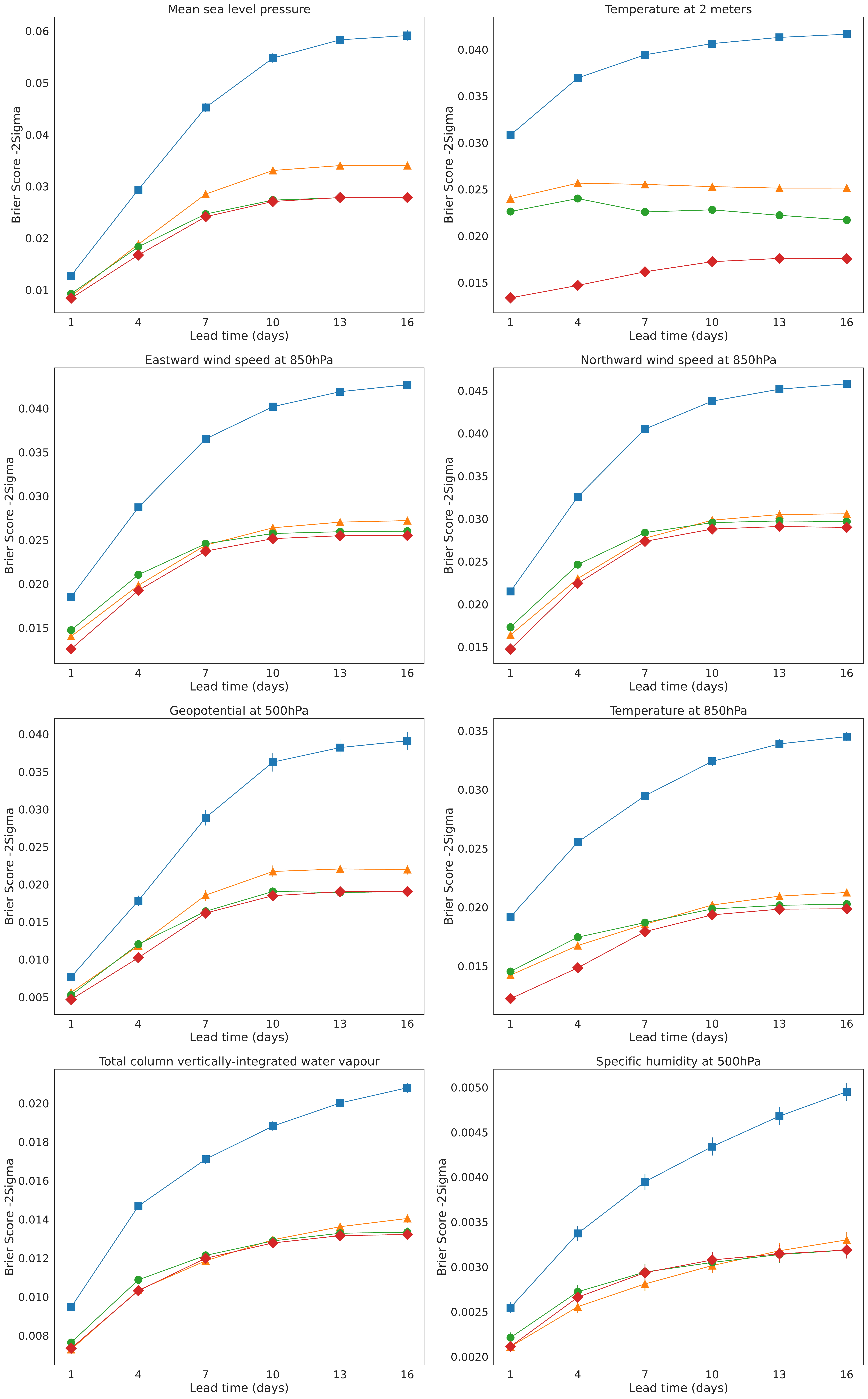}
    \includegraphics[width=0.5\columnwidth,draft=false]{figs/gee_gpp_c2/legend}  
    \caption{Brier score for $-2\sigma$ with ERA5 HRES as the  ``ground-truth'' label.}
    \label{fbrier_m2}
\end{figure}

\begin{figure}[t]
    \centering
    \includegraphics[width=0.85\columnwidth,draft=false]{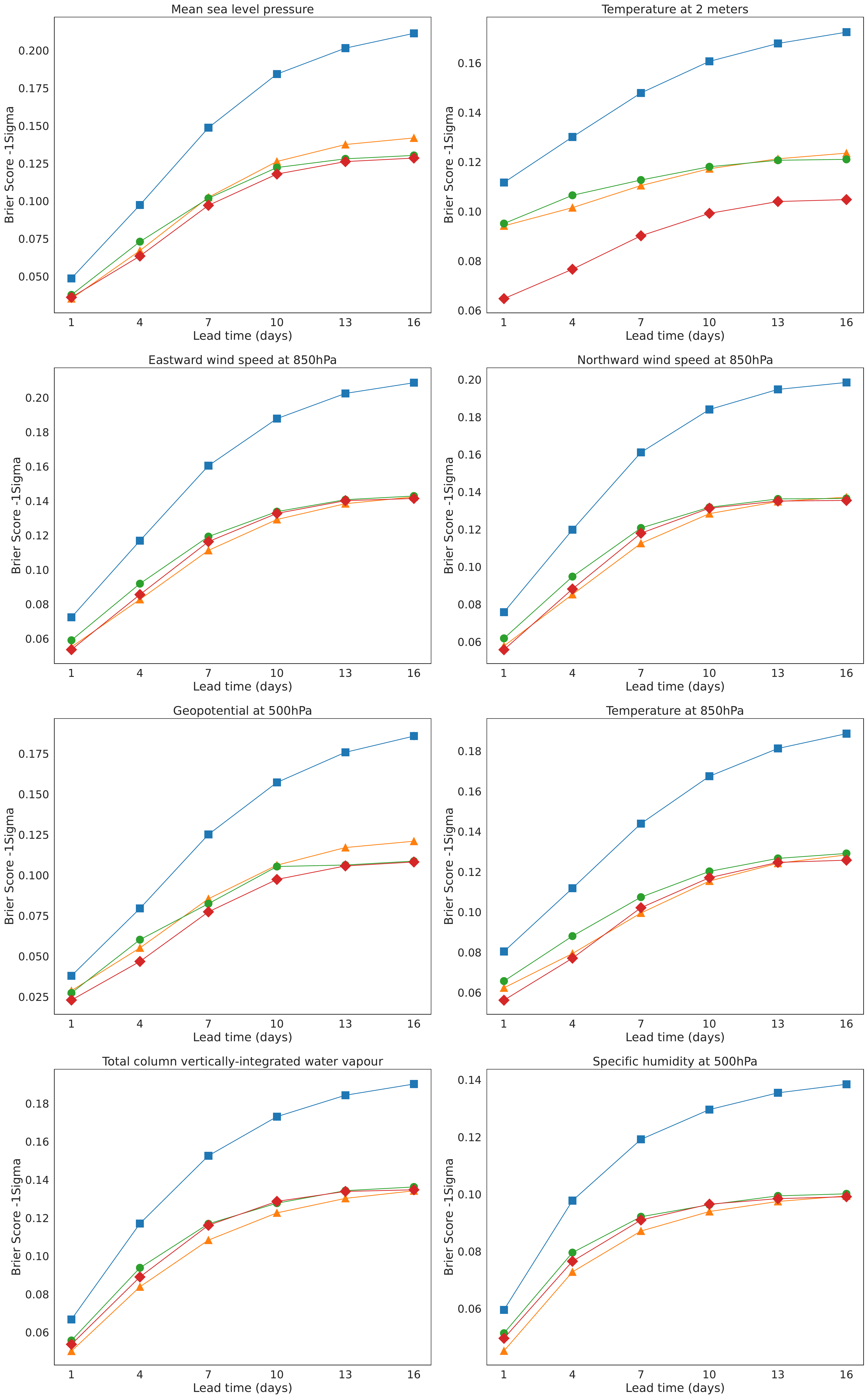}
    \includegraphics[width=0.5\columnwidth,draft=false]{figs/gee_gpp_c2/legend}  
    \caption{Brier score for $-1\sigma$ with ERA5 HRES as the  ``ground-truth'' label.}
    \label{fbrier_m1}
\end{figure}

\begin{figure}[t]
    \centering
    \includegraphics[width=0.85\columnwidth,draft=false]{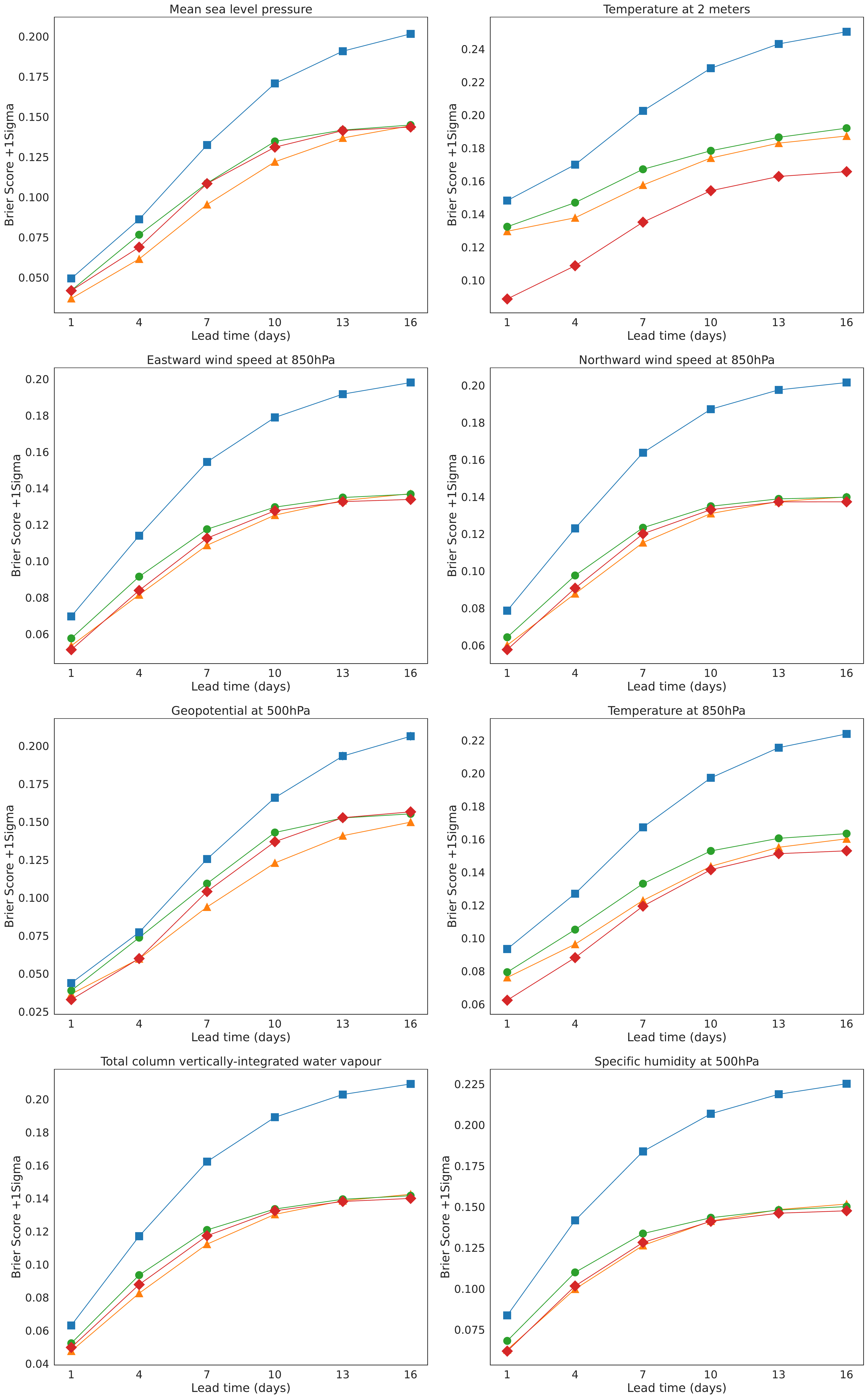}
    \includegraphics[width=0.5\columnwidth,draft=false]{figs/gee_gpp_c2/legend}  
    \caption{Brier score for $+1\sigma$ with ERA5 HRES as the  ``ground-truth'' label.}
    \label{fbrier_p1}
\end{figure}

\begin{figure}[t]
    \centering
    \includegraphics[width=0.85\columnwidth,draft=false]{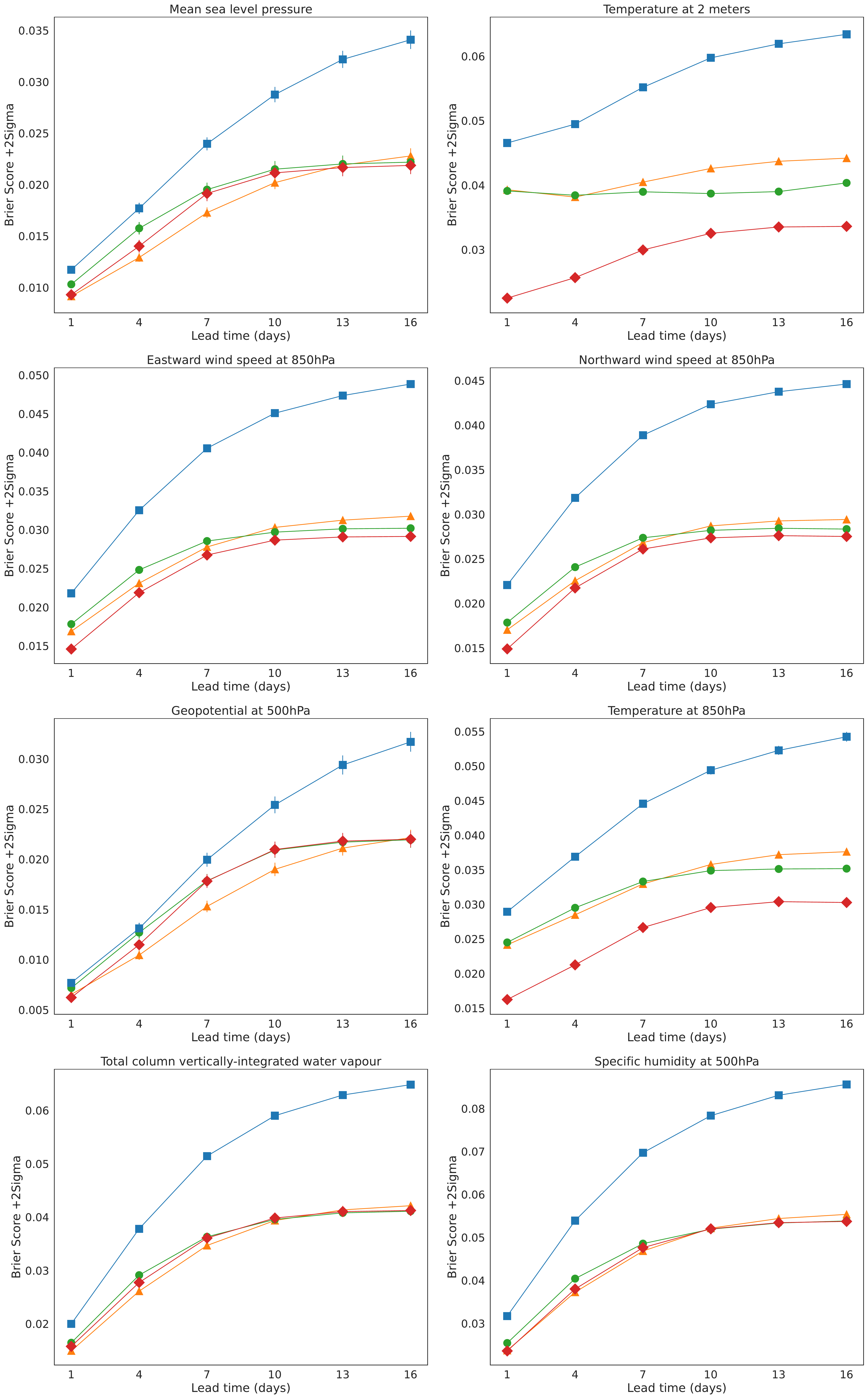}
    \includegraphics[width=0.5\columnwidth,draft=false]{figs/gee_gpp_c2/legend}  
    \caption{Brier score for $+2\sigma$ with ERA5 HRES as the  ``ground-truth'' label.}
    \label{fbrier_p2}
\end{figure}

\begin{figure}[t]
    \centering
    \includegraphics[width=0.85\columnwidth,draft=false]{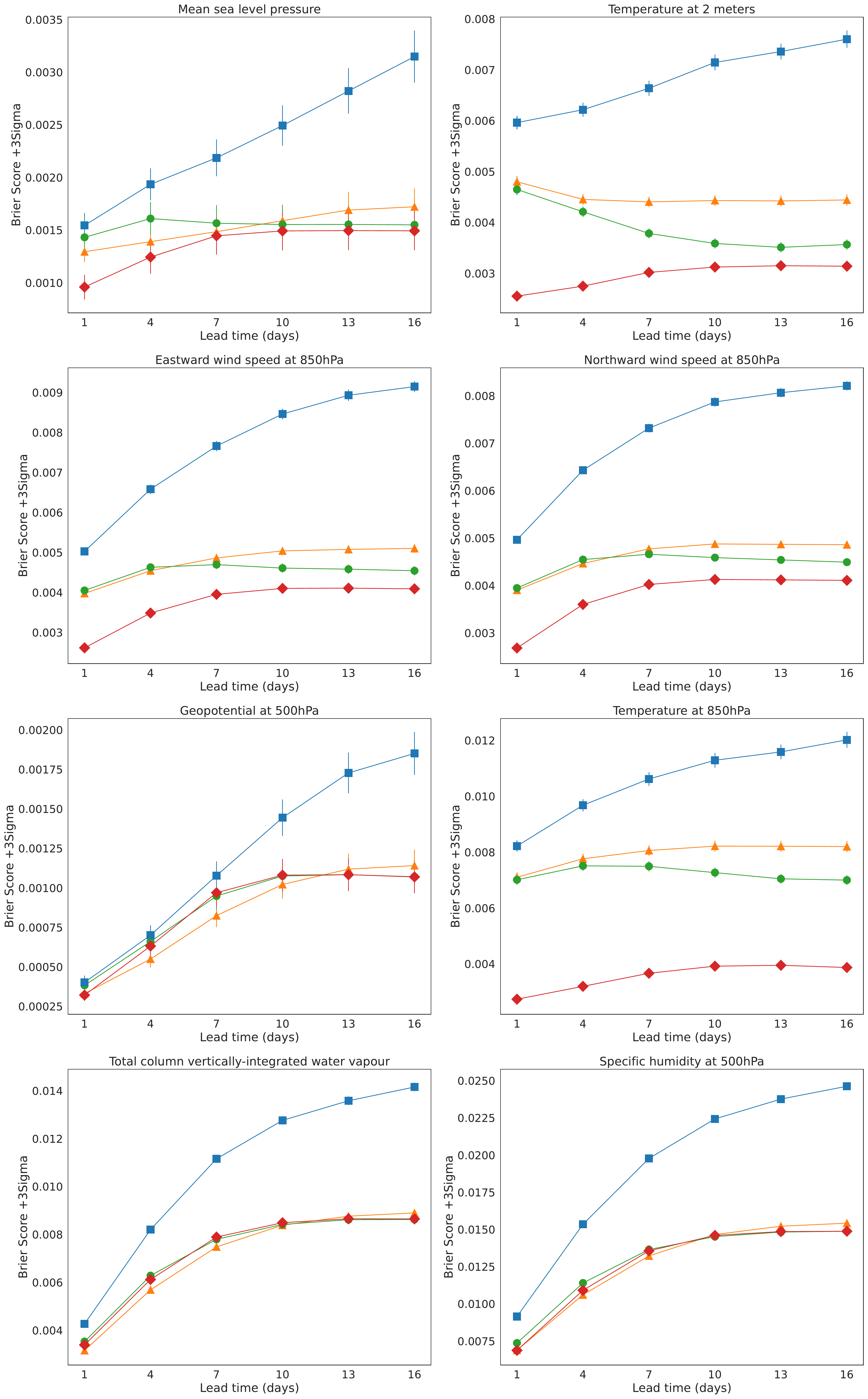}
    \includegraphics[width=0.5\columnwidth,draft=false]{figs/gee_gpp_c2/legend}  
    \caption{Brier score for $+3\sigma$ with ERA5 HRES as the  ``ground-truth'' label.}
    \label{fbrier_p3}
\end{figure}

From  Figure~\ref{fbrier_m3} to Figure~\ref{fbrier_p3}, we report the Brier score at different thresholds: $\pm 3\sigma,\pm 2\sigma, \text{and} \pm 1\sigma$, respectively.  For all thresholds, \ourgpp provides the most accurate extreme forecast ensembles among all others, for most lead times and variables. Moreover, the generative ensembles from \ourgpp and \ourgee perform particularly well for longer lead times and more extreme prediction tasks. For the most extreme thresholds considered ($\pm 3\sigma$), the generative ensembles in general outperform \gefsfull.


\subsection{Effect of $N$, $K$ and $K'$}
\label{sEnsembleHyperAppendix}
\subsubsection{Generative ensemble emulation with varying $N$ for 7-day lead time with $K=2$}
\begin{figure}[t]
    \centering
    \includegraphics[width=0.32\columnwidth,draft=false]{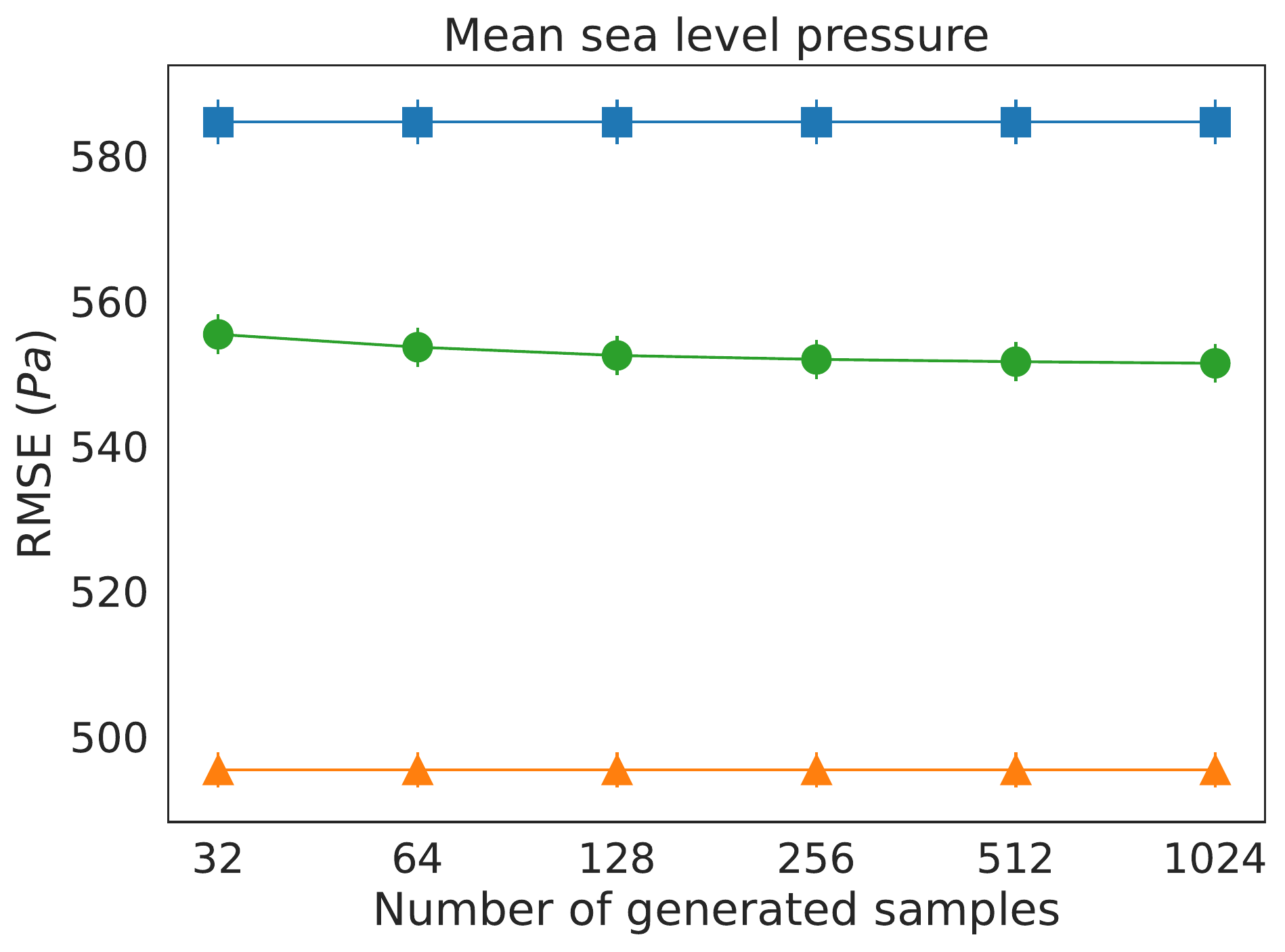}
    \includegraphics[width=0.32\columnwidth,draft=false]{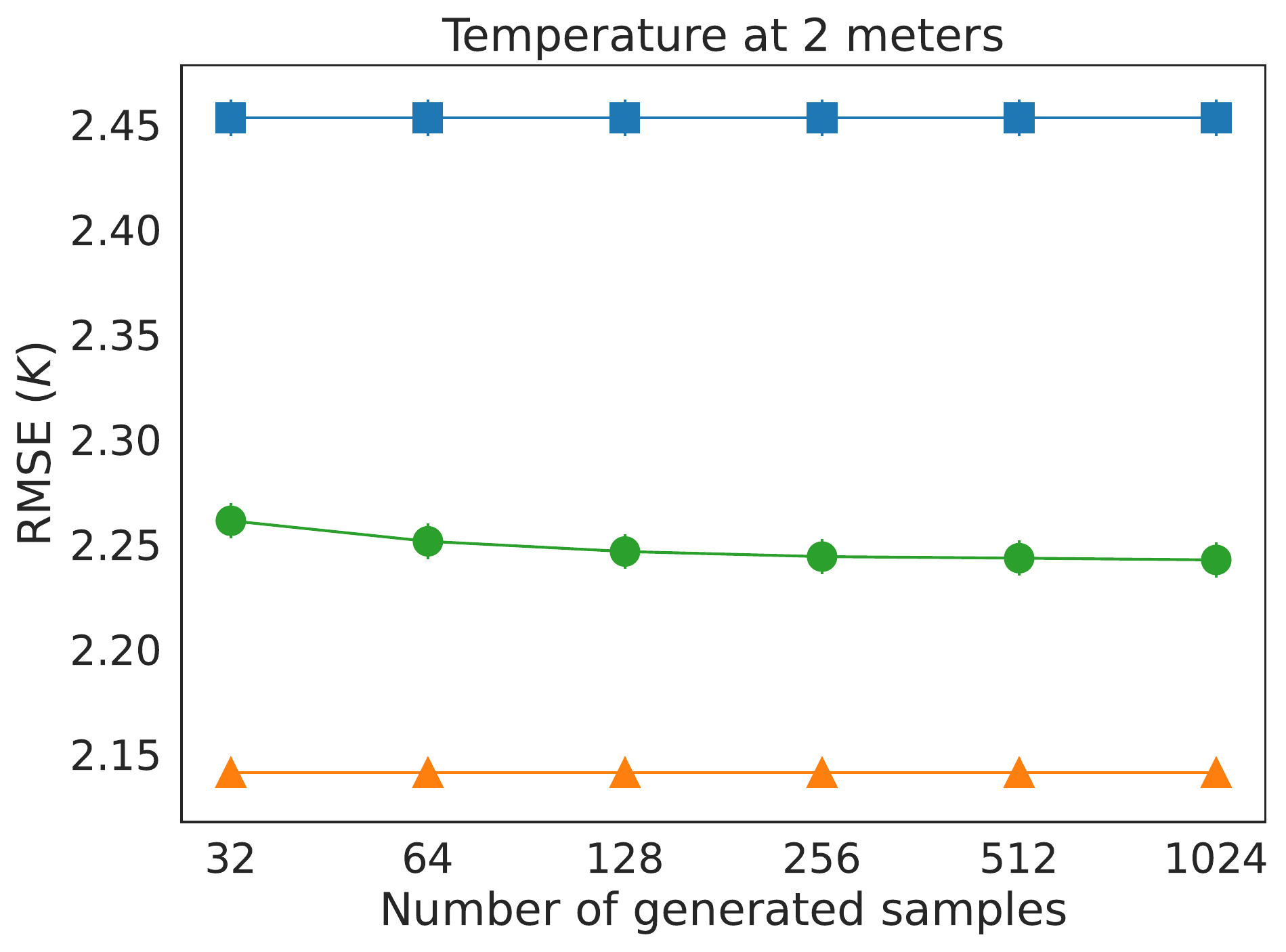}
    \includegraphics[width=0.32\columnwidth,draft=false]{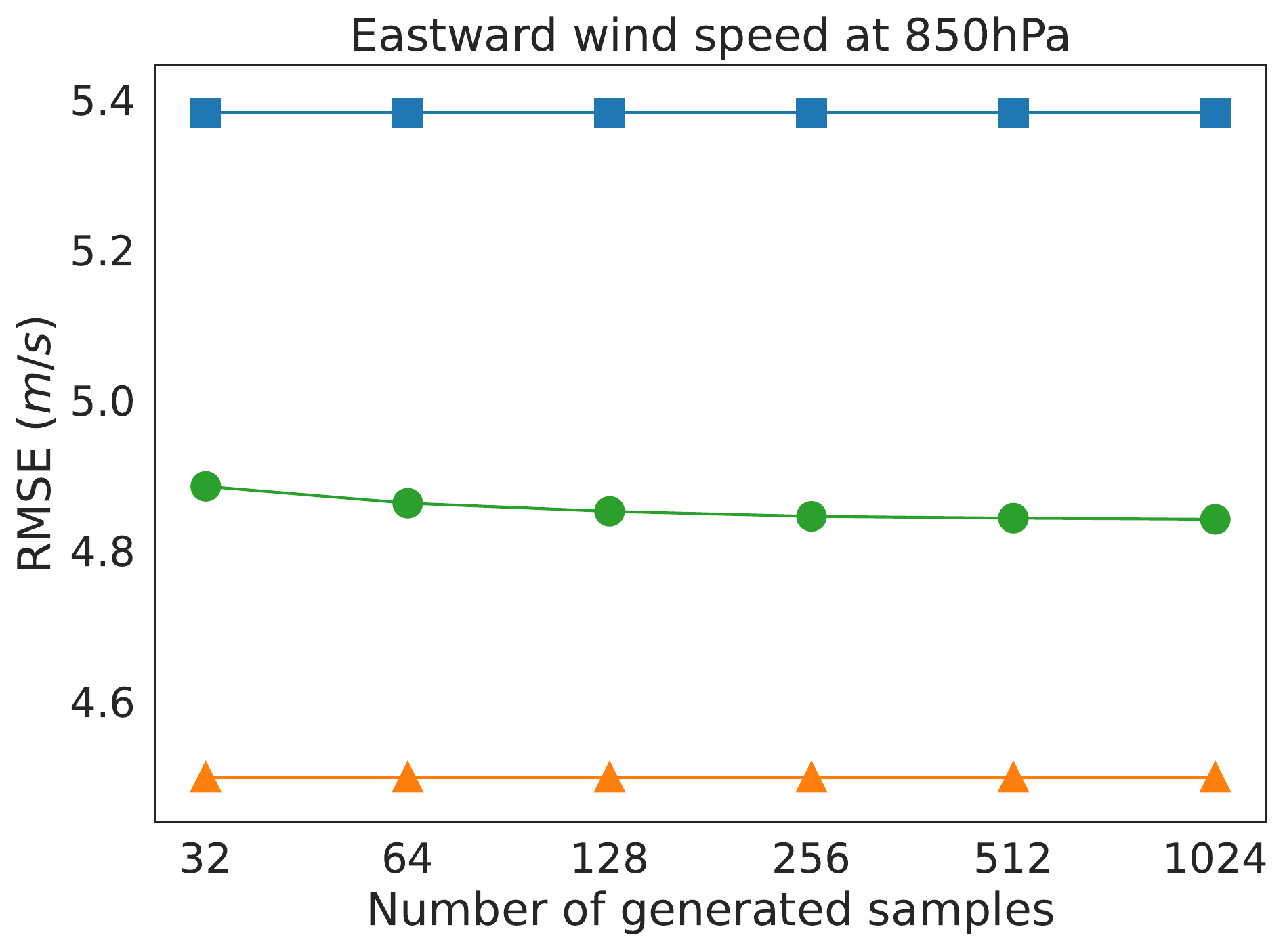}  
    \includegraphics[width=0.32\columnwidth,draft=false]{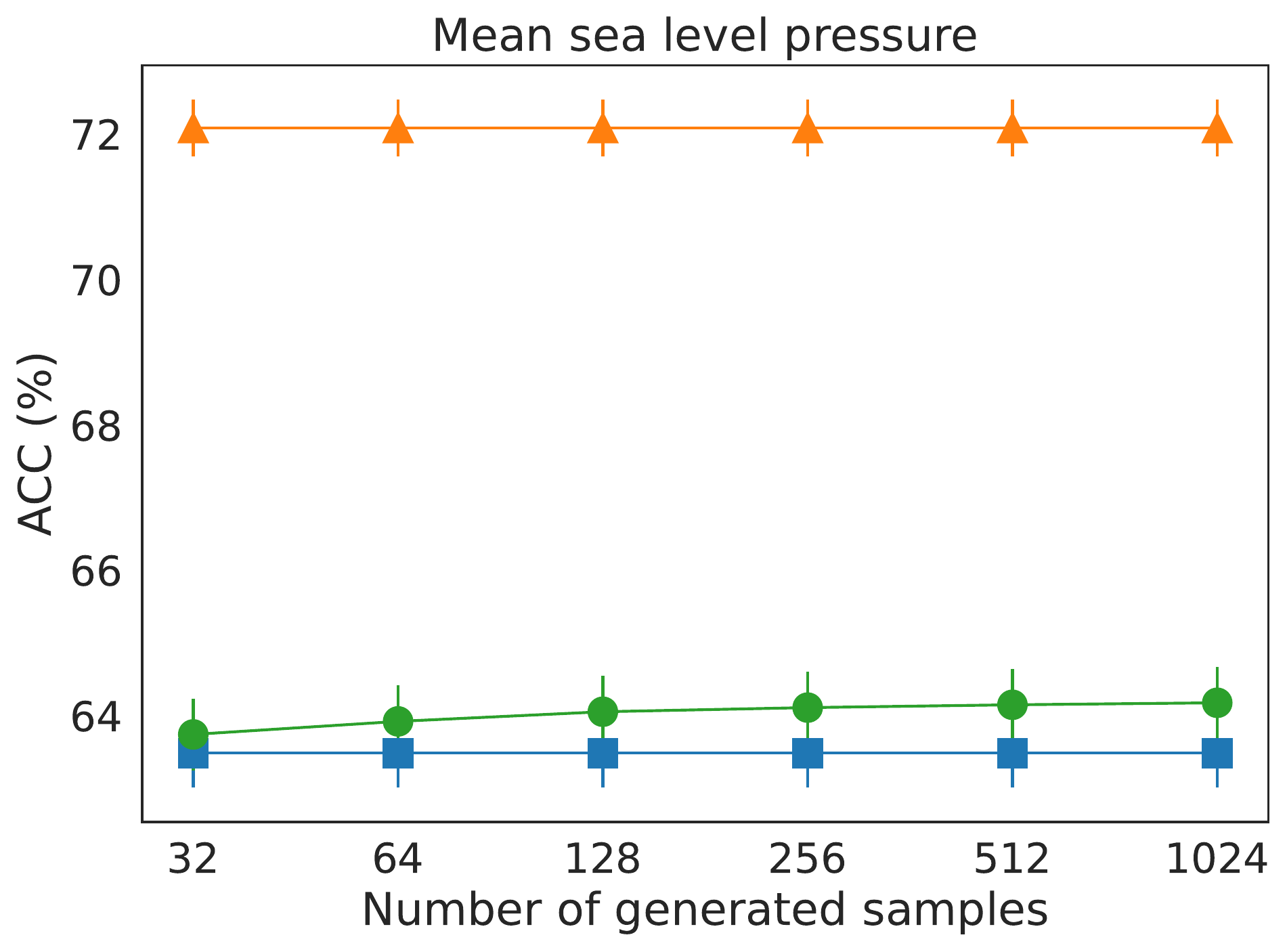}
    \includegraphics[width=0.32\columnwidth,draft=false]{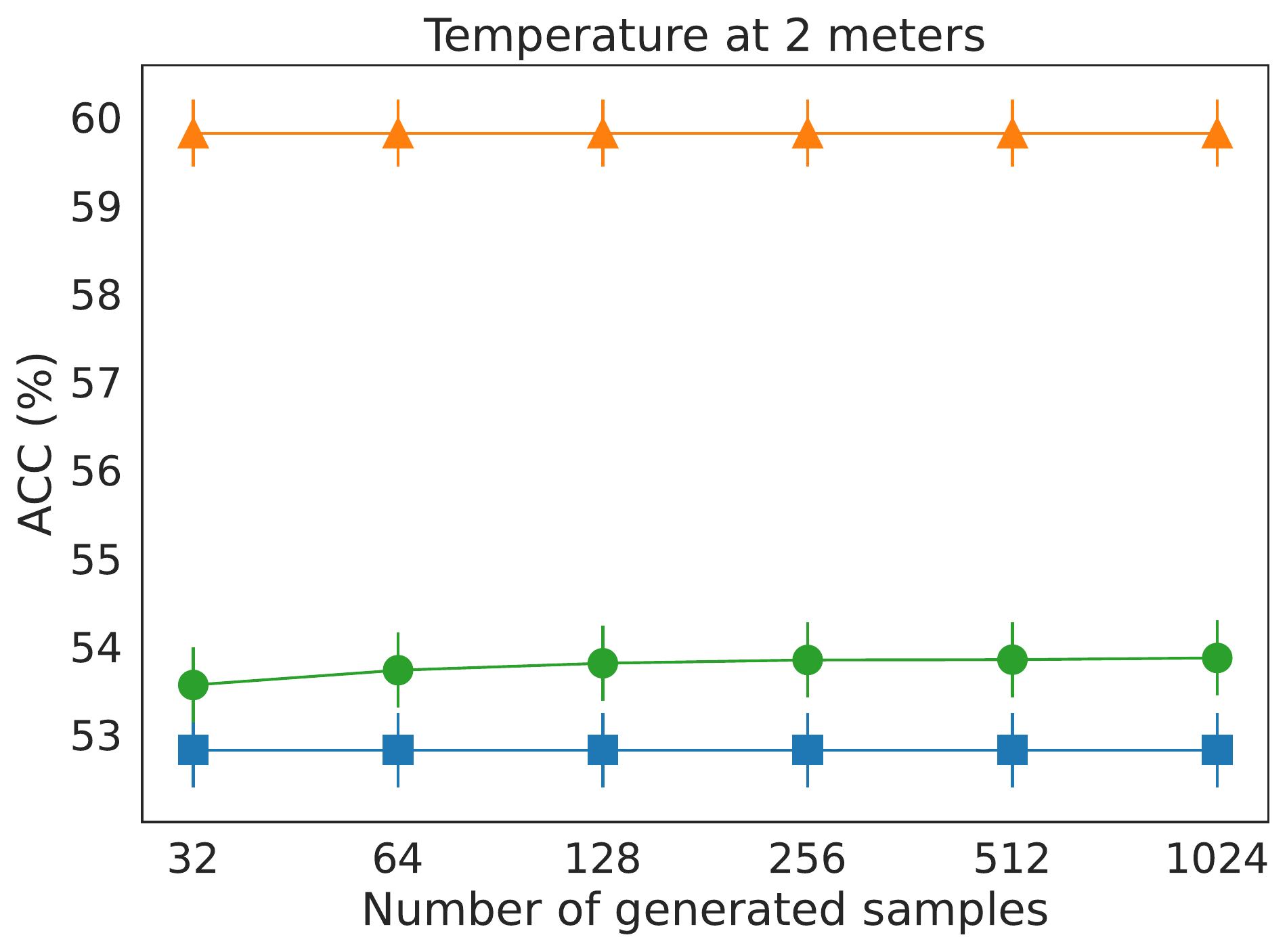}
    \includegraphics[width=0.32\columnwidth,draft=false]{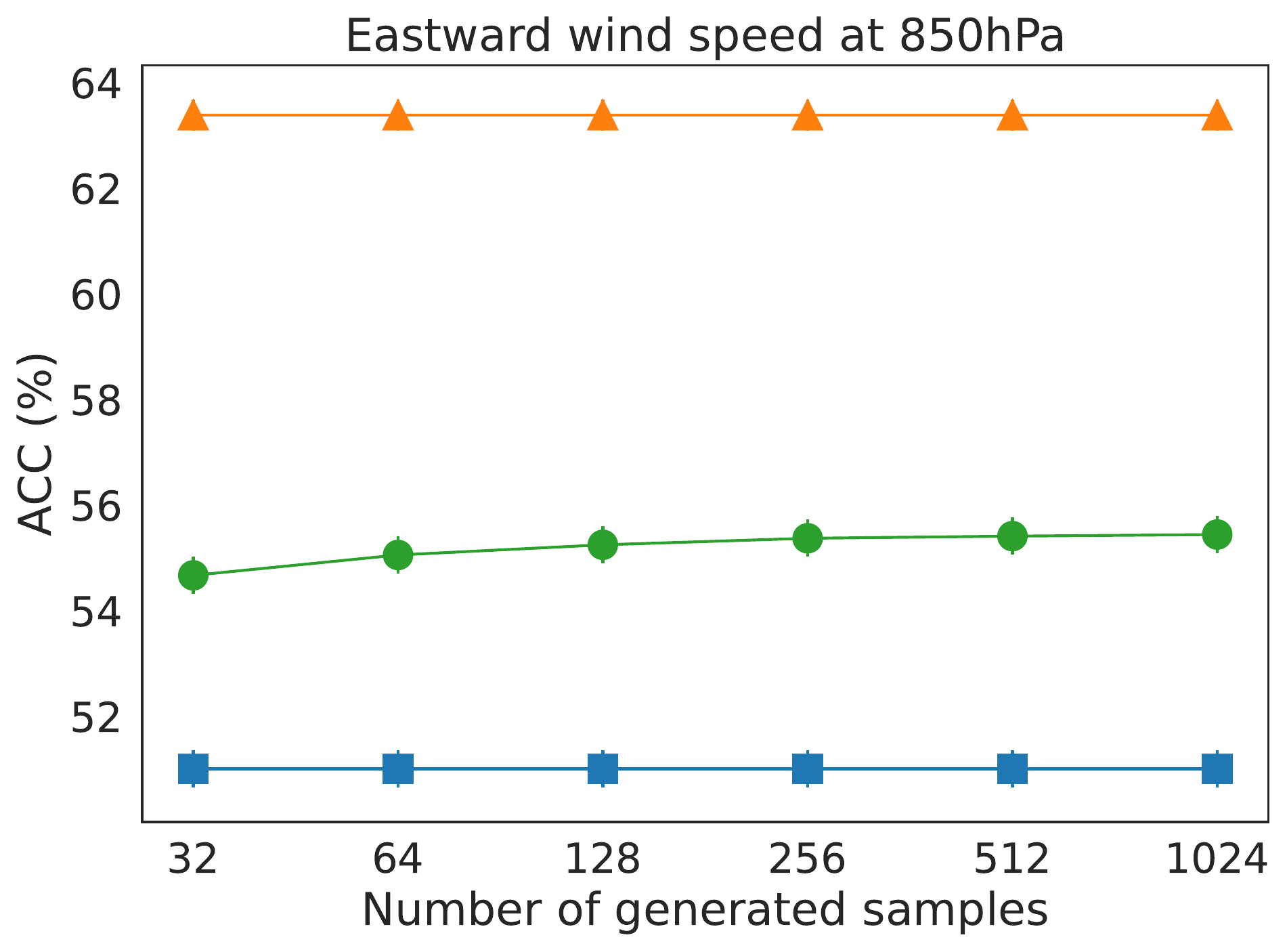}  
    \includegraphics[width=0.32\columnwidth,draft=false]{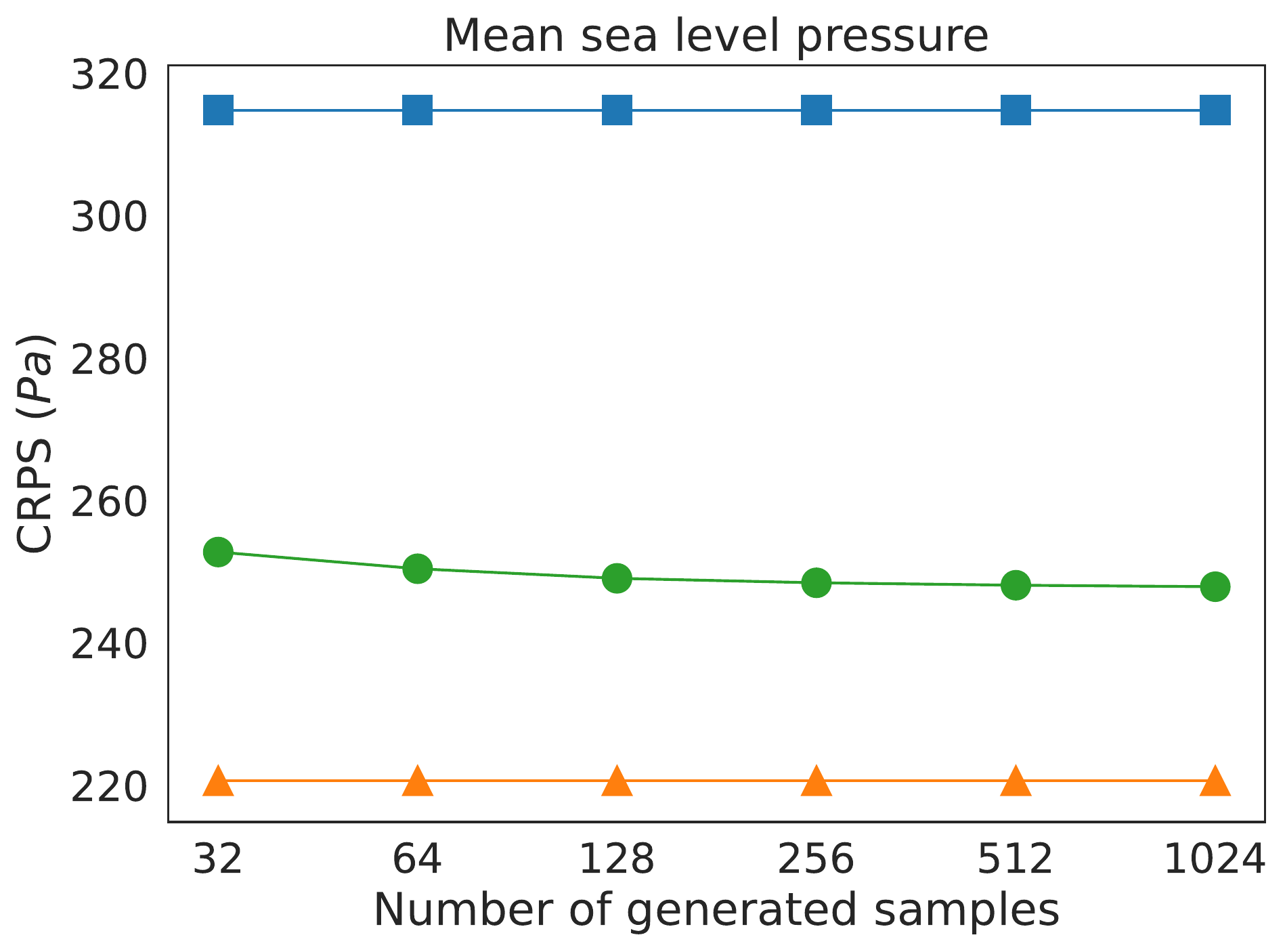}
    \includegraphics[width=0.32\columnwidth,draft=false]{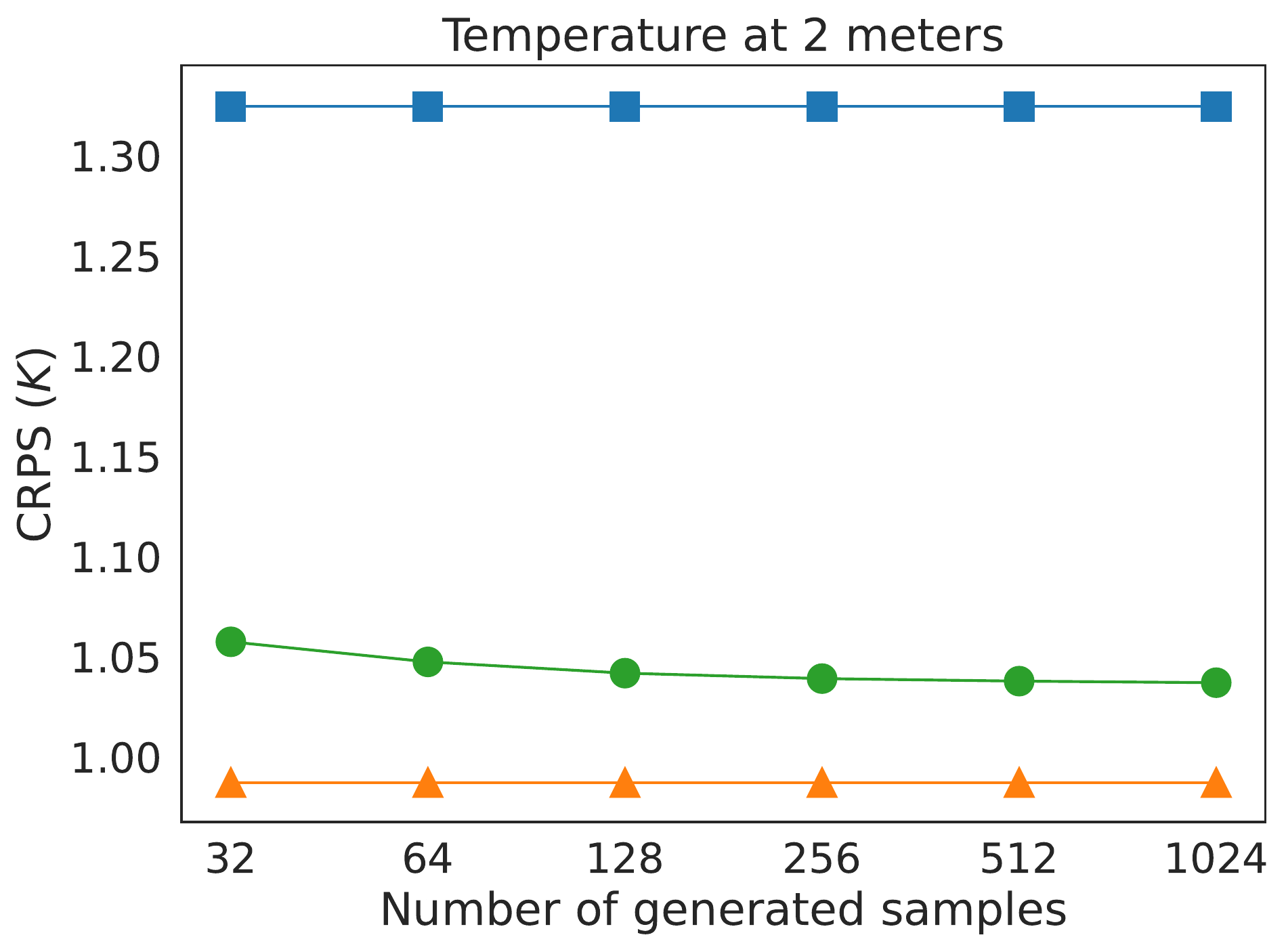}
    \includegraphics[width=0.32\columnwidth,draft=false]{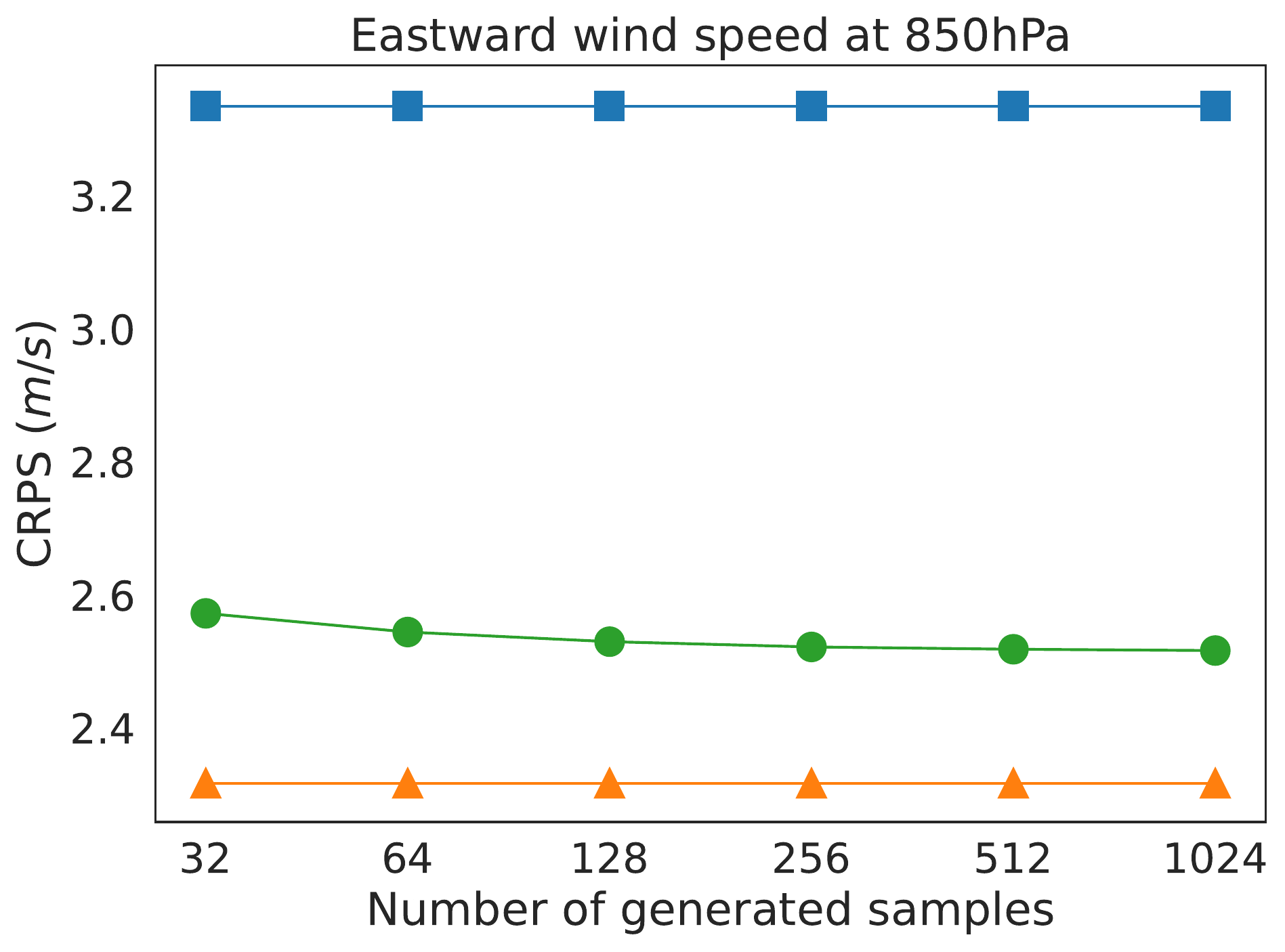}  
    \includegraphics[width=0.32\columnwidth,draft=false]{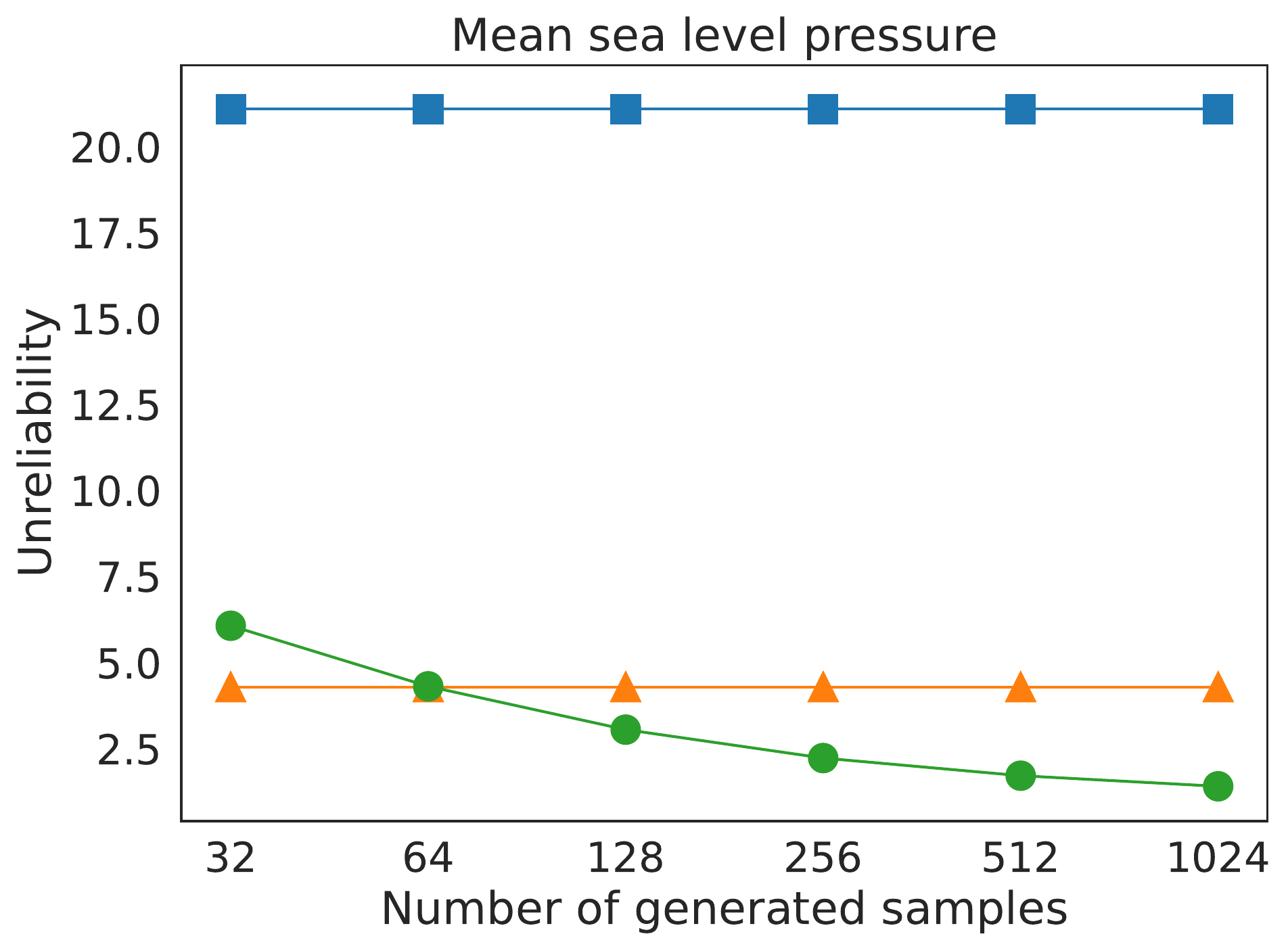}
    \includegraphics[width=0.32\columnwidth,draft=false]{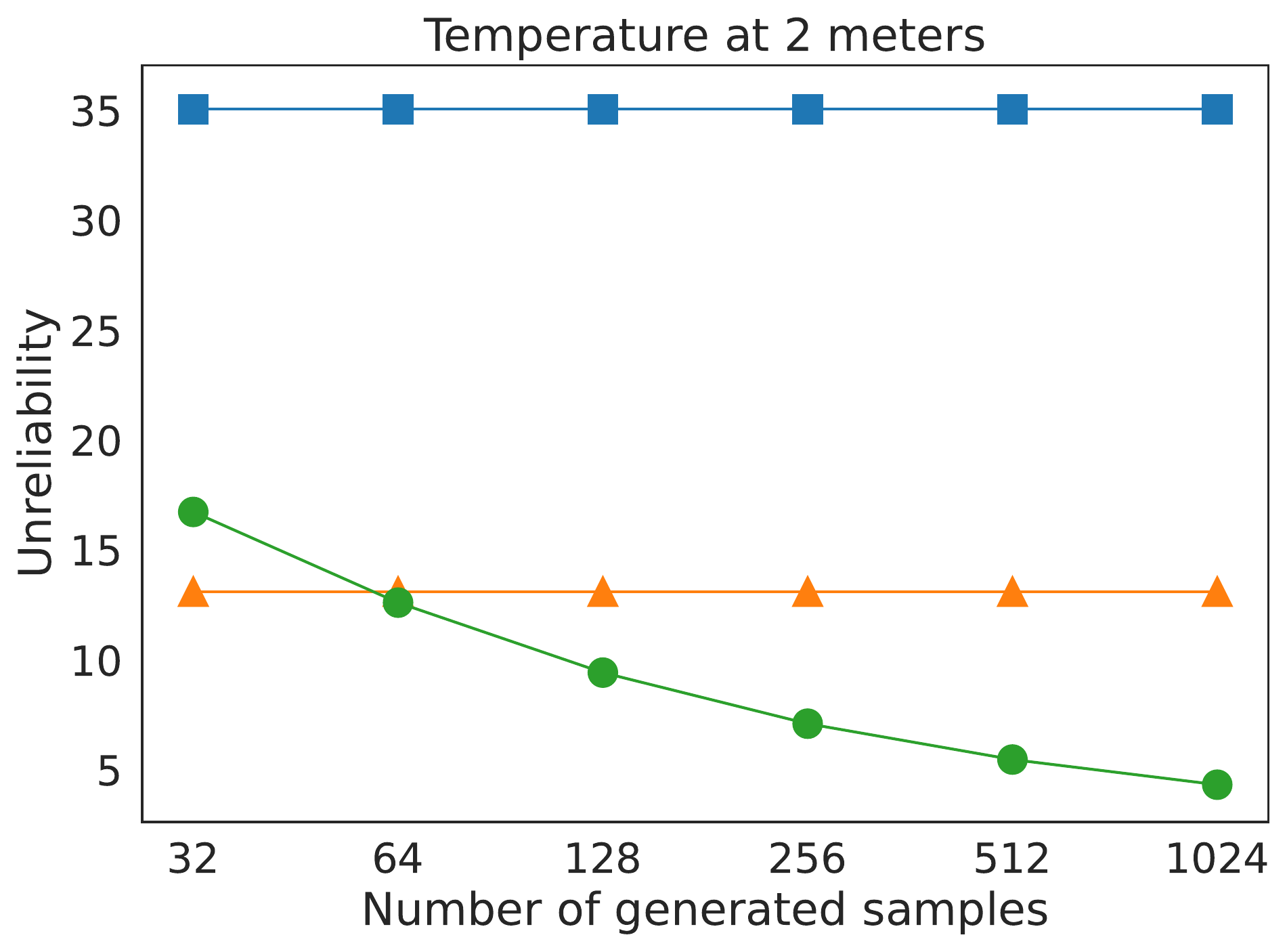}
    \includegraphics[width=0.32\columnwidth,draft=false]{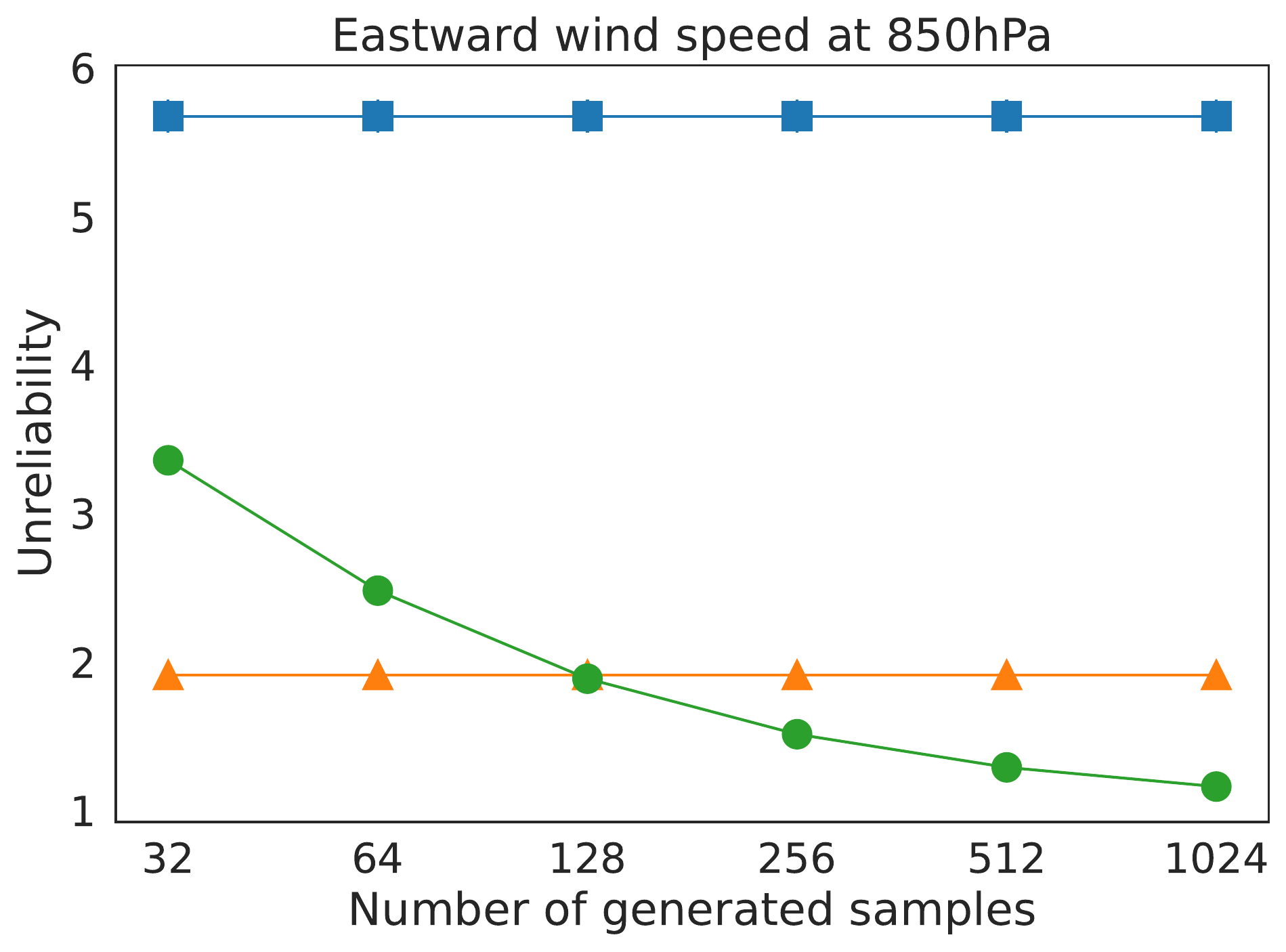}  
    \includegraphics[width=0.32\columnwidth,draft=false]{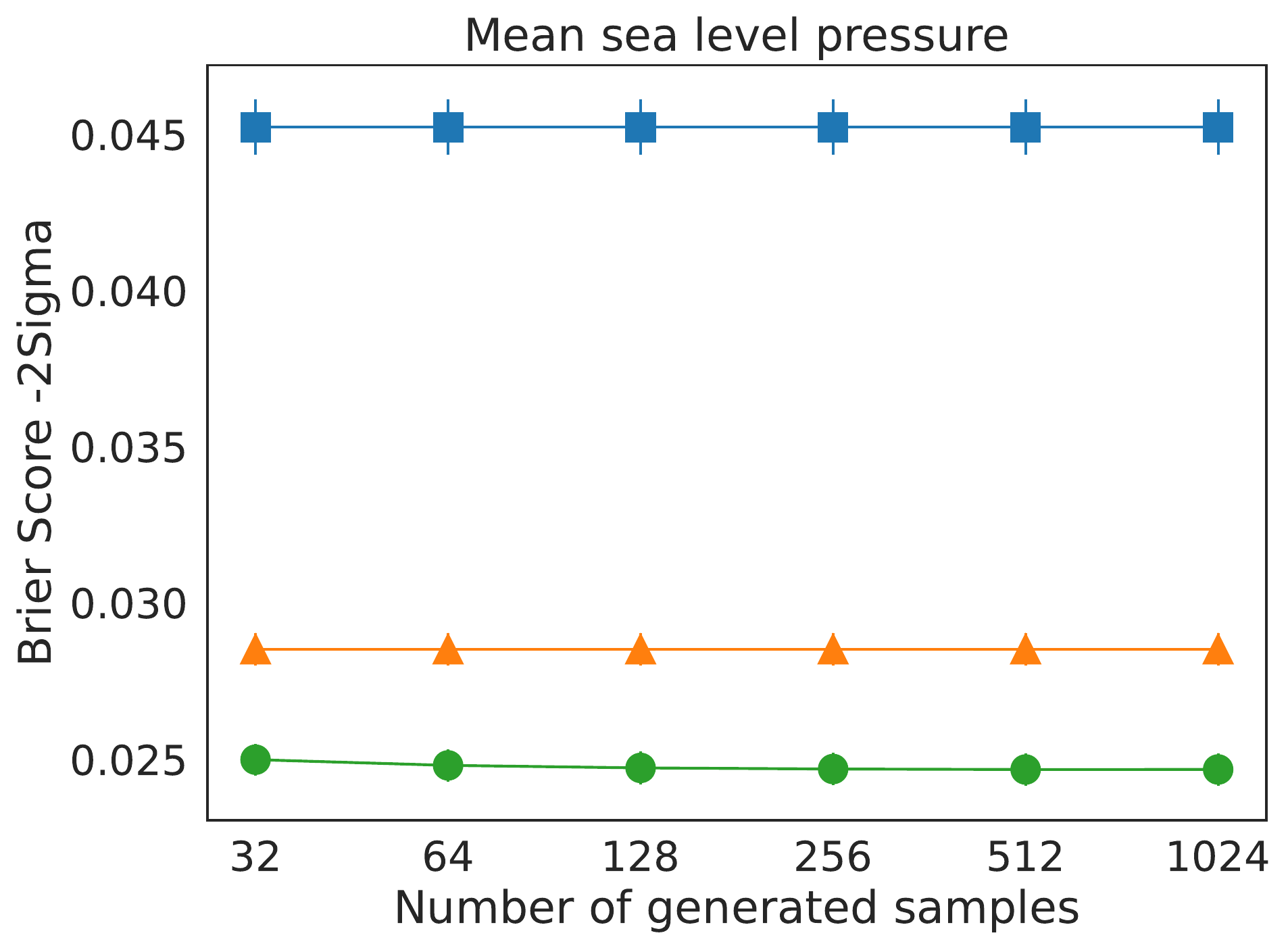}
    \includegraphics[width=0.32\columnwidth,draft=false]{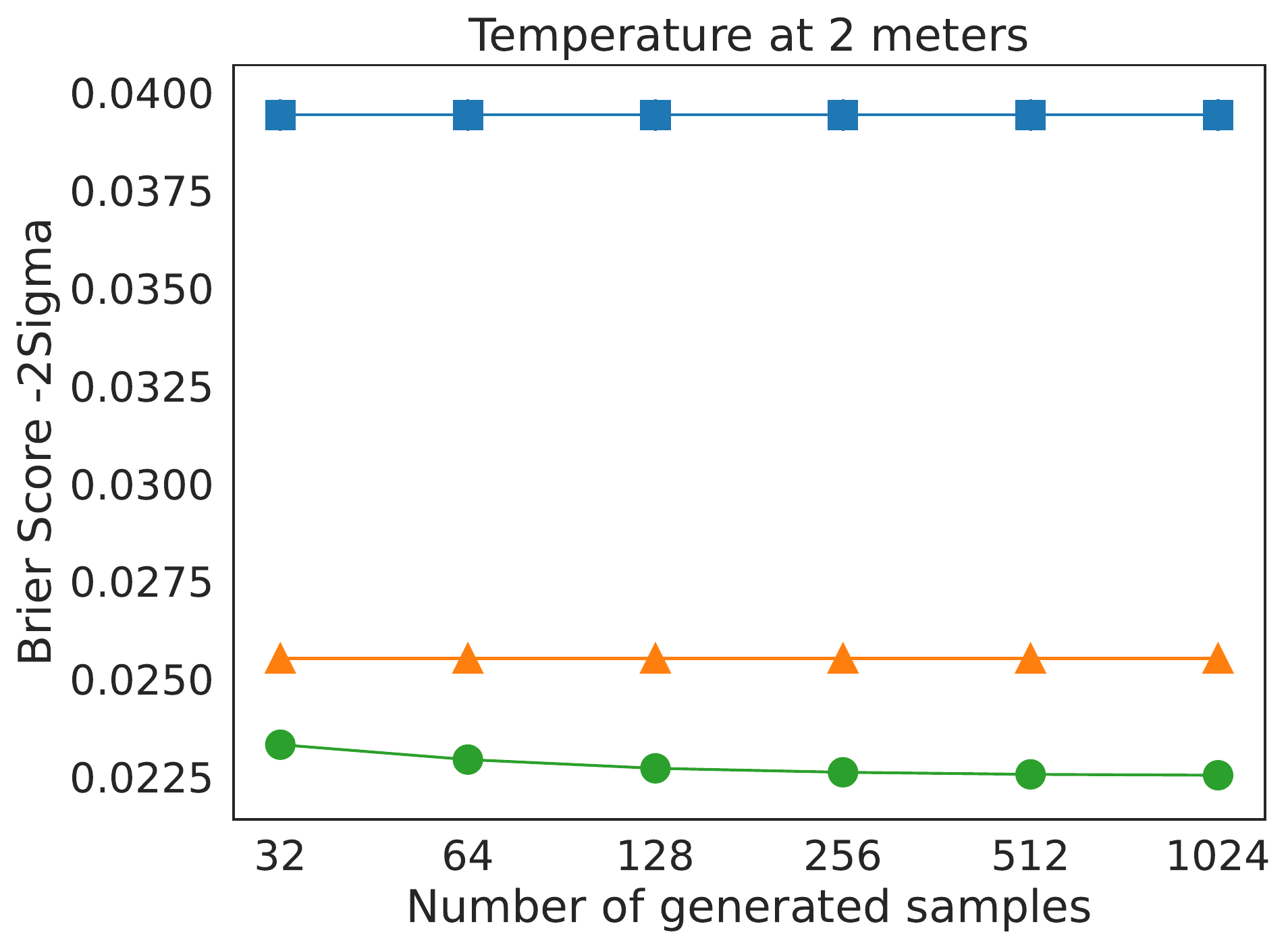}
    \includegraphics[width=0.32\columnwidth,draft=false]{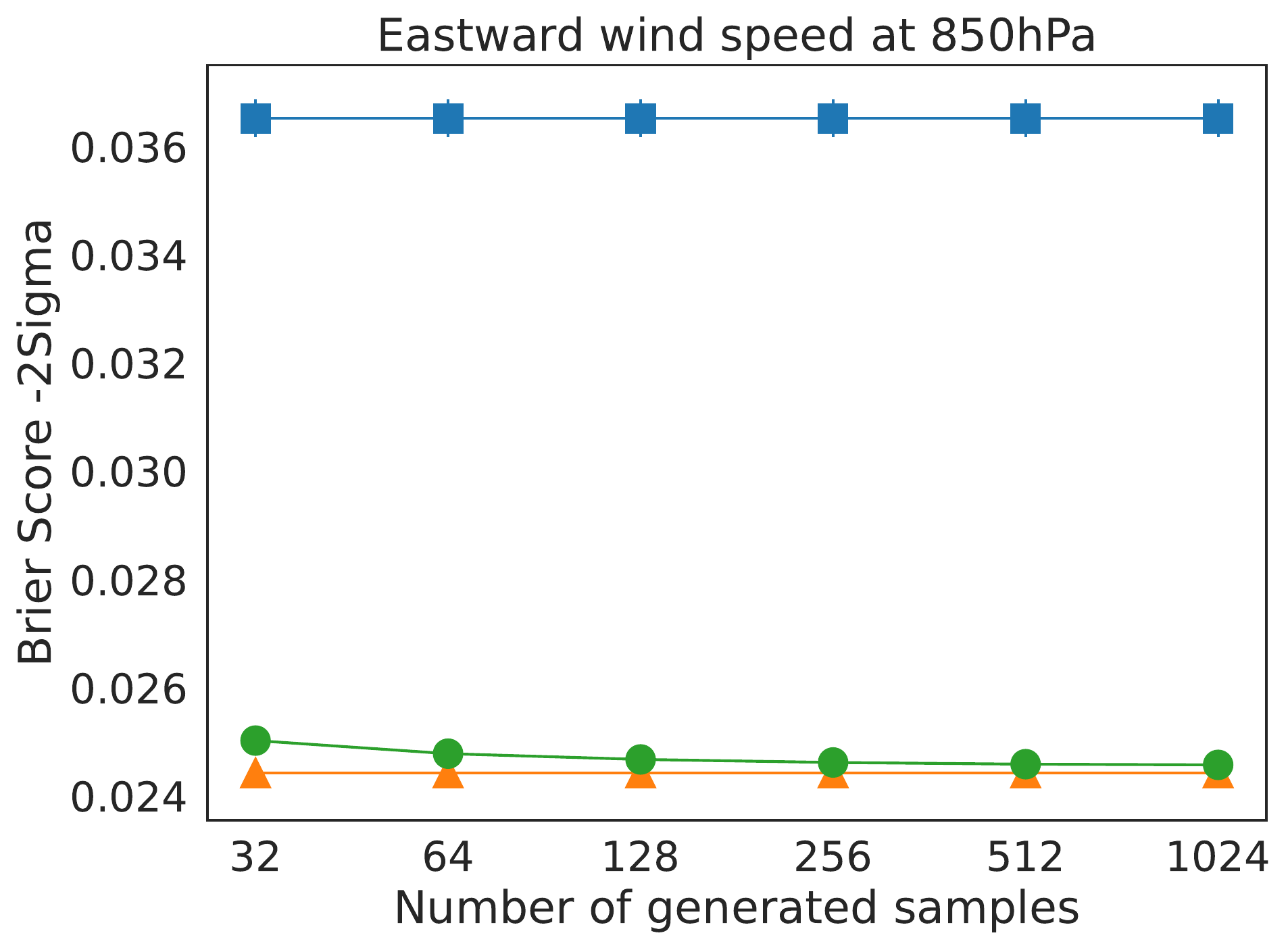}  
    \includegraphics[width=0.4\columnwidth,draft=false]{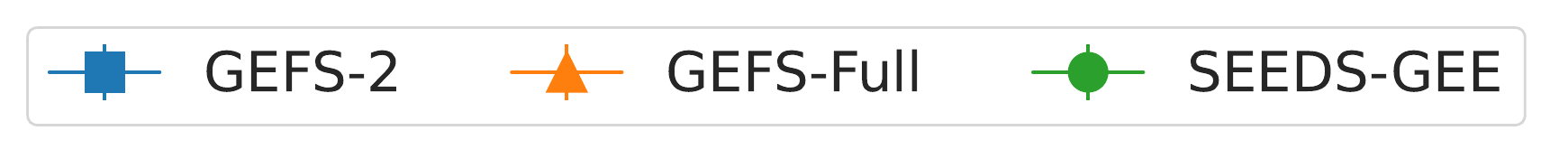}  
    \caption{The effect of $N$ on  various metrics. Top down: RMSE, ACC, CRPS, unreliability, Brier score for $-2\sigma$ events.}
    \label{fEffectOfN}
\end{figure}

Figure~\ref{fEffectOfN} shows the effect of $N$, the size of the generated ensemble, on the skill of \ourgee forecasts. (Some metrics such as Brier scores for $2\sigma$ are omitted as they convey similar information.)

For most metrics, increasing $N$ offers diminishing marginal gains beyond $N=256$. Metrics which are sensitive to the sampling coverage, such as the unreliability $\delta$ and the Brier score at $-2\sigma$, still improve with larger $N >256$.

\subsubsection{Generative ensemble emulation with varying $K$ for 7-day lead time}

\begin{figure}[t]
    \centering
    \includegraphics[width=0.32\columnwidth,draft=false]{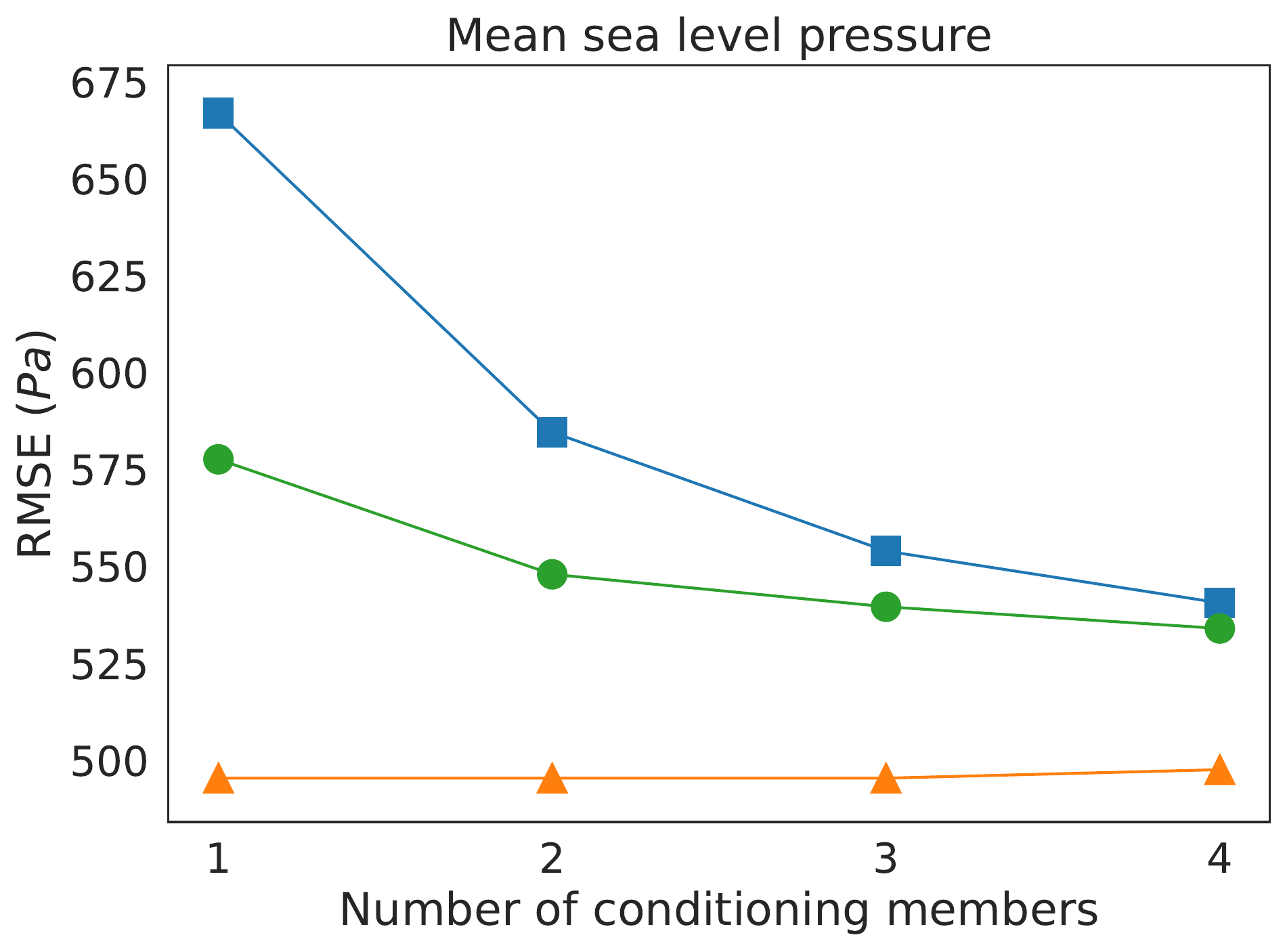}
    \includegraphics[width=0.32\columnwidth,draft=false]{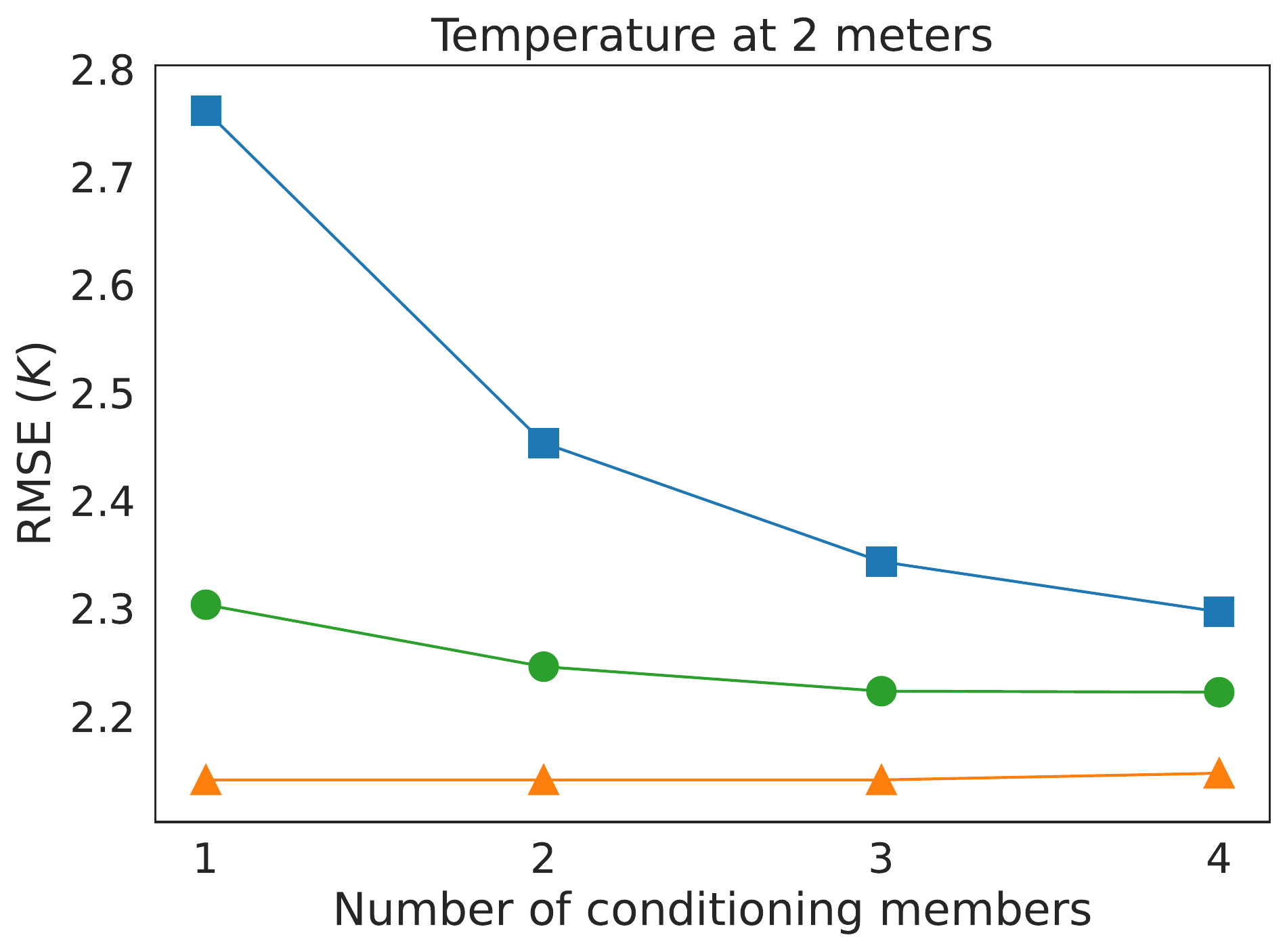}
    \includegraphics[width=0.32\columnwidth,draft=false]{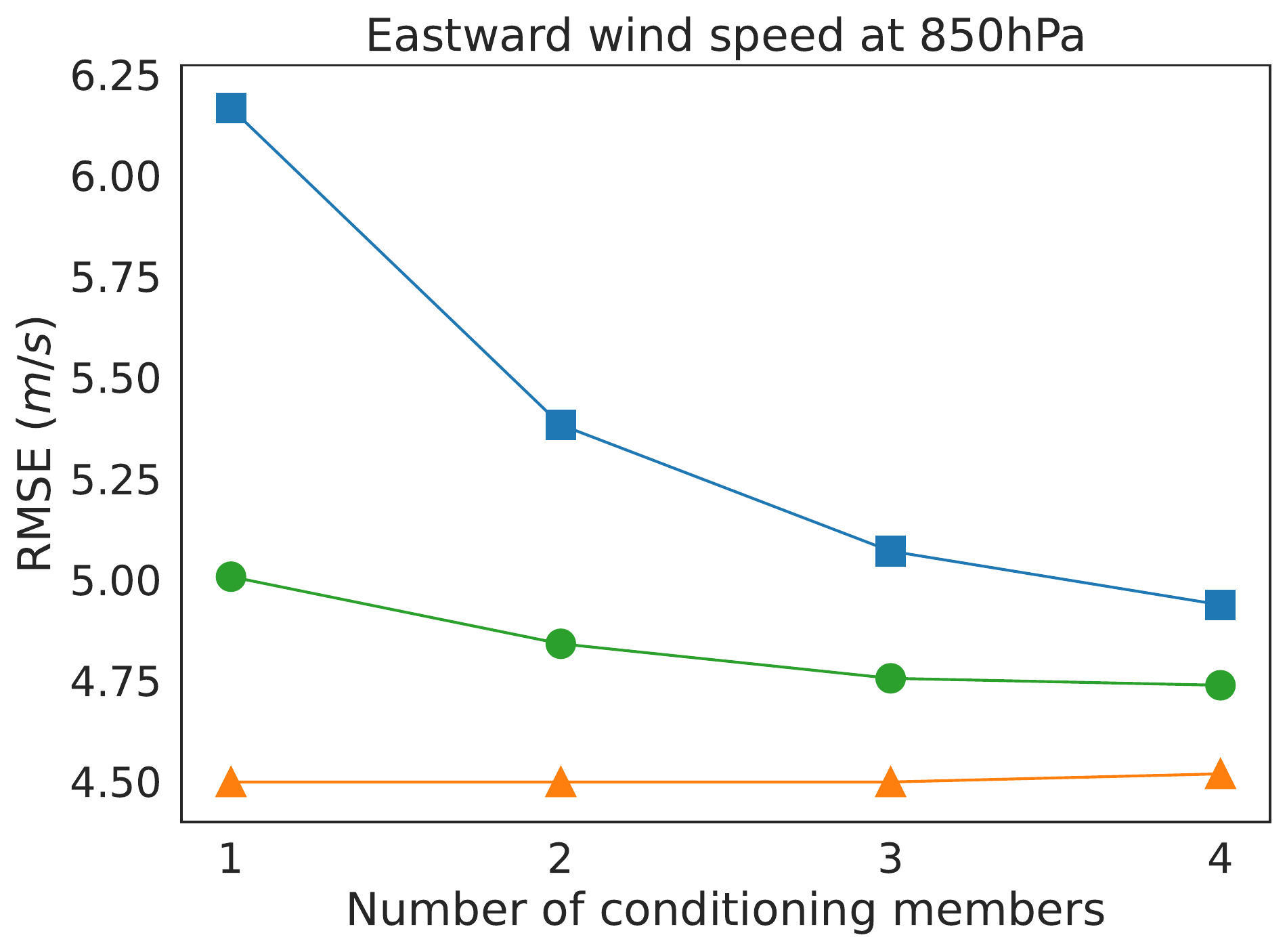}  
    \includegraphics[width=0.32\columnwidth,draft=false]{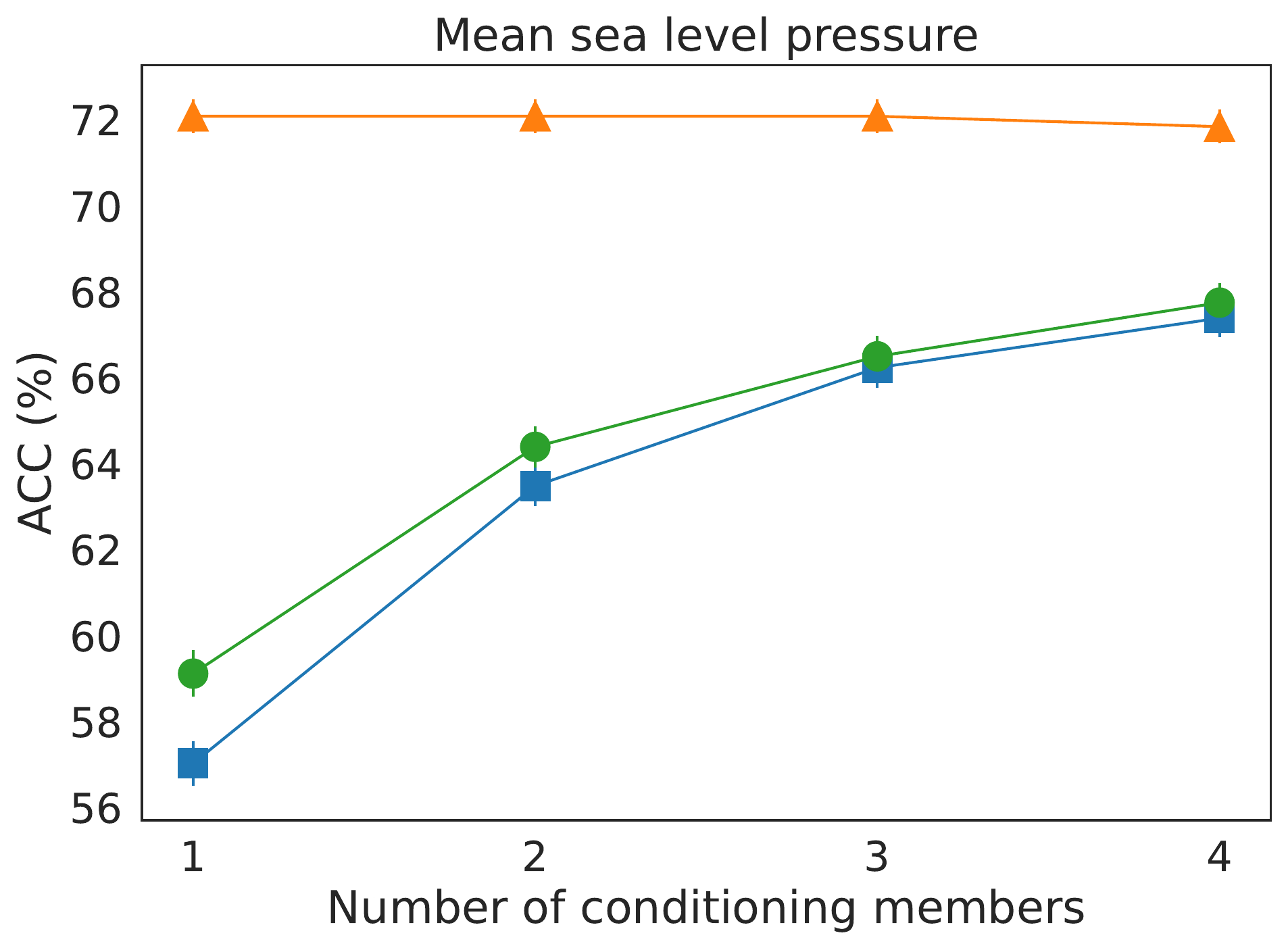}
    \includegraphics[width=0.32\columnwidth,draft=false]{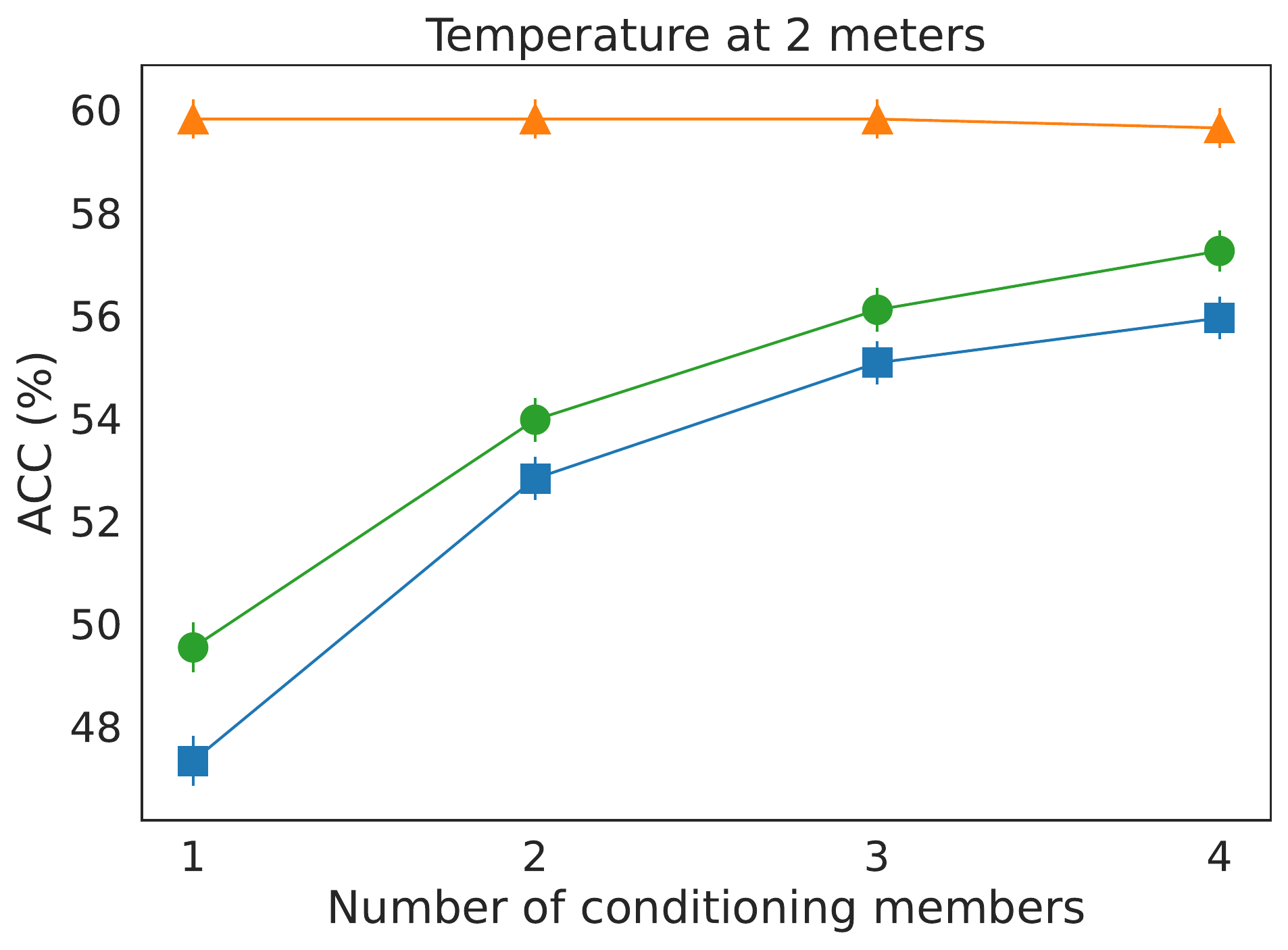}
    \includegraphics[width=0.32\columnwidth,draft=false]{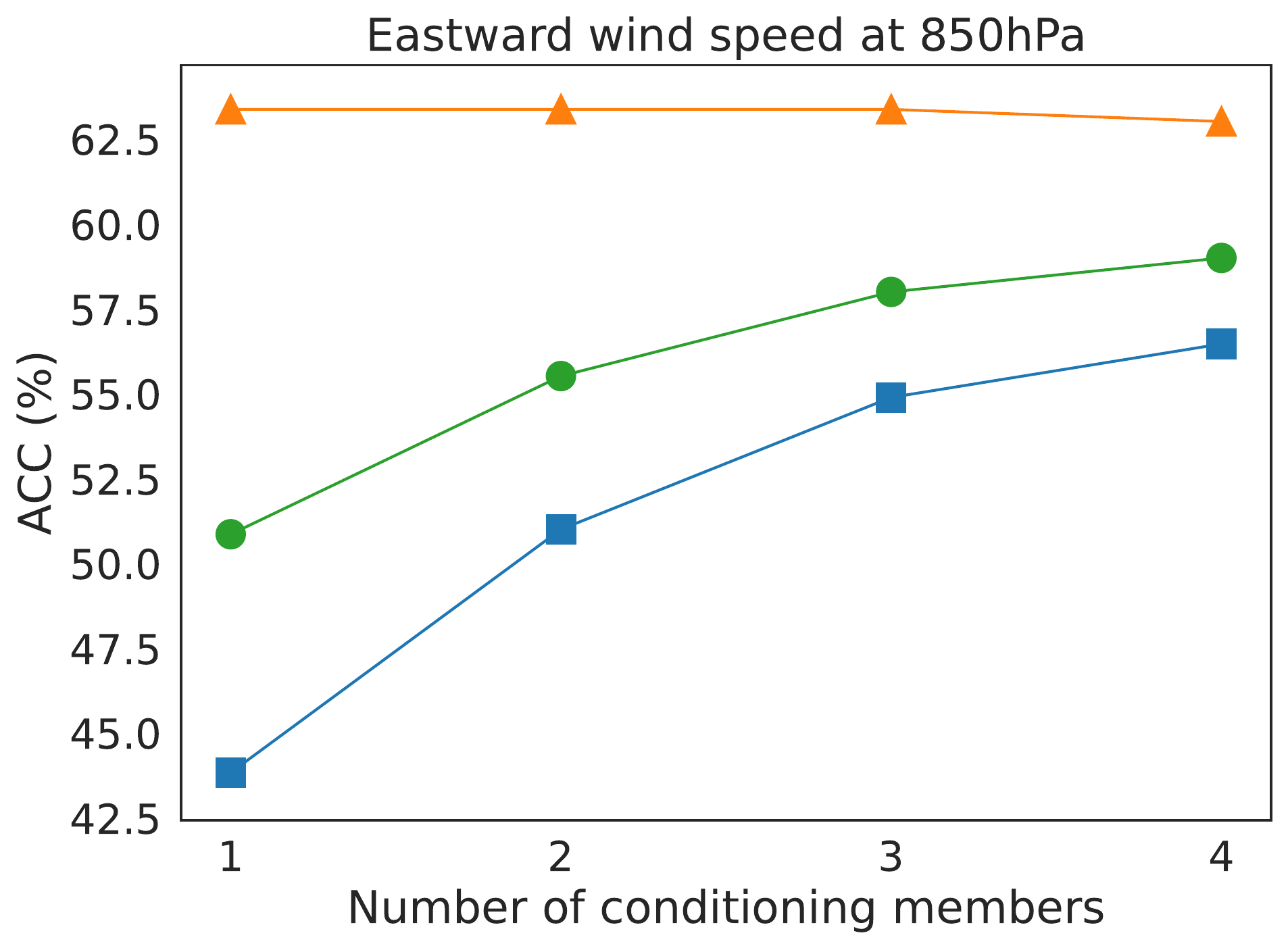}  
    \includegraphics[width=0.32\columnwidth,draft=false]{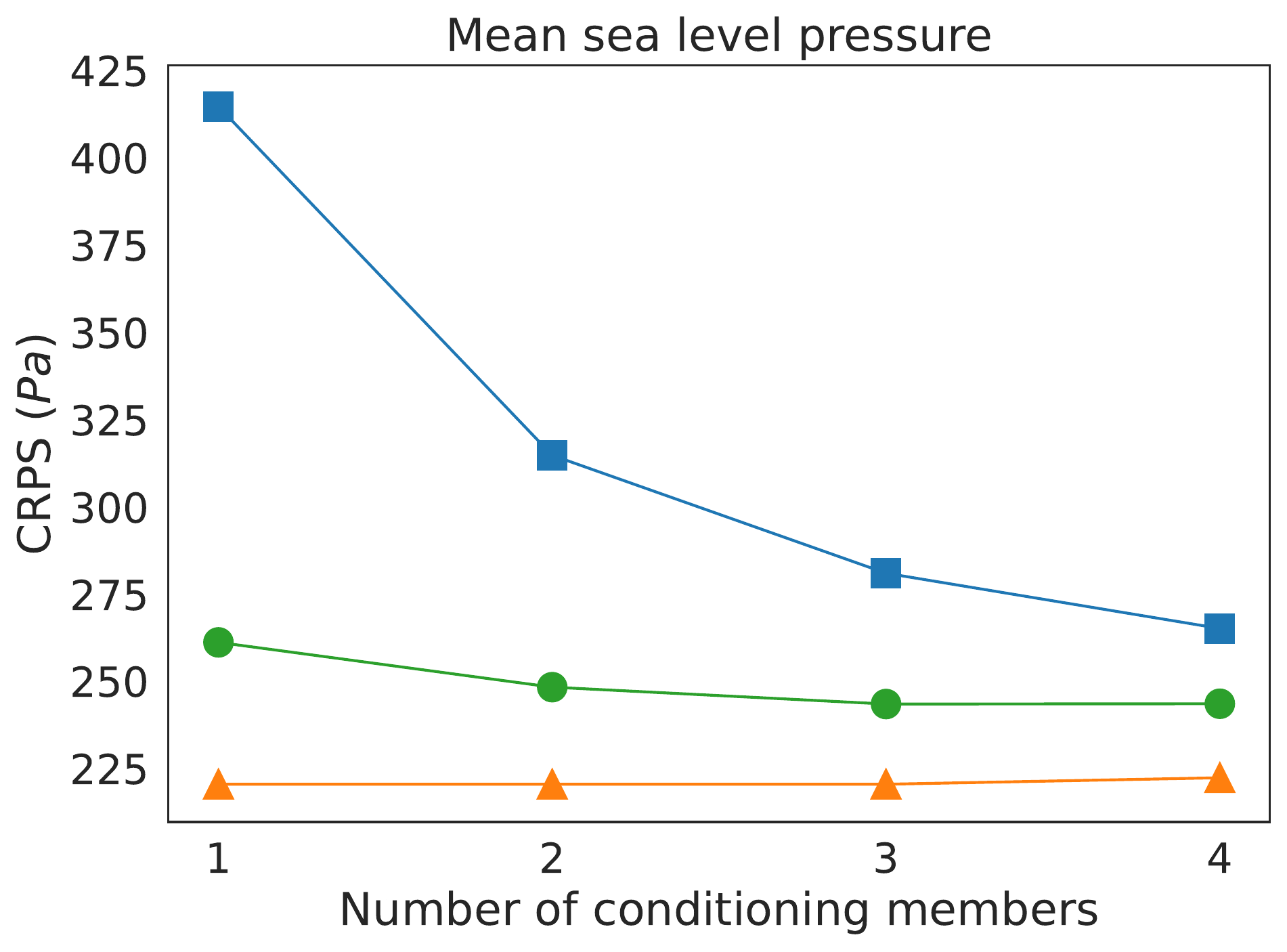}
    \includegraphics[width=0.32\columnwidth,draft=false]{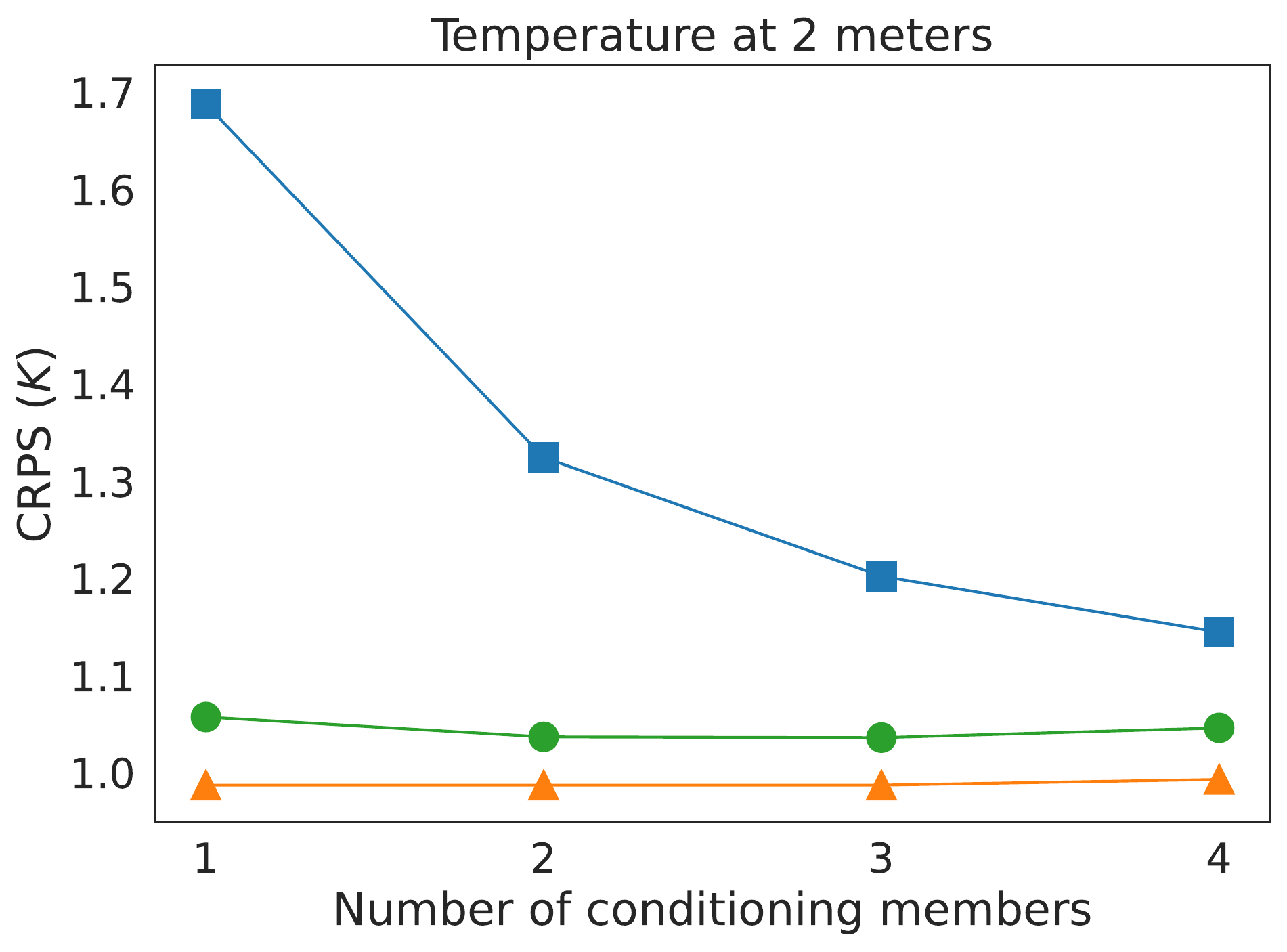}
    \includegraphics[width=0.32\columnwidth,draft=false]{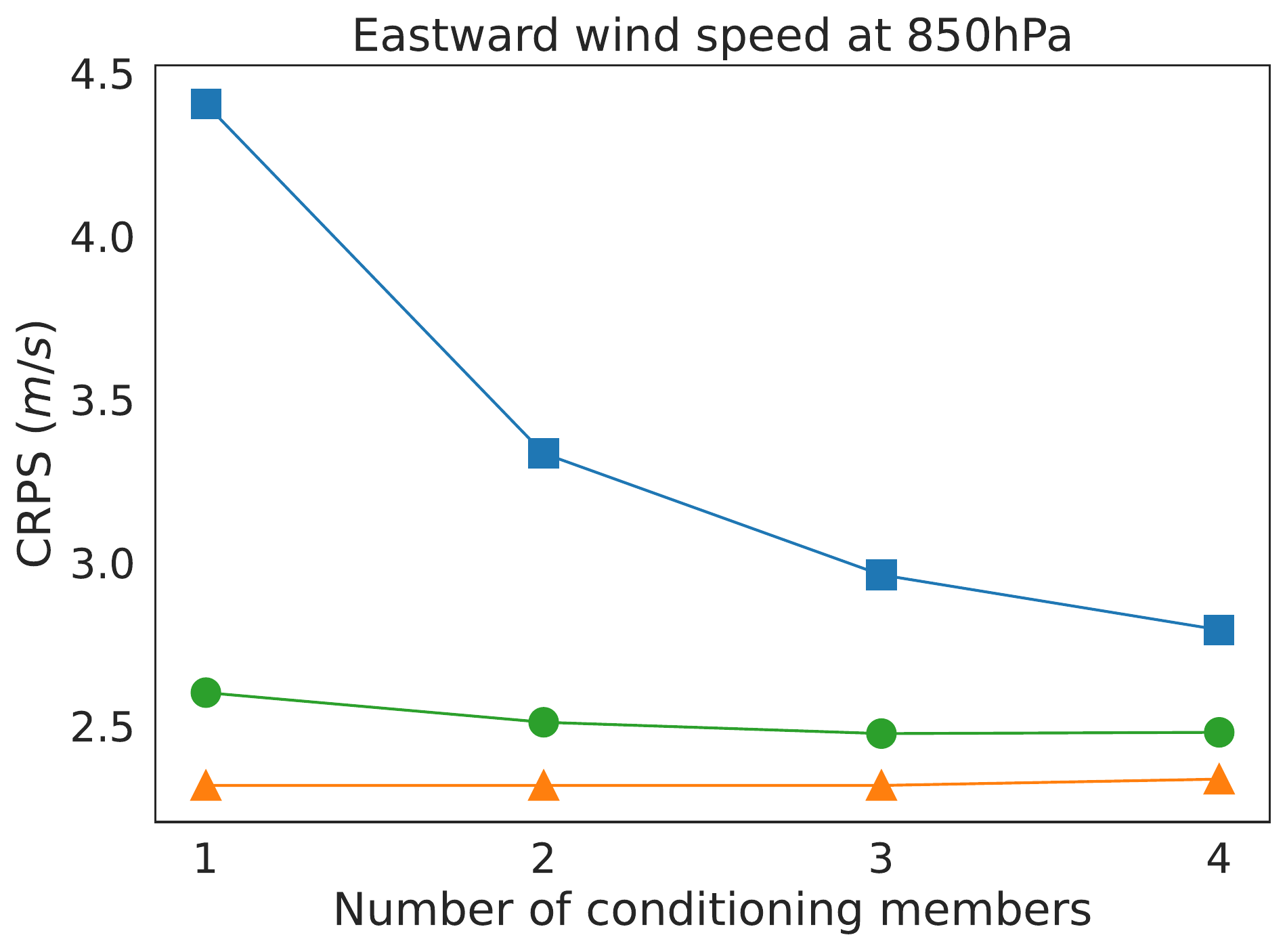}  
    \includegraphics[width=0.32\columnwidth,draft=false]{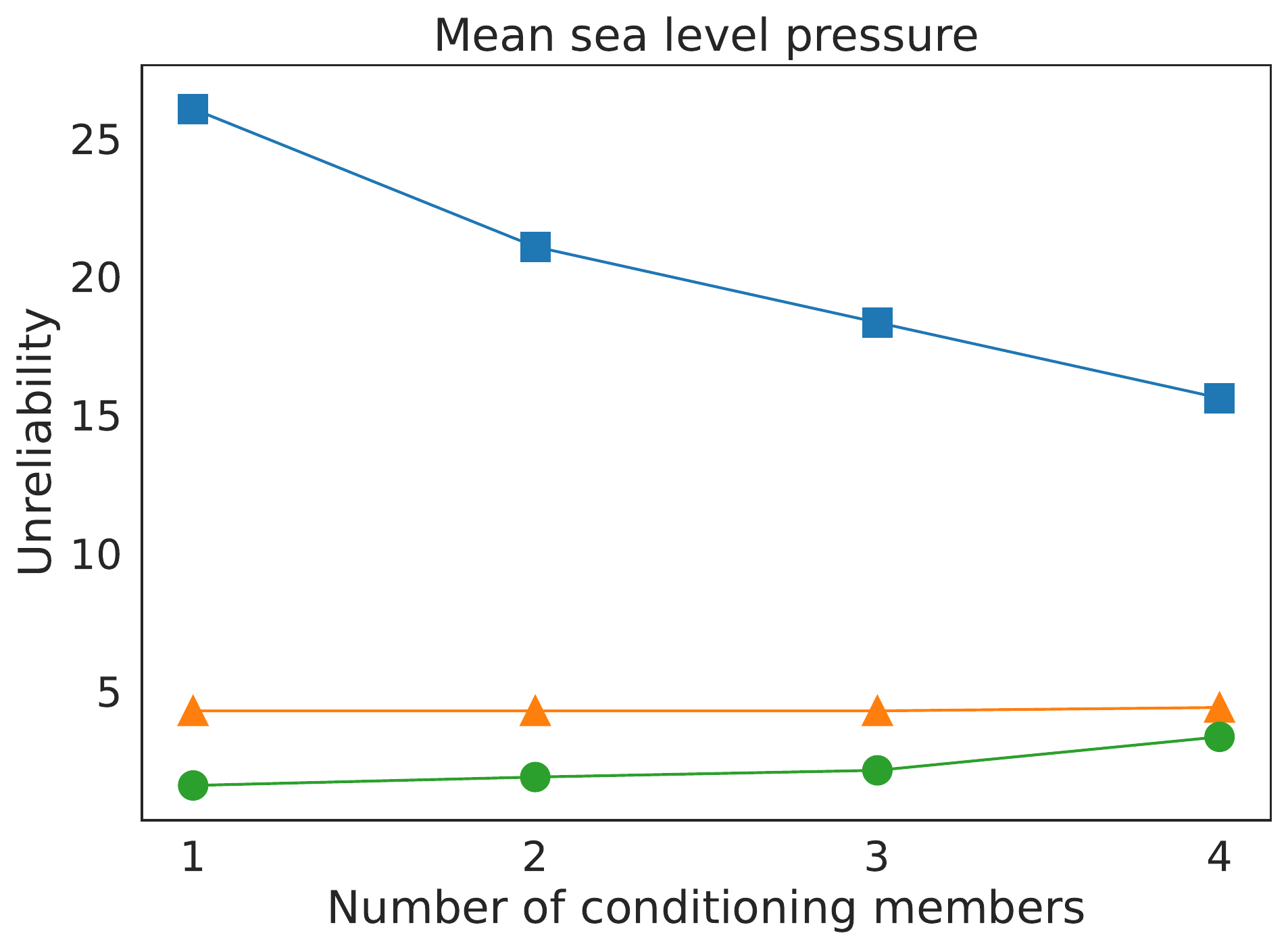}
    \includegraphics[width=0.32\columnwidth,draft=false]{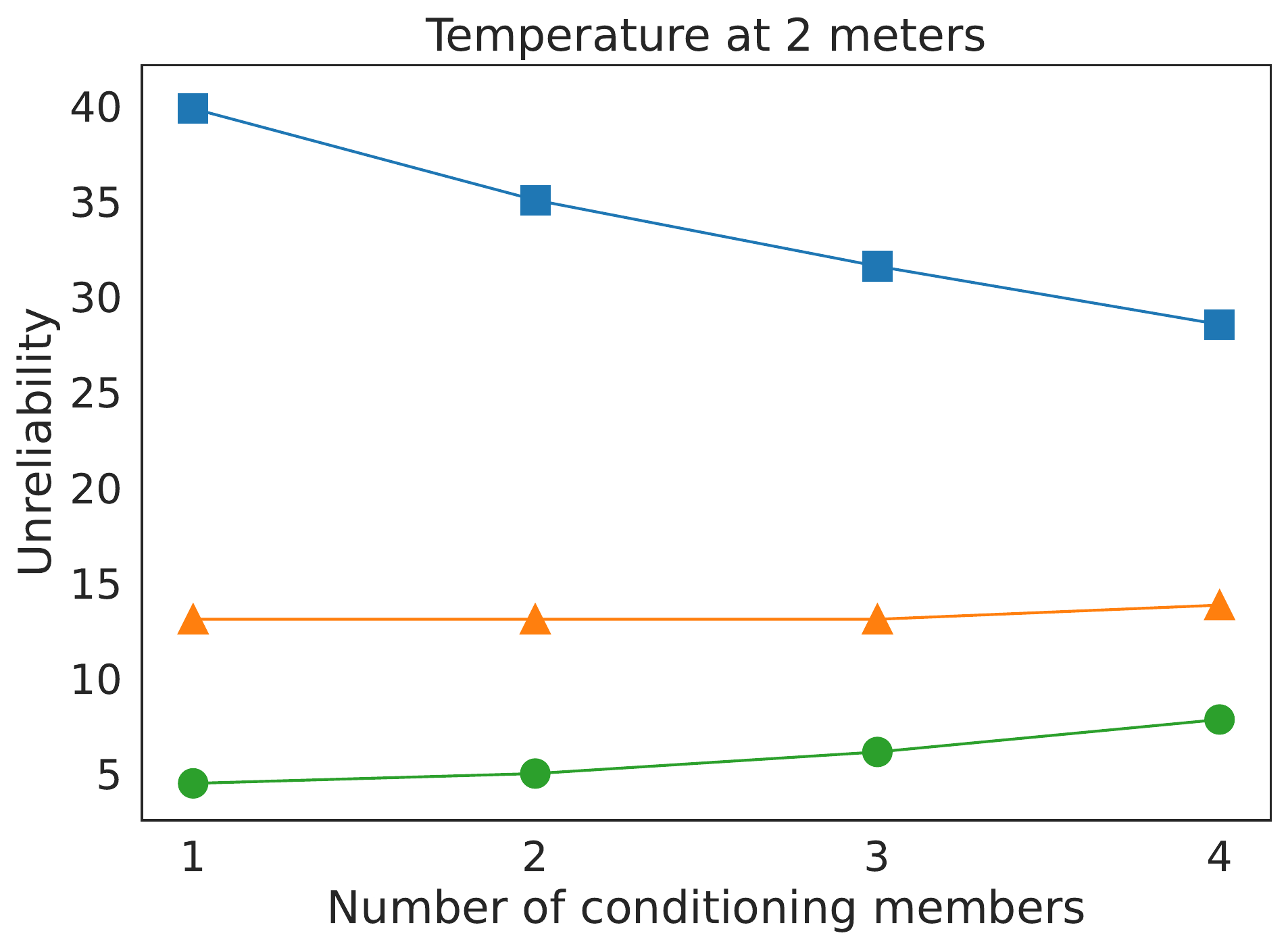}
    \includegraphics[width=0.32\columnwidth,draft=false]{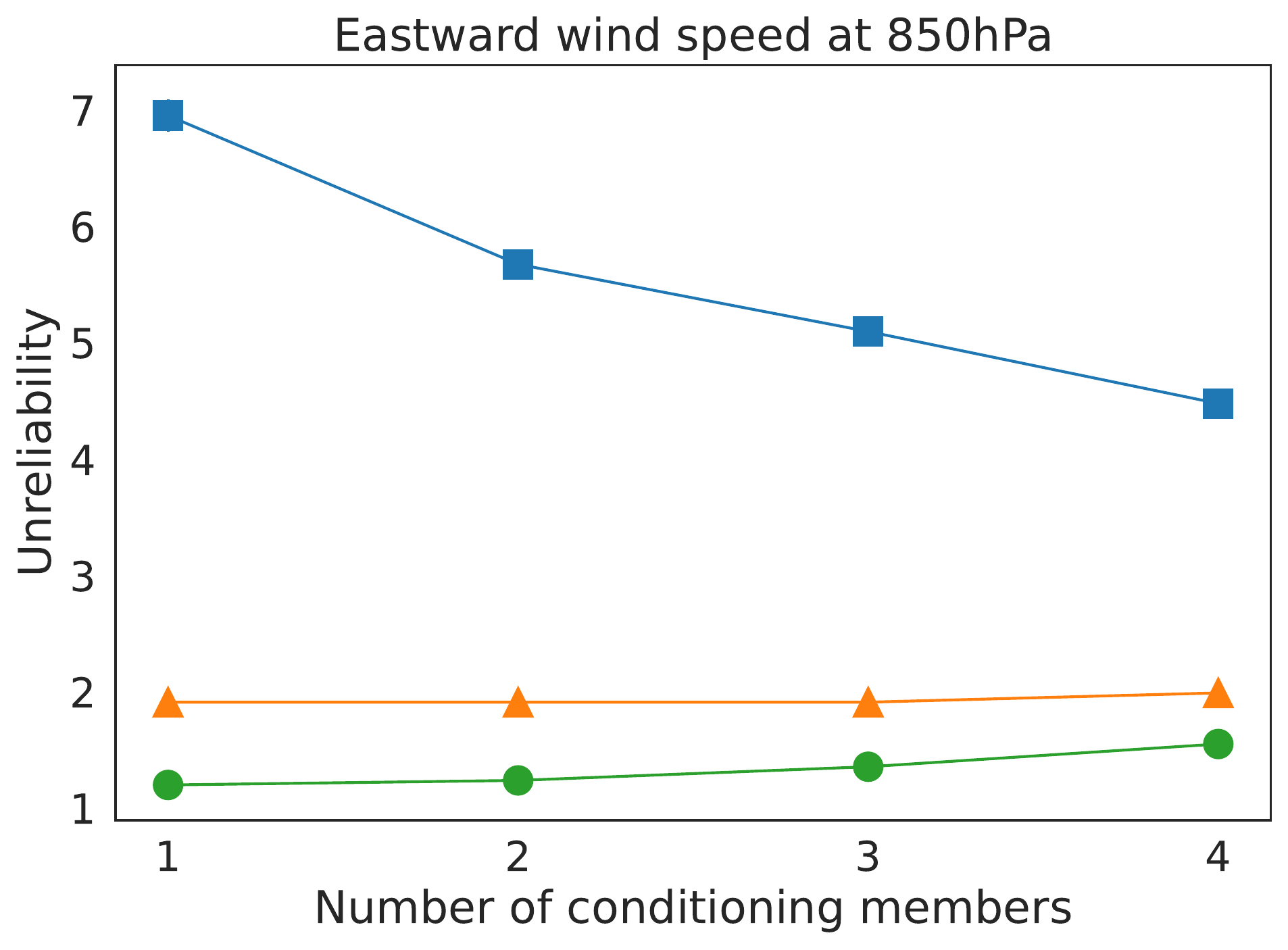}  
    \includegraphics[width=0.32\columnwidth,draft=false]{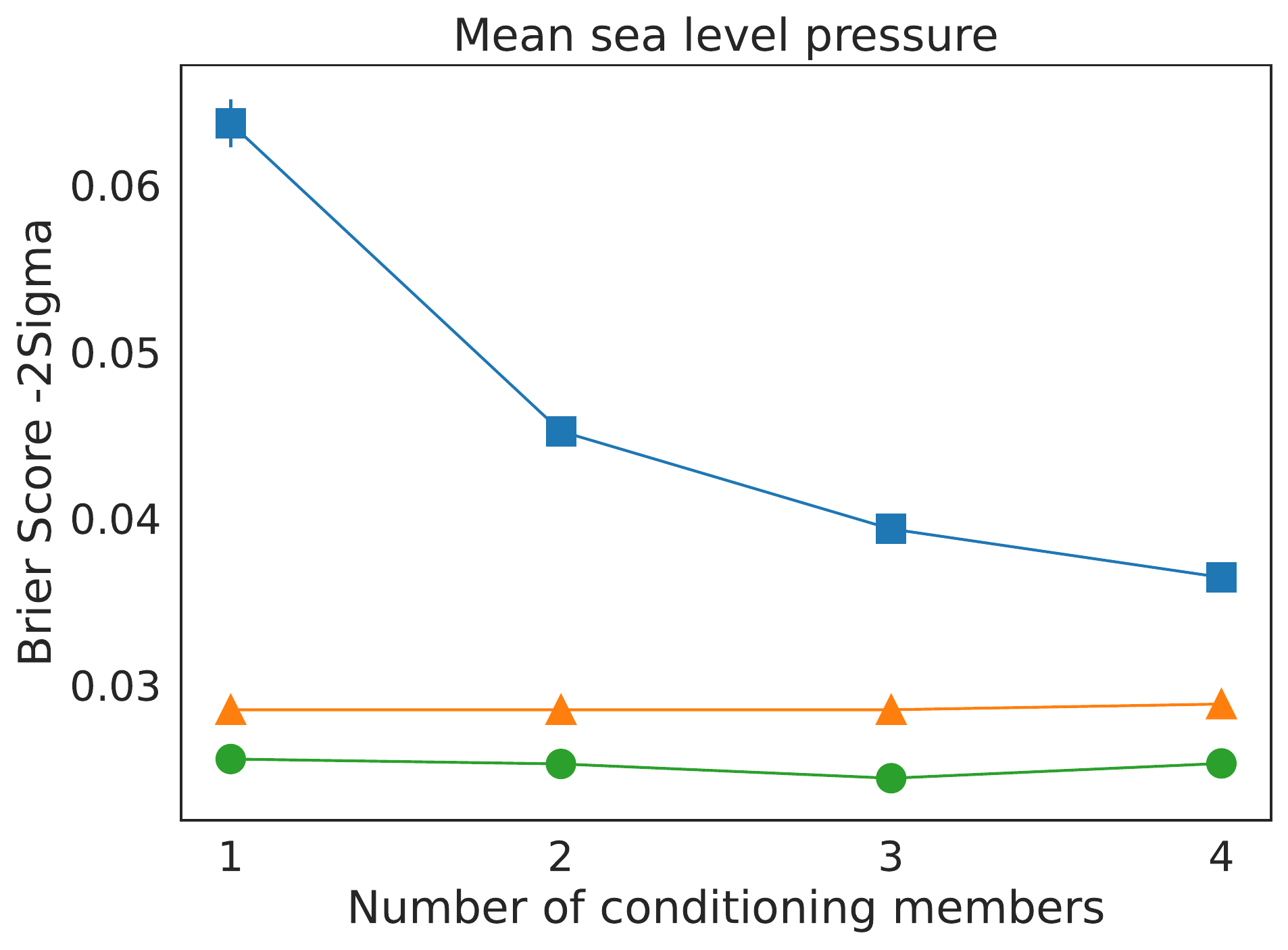}
    \includegraphics[width=0.32\columnwidth,draft=false]{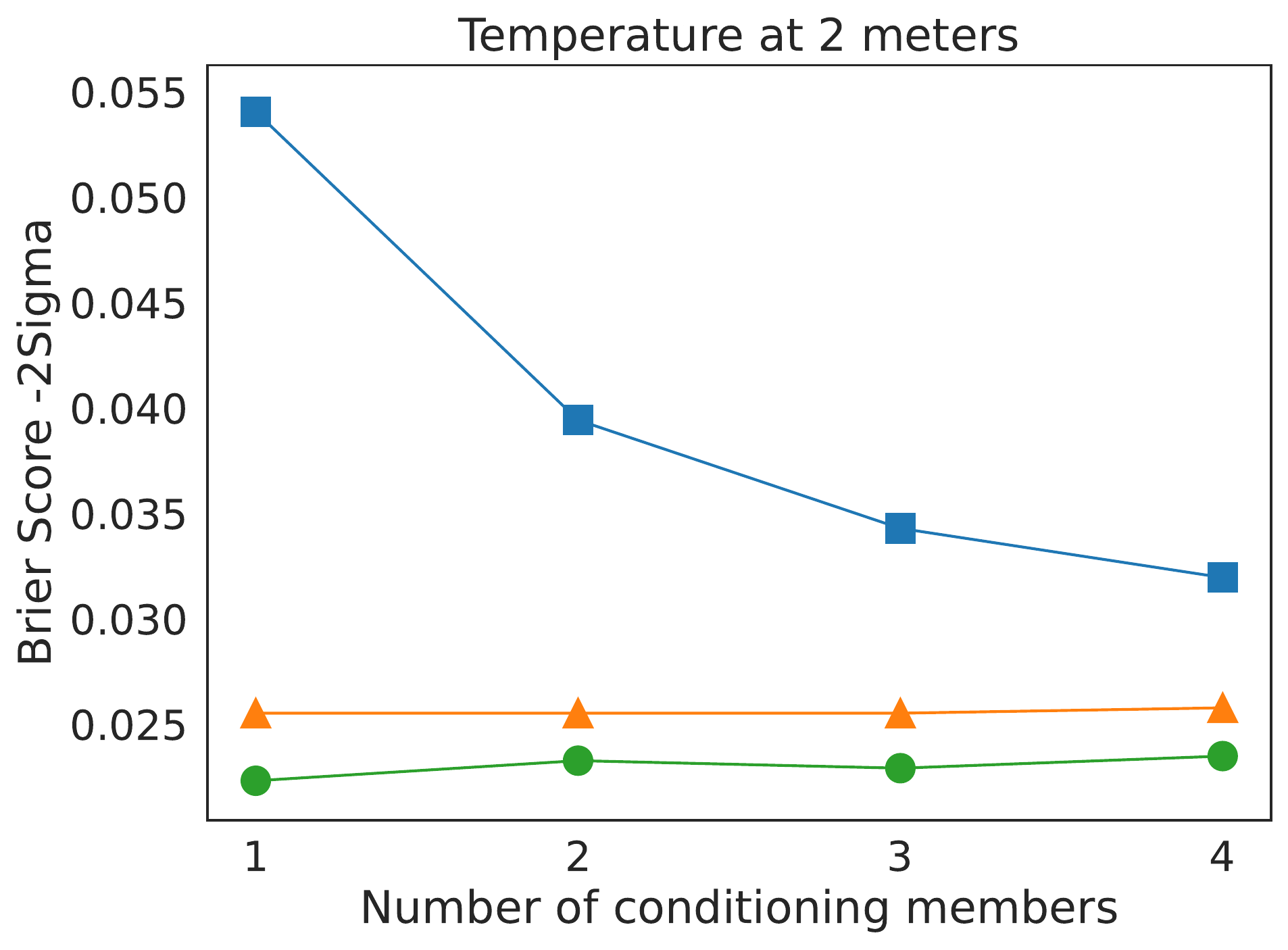}
    \includegraphics[width=0.32\columnwidth,draft=false]{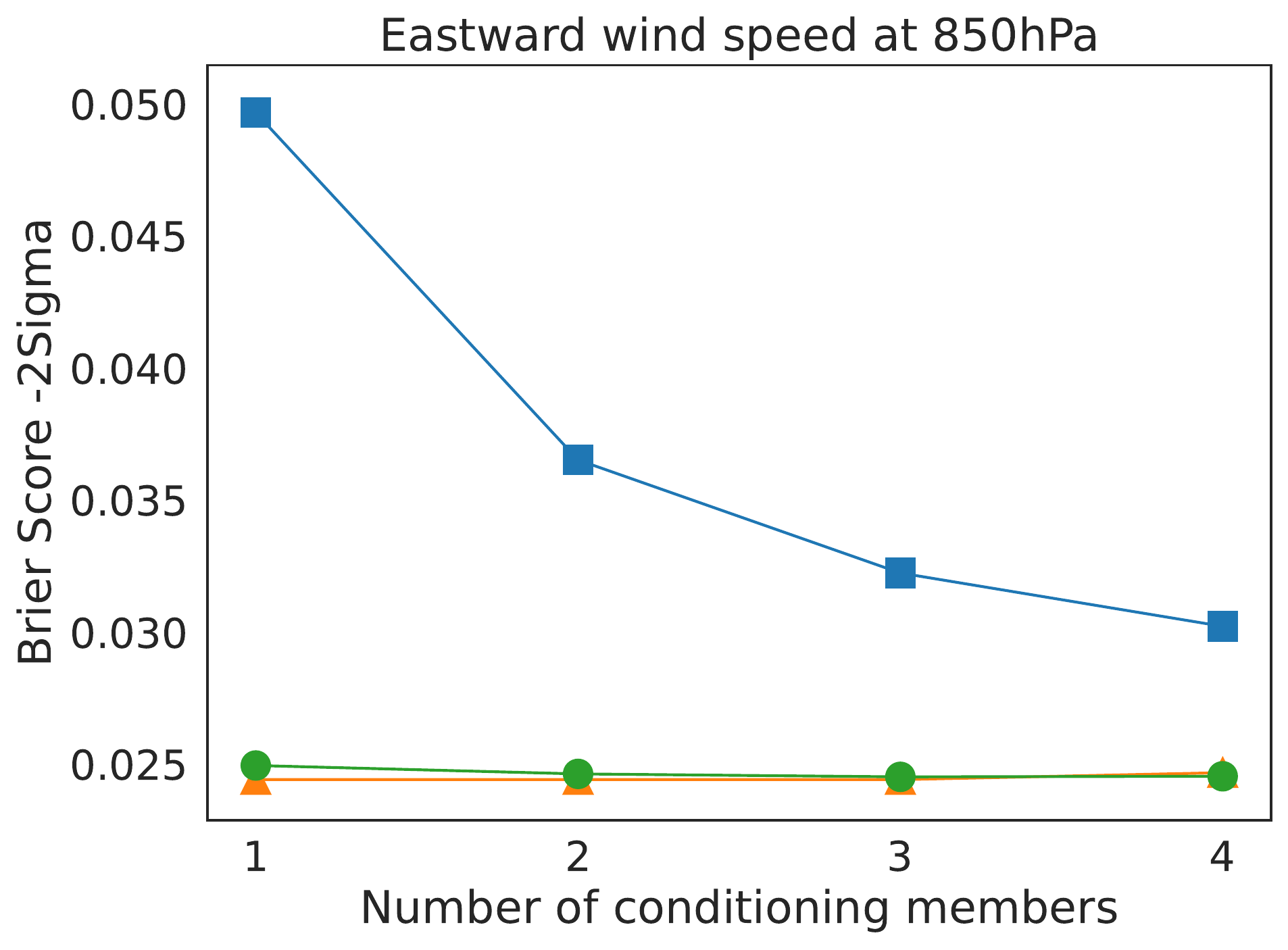}  
    \includegraphics[width=0.40\columnwidth,draft=false]{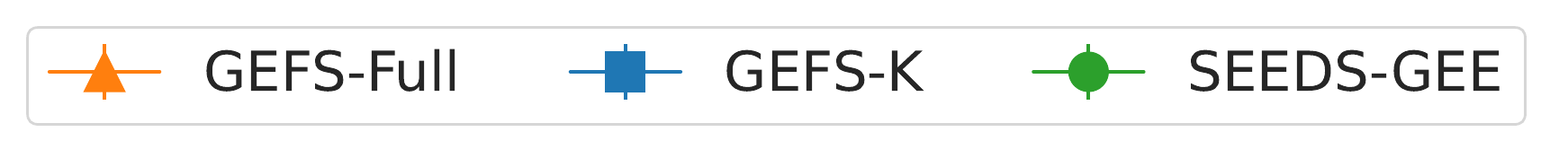}  
    \caption{The effect of $K$ on  various metrics. Top down: RMSE, ACC, CRPS, unreliability, Brier score for $-2\sigma$ event.}
    \label{fEffectOfK}
\end{figure}

Figure~\ref{fEffectOfK} studies the effect of $K$, the number of seeds,  on the skill of \ourgee forecasts. We find that the ensemble forecast skill of \ourgee is comparable to physics-based \emph{full} ensembles,  when conditioned on $K\geq2$ seeds from GEFS.  Note that \ourgee is a drastically better forecast system than physics-based ensembles formed by the conditioning members (\gefsK). 

For all values of $K$, the affordability of large ensemble generation warranted by \ourmethodshort leads to very significant improvements in extreme event classification skill with respect to the conditioning ensemble.

It is also shown that the generative emulation approach is also feasible when $K=1$. However, the skill of the generative ensemble  is closer to physics-based ensembles with just a few members (\gefsK) than to \gefsfull. These studies suggest that $K=2$ could be a good compromise between generative ensemble performance and computational savings.

\subsubsection{Generative post-processing with varying $K'$ for 7-day lead time with $K=2$}

\begin{figure}[t]
    \centering
    \includegraphics[width=0.32\columnwidth,draft=false]{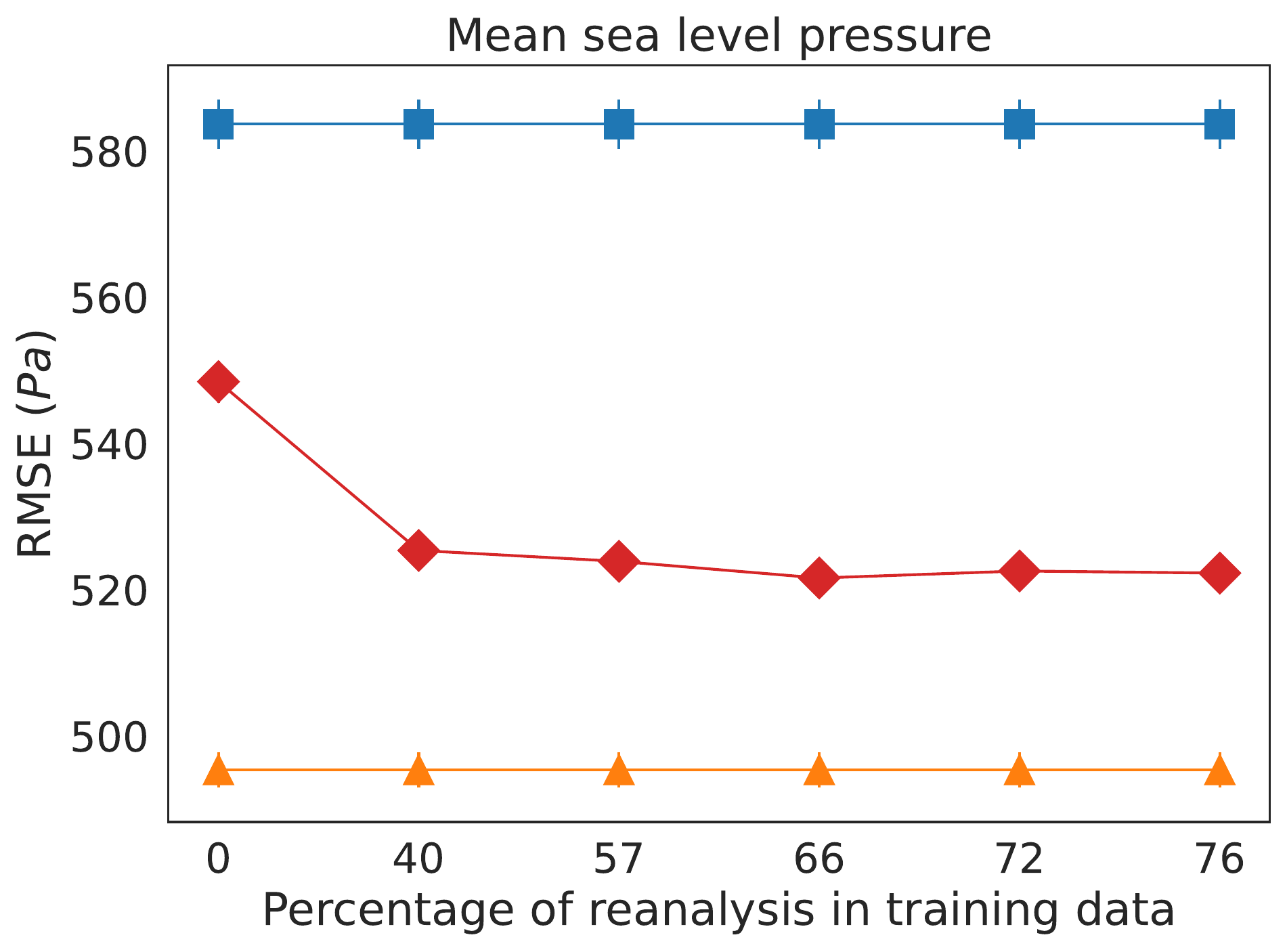}
    \includegraphics[width=0.32\columnwidth,draft=false]{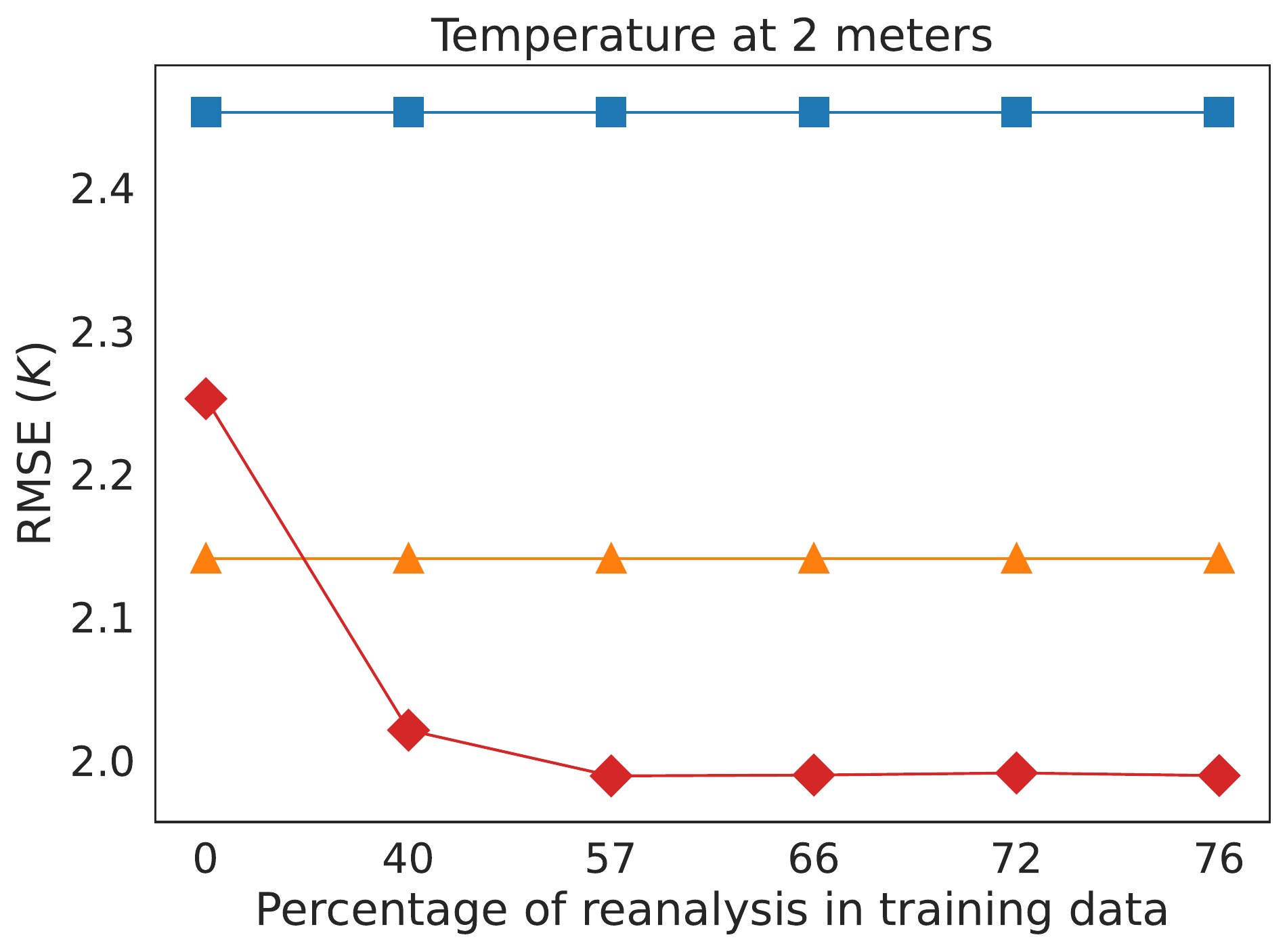}
    \includegraphics[width=0.32\columnwidth,draft=false]{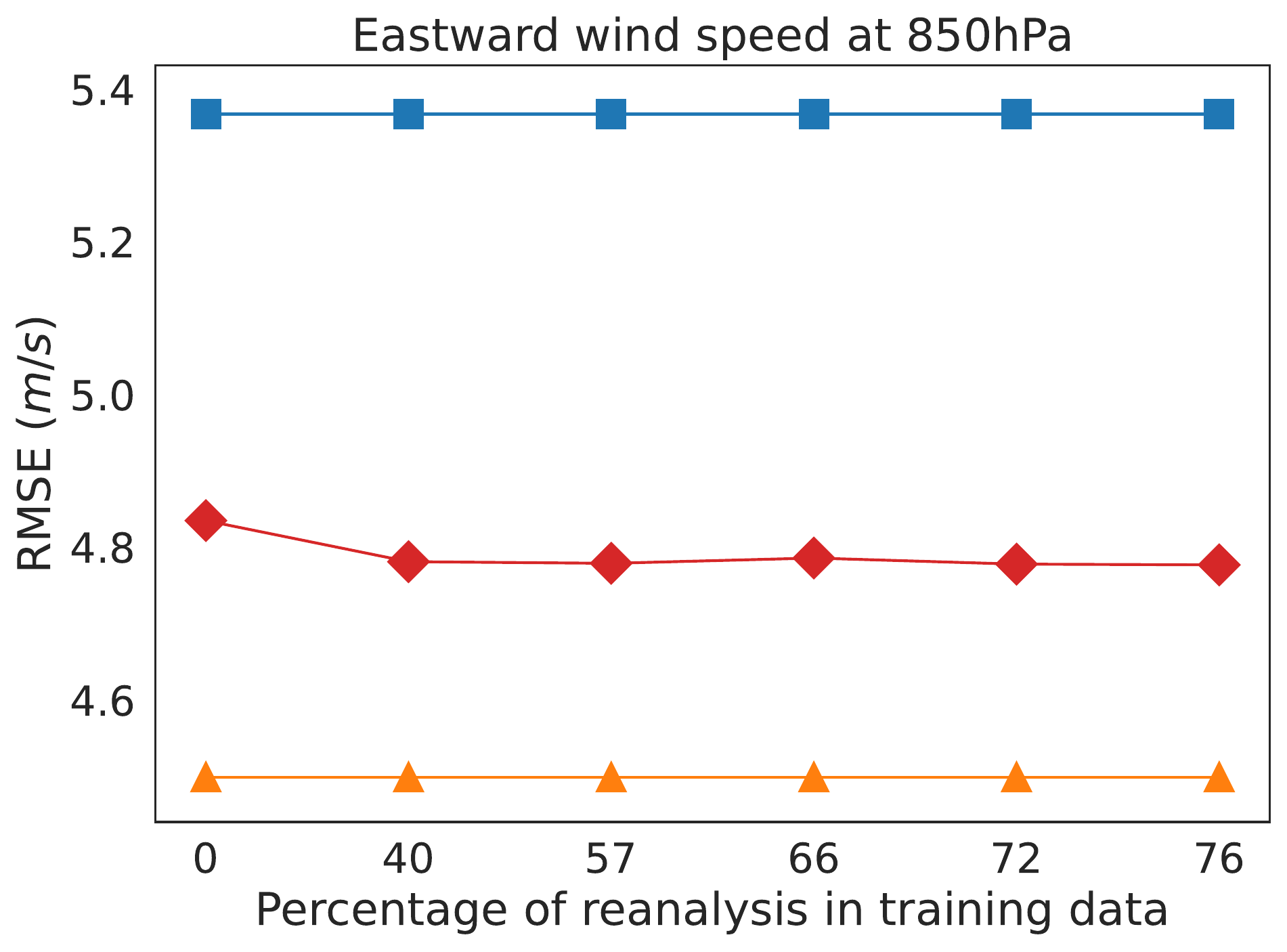}  
    \includegraphics[width=0.32\columnwidth,draft=false]{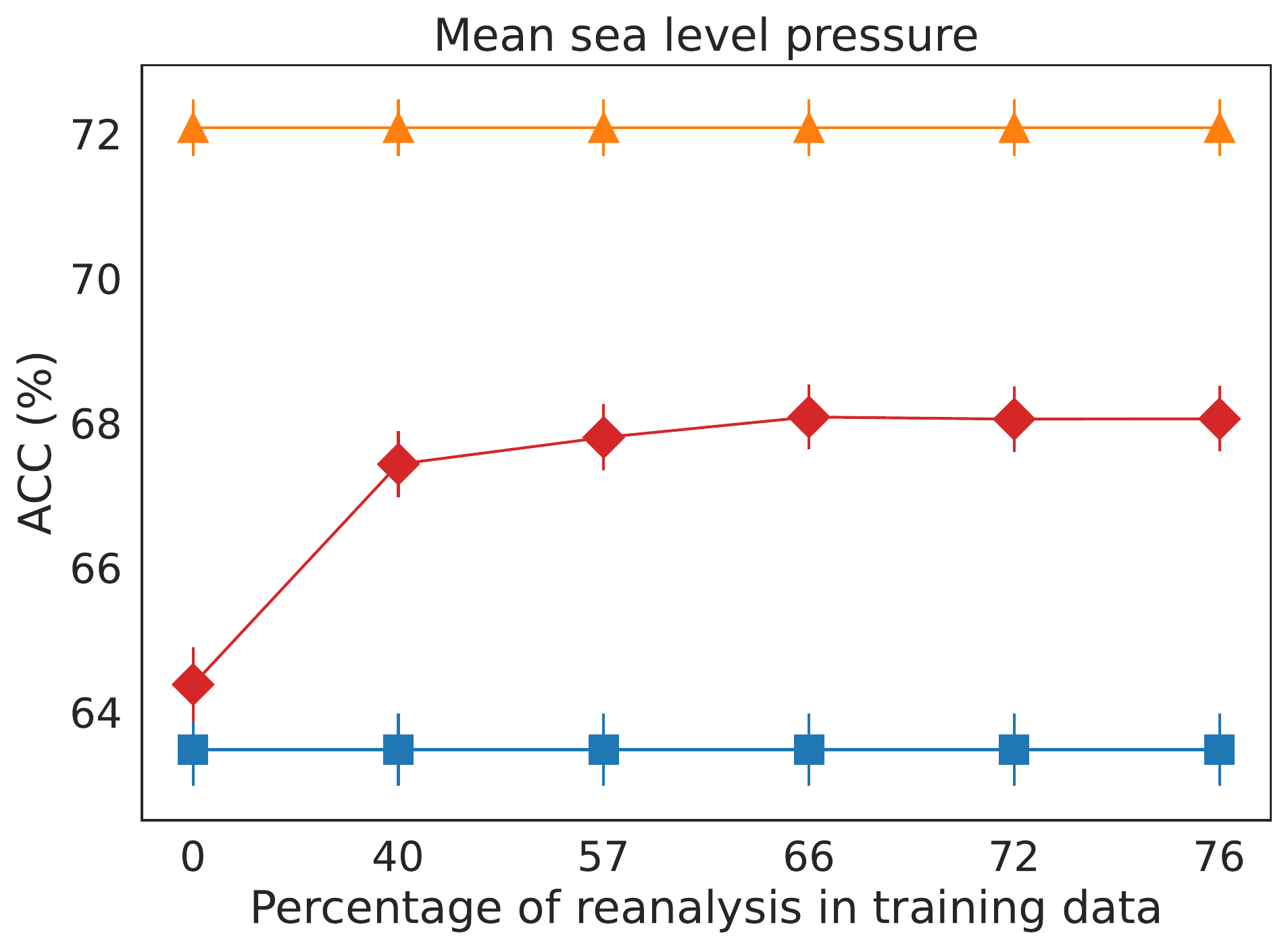}
    \includegraphics[width=0.32\columnwidth,draft=false]{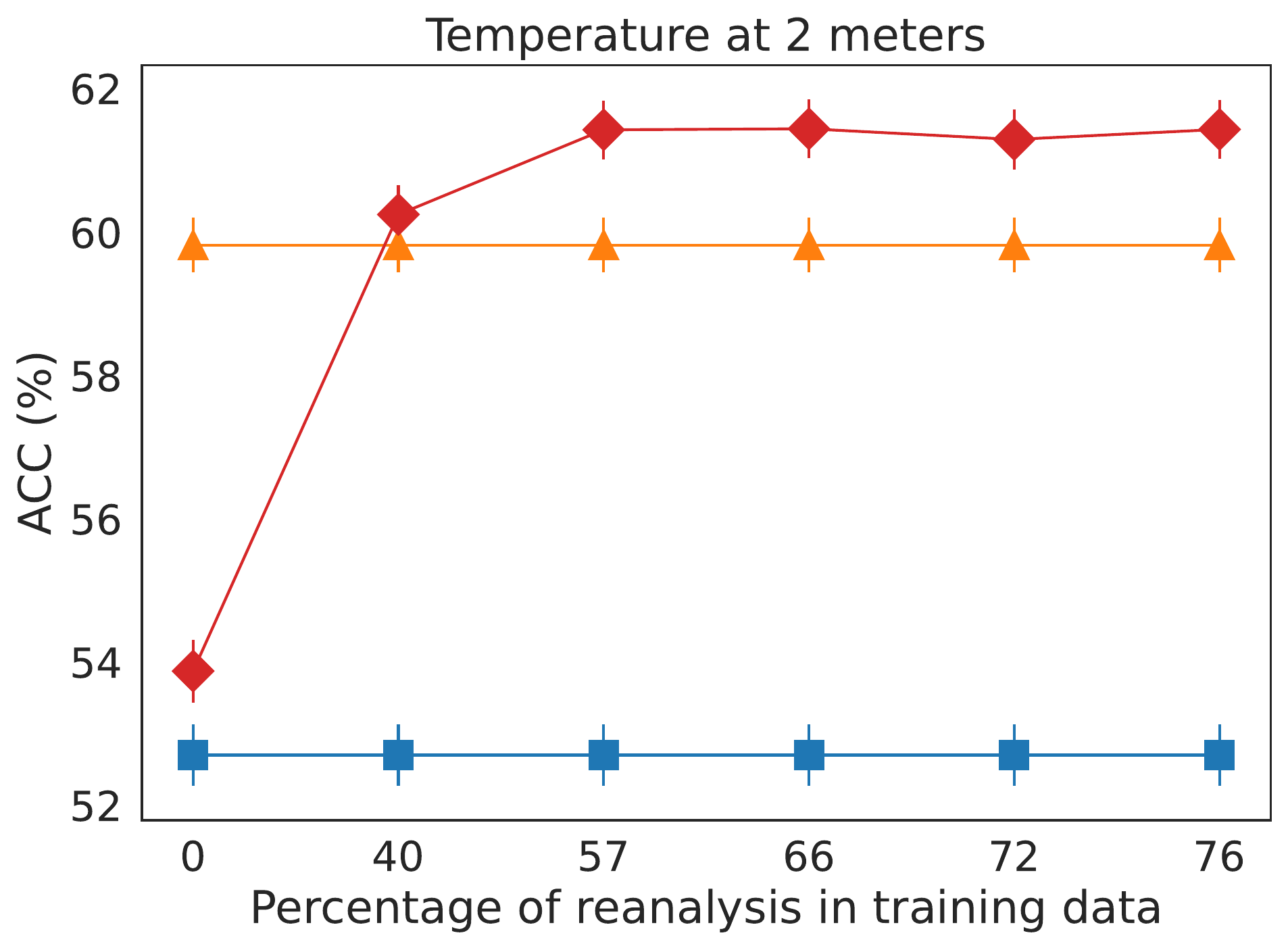}
    \includegraphics[width=0.32\columnwidth,draft=false]{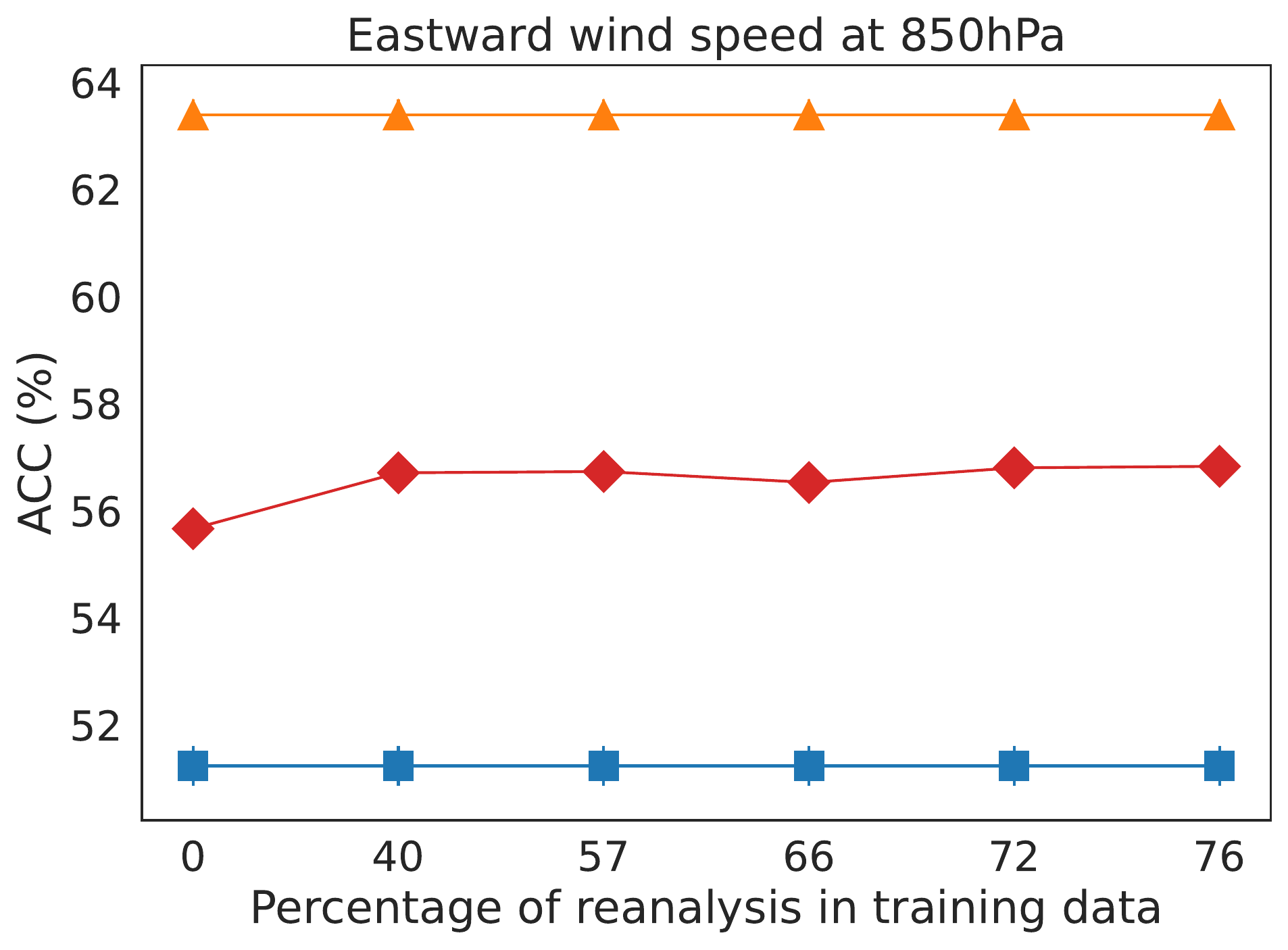}  
    \includegraphics[width=0.32\columnwidth,draft=false]{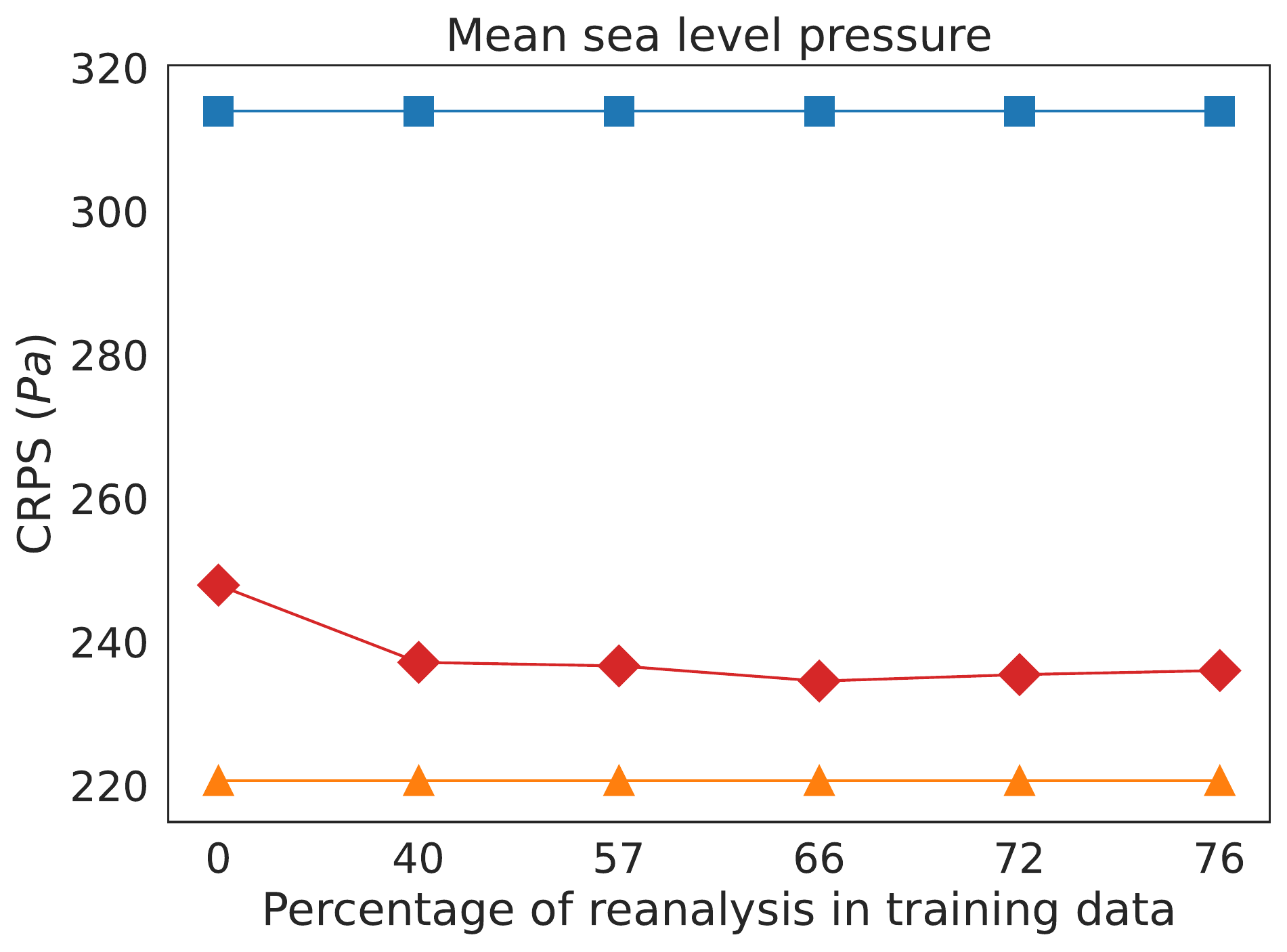}
    \includegraphics[width=0.32\columnwidth,draft=false]{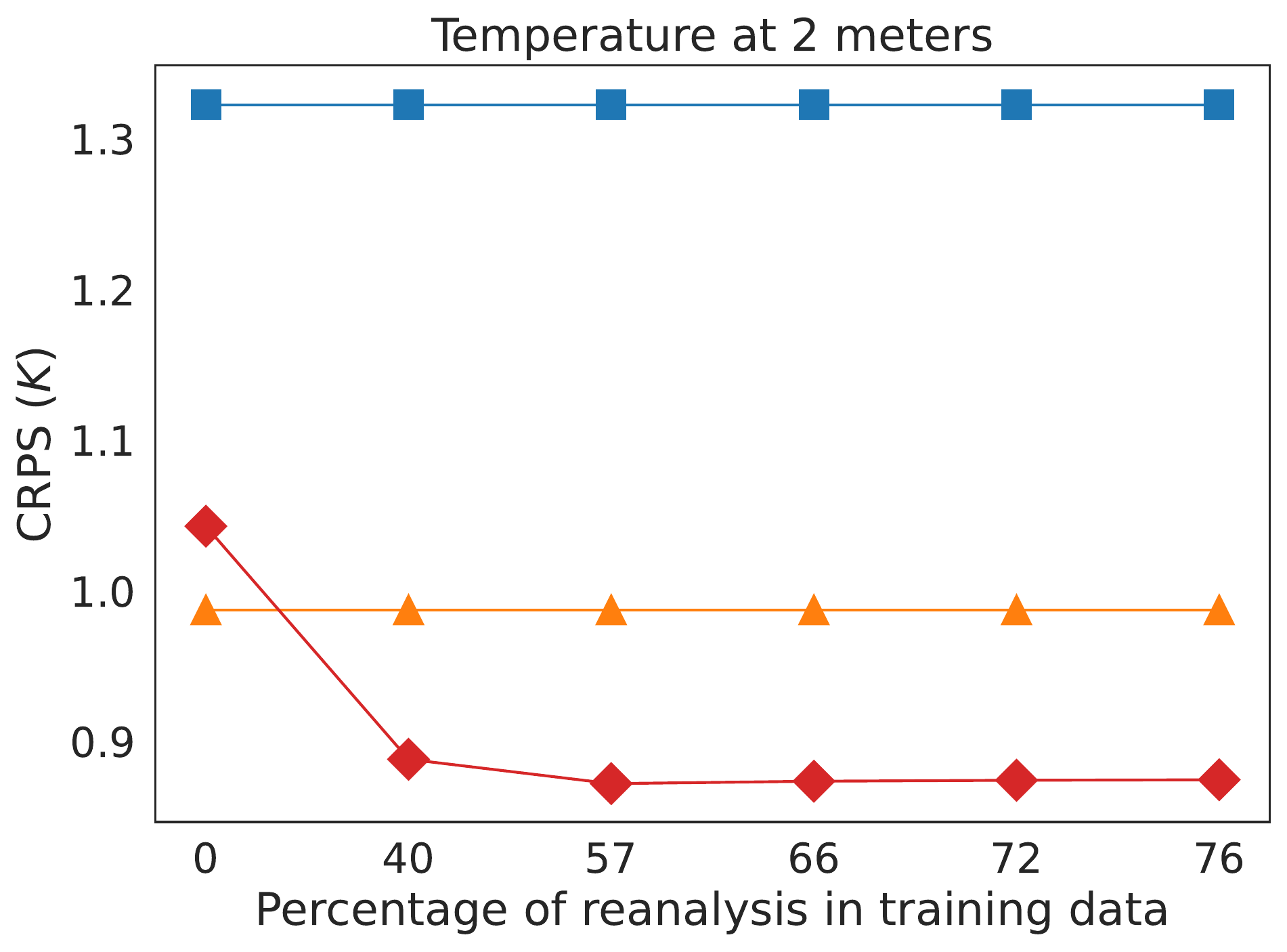}
    \includegraphics[width=0.32\columnwidth,draft=false]{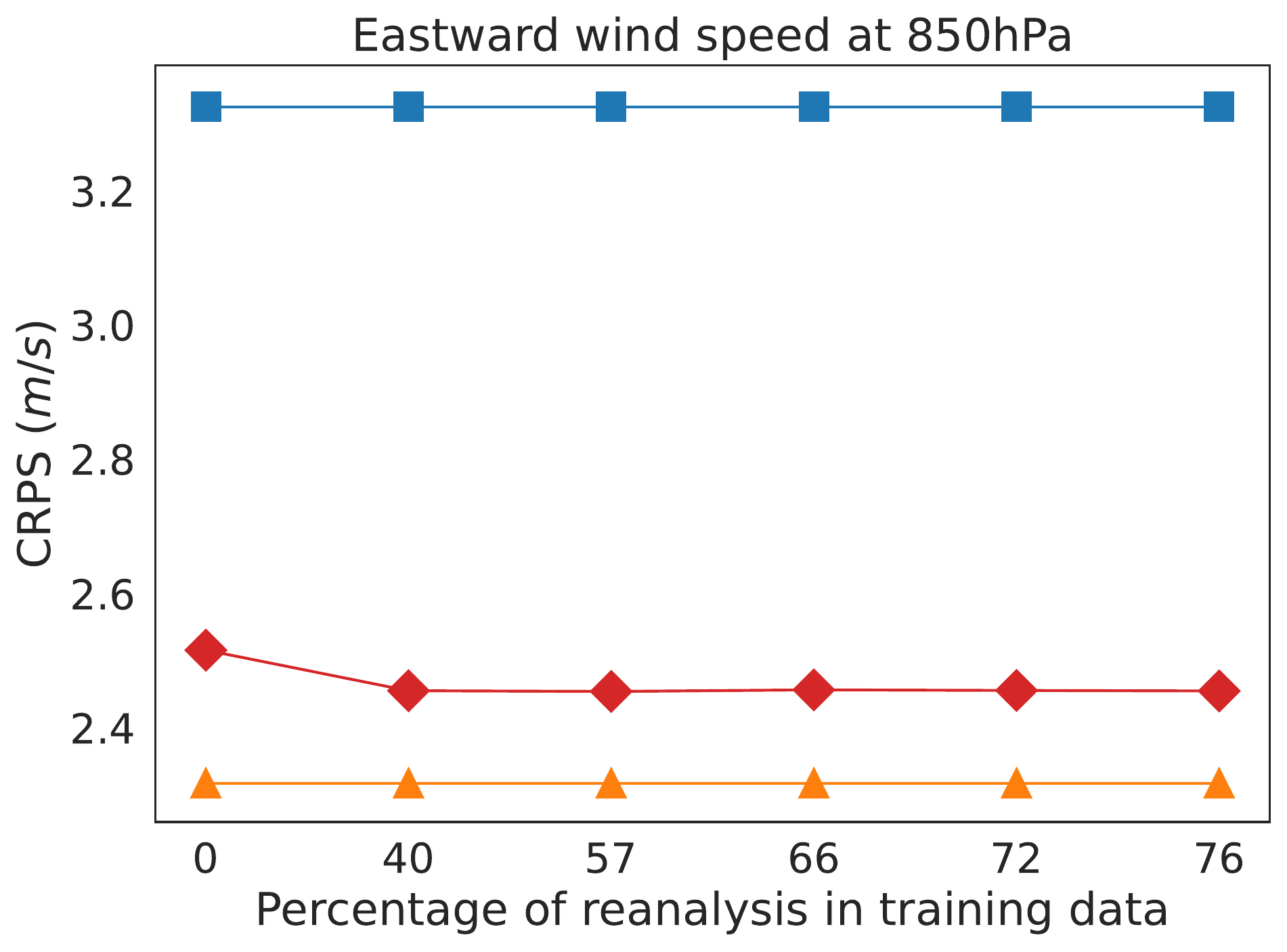}  
    \includegraphics[width=0.32\columnwidth,draft=false]{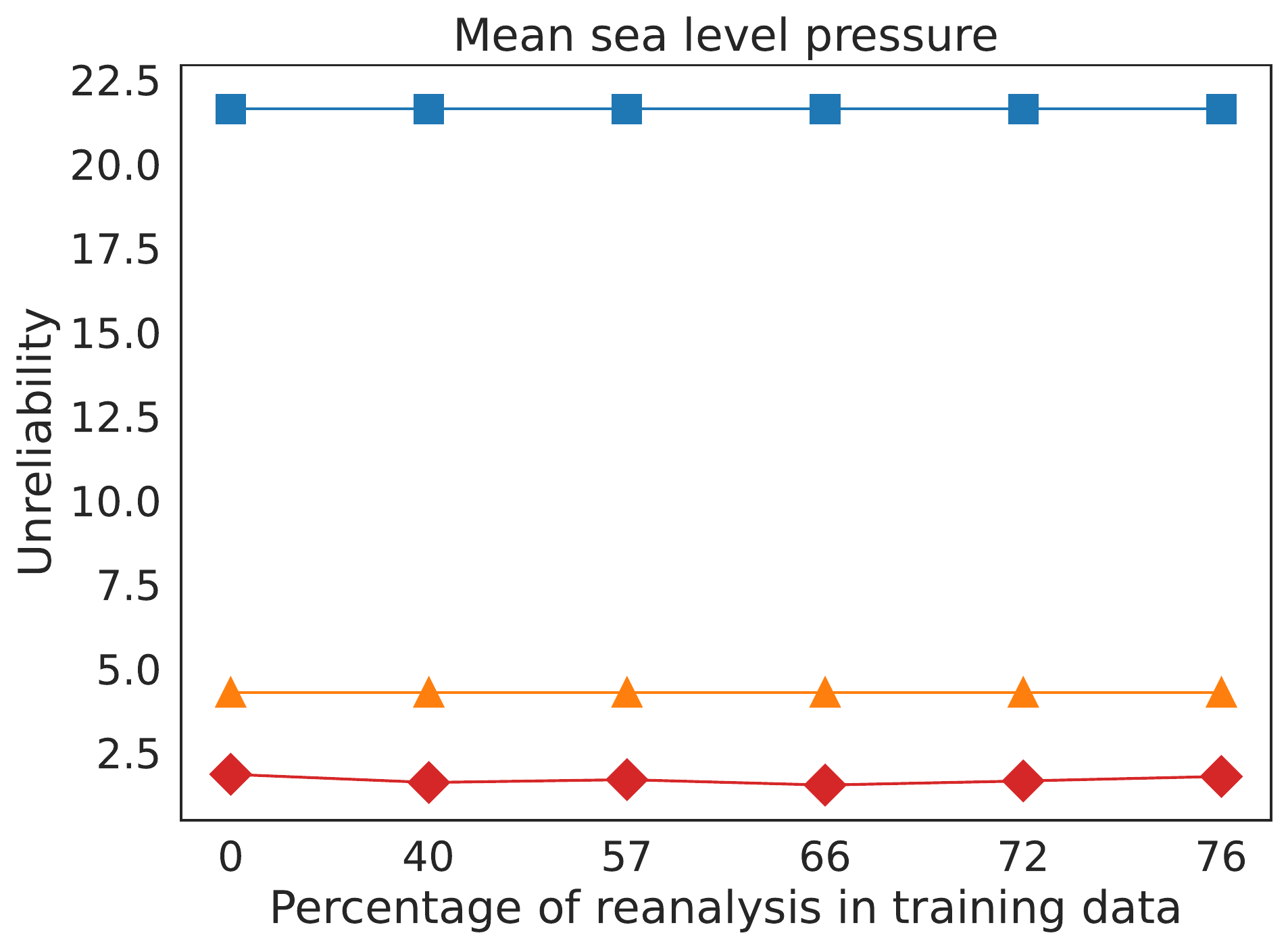}
    \includegraphics[width=0.32\columnwidth,draft=false]{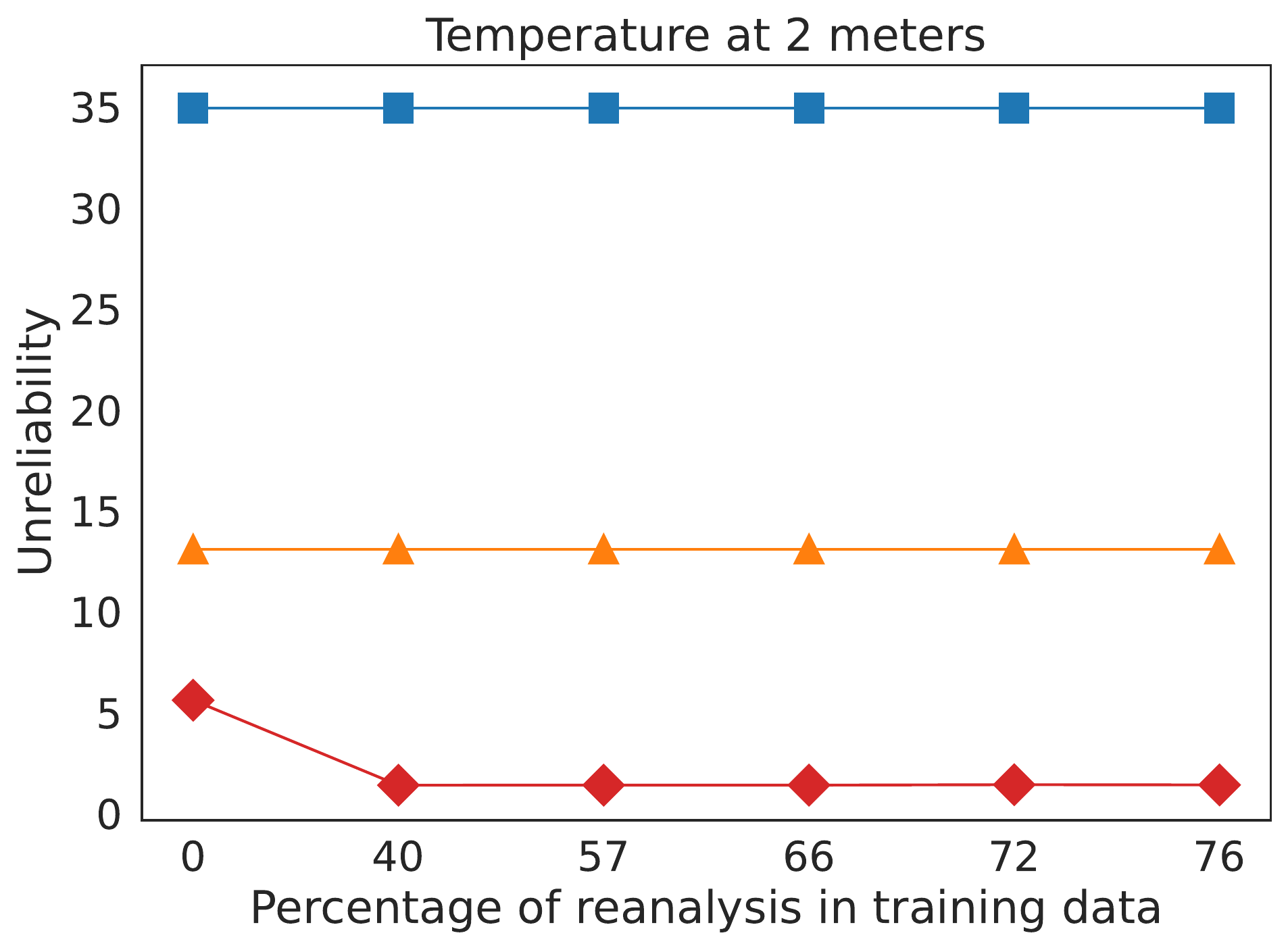}
    \includegraphics[width=0.32\columnwidth,draft=false]{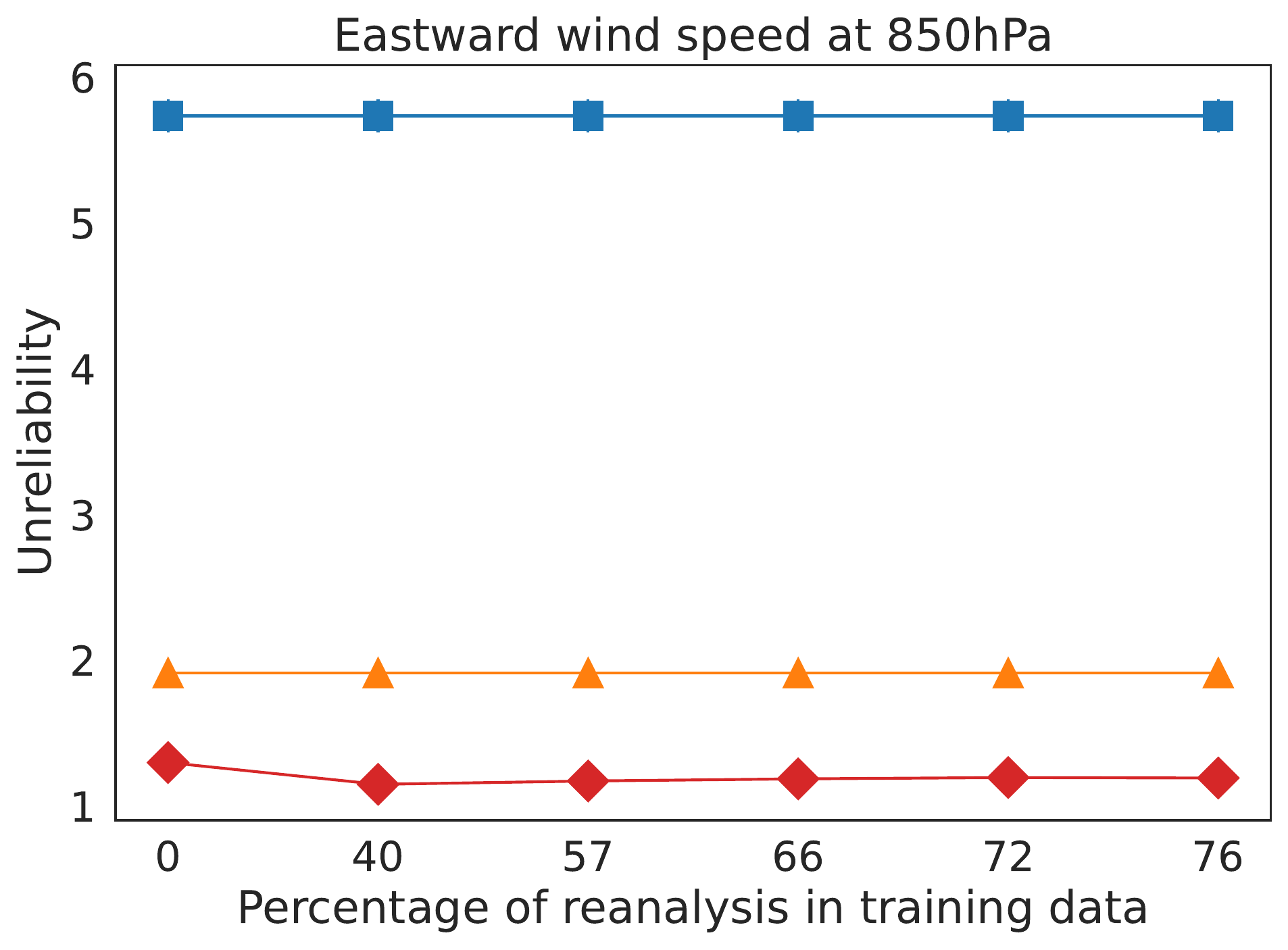}  
    \includegraphics[width=0.32\columnwidth,draft=false]{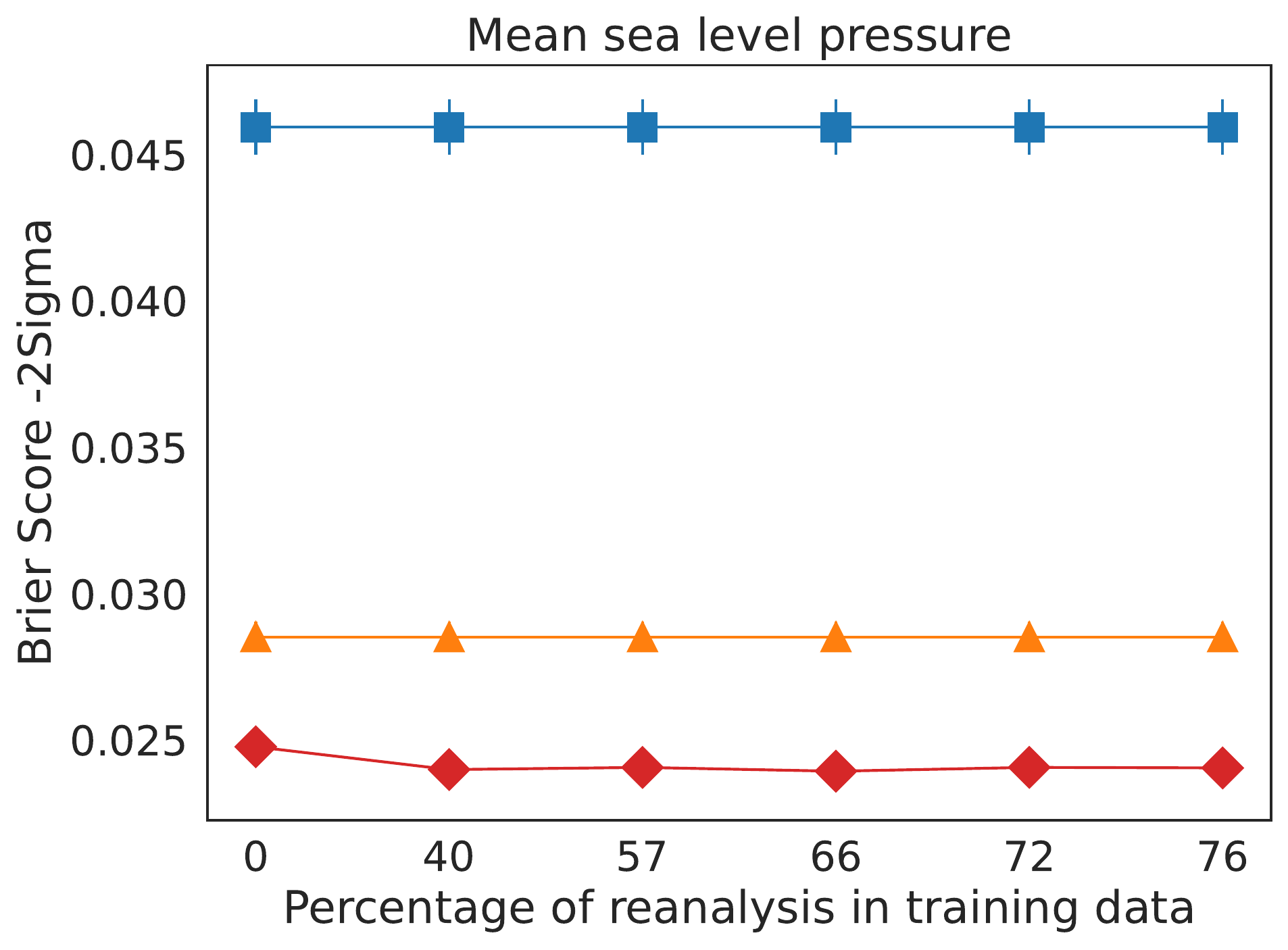}
    \includegraphics[width=0.32\columnwidth,draft=false]{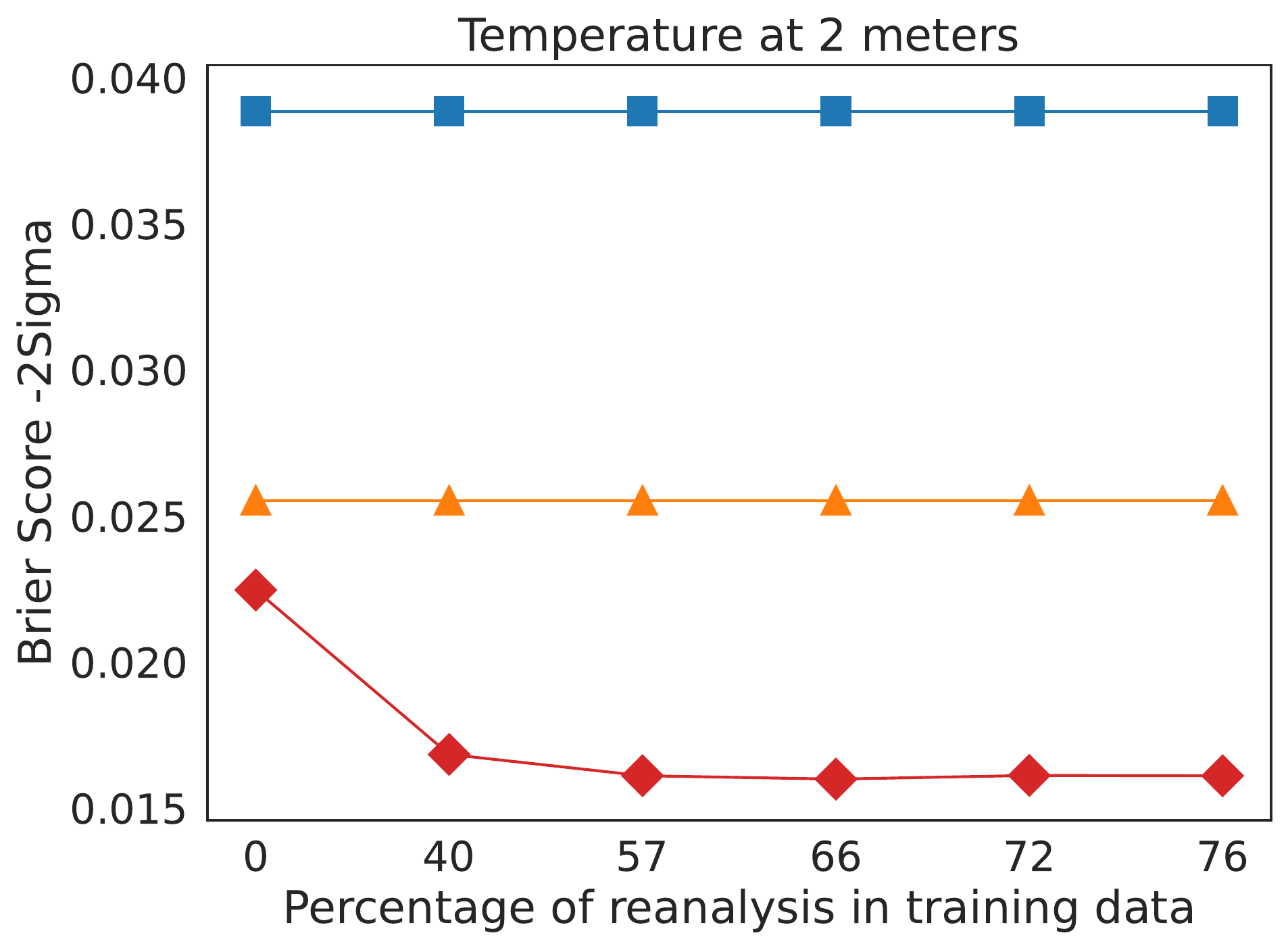}
    \includegraphics[width=0.32\columnwidth,draft=false]{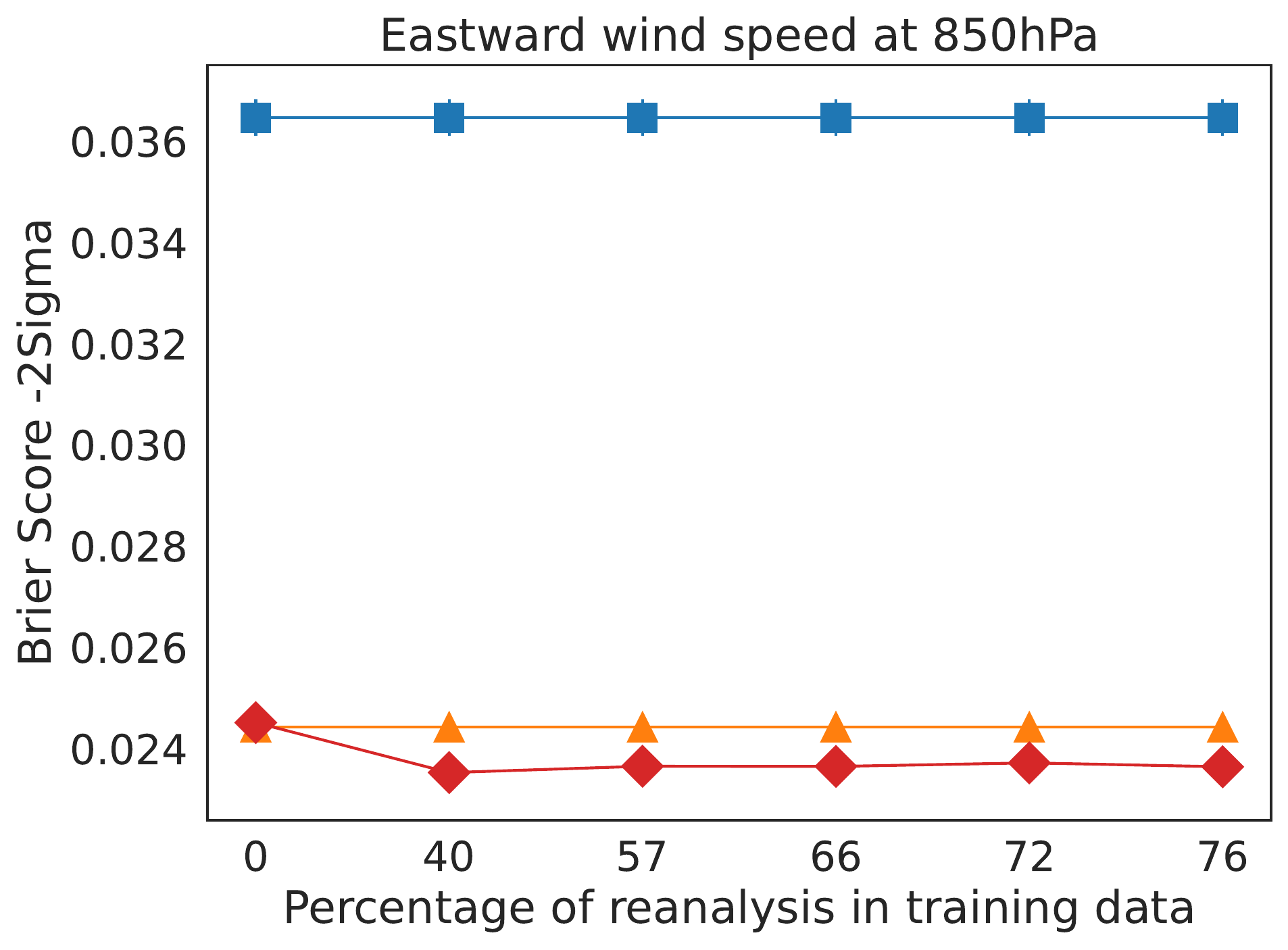}  
    \includegraphics[width=0.40\columnwidth,draft=false]{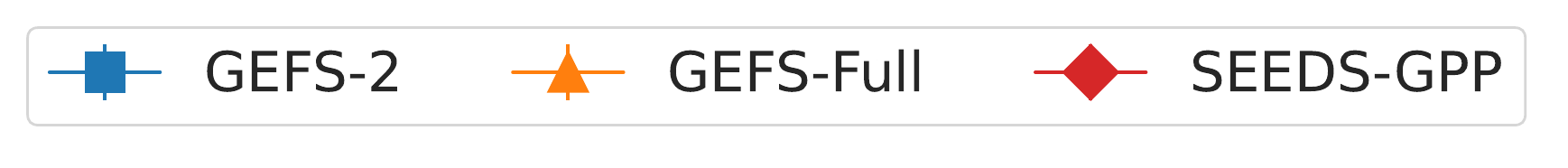}  
    \caption{The effect of $K'$ on various metrics. Top down: RMSE, ACC, CRPS, unreliability, Brier score for $-2\sigma$ event.}
       \label{fEffectOfKPrime}
\end{figure}

Figure~\ref{fEffectOfKPrime} studies the effect of $K'$, which contains the mixture weight, on the skill of \ourgpp forecasts. Since the training data has ensemble size $5$, the number of available GEFS training labels is $3$. We see no forecast skill gains blending in beyond 4 reanalyses members (around 50\% mixing ratio). In the main text, we report results with $K'=3$.

\subsubsection{Case Study:  Visualization of Generated Ensembles}

Ensemble forecasts are samples drawn from multivariate high-dimensional distributions. Making comparisons among these distributions using a limited number of samples is a challenging statistical task. In this case study, we study several ensembles by reducing the dimensions and focusing on two fields at a single spatial location.

\begin{figure}[t]
    \centering
    \begin{sideways} \hspace{0.5in}SEEDS-GEE \end{sideways}
    \includegraphics[width=0.31\columnwidth,draft=false]{figs/kde_plots/gee_c2_png/c2_case2}
    \includegraphics[width=0.31\columnwidth,draft=false]{figs/kde_plots/gee_c2_png/c2_case4}
    \includegraphics[width=0.31\columnwidth,draft=false]{figs/kde_plots/gee_c2_png/c2_case6}
    
    \begin{sideways} \hspace{0.5in}SEEDS-GPP \end{sideways}
    \includegraphics[width=0.31\columnwidth,draft=false]{figs/kde_plots/gpp_c2_png/c2_case2}
    \includegraphics[width=0.31\columnwidth,draft=false]{figs/kde_plots/gpp_c2_png/c2_case4}
    \includegraphics[width=0.31\columnwidth,draft=false]{figs/kde_plots/gpp_c2_png/c2_case6}
    
    \begin{sideways} \hspace{0.5in}SEEDS-GEE \end{sideways}
    \includegraphics[width=0.31\columnwidth,draft=false]{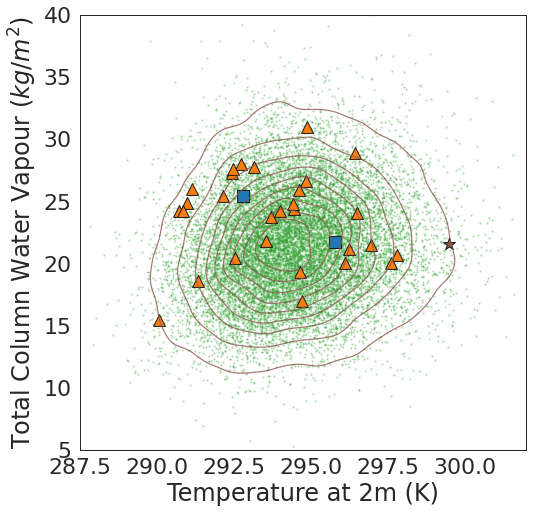}
    \includegraphics[width=0.31\columnwidth,draft=false]{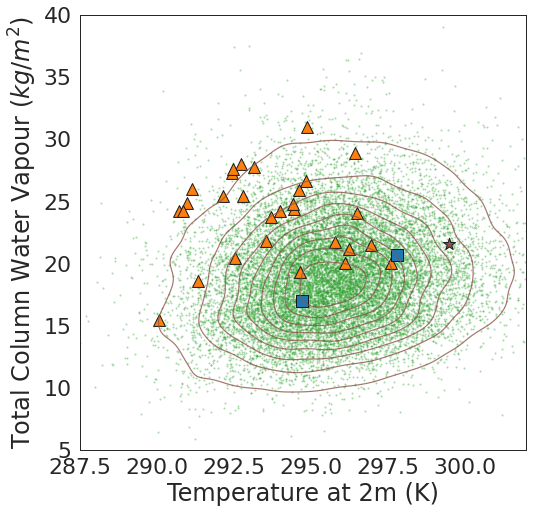}
    \includegraphics[width=0.31\columnwidth,draft=false]{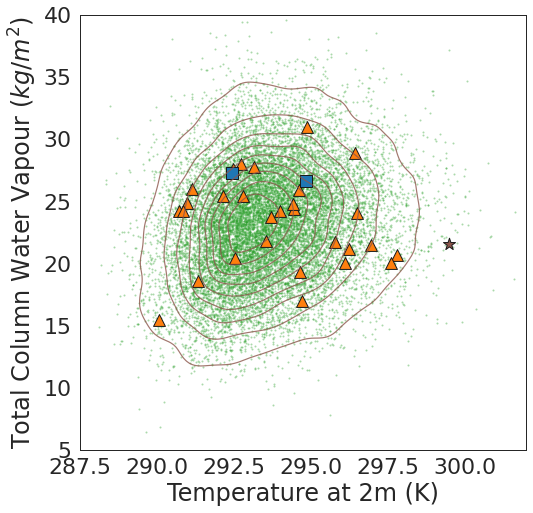}
    
    \begin{sideways} \hspace{0.5in}SEEDS-GPP \end{sideways}
    \includegraphics[width=0.31\columnwidth,draft=false]{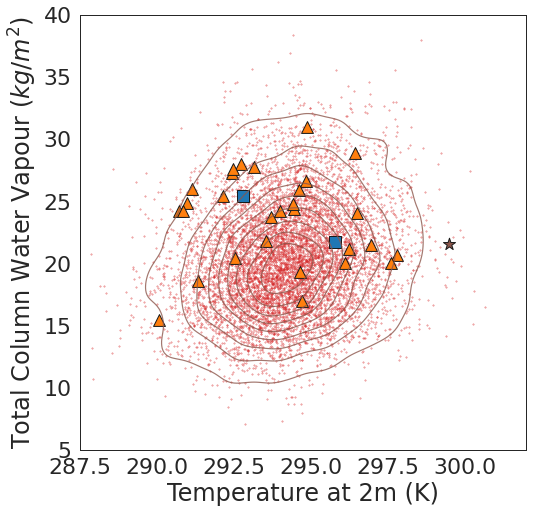}
    \includegraphics[width=0.31\columnwidth,draft=false]{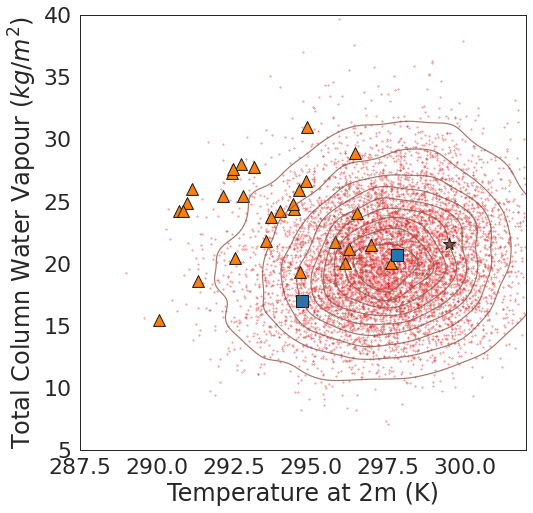}
    \includegraphics[width=0.31\columnwidth,draft=false]{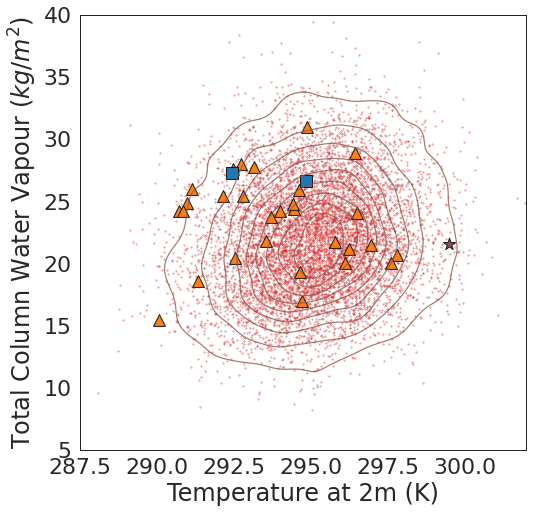}
 
    \includegraphics[width=0.5\columnwidth,draft=false]{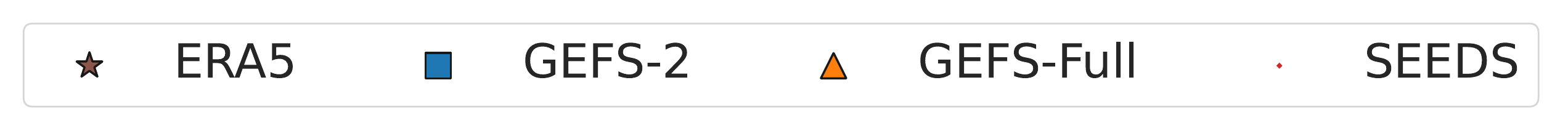}
    \caption{Temperature at 2 meters and total column water vapour of the grid point near Lisbon on 2022/07/14. SEEDS-GEE (green) and SEEDS-GPP (red) with $N=16384$ samples (dots and kernel density level sets) were conditioned on various GEFS subsamples of size $K=2$ (square).
    \label{fKDEK2}
    }
\end{figure}

\begin{figure}[t]
    \centering
    \begin{sideways} \hspace{0.3in}Conditioned on 2 \end{sideways}
    \includegraphics[width=0.31\columnwidth,draft=false]{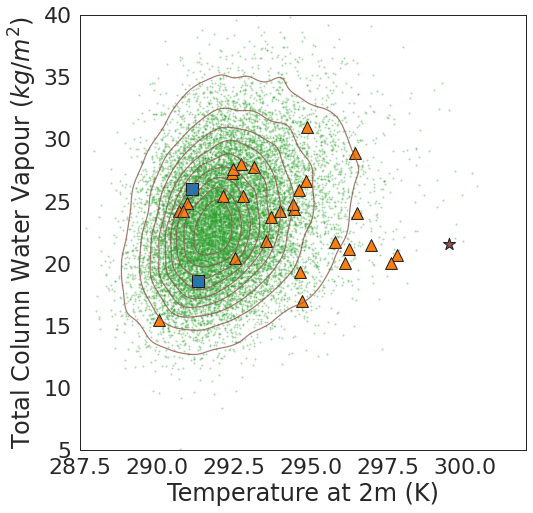}
    \includegraphics[width=0.31\columnwidth,draft=false]{figs/kde_plots/gee_c2_png/c2_case4}
    \includegraphics[width=0.31\columnwidth,draft=false]{figs/kde_plots/gee_c2_png/c2_case8}
    
    \begin{sideways} \hspace{0.3in}Conditioned on 4 \end{sideways}
    \includegraphics[width=0.31\columnwidth,draft=false]{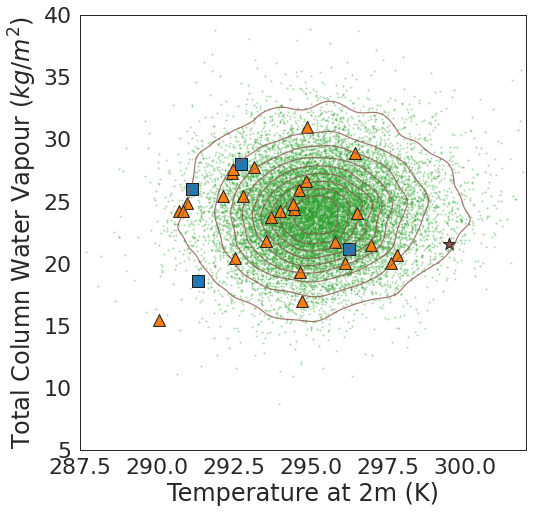}
    \includegraphics[width=0.31\columnwidth,draft=false]{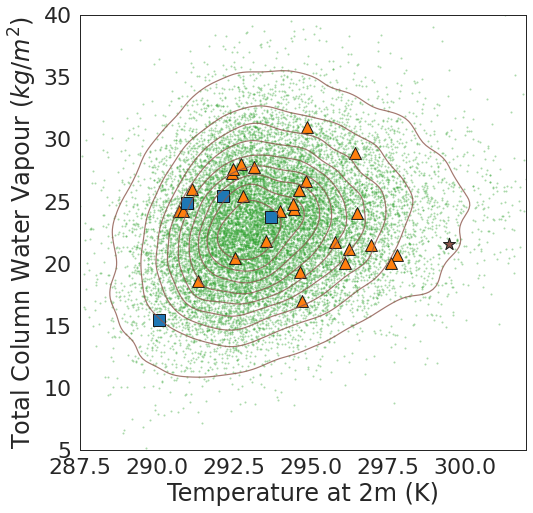}
    \includegraphics[width=0.31\columnwidth,draft=false]{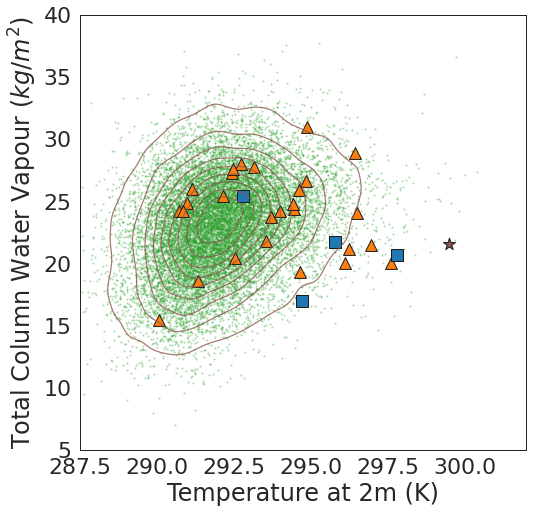}
    
    \begin{sideways} \hspace{0.3in}Conditioned on 2 \end{sideways}
    \includegraphics[width=0.31\columnwidth,draft=false]{figs/kde_plots/gee_c2_png/c2_case12}
    \includegraphics[width=0.31\columnwidth,draft=false]{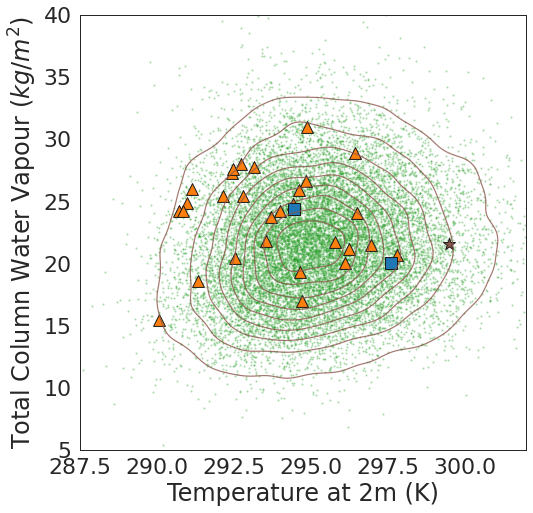}
    \includegraphics[width=0.31\columnwidth,draft=false]{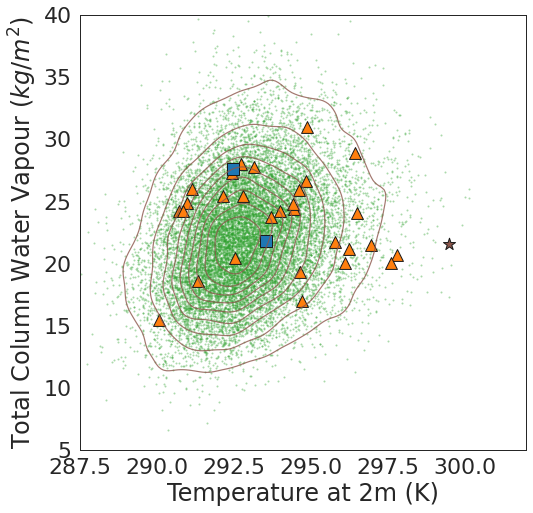}
    
    \begin{sideways} \hspace{0.3in}Conditioned on 4 \end{sideways}
    \includegraphics[width=0.31\columnwidth,draft=false]{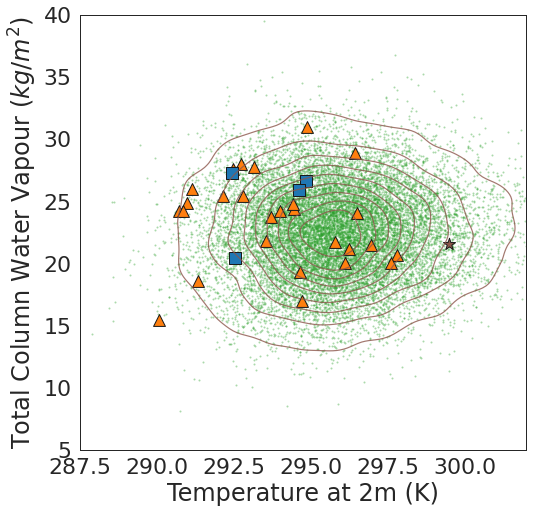}
    \includegraphics[width=0.31\columnwidth,draft=false]{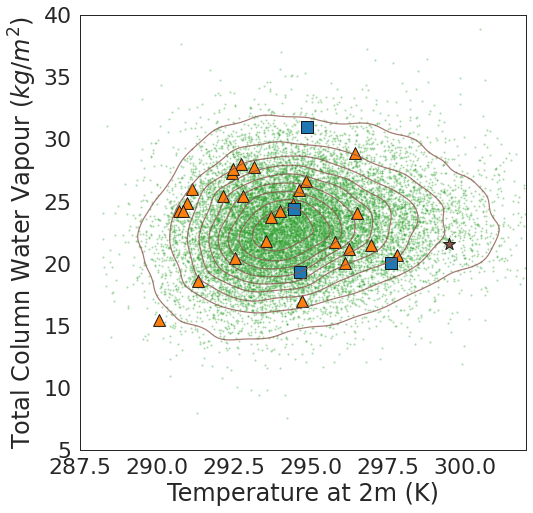}
    \includegraphics[width=0.31\columnwidth,draft=false]{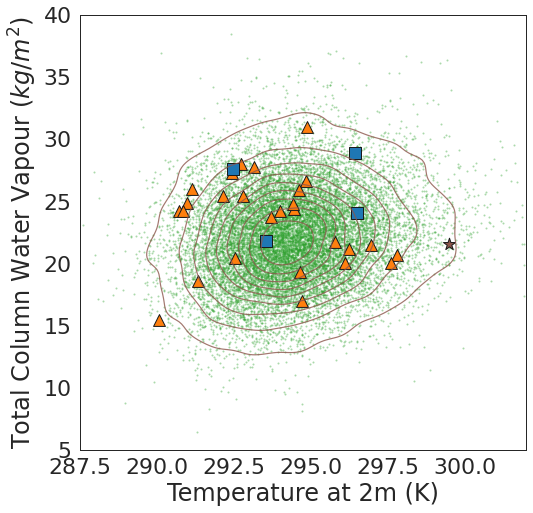}
    \\
    \includegraphics[width=0.5\columnwidth,draft=false]{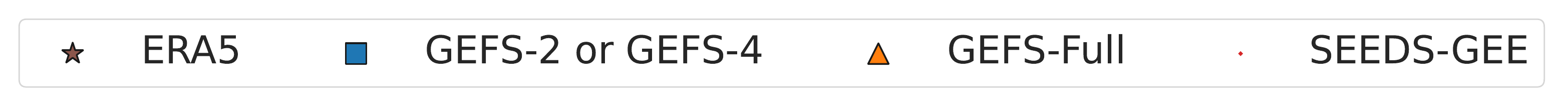}
    \caption{Temperature at 2 meters and total column water vapour of the grid point near Lisbon on 2022/07/14. SEEDS-GEE with $N=16384$ samples (dots and kernel density level sets) were conditioned on various GEFS subsamples of size $K=4$ (square).
    \label{fKDEK4}
    }
\end{figure}

Figures~\ref{fKDEK2} and \ref{fKDEK4} visualize the joint distributions of 
temperature at 2 meters and total column water vapour at the grid point near Lisbon during the extreme heat event on 2022/07/14. We used the 7-day forecasts made on 2022/07/07. For each plot, we generate 16,384-member ensembles, represented by the orange dots using different seeds from the 31-member GEFS ensemble. The density level sets are computed by fitting a kernel density estimator to the ensemble forecasts.  The observed weather event is denoted by the black star.  The operational ensembles, denoted by blue rectangles (seeds) and orange triangles,  do not cover the extreme event well, while the generated ensembles extrapolate from the two seeding forecasts, providing a better coverage of the event. 

In Figure~\ref{fKDEK2}, we compare \ourgee and \ourgpp generated samples conditioned on 6 different randomly sampled two seeding forecasts from the 31-member GEFS ensemble. The plots are arranged in 2 pairs of rows (4 rows in total) such that all the odd rows are from \ourgee and even rows are from \ourgpp and that on each paired rows, the seeds for \ourgee above and \ourgpp below are exactly the same. We see that given the same seeds, the \ourgpp distribution is correctly more biased towards warmer temperatures while maintaining the distribution shape and spread compared to \ourgee.

In Figure~\ref{fKDEK4} we compare \ourgee based on 2 seeds with \ourgee based on 4 seeds. We first generate samples conditioned on 4 randomly sampled seeding forecasts from the 31-member GEFS ensemble. Then we generate samples conditioned on 2 randomly selected seeding forecasts from the 4 chosen in the first round. The plots show 6 such pairs of samples, arranged in 2 pairs of rows (4 rows in total) such that all odd rows use 2 seeds and even rows use 4 seeds and that for each pair of rows, the 4 seeds below always contain the 2 seeds above. We observe that the distribution of the generated samples conditioned on 4 seeds are more similar to each other, which verifies the intuition that more seeds lead to more robustness to the sampling of the seeds. More seeds also lead to better coverage of the extreme heat event by the generated envelopes.

\end{document}